\documentclass[10pt,twocolumn,letterpaper]{article}

\usepackage{iccv}
\usepackage{times}
\usepackage{epsfig}
\usepackage{graphicx}
\usepackage{amsmath}
\usepackage{amssymb}

\usepackage{booktabs}
\usepackage{enumerate}
\usepackage{multirow}
\usepackage[ruled,linesnumbered]{algorithm2e}
\usepackage{color}
\usepackage{makecell}

\usepackage{animate}

\usepackage[hyphens]{url}
\usepackage{breakurl}

\usepackage[pagebackref=true,breaklinks=true,letterpaper=true,colorlinks,bookmarks=false]{hyperref}

\iccvfinalcopy 

\def\iccvPaperID{****} 

\ificcvfinal\fi

\begin{document}

\title{StyleDiffusion: Controllable Disentangled Style Transfer via Diffusion Models}

\author{Zhizhong Wang\thanks{This work was done when Zhizhong Wang was an intern at Huawei.},\hspace{0.4cm}Lei Zhao\thanks{Corresponding author.},\hspace{0.4cm}Wei Xing \\
	College of Computer Science and Technology, Zhejiang University\\
	{\tt\small \{endywon, cszhl, wxing\}@zju.edu.cn}
}

\maketitle
\ificcvfinal\thispagestyle{empty}\fi

\begin{abstract}
Content and style (C-S) disentanglement is a fundamental problem and critical challenge of style transfer. Existing approaches based on explicit definitions (\eg, Gram matrix) or implicit learning (\eg, GANs) are neither interpretable nor easy to control, resulting in entangled representations and less satisfying results. In this paper, we propose a new C-S disentangled framework for style transfer without using previous assumptions. The key insight is to explicitly extract the content information and implicitly learn the complementary style information, yielding interpretable and controllable C-S disentanglement and style transfer. A simple yet effective CLIP-based style disentanglement loss coordinated with a style reconstruction prior is introduced to disentangle C-S in the CLIP image space. By further leveraging the powerful style removal and generative ability of diffusion models, our framework achieves superior results than state of the art and flexible C-S disentanglement and trade-off control. Our work provides new insights into the C-S disentanglement in style transfer and demonstrates the potential of diffusion models for learning well-disentangled C-S characteristics.
\end{abstract}

\section{Introduction}
\label{intro}
Given a reference style image, \eg, {\em Starry Night} by Vincent Van Gogh, style transfer aims to transfer its artistic style, such as colors and brushstrokes, to an arbitrary content target. To achieve such a goal, it must first properly separate the style from the content and then transfer it to another content. This raises two fundamental challenges: (1) ``how to disentangle content and style (C-S)" and (2) ``how to transfer style to another content". 

To resolve these challenges, valuable efforts have been devoted. Gatys~\etal~\cite{gatys2016image} proposed {\em A Neural Algorithm of Artistic Style} to achieve style transfer, which {\em explicitly} defines the high-level features extracted from a pre-trained Convolutional Neural Network (CNN) (\eg, VGG~\cite{simonyan2014very}) as content, and the feature correlations (\ie, Gram matrix) as style. This approach acquires visually stunning results and inspires a large number of successors~\cite{johnson2016perceptual,huang2017arbitrary,li2017universal,an2021artflow,zhang2022exact}. Despite the successes, by diving into the essence of style transfer, we observed three problems with these approaches: (1) The C-S are not completely disentangled. Theoretically, the C-S representations are intertwined. For example, matching the content representation of an image may also match its Gram matrix, and vice versa. (2) What CNN learned is a black box rugged to interpret~\cite{zhang2021survey}, which makes the C-S definitions~\cite{gatys2016image} uninterpretable and hard to control. (3) The transfer process is modeled as a separate optimization of content loss and style loss~\cite{gatys2016image}, so there lacks a deep understanding of the relationship between C-S. These problems usually lead to unbalanced stylizations and disharmonious artifacts~\cite{chen2021artistic}, as will be shown in later Fig.~\ref{fig:quality}.

On the other hand, disentangled representation learning~\cite{higgins2018towards} provides other ideas to {\em implicitly} disentangle C-S, either supervised~\cite{kulkarni2015deep,karaletsos2015bayesian} or unsupervised~\cite{chen2016infogan,zhang2018separating}. For style transfer, Kotovenko~\etal~\cite{kotovenko2019content1} utilized fixpoint triplet style loss and disentanglement loss to enforce a GAN~\cite{goodfellow2014generative}-based framework to learn separate C-S representations in an unsupervised manner. Similarly, TPFR~\cite{svoboda2020two} learned to disentangle C-S in latent space via metric learning and two-stage peer-regularization, producing high-quality images even in the zero-shot setting. While these approaches successfully enforce properties ``encouraged” by the corresponding losses, they still have three main problems: (1) Well-disentangled models seemingly cannot be identified without supervision~\cite{locatello2019challenging,ren2021rethinking}, which means the unsupervised learning~\cite{kotovenko2019content1,svoboda2020two} may not achieve truly disentangled C-S, as will be shown in later Fig.~\ref{fig:quality}. (2) These approaches are all based on GANs and thus often confined to the GAN pre-defined domains, \eg, a specific artist's style domain~\cite{sanakoyeu2018style}. (3) The implicitly learned C-S representations are still black boxes that are hard to interpret and control~\cite{locatello2019challenging}.

Facing the challenges above, in this paper, we propose a new C-S disentangled framework for style transfer {\em without using previous assumptions} such as Gram matrix~\cite{gatys2016image} or GANs~\cite{kotovenko2019content1}. Our key insight stems from the fact that the definition of an image's style is much more complex than its content, \eg, we can easily identify the content of a painting by its structures, semantics, or shapes, but it is intractable to define the style~\cite{parker1991perception,graham2012statistics,karayev2013recognizing,wang2021evaluate}. Therefore, we can bypass such a dilemma by {\em explicitly} extracting the content information and {\em implicitly} learning its {\em complementary} style information. Since we strictly constrain style as the {\em complement} of content, the C-S can be completely disentangled, and the control of disentanglement has been transformed into the control of content extraction. It achieves both controllability and interpretability.

However, achieving plausible and controllable content extraction is also non-trivial because the contents extracted from the content images and style images should share the same content domain, and the details of the extracted contents should be easy to control. To this end, we resort to recent developed diffusion models~\cite{ho2020denoising,song2020denoising} and introduce a {\em diffusion-based style removal module} to smoothly dispel the style information of the content and style images, extracting the domain-aligned content information. Moreover, owing to the strong generative capability of diffusion models, we also introduce a {\em diffusion-based style transfer module} to better learn the disentangled style information of the style image and transfer it to the content image. The style disentanglement and transfer are encouraged via a simple yet effective {\em CLIP~\cite{radford2021learning}-based style disentanglement loss}, which induces the transfer mapping of the content image's content to its stylization (\ie, the stylized result) to be aligned with that of the style image's content to its stylization (\ie, the style image itself) in the CLIP image space. By further coordinating with a {\em style reconstruction prior}, it achieves both generalized and faithful style transfer. We conduct comprehensive comparisons and ablation study to demonstrate the effectiveness and superiority of our framework. With the well-disentangled C-S, it achieves very promising stylizations with fine style details, well-preserved contents, and a deep understanding of the relationship between C-S.

In summary, our contributions are threefold:
\begin{itemize}
	\setlength{\itemsep}{2pt}
	\setlength{\parsep}{0pt}
	\setlength{\partopsep}{0pt}
	\setlength{\parskip}{0pt}
	
	\item We propose a novel C-S disentangled framework for style transfer, which achieves more interpretable and controllable C-S disentanglement and higher-quality stylized results.
	
	\item We introduce diffusion models to our framework and demonstrate their effectiveness and superiority in controllable style removal and learning well-disentangled C-S characteristics.
	
	\item A new CLIP-based style disentanglement loss coordinated with a style reconstruction prior is introduced to disentangle C-S in the CLIP image space.
\end{itemize}

\section{Related Work}
{\bf Neural Style Transfer (NST).} The pioneering work of Gatys~\etal~\cite{gatys2016image} has opened the era of NST~\cite{jing2019neural}. Since then, this task has experienced tremendous progress, including efficiency~\cite{johnson2016perceptual,li2019learning,wang2023microast}, quality~\cite{gu2018arbitrary,wang2020glstylenet,lin2021drafting,cheng2021style,chen2021dualast,an2021artflow,kotovenko2021rethinking,wang2021rethinking,liu2021adaattn,chen2021artistic,wu2021styleformer,huo2021manifold,zuo2022style,deng2022stytr2,zhang2022domain,wang2022aesust}, generality~\cite{chen2017stylebank,huang2017arbitrary,li2017universal,park2019arbitrary,deng2020arbitrary,jing2020dynamic,hong2021domain,wang2022texture,zhang2022exact,lu2022universal,wu2022ccpl}, and diversity~\cite{ulyanov2017improved,wang2020diversified,wang2022divswapper}. Despite these successes, the essence of these approaches is mostly based on the {\em explicitly} defined C-S representations, such as Gram matrix~\cite{gatys2016image}, which have several limitations as discussed in Sec.~\ref{intro}. In our work, we propose new disentangled C-S representations {\em explicitly} extracted or {\em implicitly} learned by diffusion models, achieving more effective style transfer and higher-quality results.

{\bf Disentangled Representation Learning (DRL).} The task of DRL~\cite{higgins2018towards} aims at modeling the factors of data variations~\cite{lee2018diverse}. Earlier works used labeled data to factorize representations in a supervised manner~\cite{karaletsos2015bayesian}. Recently, unsupervised settings have been largely explored~\cite{kim2018disentangling}, especially for disentangling style from content~\cite{zhang2018separating,huang2018multimodal,lee2018diverse,kazemi2019style,wu2019disentangling,kotovenko2019content1,park2020swapping,ren2021rethinking,chen2021diverse,kwon2021diagonal}. However, due to the dependence on GANs~\cite{goodfellow2014generative}, their C-S disentanglement is usually restricted in the GAN pre-defined domains (\eg, Van Gogh's style domain). Besides, disentanglement cannot be effectively achieved without providing sufficient data~\cite{locatello2019challenging}. In contrast, our framework learns the disentangled style from a single style image, and the disentanglement can be easily achieved by providing only a few ($\sim$50) content images for training.

{\bf Diffusion Models.} Diffusion models~\cite{sohl2015deep} such as denoising diffusion probabilistic models (DDPMs)~\cite{ho2020denoising,nichol2021improved} have recently shown great success in image generation~\cite{song2020denoising,dhariwal2021diffusion,fan2023frido}, image manipulation~\cite{meng2021sdedit,avrahami2022blended,kim2022diffusionclip}, and text-conditional synthesis~\cite{nichol2022glide,saharia2022photorealistic,ramesh2022hierarchical,rombach2022high,gu2022vector,blattmann2022retrieval,li2023gligen}. These works have demonstrated the power of diffusion models to achieve higher-quality results than other generative models like VAEs~\cite{van2017neural}, auto-regressive models~\cite{esser2021taming}, flows~\cite{kingma2018glow}, and GANs~\cite{karras2019style}. Inspired by them, we introduce a diffusion-based style removal module and a style transfer module in our framework. These modules can smoothly remove the style information of images and better learn the recovery of it to achieve higher-quality style transfer results. {\em To the best of our knowledge, our work is the first to introduce diffusion models to the field of neural style transfer}.

\begin{figure*}[t]
	\centering
	\includegraphics[width=0.94\linewidth]{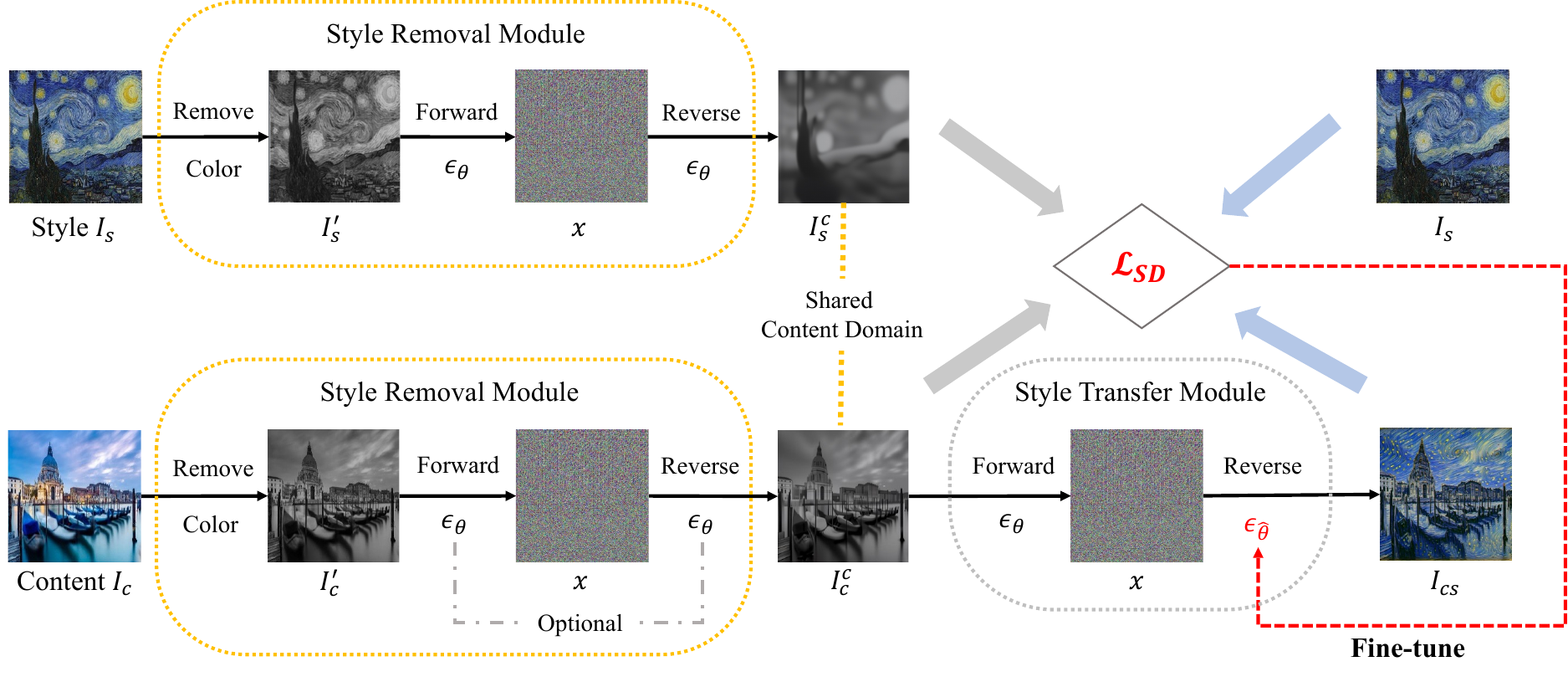}
	\caption{{\bf Overview of our proposed StyleDiffusion.} The content image $I_c$ and style image $I_s$ are first fed into a diffusion-based style removal module to explicitly extract the domain-aligned content information. Then, the content of $I_c$ is fed into a diffusion-based style transfer module to obtain the stylized result $I_{cs}$. During training, we fine-tune the style transfer module via a CLIP-based style disentanglement loss $\mathcal{L}_{SD}$ coordinated with a style reconstruction prior (see details in Sec.~\ref{sec:loss}, we omit it here for brevity) to implicitly learn the disentangled style information of $I_s$.
	}
	\label{fig:overview}
\end{figure*}

\section{Background}

Denoising diffusion probabilistic models (DDPMs)~\cite{sohl2015deep,ho2020denoising} are latent variable models that consist of two diffusion processes, \ie, a forward diffusion process and a reverse diffusion process. The forward process is a fixed Markov Chain that sequentially produces a series of latents $x_1,...,x_T$ by gradually adding Gaussian noises at each timestep $t\in[1,T]$:
\begin{equation}
	q(x_t|x_{t-1}) := \mathcal{N}(\sqrt{1-\beta_t}x_{t-1},\beta_t{\bf I}),
\end{equation}
where $\beta_t\in(0,1)$ is a fixed variance schedule. An important property of the forward process is that given clean data $x_0$, $x_t$ can be directly sampled as:
\begin{equation}
	\begin{aligned}
		q(x_t|x_0) &:= \mathcal{N}(\sqrt{\bar{\alpha}_t}x_0,(1-\bar{\alpha}_t){\bf I}), \\
		x_t &:= \sqrt{\bar{\alpha}_t} x_0 + \sqrt{1-\bar{\alpha}_t}\epsilon,
	\end{aligned}
	\label{eq:ddpmf}
\end{equation}
where $\alpha_t := 1-\beta_t$ and $\bar{\alpha}_t := \prod_{s=0}^{t} \alpha_s$. Noise $\epsilon \sim \mathcal{N}(0,{\bf I})$ has the same dimensionality as data $x_0$ and latent $x_t$.

The reverse process generates a reverse sequence by sampling the posteriors $q(x_{t-1}|x_t)$, starting from a Gaussian noise sample $x_T\sim\mathcal{N}(0,{\bf I})$. However, since $q(x_{t-1}|x_t)$ is intractable, DDPMs learn parameterized Gaussian transitions $p_\theta(x_{t-1}|x_t)$ with a learned mean $\mu_\theta(x_t,t)$ and a fixed variance $\sigma_t^2{\bf I}$~\cite{ho2020denoising}:
\begin{equation}
	p_\theta(x_{t-1}|x_t) := \mathcal{N}(\mu_\theta(x_t,t), \sigma_t^2{\bf I}),
\end{equation}
where $\mu_\theta(x_t,t)$ is the function of a noise approximator $\epsilon_\theta(x_t,t)$. Then, the reverse process can be expressed as:
\begin{equation}
	x_{t-1} := \frac{1}{\sqrt{\alpha_t}} (x_t - \frac{1-\alpha_t}{\sqrt{1-\bar{\alpha}_t}} \epsilon_\theta(x_t,t)) + \sigma_t{\bf z},
\end{equation}
where ${\bf z} \sim \mathcal{N}(0, {\bf I})$ is a standard Gaussian noise independent of $x_t$. $\epsilon_\theta(x_t,t)$ is learned by a deep neural network~\cite{ronneberger2015u} through optimizing the following loss:
\begin{equation}
	\mathop{\min}_\theta \parallel \epsilon_\theta(x_t,t) - \epsilon \parallel^2.
\end{equation}

Later, instead of using the fixed variances, Nichol and Dhariwal~\cite{nichol2021improved} presented a strategy for learning the variances. Song~\etal~\cite{song2020denoising} proposed DDIM, which formulates an alternative non-Markovian noising process that has the same forward marginals as DDPM but allows a different reverse process:
\begin{equation}
	x_{t-1} := \sqrt{\bar{\alpha}_{t-1}} f_\theta(x_t,t) +\sqrt{1-\bar{\alpha}_{t-1}-\sigma_t^2} \epsilon_\theta (x_t,t) + \sigma_t {\bf z},
	\label{eq:ddim1}	
\end{equation}
where $f_\theta(x_t,t)$ is the predicted $x_0$ at timestep $t$ given $x_t$ and $\epsilon_\theta(x_t,t)$:
\begin{equation}
	f_\theta(x_t,t):= \frac{x_t - \sqrt{1-\bar{\alpha}_t}\epsilon_\theta(x_t,t)}{\sqrt{\bar{\alpha}_t}}.	
	\label{eq:ftheta}
\end{equation}
Changing the choice of $\sigma_t$ values in Eq.~(\ref{eq:ddim1}) can achieve different reverse processes. Especially when $\sigma_t=0$, which is called DDIM~\cite{song2020denoising}, the reverse process becomes a deterministic mapping from latents to images, which enables nearly perfect inversion~\cite{kim2022diffusionclip}. Besides, it can also accelerate the reverse process with much fewer sampling steps~\cite{dhariwal2021diffusion,kim2022diffusionclip}.

\section{Method}
Our task can be described as follows: given a style image $I_s$ and an arbitrary content image $I_c$, we want to first disentangle the content and style of them and then transfer the style of $I_s$ to the content of $I_c$. To achieve so, as stated in Sec.~\ref{intro}, our key idea is to explicitly extract the content information and then implicitly learn the {\em complementary} style information. Since our framework is built upon diffusion models~\cite{ho2020denoising,song2020denoising}, we dub it {\em StyleDiffusion}.

Fig.~\ref{fig:overview} shows the overview of our StyleDiffusion, which consists of three key ingredients: I) a diffusion-based style removal module, II) a diffusion-based style transfer module, and III) a CLIP-based style disentanglement loss coordinated with a style reconstruction prior. In the following subsections, we will introduce each of them in detail.

\subsection{Style Removal Module}
\label{sec:SRM}
The style removal module aims at removing the style information of the content and style images, explicitly extracting the domain-aligned content information. Any reasonable content extraction operation can be used, depending on how the users define the content. For instance, users may want to use the structural outline as the content, so they can extract the outlines~\cite{kang2007coherent,xie2015holistically} here. However, as discussed in Sec.~\ref{intro}, one challenge is {\em controllability} since the control of C-S disentanglement has been transformed into the control of content extraction. To this end, we introduce a diffusion-based style removal module to achieve both plausible and controllable content extraction.

Given an input image, \eg, the style image $I_s$, since the color is an integral part of style~\cite{lang1987concept}, our style removal module first removes its color by a commonly used ITU-R 601-2 luma transform~\cite{gonzalez2009digital}. The obtained grayscale image is denoted as $I_s'$. Then, we leverage a pre-trained diffusion model~\cite{dhariwal2021diffusion} $\epsilon_\theta$ to remove the style details such as brushstrokes and textures of $I_s'$, extracting the content $I_s^c$. The insight is that the pre-trained diffusion model can help eliminate the domain-specific characteristics of input images and align them to the pre-trained domain~\cite{choi2021ilvr,kim2022diffusionclip}. We assume that images with different styles belong to different domains, but their contents should share the same domain. Therefore, we can pre-train the diffusion model on a surrogate domain, \eg, the photograph domain, and then use this domain to construct the contents of images. After pre-training, the diffusion model can convert the input images from diverse domains to the latents $x$ via the forward process and then inverse them to the photograph domain via the reverse process. In this way, the style characteristics can be ideally dispelled, leaving only the contents of the images.

Specifically, in order to obtain the results with fewer sampling steps and ensure that the content structures of the input images can be well preserved, we adopt the deterministic DDIM~\cite{song2020denoising} sampling as the reverse process (Eq.~(\ref{eq:ddim})), and the ODE approximation of its reversal~\cite{kim2022diffusionclip} as the forward process (Eq.~(\ref{eq:ddim_ode})):
\begin{equation}
	x_{t-1} = \sqrt{\bar{\alpha}_{t-1}}f_\theta(x_t,t) + \sqrt{1-\bar{\alpha}_{t-1}} \epsilon_\theta(x_t,t),
	\label{eq:ddim}
\end{equation}
\begin{equation}
	x_{t+1} = \sqrt{\bar{\alpha}_{t+1}}f_\theta(x_t,t) + \sqrt{1-\bar{\alpha}_{t+1}} \epsilon_\theta(x_t,t),
	\label{eq:ddim_ode}
\end{equation}
where $f_\theta(x_t,t)$ is defined in Eq.~(\ref{eq:ftheta}). The forward and reverse diffusion processes enable us to easily control the intensity of style removal by adjusting the number of return step $T_{remov}$ (see details in later Sec.~\ref{sec:impd}). With the increase of $T_{remov}$, more style characteristics will be removed, and the main content structures are retained, as will be shown in later Sec.~\ref{sec:ablation}. Note that for content images that are photographs, the diffusion processes are optional\footnote{Unless otherwise specified, we do not use the diffusion processes for content images in order to better maintain the content structures.} since they are already within the pre-trained domain, and there is almost no style except the colors to be dispelled. The superiority of diffusion-based style removal against other operations, such as Auto-Encoder (AE)~\cite{li2017universal}-based style removal, can be found in {\em supplementary material (SM)}.

\subsection{Style Transfer Module}
\label{sec:STM}
The style transfer module aims to learn the disentangled style information of the style image and transfer it to the content image. A common generative model like AEs~\cite{huang2017arbitrary} can be used here. However, inspired by the recent great success of diffusion models~\cite{dhariwal2021diffusion,kim2022diffusionclip}, we introduce a diffusion-based style transfer module, which can better learn the disentangled style information in our framework and achieve higher-quality and more flexible stylizations (see Sec.~\ref{sec:ablation}).

Given a content image $I_c$, denote $I_c^c$ is the content of $I_c$ extracted by the style removal module (Sec.~\ref{sec:SRM}). We first convert it to the latent $x$ using a pre-trained diffusion model $\epsilon_\theta$. Then, guided by a CLIP-based style disentanglement loss coordinated with a style reconstruction prior (Sec.~\ref{sec:loss}), the {\em reverse process} of the diffusion model is fine-tuned ($\epsilon_\theta \rightarrow \epsilon_{\hat{\theta}}$) to generate the stylized result $I_{cs}$ referenced by the style image $I_s$. Once the fine-tuning is completed, {\em any content image can be manipulated into the stylized result with the disentangled style of the style image $I_s$}. To make the training easier and more stable, we adopt the deterministic DDIM forward and reverse processes in Eq.~(\ref{eq:ddim}) and Eq.~(\ref{eq:ddim_ode}) during the fine-tuning. However, at inference, the stochastic DDPM~\cite{ho2020denoising} forward process (Eq.~(\ref{eq:ddpmf})) can also be used directly to help obtain diverse results~\cite{wang2020diversified} (Sec.~\ref{sec:ablation}). 

\begin{figure*}[t]
	\centering
	\includegraphics[width=1\linewidth]{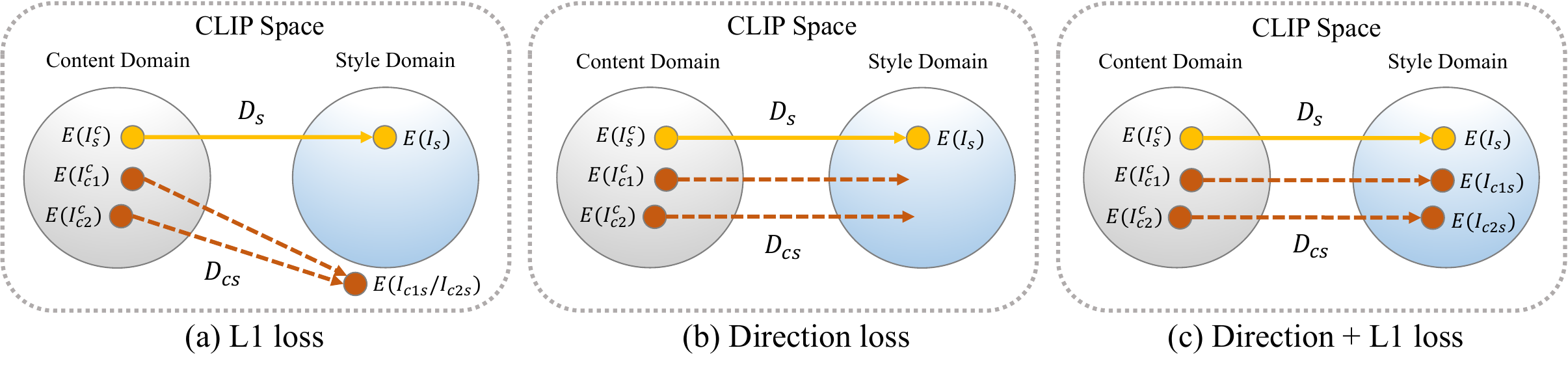}
	\caption{{\bf Illustration of different loss functions} to transfer the disentangled style information. (a) L1 loss cannot guarantee the stylized results are within the style domain and may suffer from a collapse problem. (b) Direction loss aligns the disentangled directions but cannot realize accurate mappings. (c) Combining L1 loss and direction loss is able to achieve accurate one-to-one mappings from the content domain to the style domain.
	}
	\label{fig:loss}
\end{figure*}

\subsection{Loss Functions and Fine-tuning}
\label{sec:loss}
Enforcing the style transfer module (Sec.~\ref{sec:STM}) to learn and transfer the disentangled style information should address two key questions: (1) ``how to regularize the learned style is disentangled'' and (2) ``how to aptly transfer it to other contents''. To answer these questions, we introduce a novel CLIP-based style disentanglement loss coordinated with a style reconstruction prior to train the networks.

{\bf CLIP-based Style Disentanglement Loss.} Denote $I_c^c$ and $I_s^c$ are the respective contents of the content image $I_c$ and the style image $I_s$ extracted by the style removal module (Sec.~\ref{sec:SRM}). We aim to learn the disentangled style information of the style image $I_s$ {\em complementary} to its content $I_s^c$. Therefore, a straightforward way to obtain the disentangled style information is a direct subtraction:
\begin{equation}
	D_s^{px} = I_s - I_s^c.
\end{equation}
However, the simple pixel differences do not contain meaningful semantic information, thus cannot achieve plausible results~\cite{gatys2016image,kotovenko2019content1}. To address this problem, we can formulate the disentanglement in a latent semantic space:
\begin{equation}
	D_s = E(I_s) - E(I_s^c),
\end{equation}
where $E$ is a well-pre-trained projector. Specifically, since $I_s$ and $I_s^c$ have similar contents but with different styles, the projector $E$ must have the ability to distinguish them in terms of the style characteristics. In other words, as we define that images with different styles belong to different domains, the projector $E$ should be able to distinguish the domains of $I_s$ and $I_s^c$. Fortunately, inspired by the recent vision-language model CLIP~\cite{radford2021learning} that encapsulates knowledgeable semantic information of not only the photograph domain but also the artistic domain~\cite{gal2022stylegan,ramesh2022hierarchical,kwon2022clipstyler}, we can use its image encoder as our projector $E$ off the shelf. The open-domain CLIP space here serves as a good metric space to measure the ``style distance'' between content and its stylized result. This ``style distance'' thus can be interpreted as the disentangled style information. Note that here the style is implicitly defined as the {\em complement} of content, which is fundamentally different from the Gram matrix~\cite{gatys2016image} that is an explicit style definition independent of content (see comparisons in Sec.~\ref{sec:ablation}). The comparisons between CLIP space and other possible spaces can be found in {\em SM}.

After obtaining the disentangled style information $D_s$, the next question is how to properly transfer it to other contents. A possible solution is directly optimizing the L1 loss:
\begin{equation}
	\begin{aligned}
		D_{cs} & = E(I_{cs}) - E(I_c^c), \\
		\mathcal{L}_{SD}^{L1} & = \parallel D_{cs} - D_s \parallel,
	\end{aligned}
\end{equation}
where $I_{cs}$ is the stylized result, $D_{cs}$ is the disentangled style information of $I_{cs}$. However, as illustrated in Fig.~\ref{fig:loss}~(a) and further validated in later Sec.~\ref{sec:ablation}, minimizing the L1 loss cannot guarantee the stylized result $I_{cs}$ is within the style domain of the style image $I_s$. It is because L1 loss only minimizes the absolute pixel difference (\ie, Manhattan distance); thus, it may produce stylized images that satisfy the Manhattan distance but deviate from the target style domain in the transfer direction. Besides, it may also lead to a collapse problem where a stylized output meets the same Manhattan distance with different contents in the latent space.

To address these problems, we can further constrain the disentangled directions as follows:
\begin{equation}
	\mathcal{L}_{SD}^{dir} = 1 - \frac{D_{cs} \cdot D_s}{\parallel D_{cs} \parallel \parallel D_s \parallel}.
\end{equation}
This direction loss aligns the transfer direction of the content image's content to its stylization (\ie, the stylized result) with the direction of the style image's content to its stylization (\ie, the style image itself), as illustrated in Fig.~\ref{fig:loss}~(b). Collaborated with this loss, the L1 loss $\mathcal{L}_{SD}^{L1}$ thus can achieve accurate one-to-one mappings from contents in the content domain to their stylizations in the style domain, as illustrated in Fig.~\ref{fig:loss}~(c). 

\begin{figure*}
	\small
	\centering
	\setlength{\tabcolsep}{0.015cm}
	\renewcommand\arraystretch{0.4}
	\begin{tabular}{cccccccccccc}
		\includegraphics[width=0.082\linewidth]{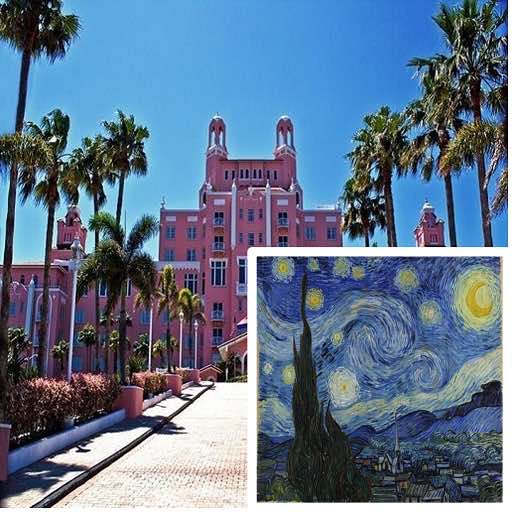}&
		\includegraphics[width=0.082\linewidth]{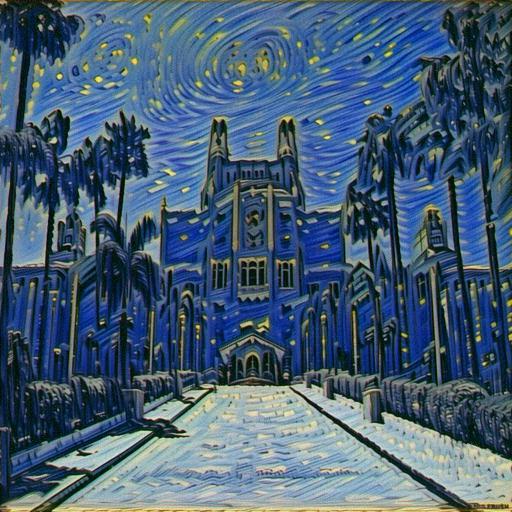}&
		\includegraphics[width=0.082\linewidth]{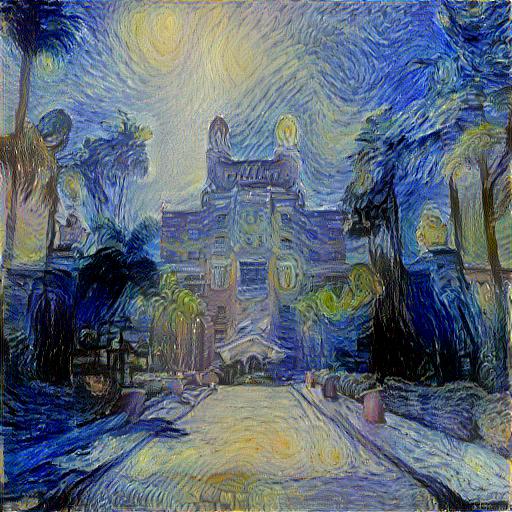}&
		\includegraphics[width=0.082\linewidth]{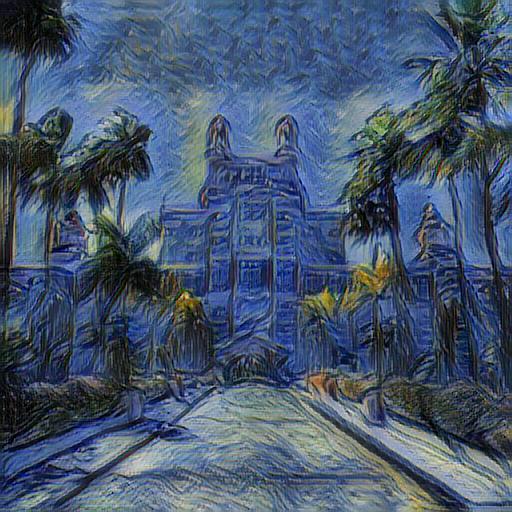}&
		\includegraphics[width=0.082\linewidth]{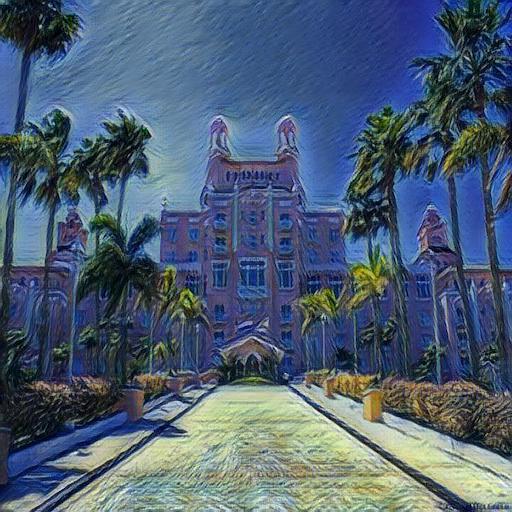}&
		\includegraphics[width=0.082\linewidth]{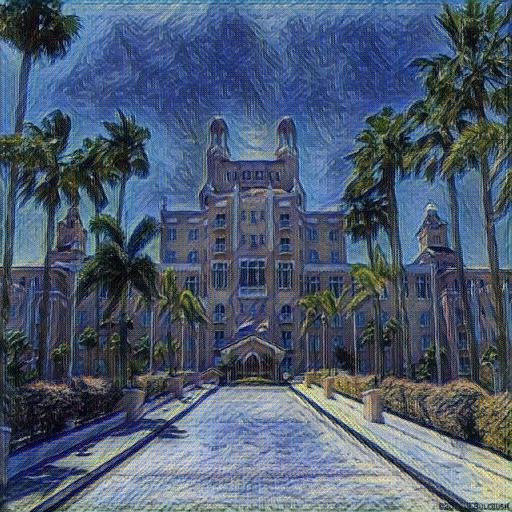}&
		\includegraphics[width=0.082\linewidth]{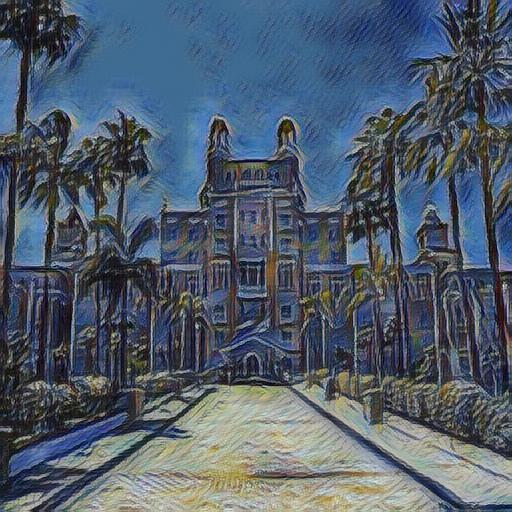}&
		\includegraphics[width=0.082\linewidth]{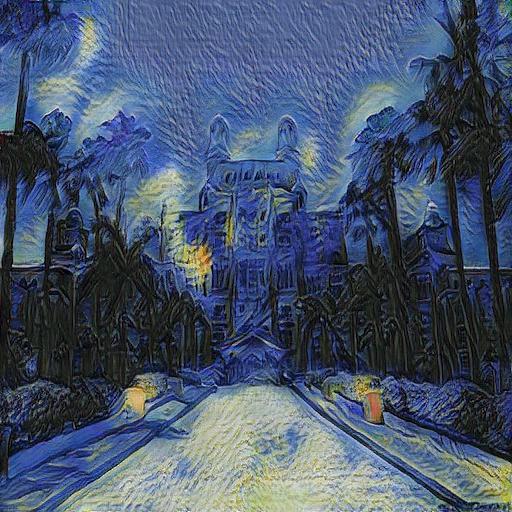}&
		\includegraphics[width=0.082\linewidth]{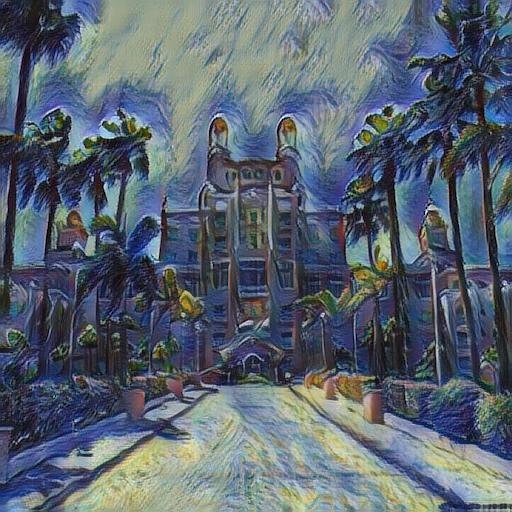}&
		\includegraphics[width=0.082\linewidth]{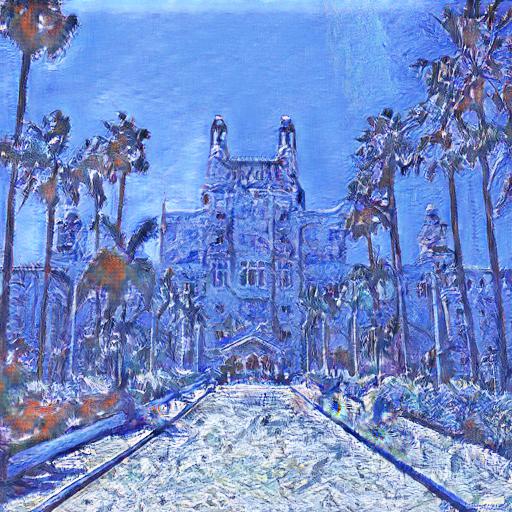}&
		\includegraphics[width=0.082\linewidth]{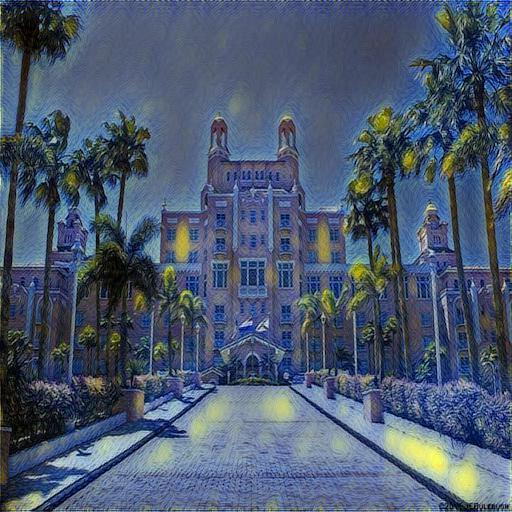}&
		\includegraphics[width=0.082\linewidth]{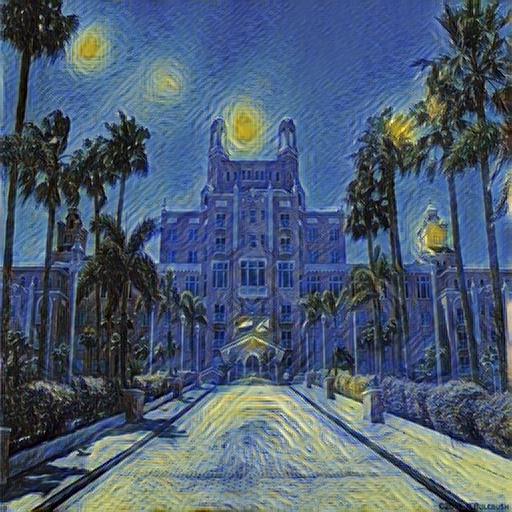}
		\\

		\includegraphics[width=0.082\linewidth]{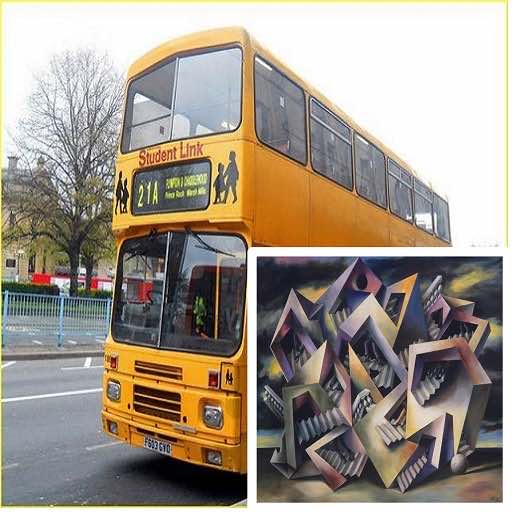}&
		\includegraphics[width=0.082\linewidth]{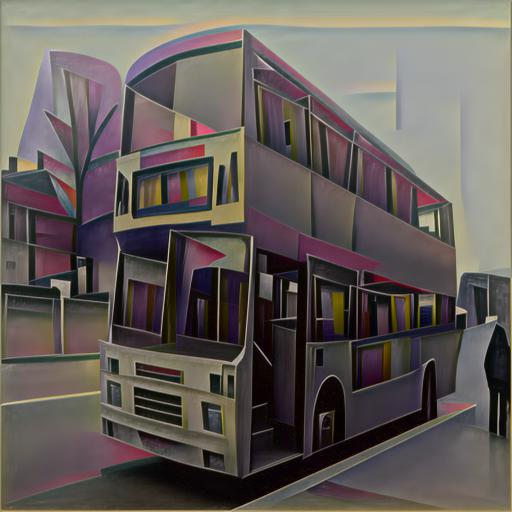}&
		\includegraphics[width=0.082\linewidth]{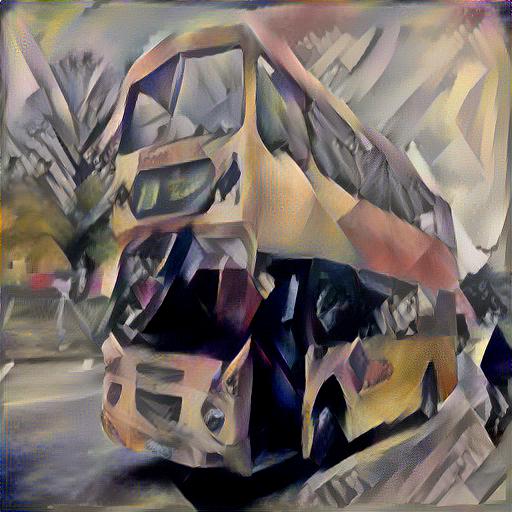}&
		\includegraphics[width=0.082\linewidth]{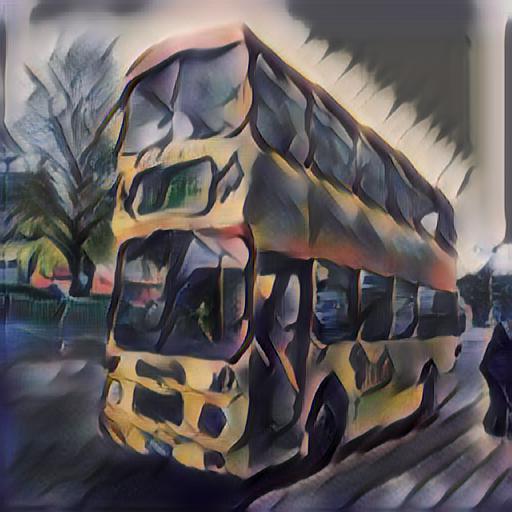}&
		\includegraphics[width=0.082\linewidth]{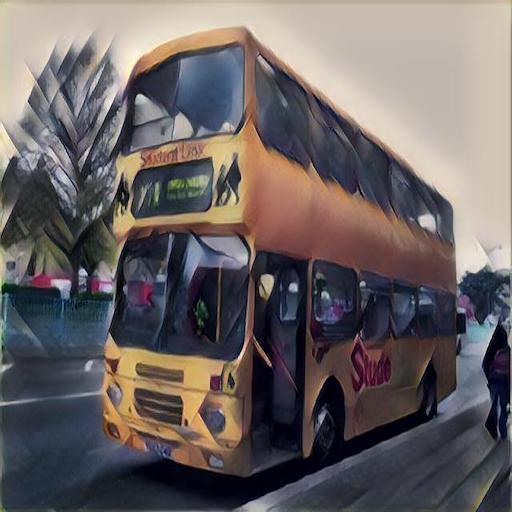}&
		\includegraphics[width=0.082\linewidth]{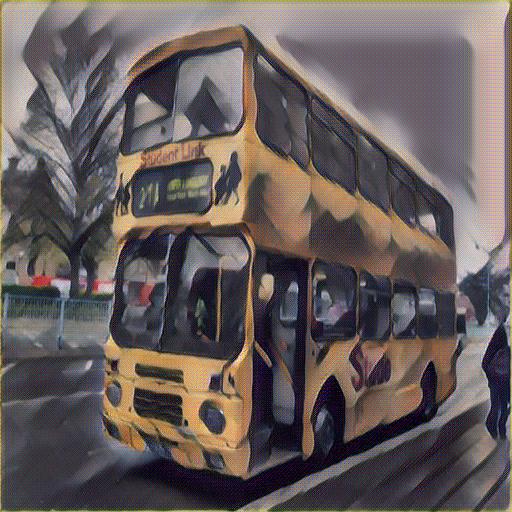}&
		\includegraphics[width=0.082\linewidth]{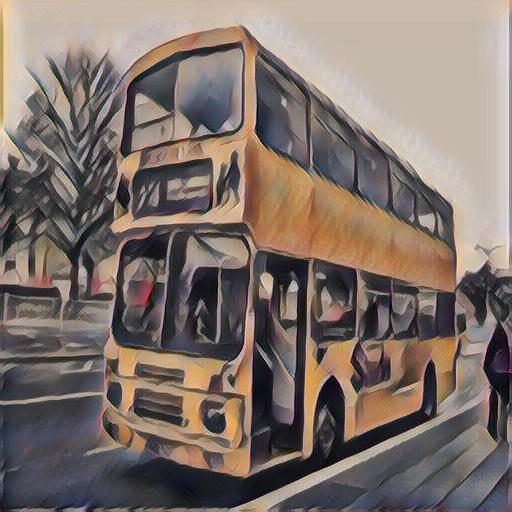}&
		\includegraphics[width=0.082\linewidth]{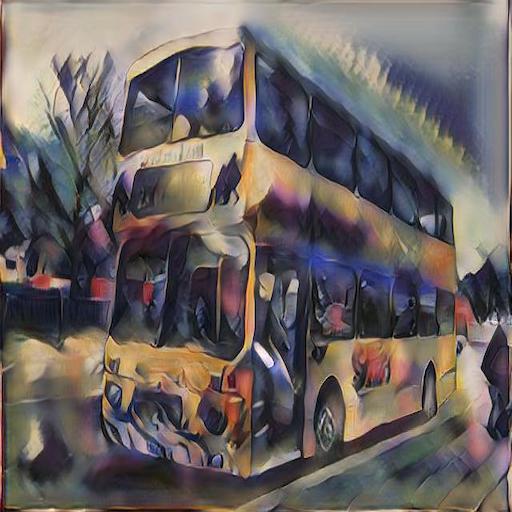}&
		\includegraphics[width=0.082\linewidth]{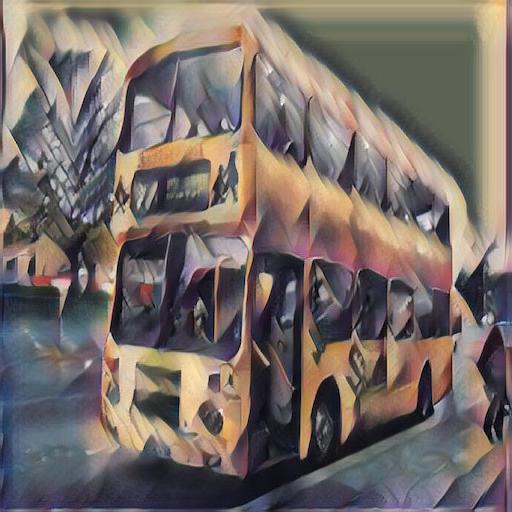}&
		\includegraphics[width=0.082\linewidth]{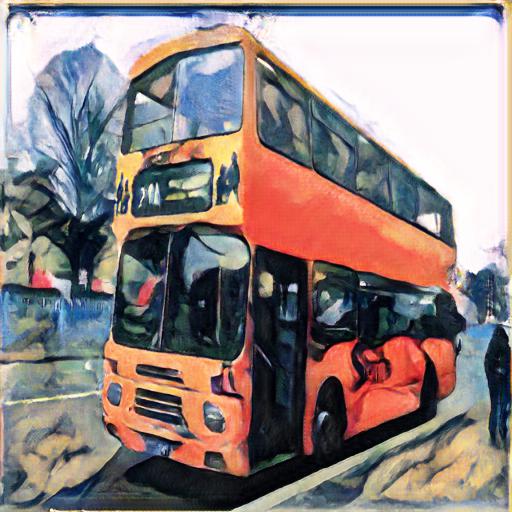}&
		\includegraphics[width=0.082\linewidth]{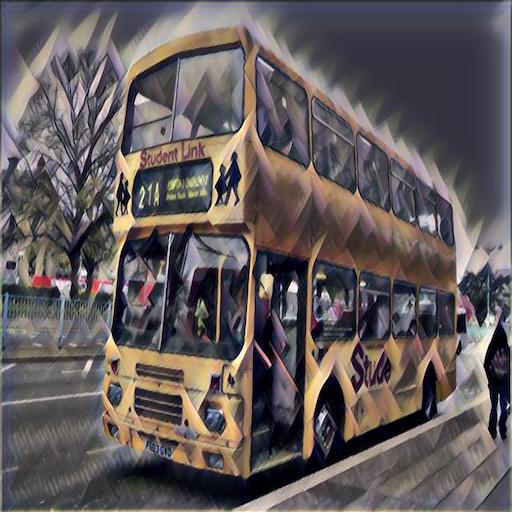}&
		\includegraphics[width=0.082\linewidth]{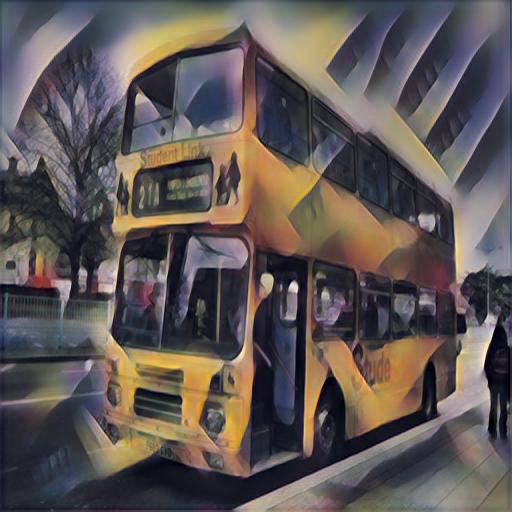}
		\\
		
		\includegraphics[width=0.082\linewidth]{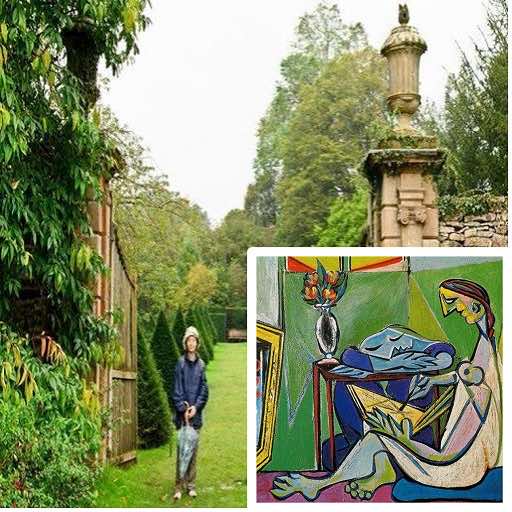}&
		\includegraphics[width=0.082\linewidth]{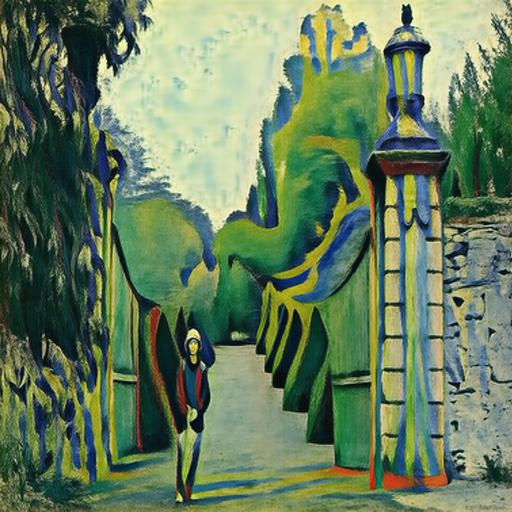}&
		\includegraphics[width=0.082\linewidth]{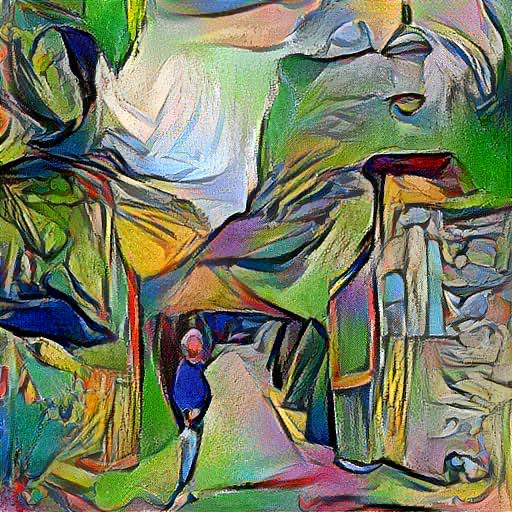}&
		\includegraphics[width=0.082\linewidth]{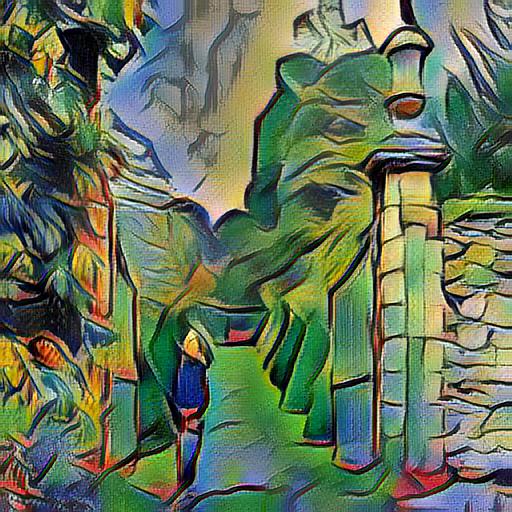}&
		\includegraphics[width=0.082\linewidth]{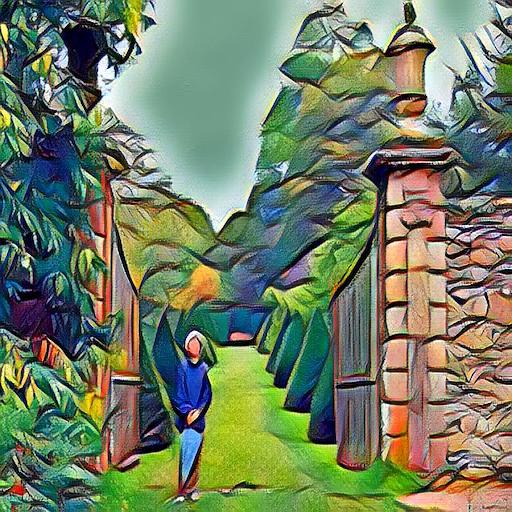}&
		\includegraphics[width=0.082\linewidth]{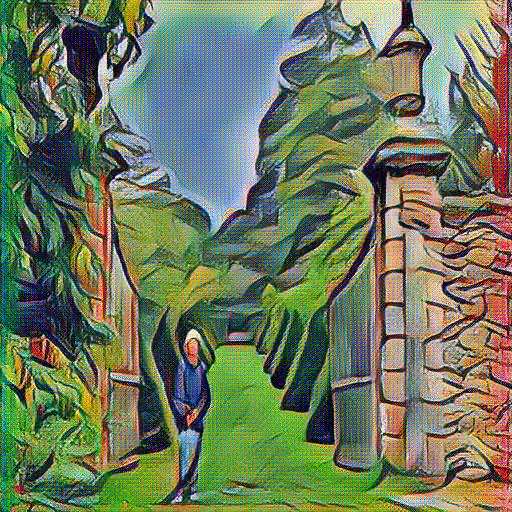}&
		\includegraphics[width=0.082\linewidth]{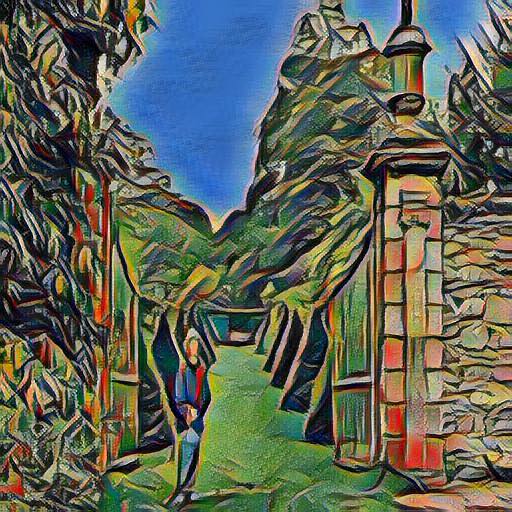}&
		\includegraphics[width=0.082\linewidth]{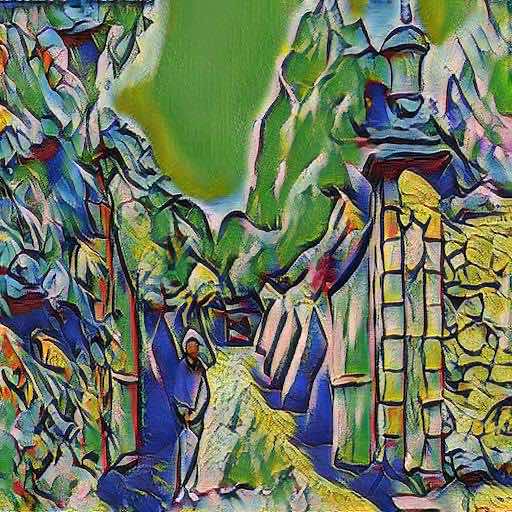}&
		\includegraphics[width=0.082\linewidth]{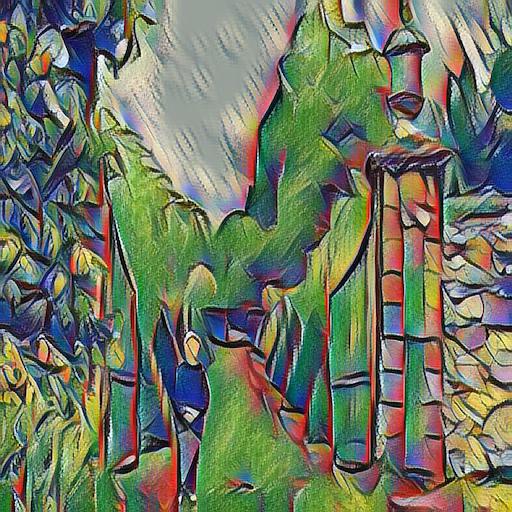}&
		\includegraphics[width=0.082\linewidth]{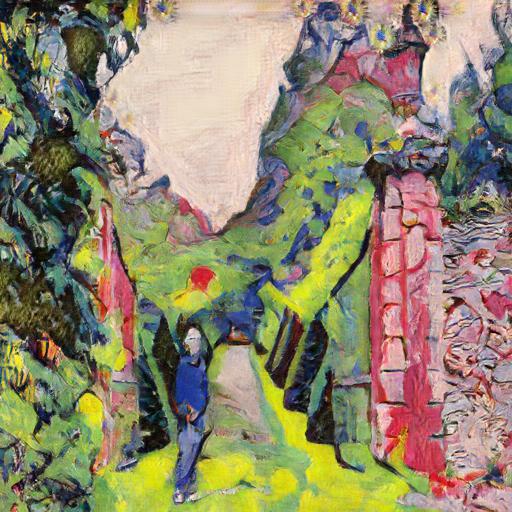}&
		\includegraphics[width=0.082\linewidth]{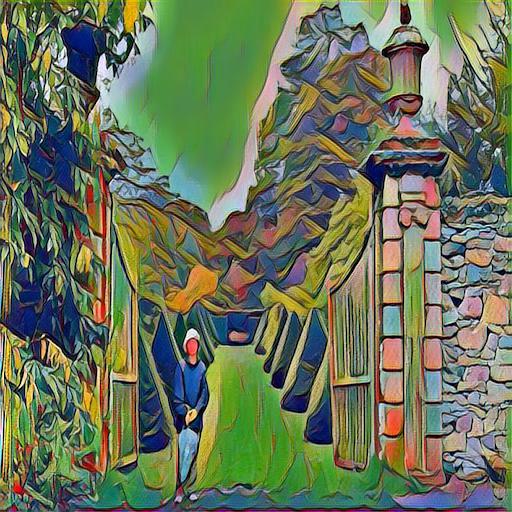}&
		\includegraphics[width=0.082\linewidth]{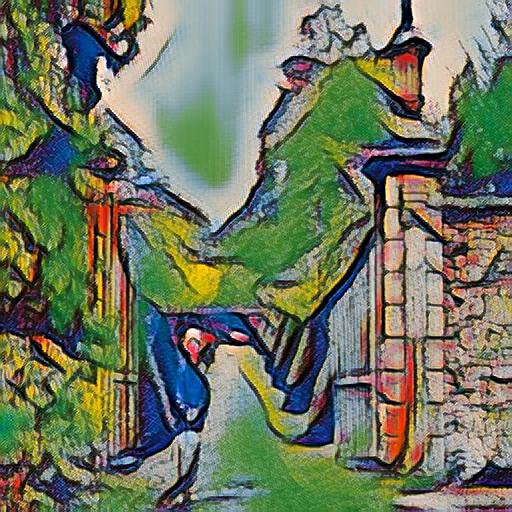}
		\\

		\includegraphics[width=0.082\linewidth]{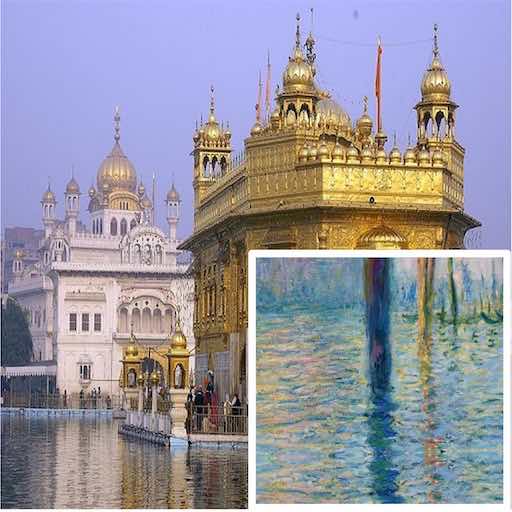}&
		\includegraphics[width=0.082\linewidth]{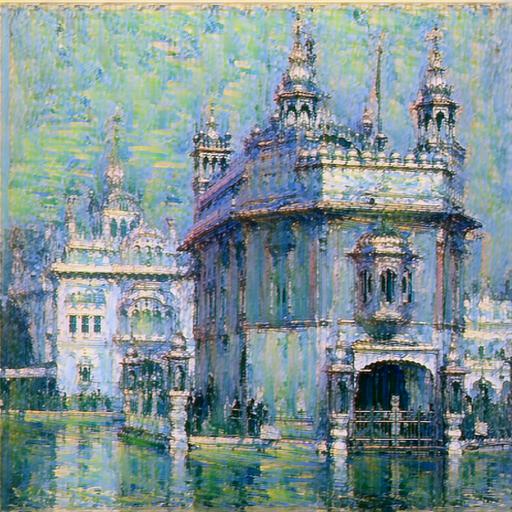}&
		\includegraphics[width=0.082\linewidth]{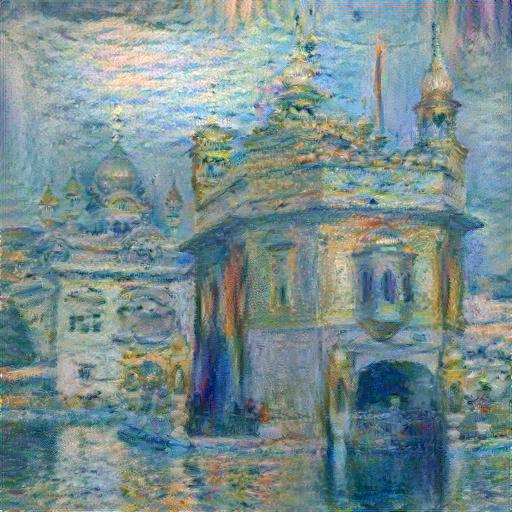}&
		\includegraphics[width=0.082\linewidth]{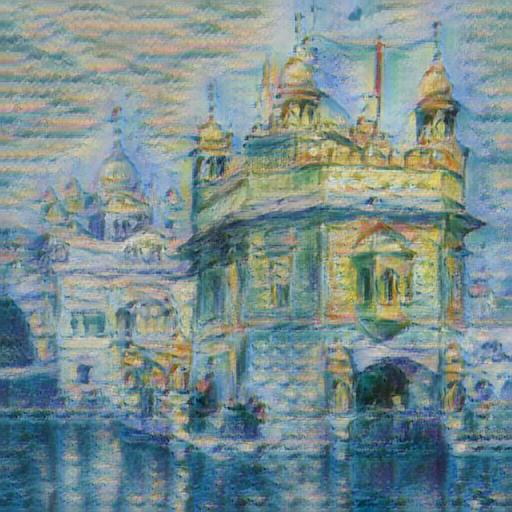}&
		\includegraphics[width=0.082\linewidth]{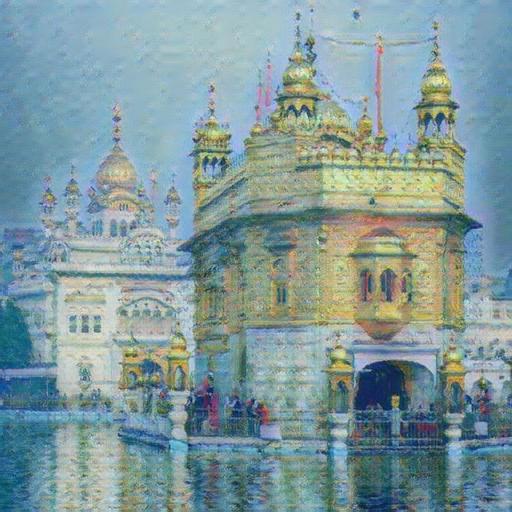}&
		\includegraphics[width=0.082\linewidth]{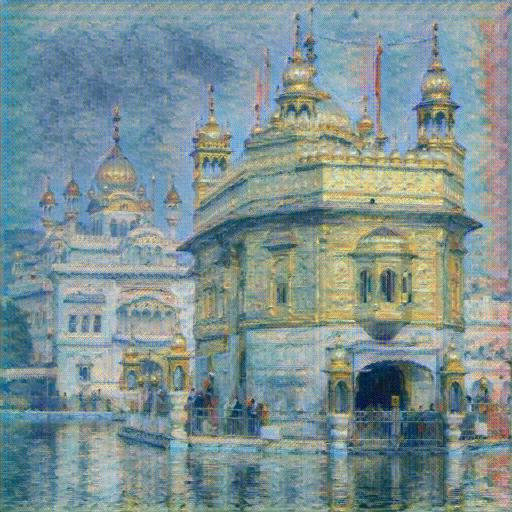}&
		\includegraphics[width=0.082\linewidth]{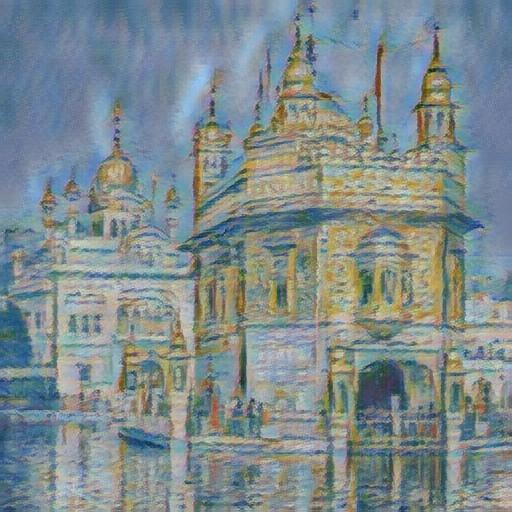}&
		\includegraphics[width=0.082\linewidth]{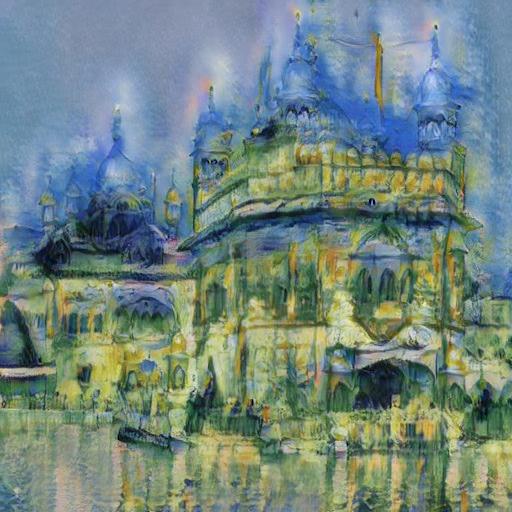}&
		\includegraphics[width=0.082\linewidth]{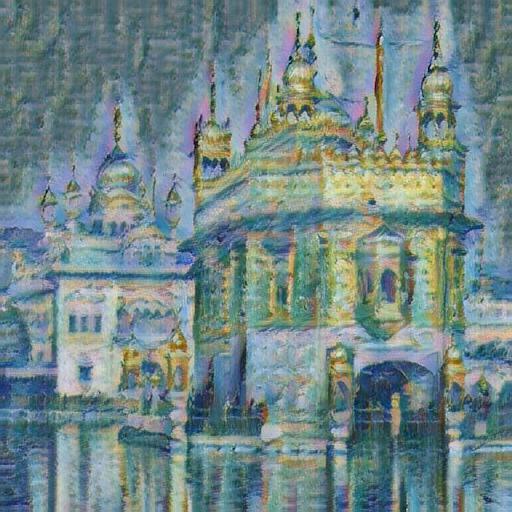}&
		\includegraphics[width=0.082\linewidth]{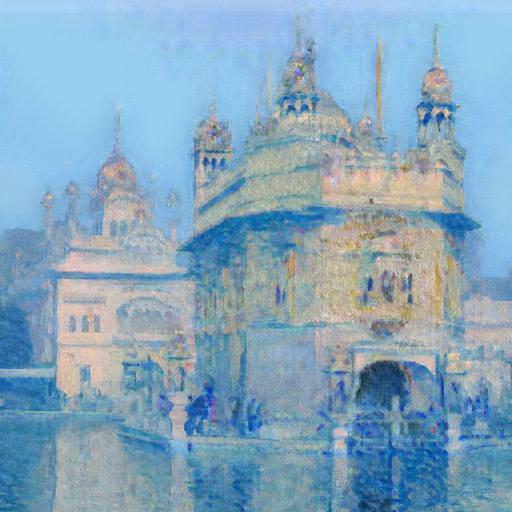}&
		\includegraphics[width=0.082\linewidth]{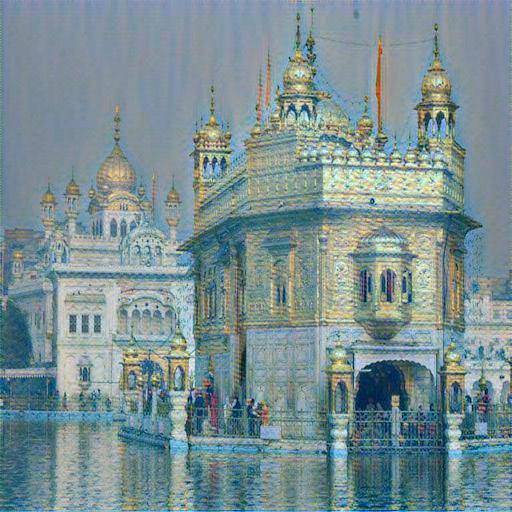}&
		\includegraphics[width=0.082\linewidth]{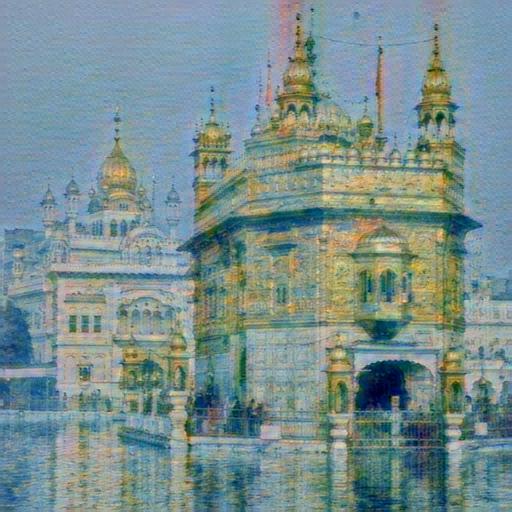}
		\\
		
		\includegraphics[width=0.082\linewidth]{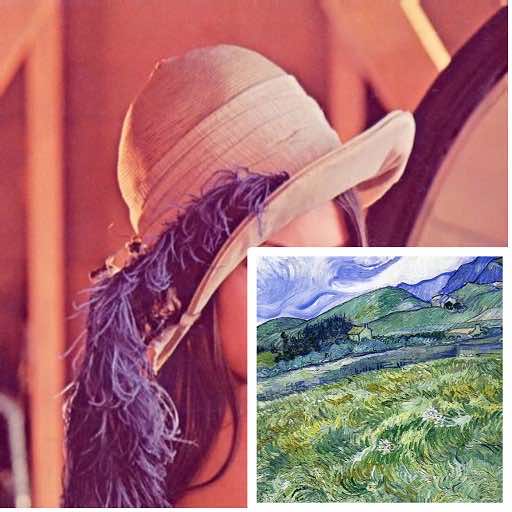}&
		\includegraphics[width=0.082\linewidth]{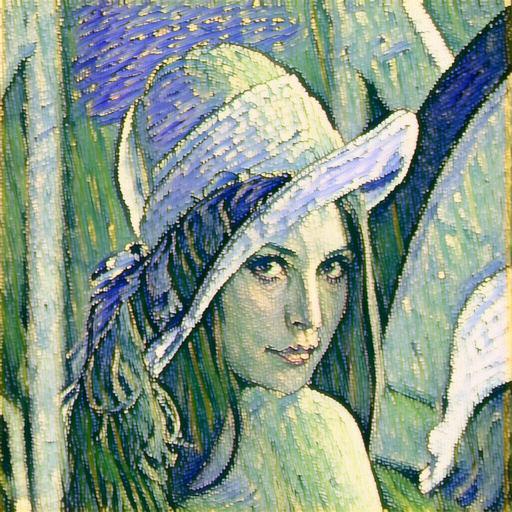}&
		\includegraphics[width=0.082\linewidth]{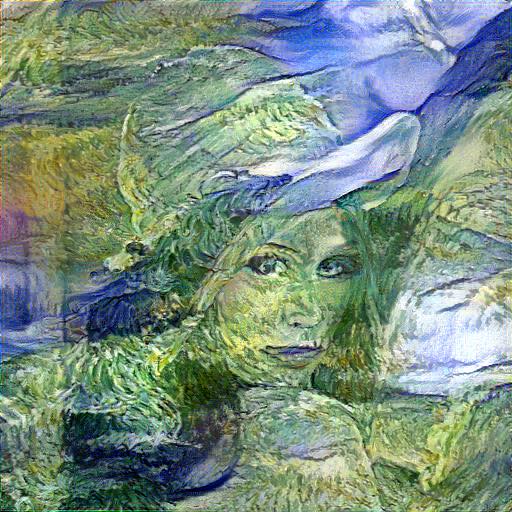}&
		\includegraphics[width=0.082\linewidth]{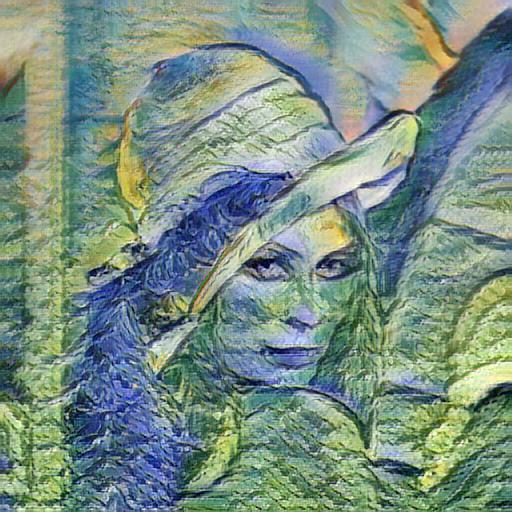}&
		\includegraphics[width=0.082\linewidth]{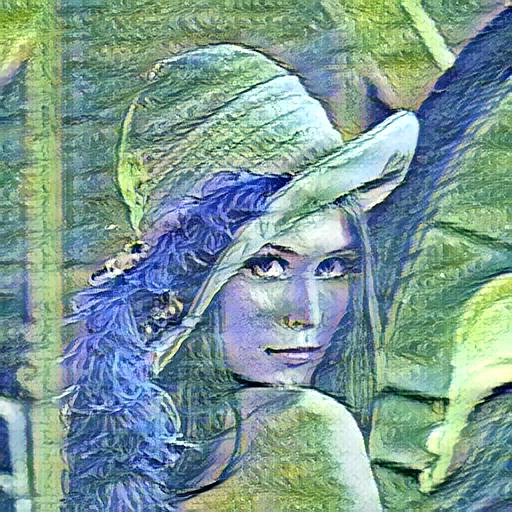}&
		\includegraphics[width=0.082\linewidth]{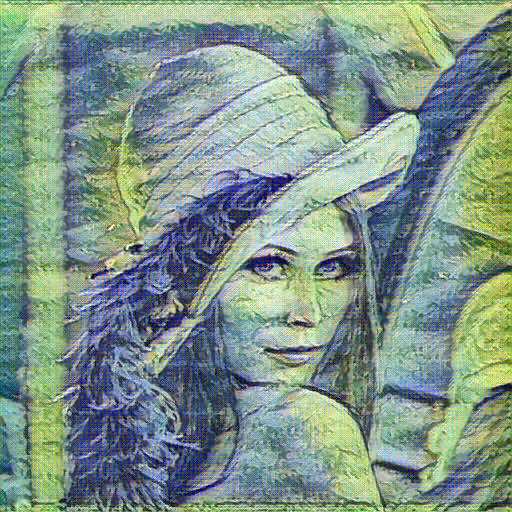}&
		\includegraphics[width=0.082\linewidth]{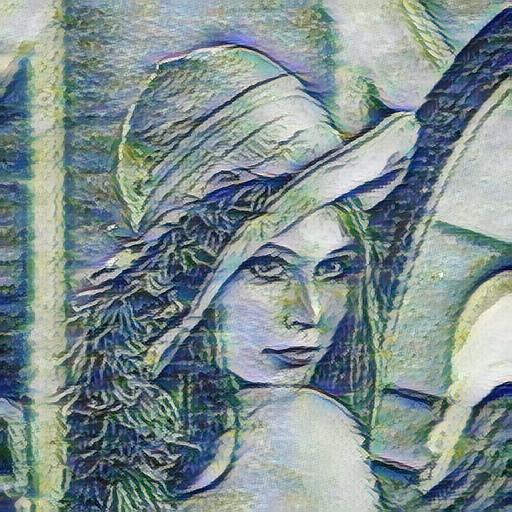}&
		\includegraphics[width=0.082\linewidth]{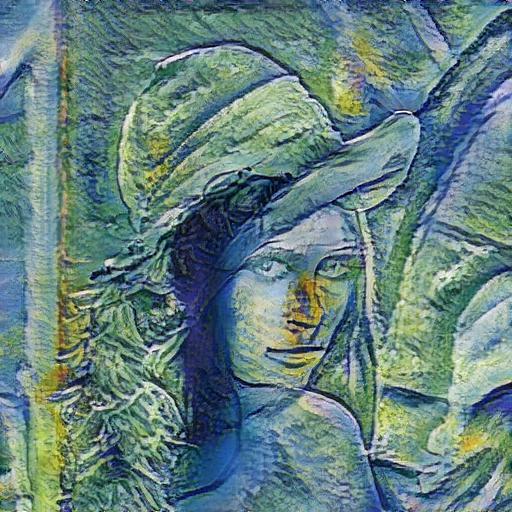}&
		\includegraphics[width=0.082\linewidth]{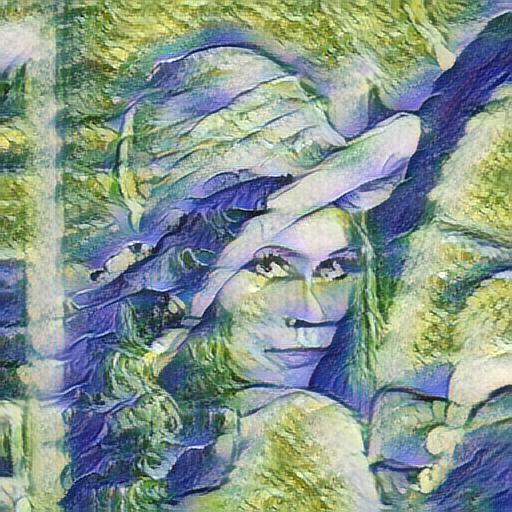}&
		\includegraphics[width=0.082\linewidth]{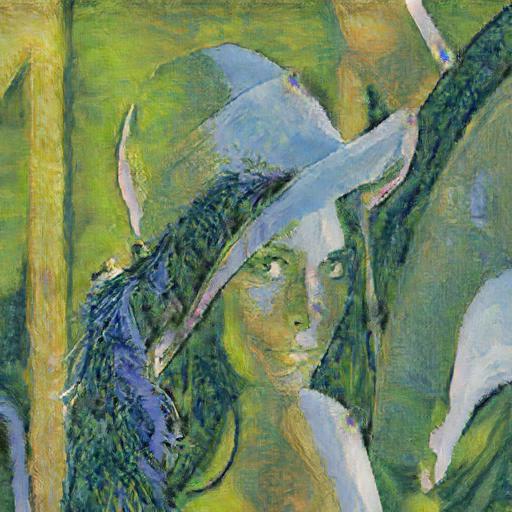}&
		\includegraphics[width=0.082\linewidth]{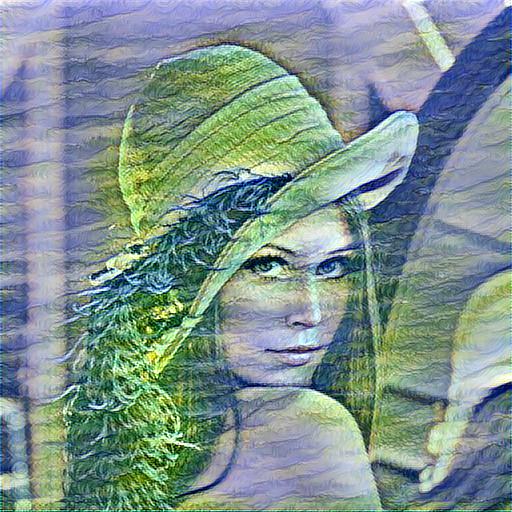}&
		\includegraphics[width=0.082\linewidth]{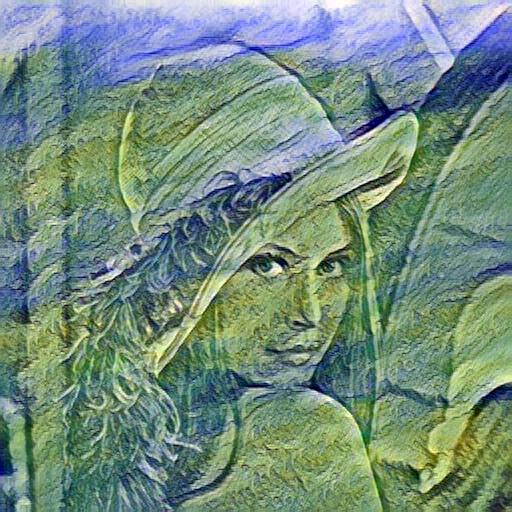}
		\\
		
		\includegraphics[width=0.082\linewidth]{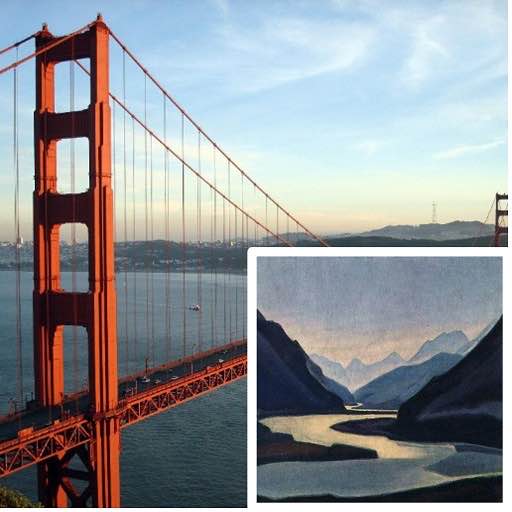}&
		\includegraphics[width=0.082\linewidth]{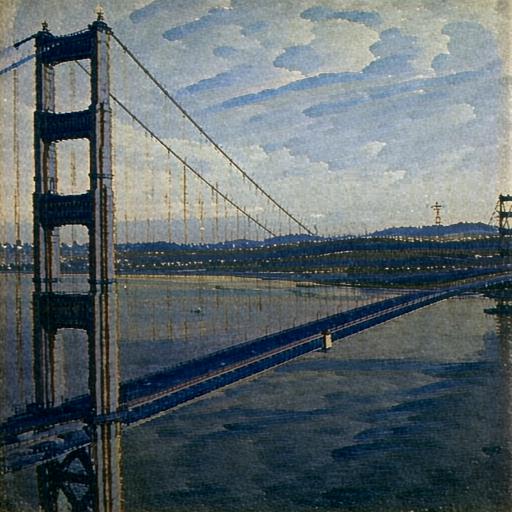}&
		\includegraphics[width=0.082\linewidth]{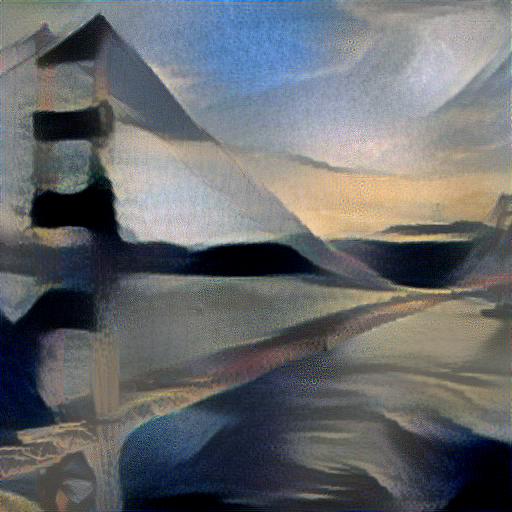}&
		\includegraphics[width=0.082\linewidth]{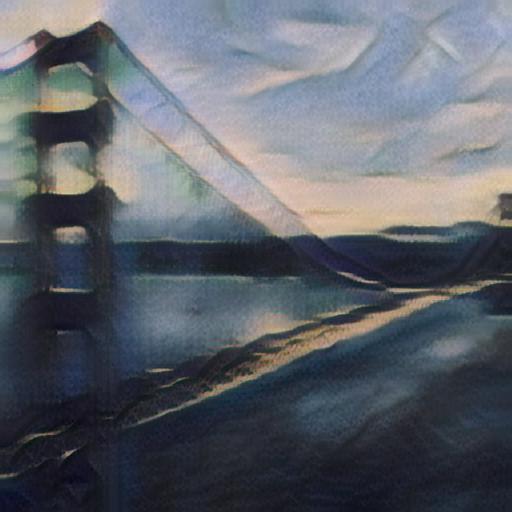}&
		\includegraphics[width=0.082\linewidth]{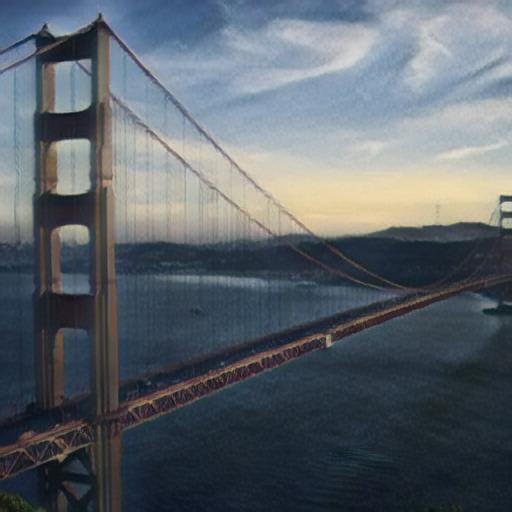}&
		\includegraphics[width=0.082\linewidth]{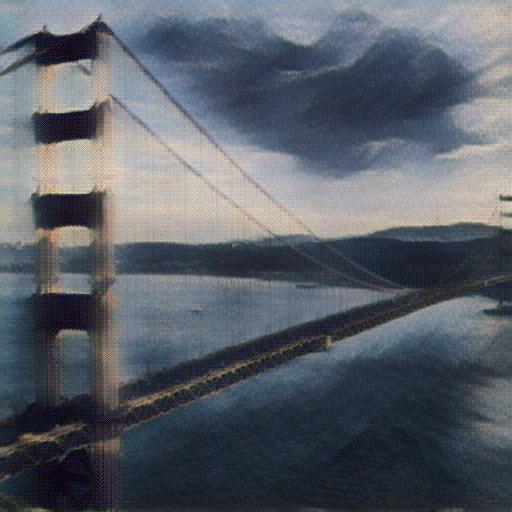}&
		\includegraphics[width=0.082\linewidth]{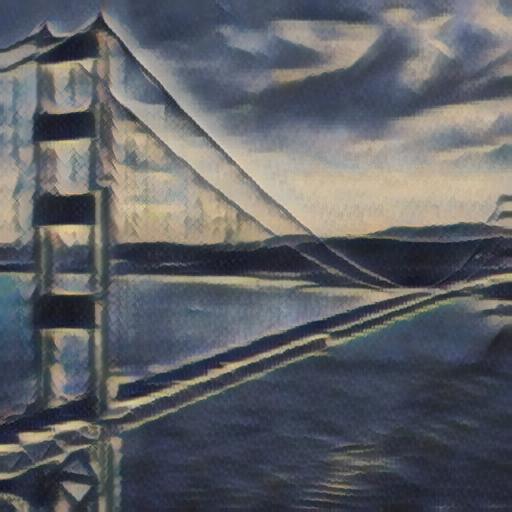}&
		\includegraphics[width=0.082\linewidth]{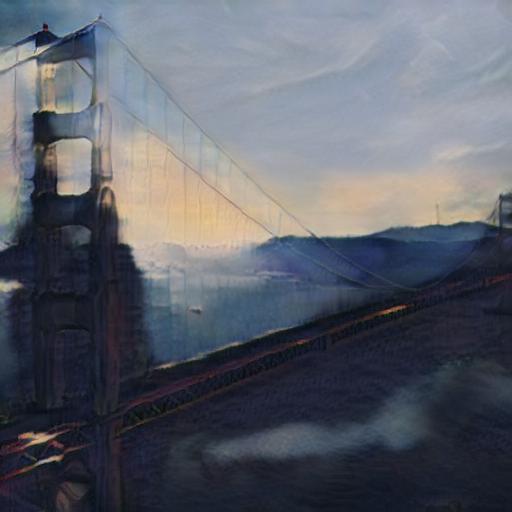}&
		\includegraphics[width=0.082\linewidth]{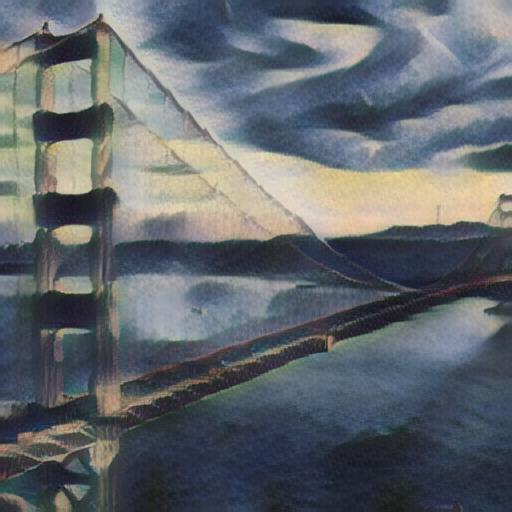}&
		\includegraphics[width=0.082\linewidth]{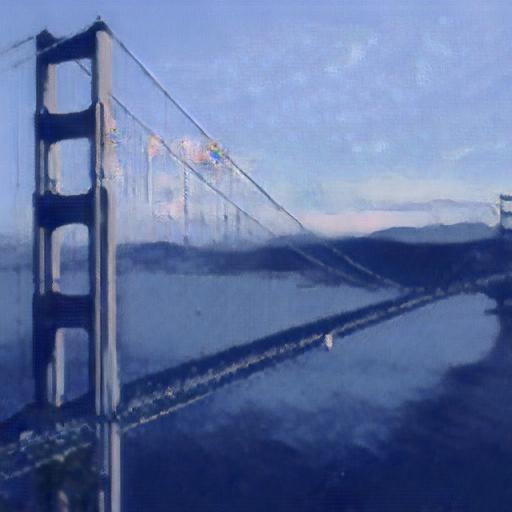}&
		\includegraphics[width=0.082\linewidth]{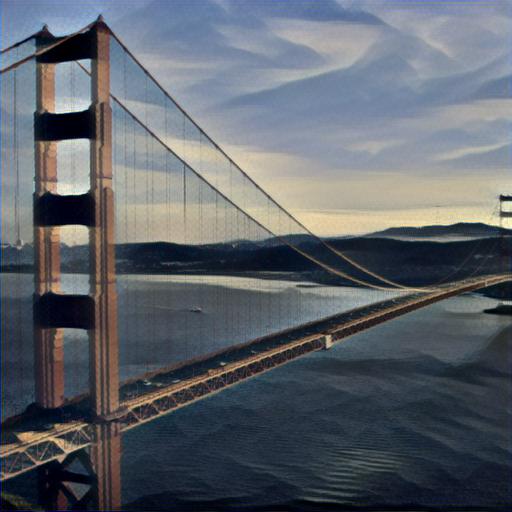}&
		\includegraphics[width=0.082\linewidth]{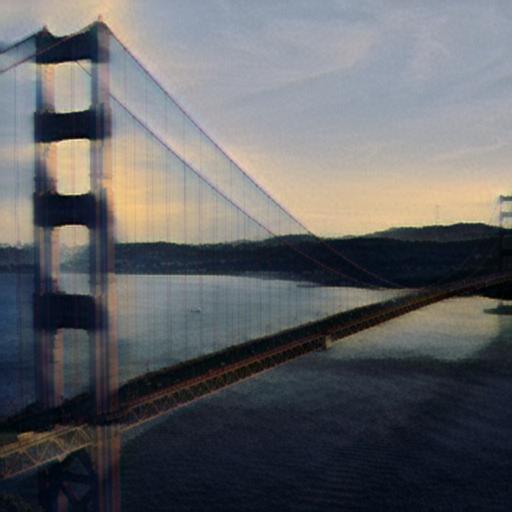}
		\\
		
		\includegraphics[width=0.082\linewidth]{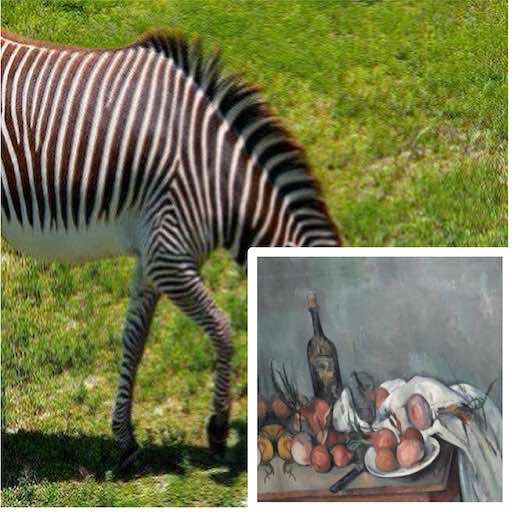}&
		\includegraphics[width=0.082\linewidth]{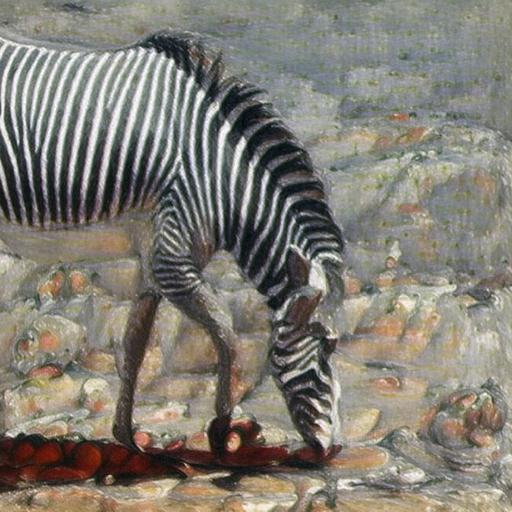}&
		\includegraphics[width=0.082\linewidth]{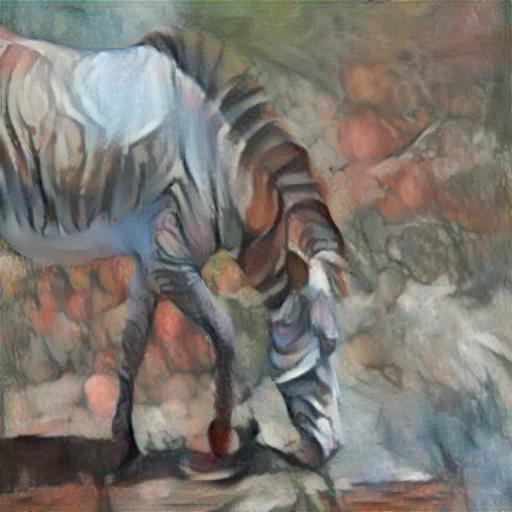}&
		\includegraphics[width=0.082\linewidth]{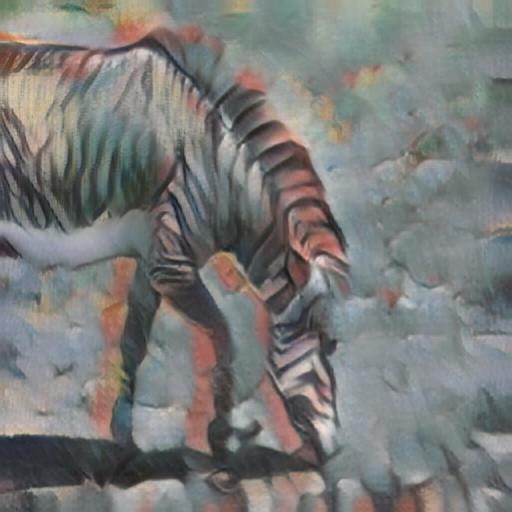}&
		\includegraphics[width=0.082\linewidth]{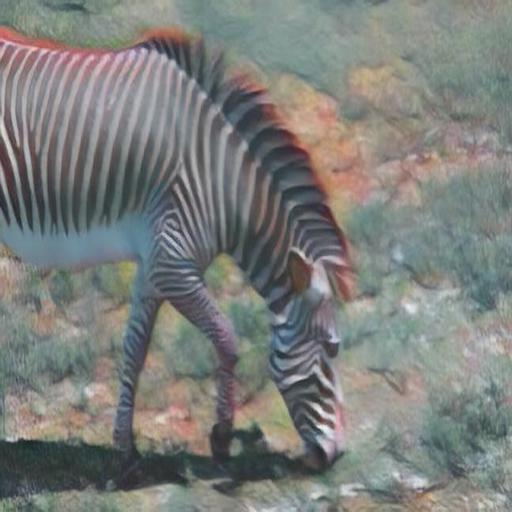}&
		\includegraphics[width=0.082\linewidth]{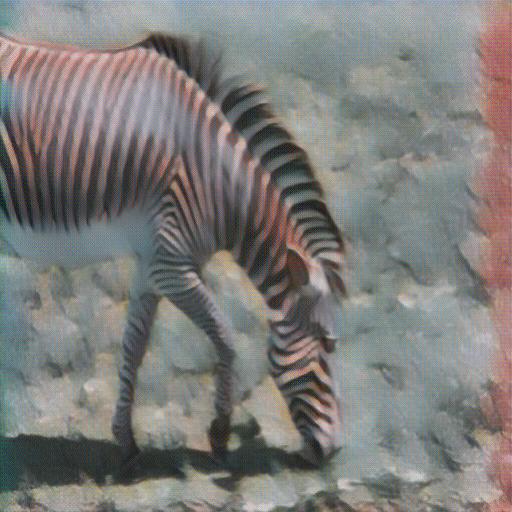}&
		\includegraphics[width=0.082\linewidth]{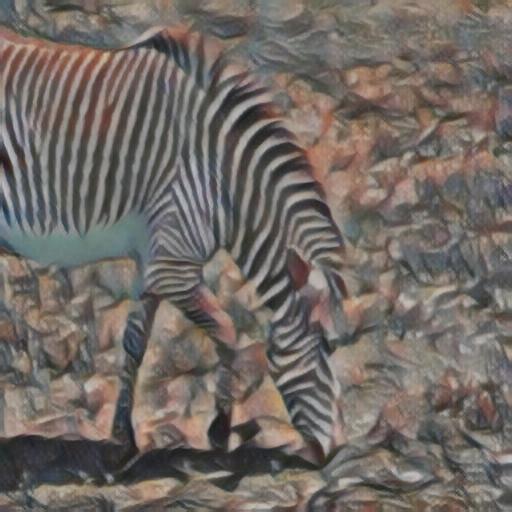}&
		\includegraphics[width=0.082\linewidth]{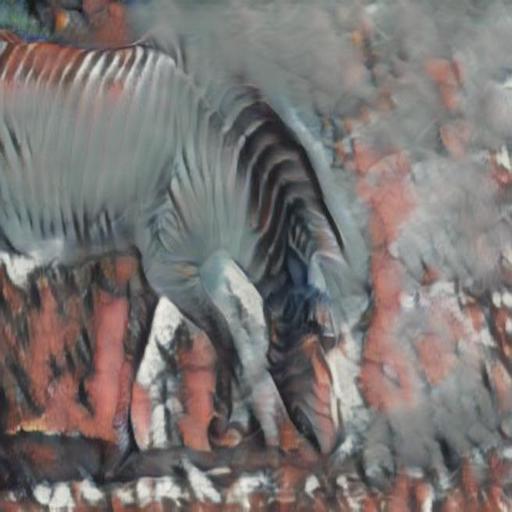}&
		\includegraphics[width=0.082\linewidth]{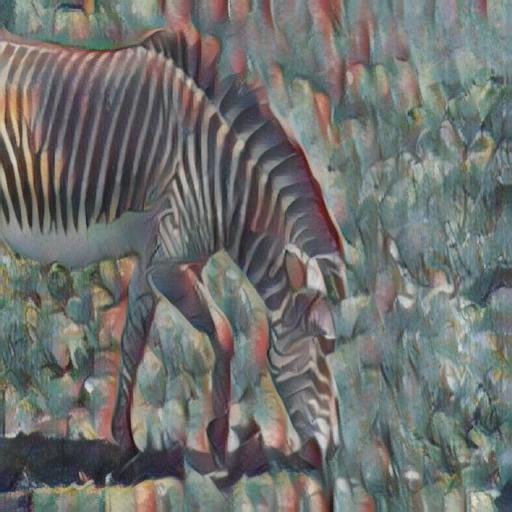}&
		\includegraphics[width=0.082\linewidth]{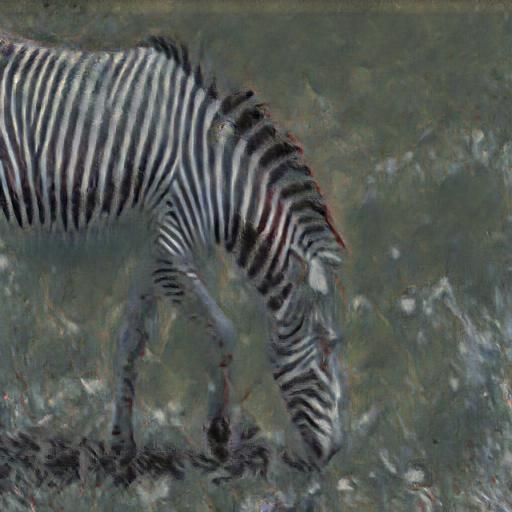}&
		\includegraphics[width=0.082\linewidth]{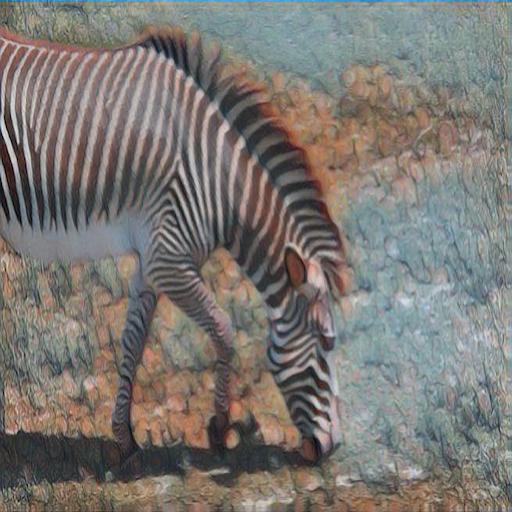}&
		\includegraphics[width=0.082\linewidth]{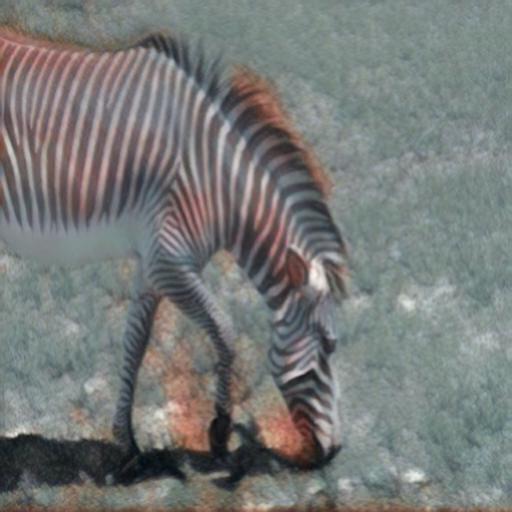}
		\\
		
		\includegraphics[width=0.082\linewidth]{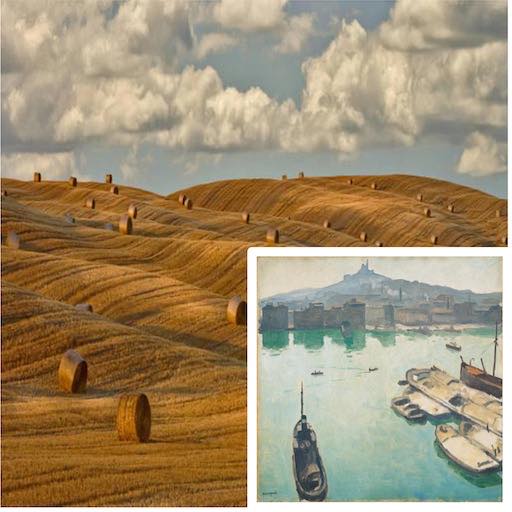}&
		\includegraphics[width=0.082\linewidth]{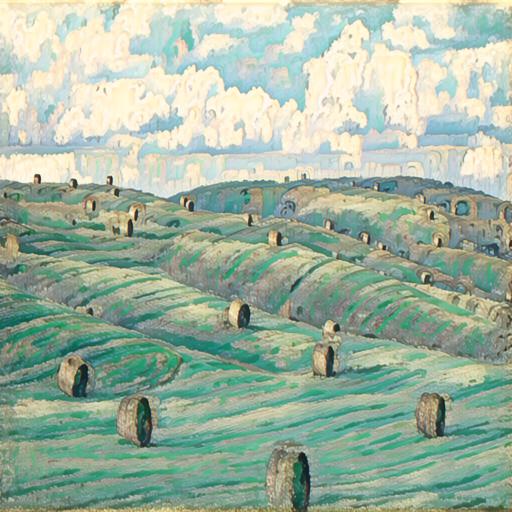}&
		\includegraphics[width=0.082\linewidth]{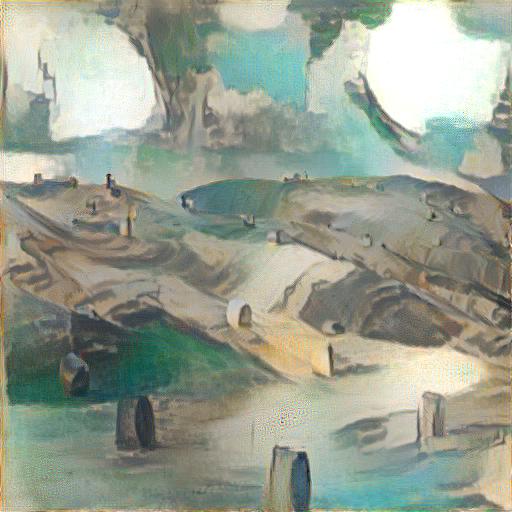}&
		\includegraphics[width=0.082\linewidth]{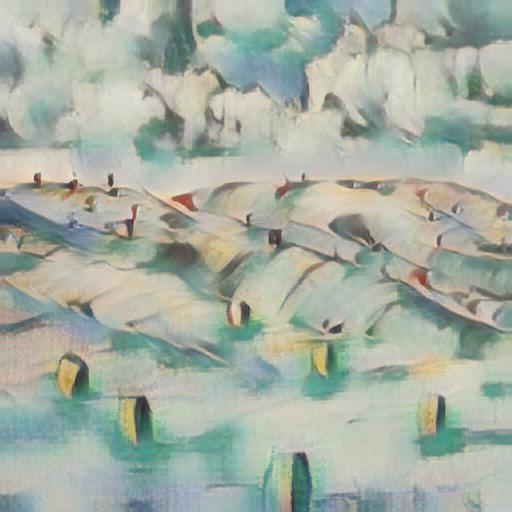}&
		\includegraphics[width=0.082\linewidth]{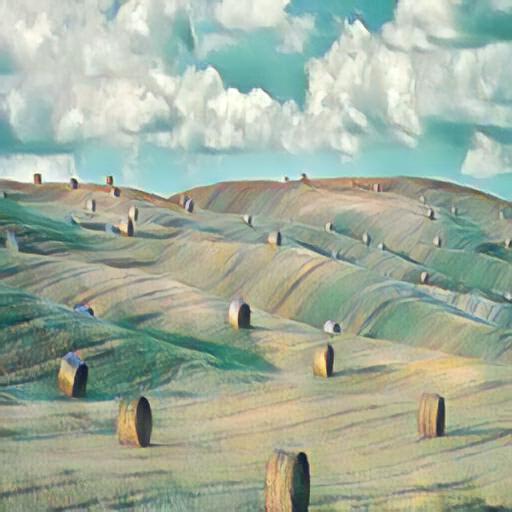}&
		\includegraphics[width=0.082\linewidth]{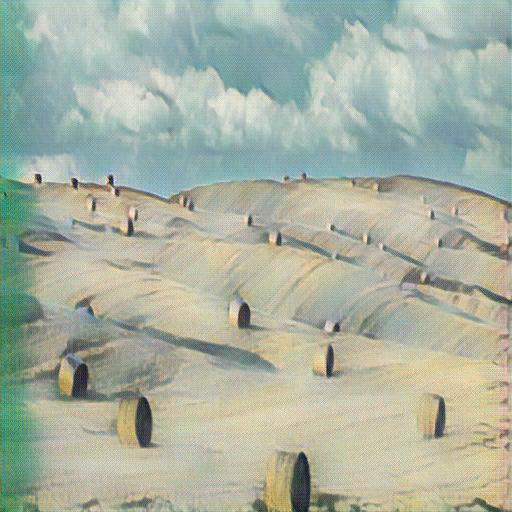}&
		\includegraphics[width=0.082\linewidth]{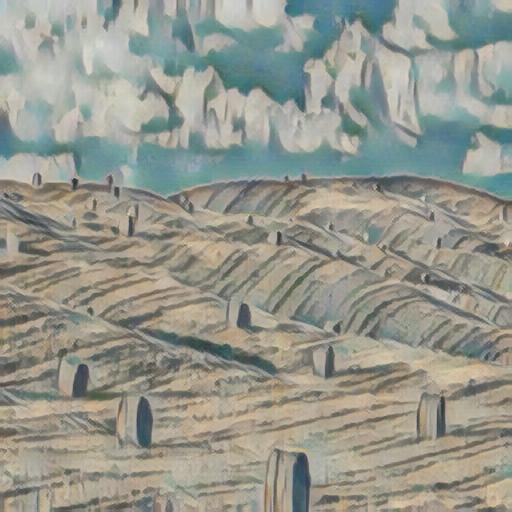}&
		\includegraphics[width=0.082\linewidth]{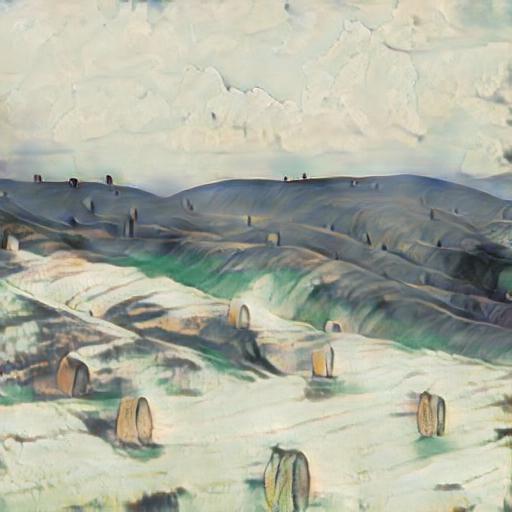}&
		\includegraphics[width=0.082\linewidth]{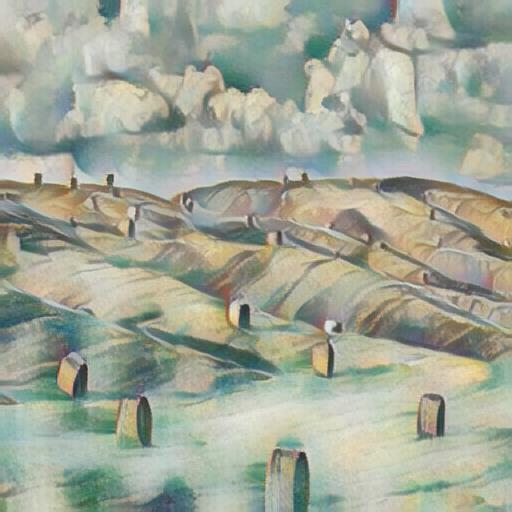}&
		\includegraphics[width=0.082\linewidth]{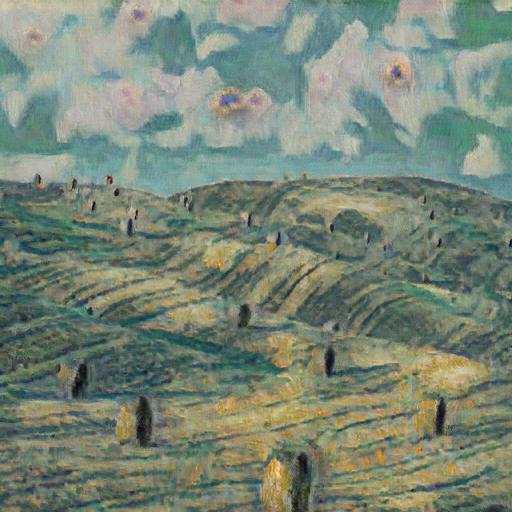}&
		\includegraphics[width=0.082\linewidth]{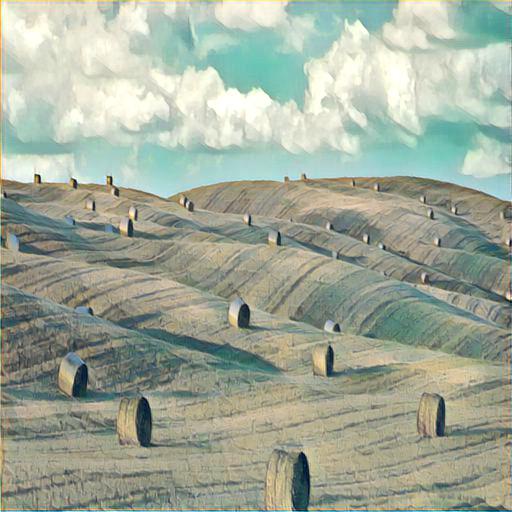}&
		\includegraphics[width=0.082\linewidth]{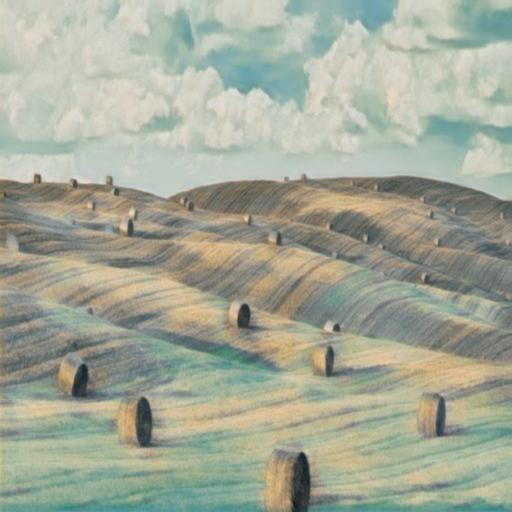}
		\\

		Inputs & \bf Ours & Gatys & EFDM & StyTr$^2$ & ArtFlow & AdaAttN & IECAST & MAST  & TPFR & Johnson & LapStyle

	\end{tabular}
\vspace{0.1em}
	\caption{ {\bf Qualitative comparisons} with state of the art. Zoom-in for better comparison. Please see more in {\em SM}.}
	\label{fig:quality}
\end{figure*}

Finally, our style disentanglement loss is defined as a compound of $\mathcal{L}_{SD}^{L1}$ and $\mathcal{L}_{SD}^{dir}$:
\begin{equation}
	\mathcal{L}_{SD} = \lambda_{L1} \mathcal{L}_{SD}^{L1} + \lambda_{dir} \mathcal{L}_{SD}^{dir},
\end{equation}
where $\lambda_{L1}$ and $\lambda_{dir}$ are hyper-parameters set to 10 and 1 in our experiments. Since our style information is induced by the difference between content and its stylized result, we can deeply understand the relationship between C-S through learning. As a result, the style can be naturally and harmoniously transferred to the content, leading to better stylized images, as will be shown in later Fig.~\ref{fig:quality}.

{\bf Style Reconstruction Prior.} To fully use the prior information provided by the style image and further elevate the stylization effects, we integrate a style reconstruction prior into the fine-tuning of the style transfer module. Intuitively, given the content $I_s^c$ of the style image $I_s$, the style transfer module should be capable of recovering it to the original style image as much as possible. Therefore, we can define a style reconstruction loss as follows:
\begin{equation}
	\mathcal{L}_{SR} = \parallel I_{ss} - I_s \parallel,
\end{equation}
where $I_{ss}$ is the stylized result given $I_s^c$ as content. We optimize it separately before optimizing the style disentanglement loss $\mathcal{L}_{SD}$. The detailed fine-tuning procedure can be found in {\em SM}. The style reconstruction prior helps our model recover the style information more sufficiently. It also provides a good initialization for the optimization of $\mathcal{L}_{SD}$, which helps the latter give full play to its ability, thus producing higher-quality results (see later Sec.~\ref{sec:ablation}).

\section{Experimental Results}
\subsection{Implementation Details}
\label{sec:impd}
We use ADM diffusion model~\cite{dhariwal2021diffusion} pre-trained on ImageNet~\cite{russakovsky2015imagenet} and adopt a fast sampling strategy~\cite{kim2022diffusionclip}. Specifically, instead of sequentially conducting the diffusion processes until the last timestep $T$ (\eg, 1000), we accelerate them by performing up to $T_{\{\cdot\}}<T$ (which is called return step), \ie, $T_{remov} = 601$ for style removal and $T_{trans} = 301$ for style transfer. Moreover, as suggested by~\cite{kim2022diffusionclip}, we further accelerate the forward and reverse processes with fewer discretization steps, \ie, $(S_{for},S_{rev}) = (40, 40)$ ($S_{for}$ for forward process and $S_{rev}$ for reverse process) for style removal, and $(S_{for}, S_{rev}) = (40, 6)$ for style transfer. When fine-tuning or inference, we can adjust $T_{remov}$ or $T_{trans}$ to flexibly control the degree of style removal and C-S disentanglement, as will be shown in Sec.~\ref{sec:ablation}. To fine-tune the model for a target style image, we randomly sample 50 images from ImageNet as the content images. We use Adam optimizer~\cite{kingma2014adam} with an initial learning rate of \mbox{4e-6} and increase it linearly by 1.2 per epoch. All models are fine-tuned with 5 epochs. See more details in {\em SM}.

\renewcommand\arraystretch{0.1}
\begin{table*}[t]
	\small
	\centering
	\setlength{\tabcolsep}{0.19cm}
	\begin{tabular}{c|c|ccccccccc|cc}
		\toprule
		\multicolumn{2}{c|}{}& \bf Ours  & Gatys & EFDM & StyTr$^2$ & ArtFlow & AdaAttN & IECAST & MAST  & TPFR & Johnson & LapStyle
		
		\\
		\midrule
		\multicolumn{2}{c|}{SSIM $\uparrow$} & \bf 0.672 & 0.311 & 0.316 & 0.537 & 0.501  & 0.542 & 0.365 & 0.392 & 0.536 & 0.634 & 0.657
		\\
		
		\multicolumn{2}{c|}{CLIP Score $\uparrow$} & \bf 0.741  & 0.677 & 0.607 & 0.531 & 0.546 & 0.577 & 0.646 & 0.590 & 0.644 & 0.537 & 0.595
		\\
		
		\multicolumn{2}{c|}{Style Loss $\downarrow$} & 0.837 & \bf 0.111 & 0.178  & 0.216  & 0.258 & 0.310 & 0.284 & 0.229  & 0.989 & 0.364 & 0.274
		\\

		\midrule
		User &Style  & -  & 43.1\% & 41.2\% & 39.3\% & 36.4\% & 37.2\% & 33.8\% & 39.1\% & 14.5\% & 42.8\%& 47.3\%
		\\
		
		Study & Overall  & - & 26.0\% & 38.1\% & 44.0\% & 34.2\% & 43.9\% & 32.7\% & 32.2\% & 22.6\% &  43.4\%  & 46.2\% 
		\\
		
		\midrule
		\multicolumn{2}{c|}{Training Time/h}& $\sim$0.4 & -  & $\sim$3 & $\sim$4  & $\sim$3 &  $\sim$3  & $\sim$3 & $\sim$3 & $\sim$10 & $\sim$1 & $\sim$3
		\\
		
		\multicolumn{2}{c|}{Testing Time/s}& 5.612 & 10.165 & 0.028 & 0.168 & 0.204 & 0.076 & 0.034 & 0.066 & 0.302 &  0.015 & 0.008
		\\
		
		\bottomrule
	\end{tabular}
	\vspace{0.5em}
	\caption{{\bf Quantitative comparisons} with state of the art. The training/testing time is measured with an Nvidia Tesla A100 GPU, and the testing time is averaged on images of size 512$\times$512 pixels. $\uparrow$: Higher is better. $\downarrow$: Lower is better.}
	\label{tab:quantity}
\end{table*}

\subsection{Comparisons with Prior Arts}
We compare our StyleDiffusion against ten state-of-the-art (SOTA) methods~\cite{gatys2016image,zhang2022exact,deng2022stytr2,an2021artflow,liu2021adaattn,chen2021artistic,deng2020arbitrary,svoboda2020two,johnson2016perceptual,lin2021drafting}. For fair comparisons, all these methods are fine-tuned or trained on the target styles similar to our approach.

{\bf Qualitative Comparisons.} As can be observed in Fig.~\ref{fig:quality}, due to the entangling of C-S representations, Gatys~\cite{gatys2016image} and EFDM~\cite{zhang2022exact} often produce unsatisfying results with distorted contents (\eg, rows 1-3) and messy textures (\eg, rows 4-8). StyTr$^2$~\cite{deng2022stytr2} and ArtFlow~\cite{an2021artflow} improve the results by adopting more advanced networks~\cite{vaswani2017attention,kingma2018glow}, but they may still produce inferior results with halo boundaries (\eg, rows 2-3) or dirty artifacts (\eg, rows 4-6). AdaAttN~\cite{liu2021adaattn} performs per-point attentive normalization to preserve the content structures better, but the stylization effects may be degraded in some cases (\eg, rows 1, 2, 4, and 5). IECAST~\cite{chen2021artistic} utilizes contrastive learning and external learning for style transfer, so fine-tuning it on a single style image would result in degraded results. MAST~\cite{deng2020arbitrary} uses multi-adaptation networks to disentangle C-S. However, since it still relies on the C-S representations of~\cite{gatys2016image}, the results usually exhibit messy textures and conspicuous artifacts. TPFR~\cite{svoboda2020two} is a GAN-based framework that learns to disentangle C-S in latent space. As the results show, it cannot recover correct style details and often generates deviated stylizations, which signifies that it may not learn truly disentangled C-S representations~\cite{locatello2019challenging}. Like our method, Johnson~\cite{johnson2016perceptual} and LapStyle~\cite{lin2021drafting} also train separate models for each style. However, due to the trade-off between C-S losses of~\cite{gatys2016image}, they may produce less-stylized results or introduce unnatural patterns (\eg, rows 1-6).

By contrast, our StyleDiffusion completely disentangles C-S based on diffusion models. Therefore, it can generate high-quality results with sufficient style details (\eg, rows 1-4) and well-preserved contents (\eg, rows 5-8). Compared with the previous methods that tend to produce mixed results of content and style, our approach can better consider the relationship between them. Thus, the stylizations are more natural and harmonious, especially for challenging styles such as cubism (\eg, row 2) and oil painting (\eg, rows 1, 3, 4, and 5).

{\bf Quantitative Comparisons.} We also resort to quantitative metrics to better evaluate our method, as shown in Tab.~\ref{tab:quantity}. We collect 32 content and 12 style images to synthesize 384 stylized results and compute the average Structural Similarity Index (SSIM)~\cite{an2021artflow} to assess the content similarity. To evaluate the style similarity, we calculate the CLIP image similarity score~\cite{radford2021learning} and Style Loss~\cite{gatys2016image,huang2017arbitrary} between the style images and the corresponding stylized results. As shown in Tab.~\ref{tab:quantity}, our method obtains the highest SSIM and CLIP Score while the Style Loss is relatively higher than other methods. It is because these methods are directly trained to optimize Style Loss. Nevertheless, the Style Loss achieved by our method is still comparable and lower than the GAN-based TPFR~\cite{svoboda2020two}. Furthermore, it is noteworthy that our method can also incorporate Style Loss to enhance the performance in this regard (see later Sec.~\ref{sec:ablation}).

{\bf User Study.} As style transfer is highly subjective and CLIP Score and Style Loss are biased to the training objective, we additionally resort to user study to evaluate the style similarity and overall stylization quality. We randomly select 50 C-S pairs for each user. Given each C-S pair, we show the stylized results generated by our method and a randomly selected SOTA method side by side in random order. The users are asked to choose (1) which result transfers the style patterns better and (2) which result has overall better stylization effects. We obtain 1000 votes for each question from 20 users and show the percentage of votes that existing methods are preferred to ours in Tab.~\ref{tab:quantity}. The lower numbers indicate our method is more preferred than the competitors. As the results show, our method is superior to others in both style consistency and overall quality.

{\bf Efficiency.} As shown in the bottom two rows of Tab.~\ref{tab:quantity}, our approach requires less training time than others as it is fine-tuned on only a few ($\sim$50) content images. When testing, our approach is faster than the optimization-based method Gatys~\cite{gatys2016image}, albeit slower than the remaining feed-forward methods due to the utilization of diffusion models. We discuss it in later Sec.~\ref{sec:conclusion}, and more timing and resource details can be found in {\em SM}.

\subsection{Ablation Study}
\label{sec:ablation}

\begin{figure}
	\centering
	\setlength{\tabcolsep}{0.02cm}
	\renewcommand\arraystretch{0.4}
	
	\begin{tabular}{ccccc}
		
		\includegraphics[width=0.195\linewidth]{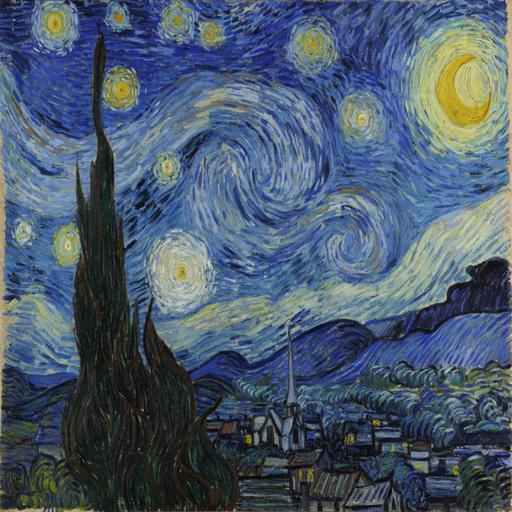}&
		\includegraphics[width=0.195\linewidth]{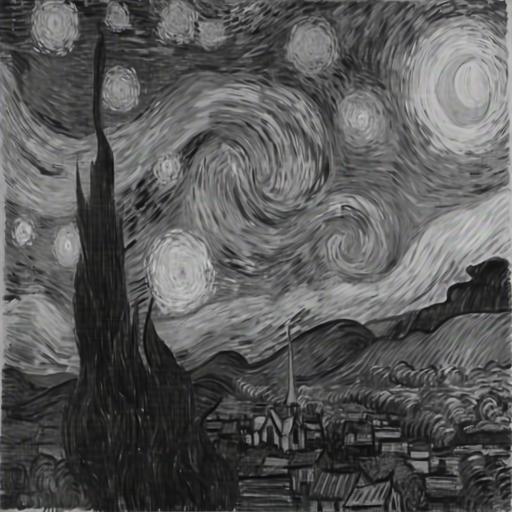}&
		\includegraphics[width=0.195\linewidth]{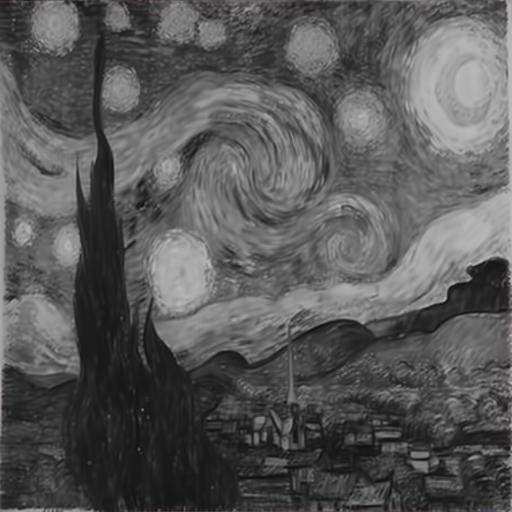}&
		\includegraphics[width=0.195\linewidth]{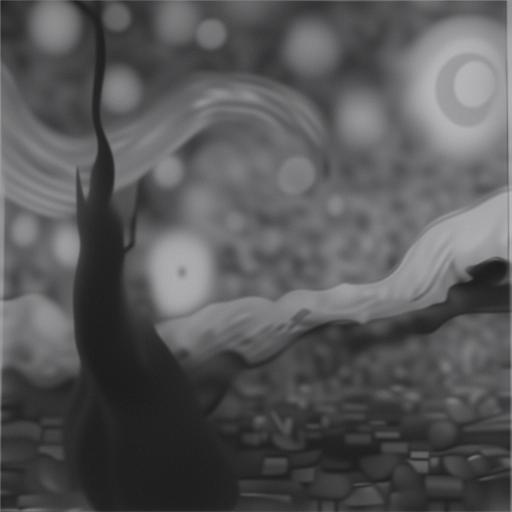}&
		\includegraphics[width=0.195\linewidth]{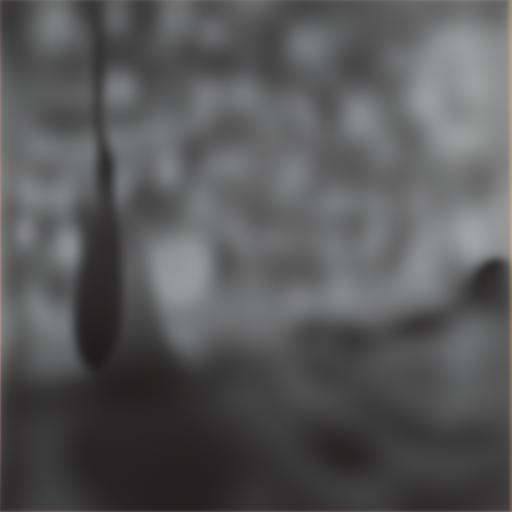}
		
		\\
		\includegraphics[width=0.195\linewidth]{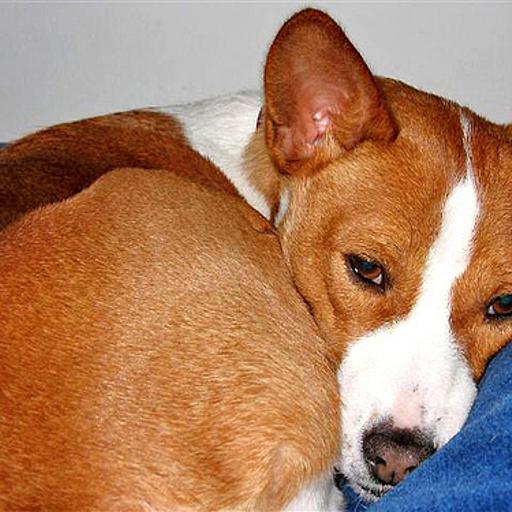}&
		\includegraphics[width=0.195\linewidth]{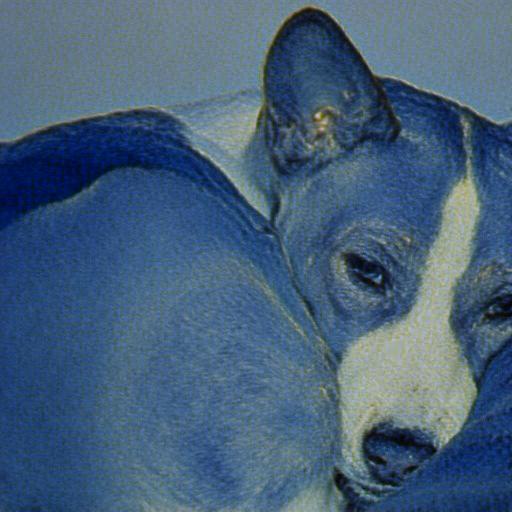}&
		\includegraphics[width=0.195\linewidth]{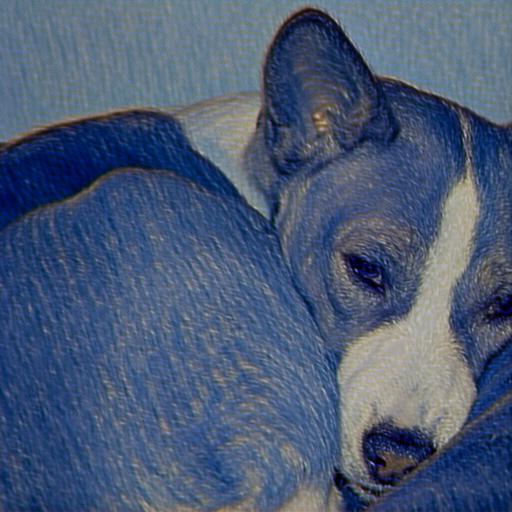}&
		\includegraphics[width=0.195\linewidth]{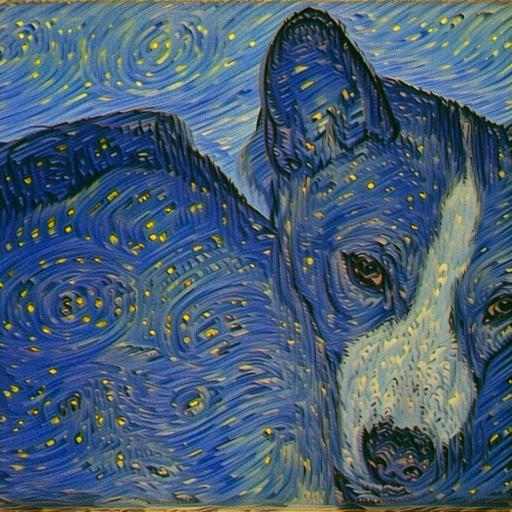}&
		\includegraphics[width=0.195\linewidth]{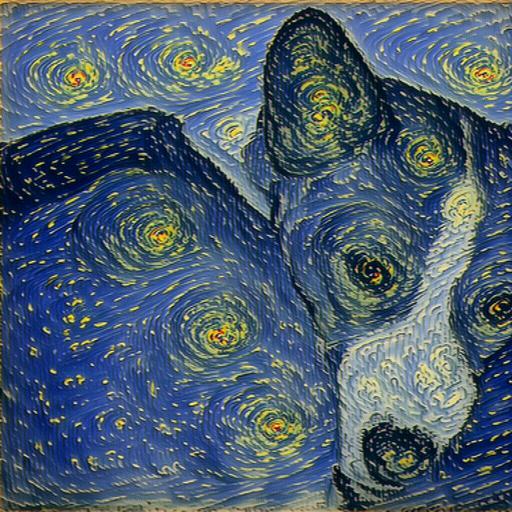}
		\\
		Inputs & 201 & 401 & 601* & 801

	\end{tabular}
	\vspace{0.1em}
	\caption{ {\bf Control of C-S disentanglement} by adjusting the return step $T_{remov}$ of the {\em style removal module}. The top row shows the extracted contents of the style image. The bottom row shows the corresponding stylized results. * denotes our default setting. Zoom-in for better comparison. {\em See SM for quantitative analyses.}}
	\label{fig:disent}
\end{figure}

{\bf Control of C-S Disentanglement.} A prominent advantage of our StyleDiffusion is that we can flexibly control the C-S disentanglement by adjusting the content extraction of the style removal module (Sec.~\ref{sec:SRM}). Fig.~\ref{fig:disent} demonstrates the continuous control achieved by adjusting the return step $T_{remov}$ of the style removal module. As shown in the top row, with the increase of $T_{remov}$, more style characteristics are dispelled, and the main content structures are retained. Correspondingly, when more style is removed in the top row, it will be aptly transferred to the stylized results in the bottom row, \eg, the twisted brushstrokes and the star patterns. It validates that our method successfully separates style from content in a controllable manner and properly transfers it to other contents. Moreover, the flexible C-S disentanglement also makes our StyleDiffusion versatile for other tasks, such as photo-realistic style transfer (see {\em SM}).

\begin{figure}
	\centering
	\setlength{\tabcolsep}{0.02cm}
	\renewcommand\arraystretch{0.4}
	\begin{tabular}{ccccc}
		
		\includegraphics[width=0.195\linewidth]{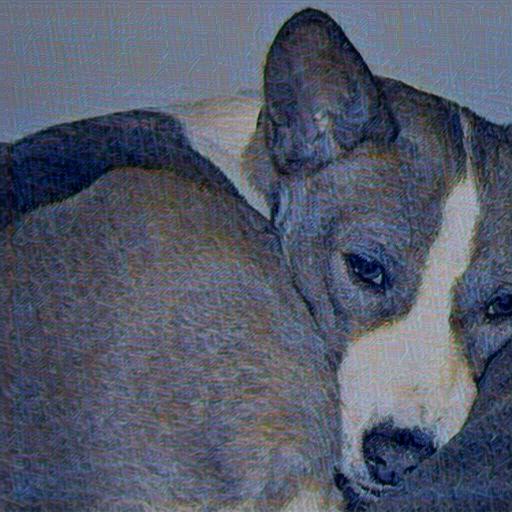}&
		\includegraphics[width=0.195\linewidth]{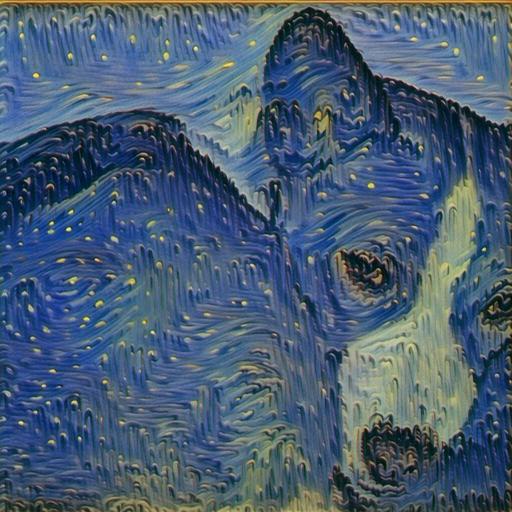}&
		\includegraphics[width=0.195\linewidth]{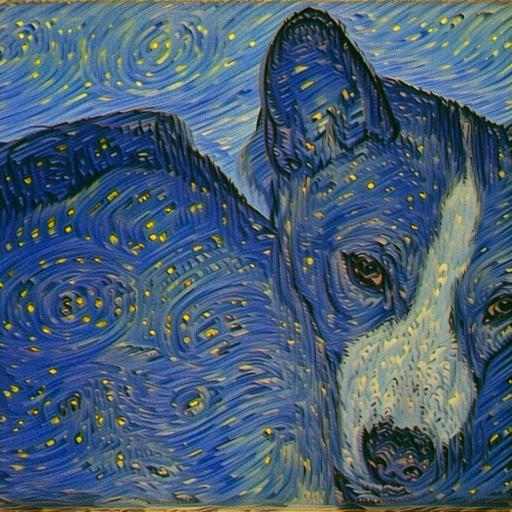}&
		\includegraphics[width=0.195\linewidth]{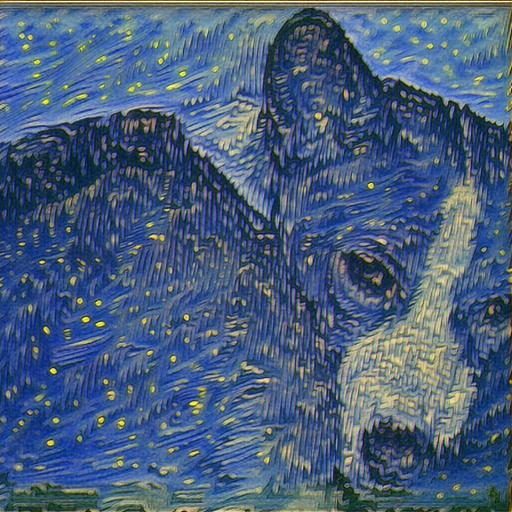}&
		\includegraphics[width=0.195\linewidth]{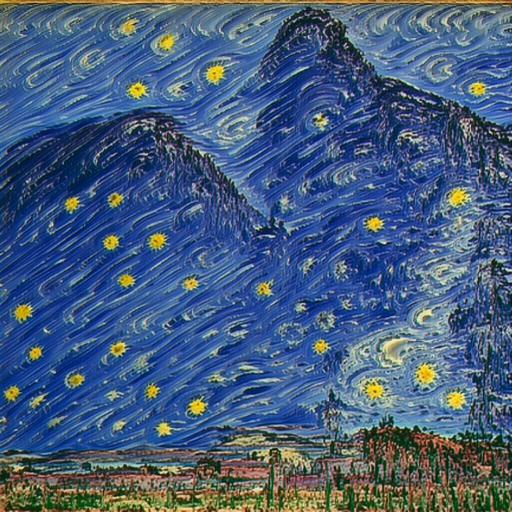}
		\\
		
		\includegraphics[width=0.195\linewidth]{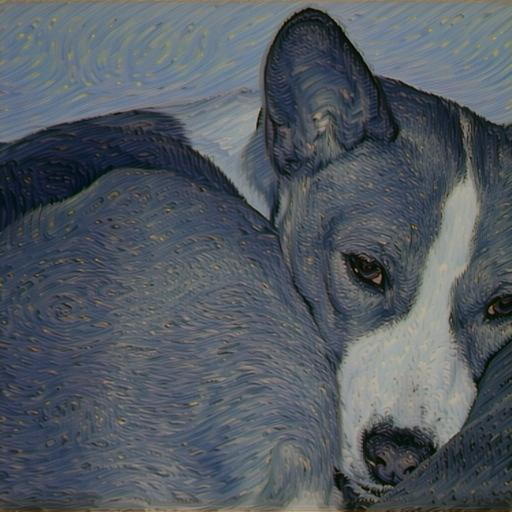}&
		\includegraphics[width=0.195\linewidth]{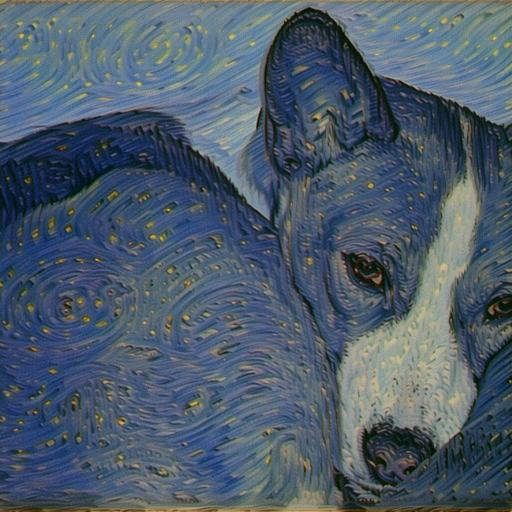}&
		\includegraphics[width=0.195\linewidth]{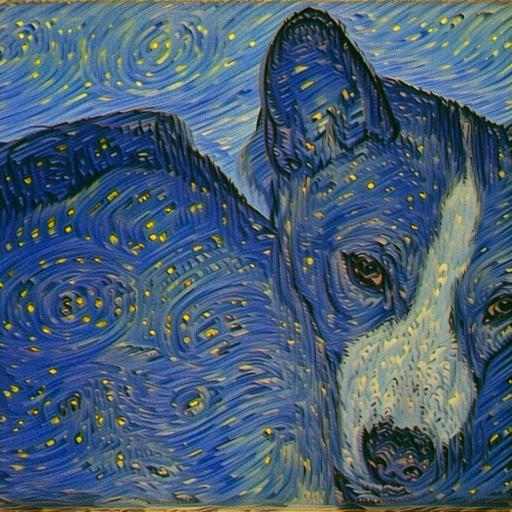}&
		\includegraphics[width=0.195\linewidth]{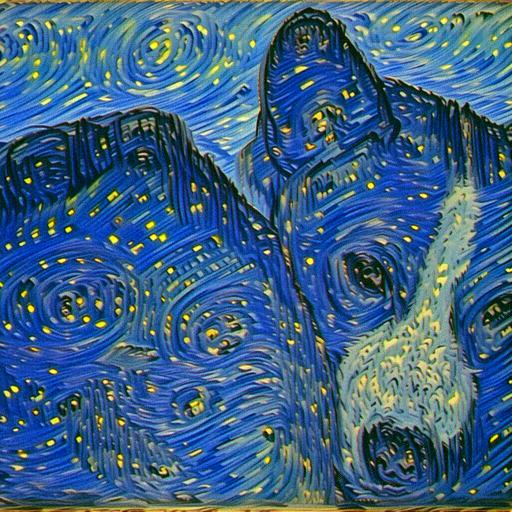}&
		\includegraphics[width=0.195\linewidth]{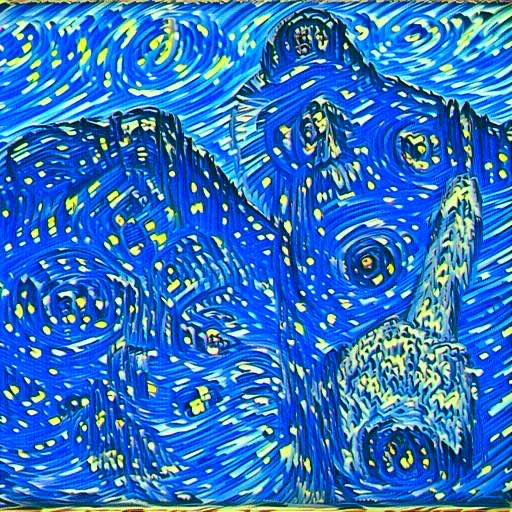}
		\\
		
		101 & 201 & 301* & 401 & 601
		
	\end{tabular}
	\vspace{0.1em}
	\caption{ {\bf Control of C-S trade-off} by adjusting the return step $T_{trans}$ of the {\em style transfer module}. The top row shows adjusting $T_{trans}$ at the {\bf training} stage while fixing $T_{trans}=301$ at the testing stage. The bottom row shows adjusting $T_{trans}$ at the {\bf testing} stage while fixing $T_{trans}=301$ at the training stage. * denotes our default setting. Zoom-in for better comparison. {\em See SM for quantitative analyses.}}
	\label{fig:tradeoff}
\end{figure}

\begin{figure}
	\centering
	\setlength{\tabcolsep}{0.02cm}
	\renewcommand\arraystretch{0.4}
	\begin{tabular}{ccp{0.15em}|p{0.15em}cc}
		
		\includegraphics[width=0.24\linewidth]{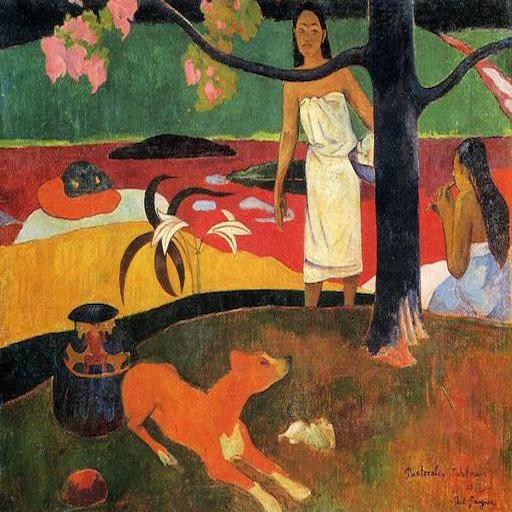}&
		\includegraphics[width=0.24\linewidth]{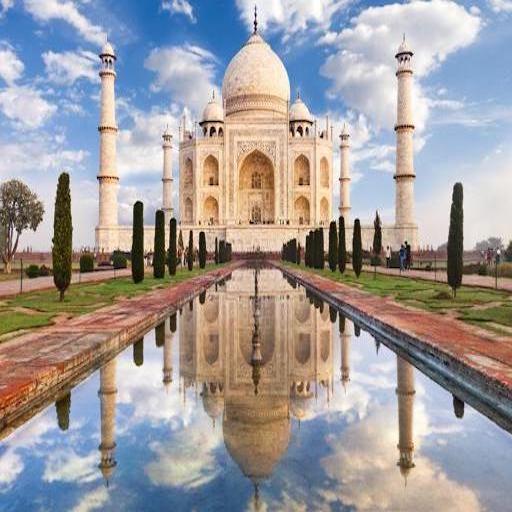}& &&
		\includegraphics[width=0.24\linewidth]{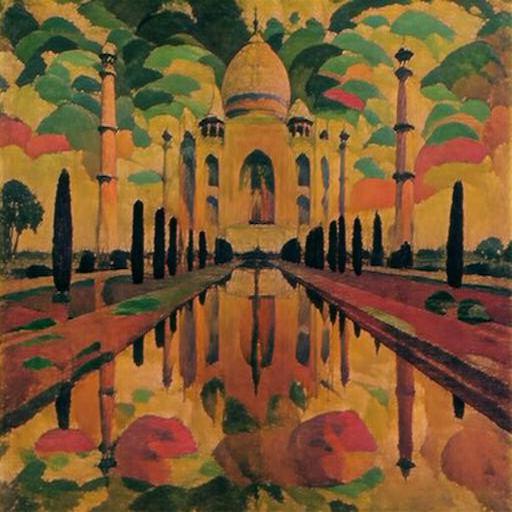}&
		\includegraphics[width=0.24\linewidth]{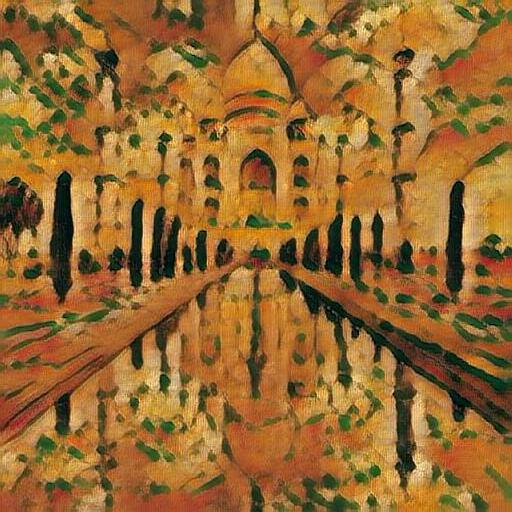}
		\\
		
		\includegraphics[width=0.24\linewidth]{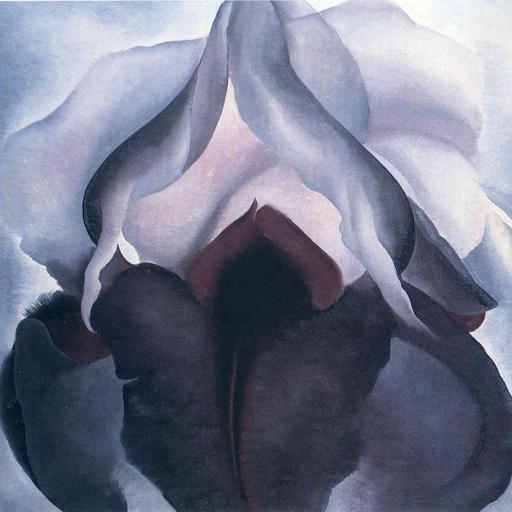}&
		\includegraphics[width=0.24\linewidth]{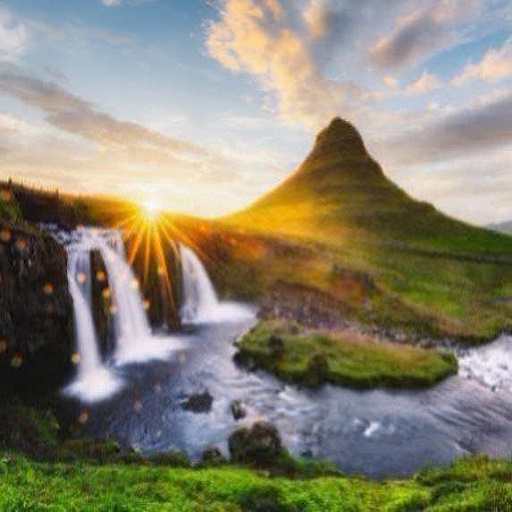}& &&
		\includegraphics[width=0.24\linewidth]{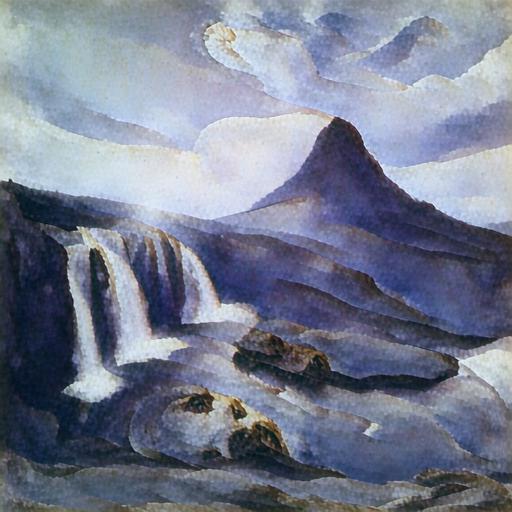}&
		\includegraphics[width=0.24\linewidth]{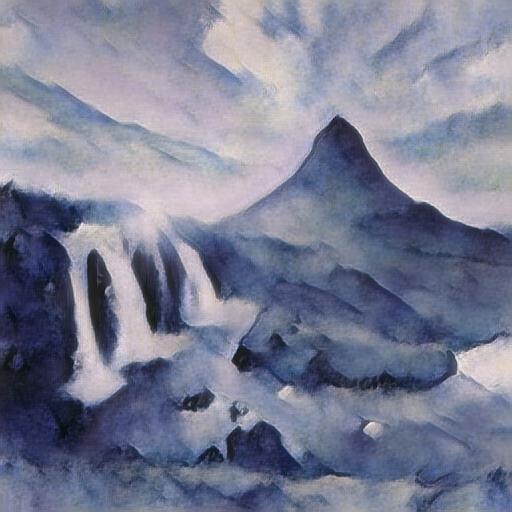}
		\\
		
		Style & Content &&& (a) Diffusion & (b) AE
		
		\\
		\vspace{-0.3em}
		\\
		\hline
		\vspace{-0.3em}
		\\
		\multicolumn{2}{c}{ SSIM / CLIP Score:} &&& 0.672 / 0.741  &  0.526 / 0.702
		
	\end{tabular}
	\vspace{0.5em}
	\caption{ {\bf Diffusion-based vs. AE-based style transfer.}}
	\label{fig:models}
\end{figure}

{\bf Superiority of Diffusion-based Style Transfer.} Although our style transfer module is not limited to the diffusion model, using it offers three main advantages: {\bf (1)} {\em Flexible C-S trade-off control.} As shown in Fig.~\ref{fig:tradeoff}, we can flexibly control the C-S trade-off at both the training stage (top row) and the testing stage (bottom row) by adjusting the return step $T_{trans}$ of the diffusion model. With the increase of $T_{trans}$, more style characteristics are transferred, yet the content structures may be ruined (\eg, the last column). When proper $T_{trans}$ is adopted, \eg, $T_{trans}=301$, the sweet spot can be well achieved. Interestingly, as shown in the last two columns of the bottom row, though the model is trained on $T_{trans}=301$, we can extrapolate the style by using larger $T_{trans}$ (\eg, 401) at the testing stage (but the results may be degraded when using too large $T_{trans}$, \eg, 601). It provides a very flexible way for users to adjust the results according to their preferences. This property, however, cannot be simply achieved by using other models, \eg, the widely used AEs~\cite{huang2017arbitrary,li2017universal}, since our framework does not involve any feature transforms~\cite{huang2017arbitrary,li2017universal} or C-S losses trade-off~\cite{babaeizadeh2019adjustable}. {\bf (2)} {\em Higher-quality stylizations.} Owing to the strong generative ability of the diffusion model, it can achieve higher-quality stylizations than other models. For comparison, we use the pre-trained VGG-AE~\cite{huang2017arbitrary,kwon2022clipstyler} as the style transfer module and fine-tune its decoder network for each style. As shown in column (b) of Fig.~\ref{fig:models}, though the results are still acceptable, they may produce distorted contents and inferior textures, clearly worse than the results generated by the diffusion model in column (a). This is also validated by the bottom quantitative scores. It signifies that the diffusion model can better learn the disentangled content and style characteristics in our framework, helping produce better style transfer results. {\bf (3)} {\em Diversified style transfer.} As mentioned in Sec.~\ref{sec:STM}, during inference, we can directly adopt the stochastic DDPM~\cite{ho2020denoising} forward process (Eq.~(\ref{eq:ddpmf})) to obtain diverse results (see {\em SM}). The diverse results can give users endless choices to obtain more satisfactory results. However, using other models like AEs in our framework cannot easily achieve it~\cite{wang2020diversified}.

\begin{figure*}
	\centering
	\setlength{\tabcolsep}{0.023cm}
	\renewcommand\arraystretch{0.4}
	\begin{tabular}{ccp{0.15em}|p{0.15em}ccccp{0.1em}|p{0.1em}cc}
		
		\includegraphics[width=0.12\linewidth]{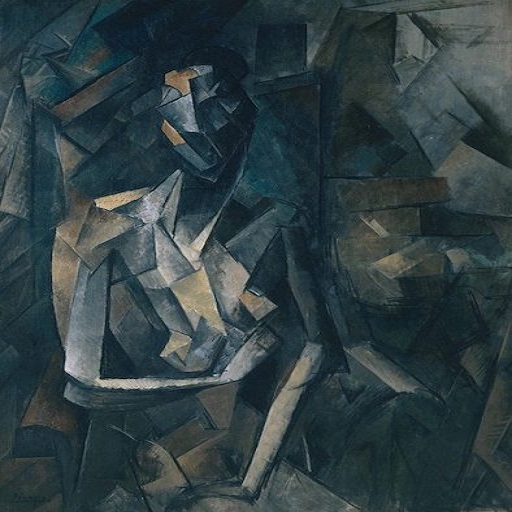}&
		\includegraphics[width=0.12\linewidth]{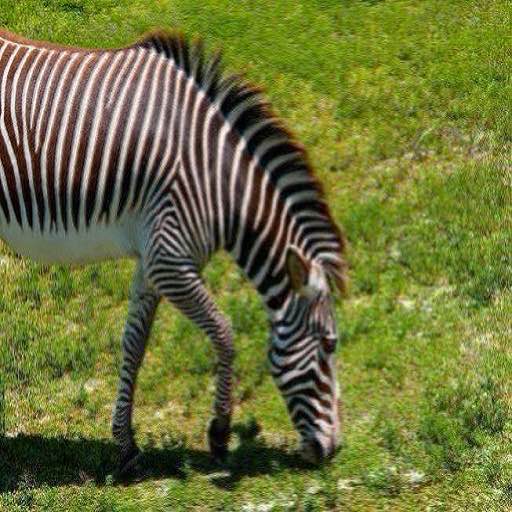}&&&
		\includegraphics[width=0.12\linewidth]{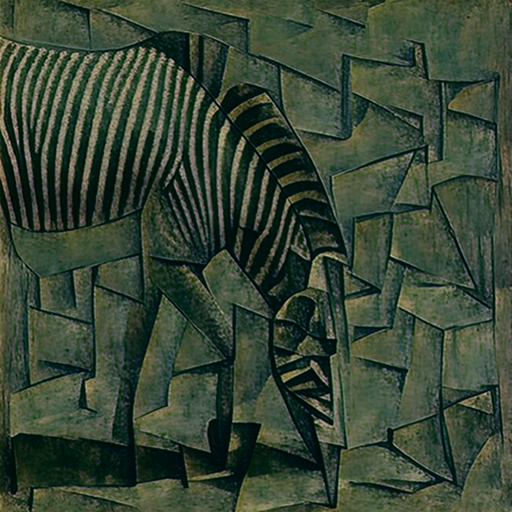}&
		\includegraphics[width=0.12\linewidth]{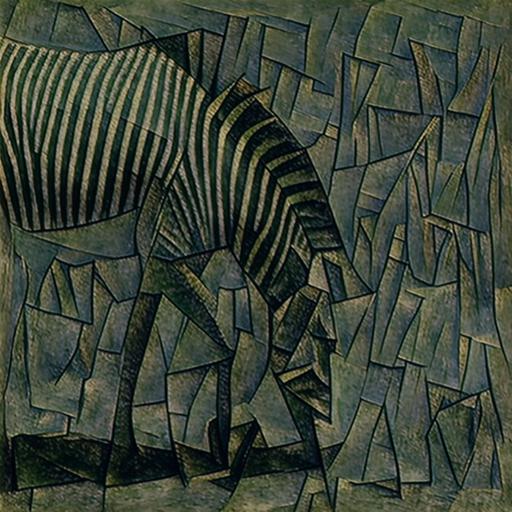}&
		\includegraphics[width=0.12\linewidth]{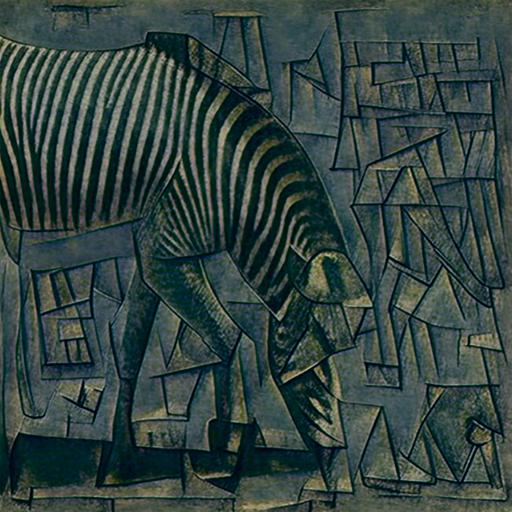}&
		\includegraphics[width=0.12\linewidth]{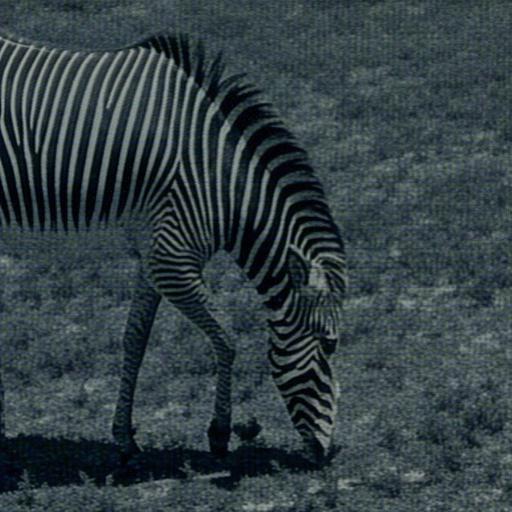}&&&
		\includegraphics[width=0.12\linewidth]{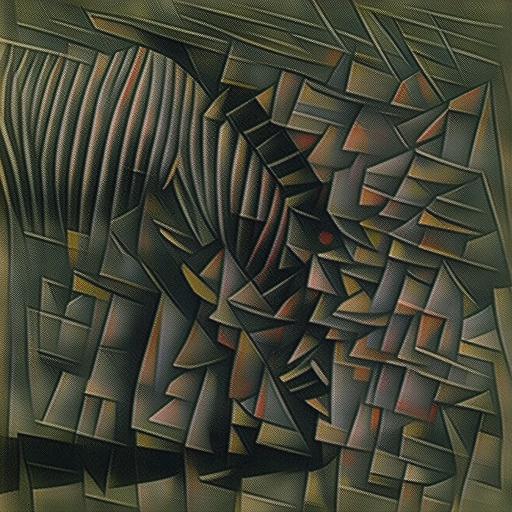}&
		\includegraphics[width=0.12\linewidth]{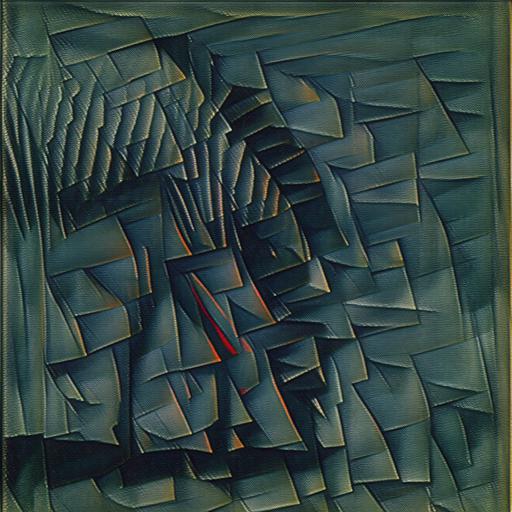}
		\\
		
		\includegraphics[width=0.12\linewidth]{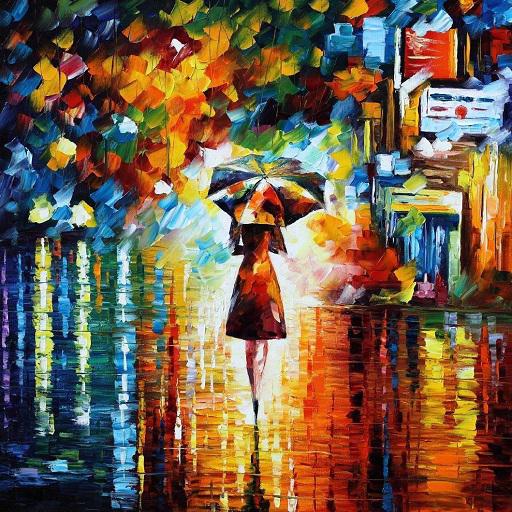}&
		\includegraphics[width=0.12\linewidth]{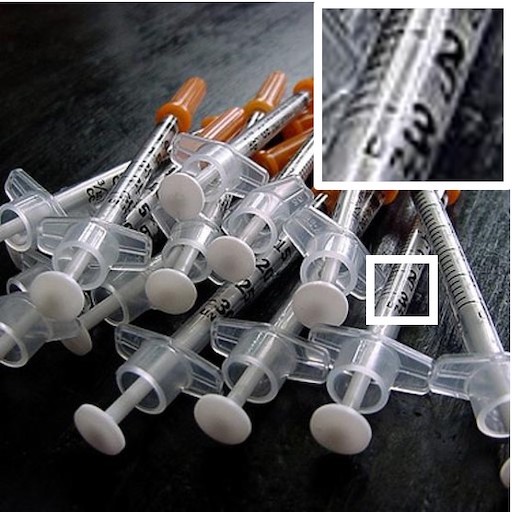}&&&
		\includegraphics[width=0.12\linewidth]{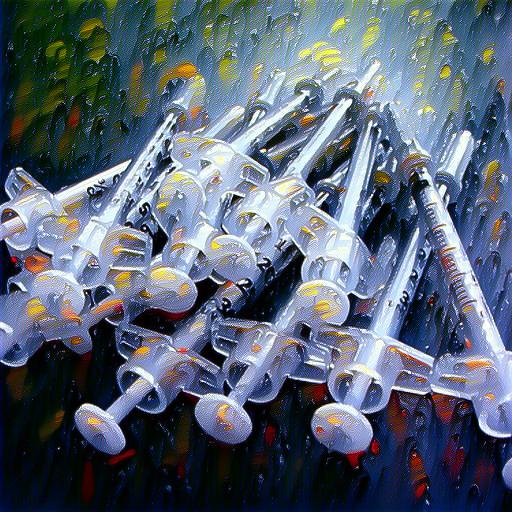}&
		\includegraphics[width=0.12\linewidth]{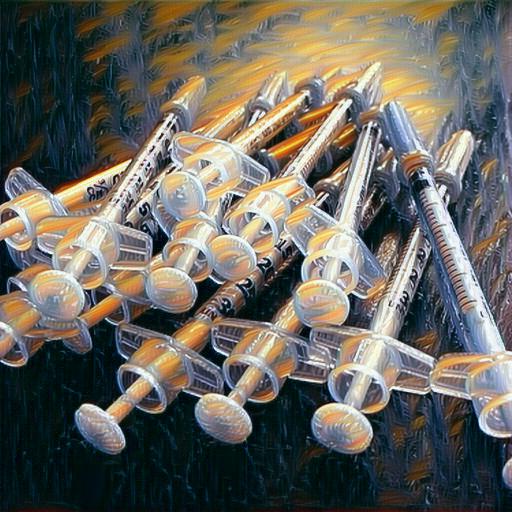}&
		\includegraphics[width=0.12\linewidth]{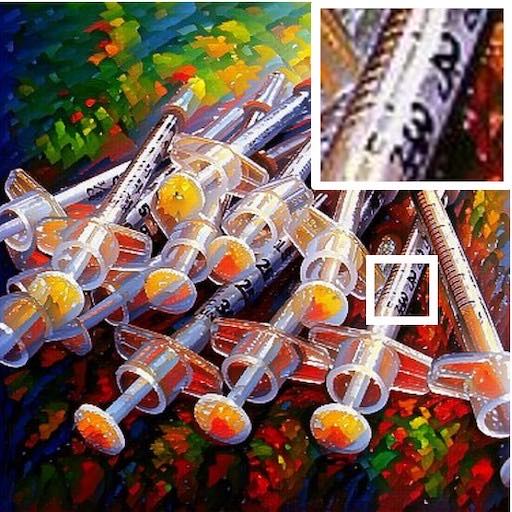}&
		\includegraphics[width=0.12\linewidth]{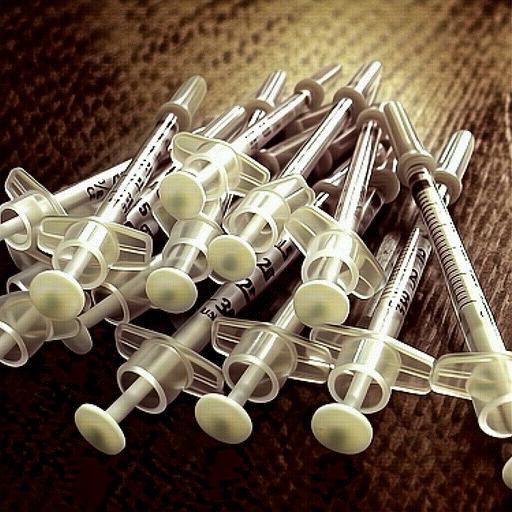}&&&
		\includegraphics[width=0.12\linewidth]{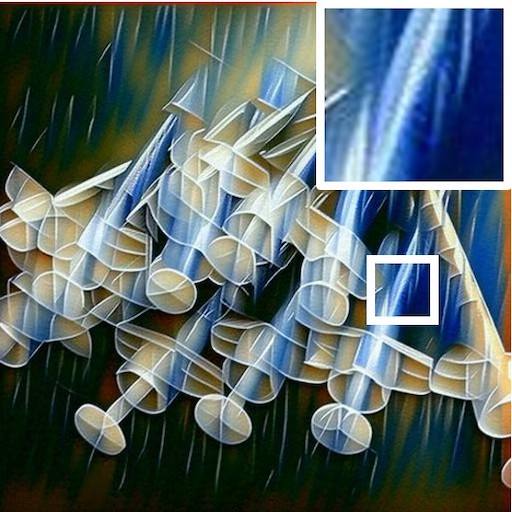}&
		\includegraphics[width=0.12\linewidth]{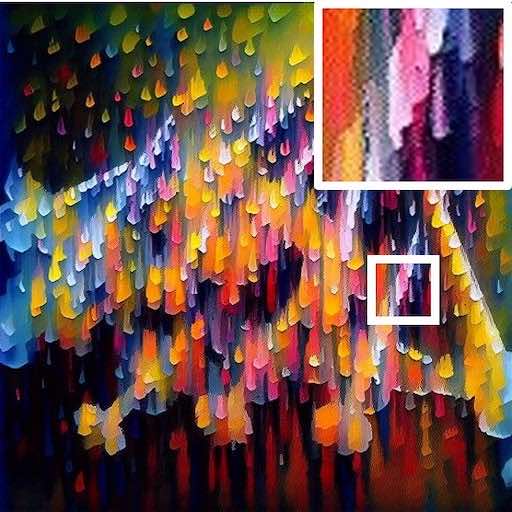}
		\\
		
		Style & Content &&&
		(a) $\mathcal{L}_{SD}^{L1}$ & (b) + $\mathcal{L}_{SD}^{dir}$ & (c) + $\mathcal{L}_{SR}$* & (d) $\mathcal{L}_{SR}$ &&&
		(e) $\mathcal{L}_{Gram}$ & (f) + $\mathcal{L}_{SR}$ 
		
		\\
		\vspace{-0.3em}
		\\
		\hline
		\vspace{-0.3em}
		\\
		\multicolumn{2}{c}{ SSIM / CLIP Score:} &&&  0.660 / 0.652    &  0.693 / 0.705    & 0.672 / 0.741  &  0.793 / 0.488  &&&  0.429 / 0.712  &  0.367 / 0.763
		
	\end{tabular}
	\vspace{0.1em}
	\caption{ {\bf Ablation study on loss functions.} * denotes our full model. Zoom-in for better comparison.}
	\label{fig:loss_ablation}
\end{figure*}

{\bf Loss Analyses.} To verify the effectiveness of each loss term used for fine-tuning our StyleDiffusion, we present ablation study results in Fig.~\ref{fig:loss_ablation}~(a-d). {\bf (1)} Using L1 loss $\mathcal{L}_{SD}^{L1}$ successfully transfers the cubism style like the blocky patterns in the top row, but the colors stray from the style images, especially in the bottom row. It is consistent with our earlier analyses in Sec.~\ref{sec:loss} that the L1 loss is prone to produce implausible results outside the style domain. {\bf (2)} Adding direction loss $\mathcal{L}_{SD}^{dir}$ helps pull the results closer to the style domain. The textures are enhanced in the top row, and the colors are more plausible in the top and bottom rows. {\bf (3)} By further coordinating with the style reconstruction prior $\mathcal{L}_{SR}$, the stylization effects are significantly elevated where the style information is recovered more sufficiently. It may be because it provides a good initialization for the optimization of $\mathcal{L}_{SD}^{L1}$ and $\mathcal{L}_{SD}^{dir}$, which helps them give full play to their abilities. As verified in Fig.~\ref{fig:loss_ablation}~(d), using the style reconstruction alone cannot learn meaningful style patterns except for basic colors. All the above analyses are also supported by the bottom quantitative scores.

\begin{figure}
	\centering
	\setlength{\tabcolsep}{0.015cm}
	\renewcommand\arraystretch{0.5}
	\begin{tabular}{ccp{0.02em}|p{0.02em}ccc}
		\includegraphics[width=0.195\linewidth]{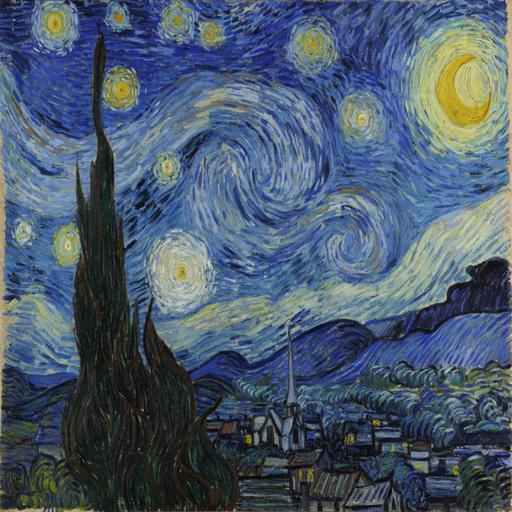}& 
		\includegraphics[width=0.195\linewidth]{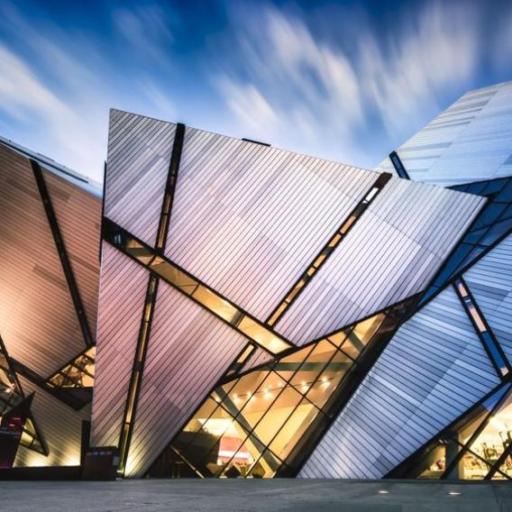}&&&
		\includegraphics[width=0.195\linewidth]{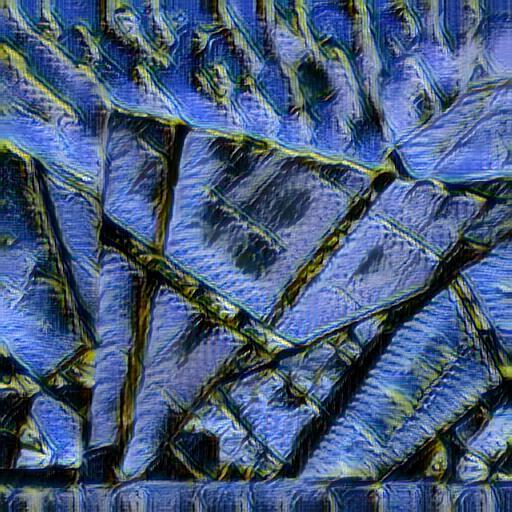}&
		\includegraphics[width=0.195\linewidth]{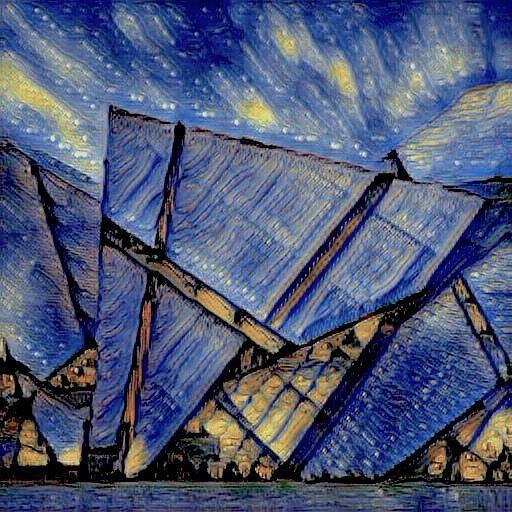}&
		\includegraphics[width=0.195\linewidth]{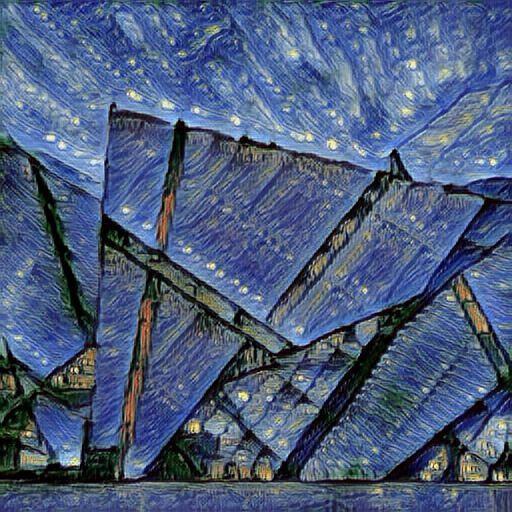}
		
		\\
		\includegraphics[width=0.195\linewidth]{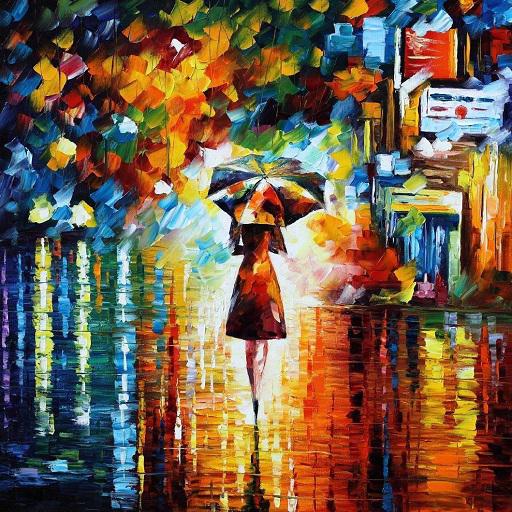}& 
		\includegraphics[width=0.195\linewidth]{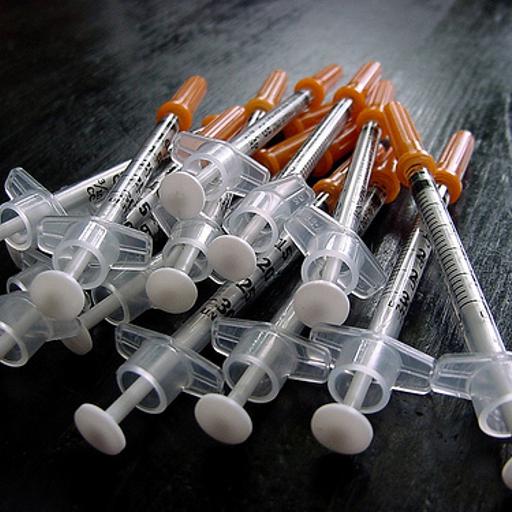}&&&
		\includegraphics[width=0.195\linewidth]{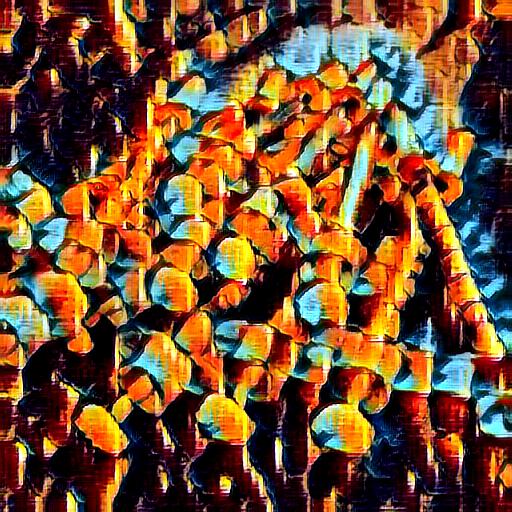}&
		\includegraphics[width=0.195\linewidth]{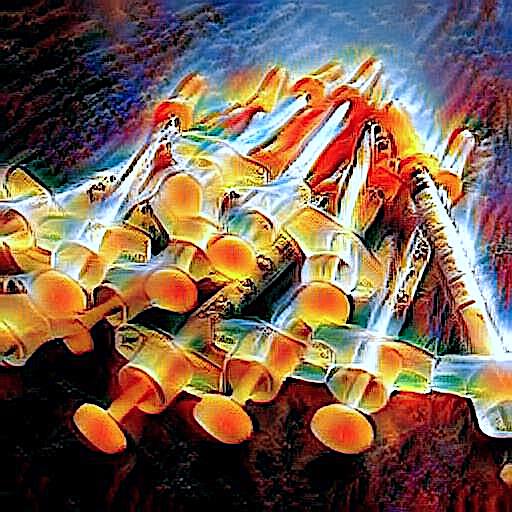}&
		\includegraphics[width=0.195\linewidth]{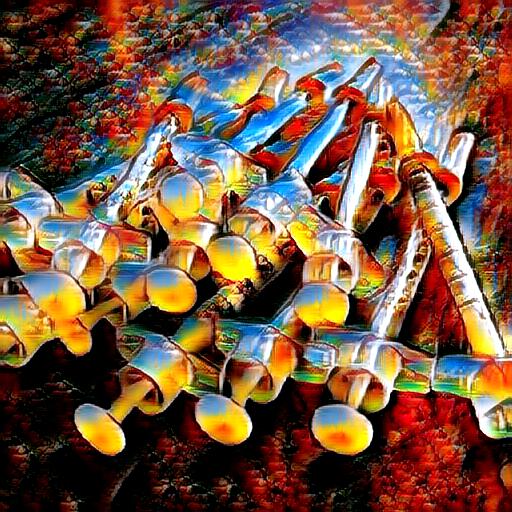}
		\\
		
		Style &  Content  &&& (a) $\mathcal{L}_{Gram}$ &  (b) Ours  & (c) Both
		
		\\
		\vspace{-0.3em}
		\\
		\hline
		\vspace{-0.3em}
		\\
		\multicolumn{2}{c}{ SSIM:} &&& 0.306  &  {\bf 0.526}  &  0.464   \\
		
		\multicolumn{2}{c}{ CLIP Score:} &&& 0.586  &  0.702  &  {\bf  0.728}   \\
		
		\multicolumn{2}{c}{ Style Loss:} &&& 0.263  &  0.732  &  {\bf  0.231}

	\end{tabular}
	\vspace{0.05em}
	\caption{ {\bf More loss function ablation study} on the AE baseline. }
	\label{fig:abs_AEs}
\end{figure}

{\bf Comparison with Gram Loss.} To further verify the superiority of our proposed losses, we replace them with the widely used Gram Loss~\cite{gatys2016image,huang2017arbitrary} in Fig.~\ref{fig:loss_ablation}~(e-f). As can be observed, Gram Loss destroys the content structures severely, \eg, the zebra head in the top row and the enlarged area in the bottom row. This is because it does not disentangle C-S and only matches the global statistics without considering the relationship between C-S. In contrast, our losses focus on learning the disentangled style information apart from the content, which is induced by the difference between content and its stylized result. Therefore, they can better understand the relationship between C-S, achieving more satisfactory results with fine style details and better-preserved contents, as validated by Fig.~\ref{fig:loss_ablation}~(c) and the bottom quantitative scores. Furthermore, we also conduct comparisons between our proposed losses and Gram Loss~\cite{gatys2016image,huang2017arbitrary} on the AE baseline~\cite{huang2017arbitrary,kwon2022clipstyler} to eliminate the impact of diffusion models. As shown in Fig.~\ref{fig:abs_AEs}~(a-b), our losses can achieve more satisfactory results than Gram Loss, which is consistent with the results in Fig.~\ref{fig:loss_ablation}. Moreover, as shown in Fig.~\ref{fig:abs_AEs}~(c), they can also be combined with Gram Loss to improve the performance on the Style Loss metric. However, it may affect the full disentanglement of C-S in our framework, which strays from our target and decreases the content preservation (see SSIM score in Fig.~\ref{fig:abs_AEs}~(c)). Therefore, we do not incorporate Gram Loss in our framework by default.

\section{Conclusion and Limitation}
\label{sec:conclusion}
In this work, we present a new framework for more interpretable and controllable C-S disentanglement and style transfer. Our framework, termed {\em StyleDiffusion}, leverages diffusion models to explicitly extract the content information and implicitly learn the complementary style information. A novel CLIP-based style disentanglement loss coordinated with a style reconstruction prior is also introduced to encourage the disentanglement and style transfer. Our method yields very encouraging stylizations, especially for challenging styles, and the experimental results verify its effectiveness and superiority against state of the art. 

Currently, the framework still suffers from several limitations: (1) The model needs to be fine-tuned for each style, and arbitrary style transfer is left to our future work. (2) The efficiency is not fast enough due to the use of diffusion models. Further research in accelerating diffusion sampling would be helpful. (3) There are some failure cases analyzed in {\em SM}, which may help inspire future improvements. Moreover, our framework may also be applied to other image translation~\cite{huang2018multimodal} or manipulation~\cite{park2020swapping} tasks, and we would like to explore them in our future work.

\let\thefootnote\relax
\footnotetext{{\bf Acknowledgments.} We thank Zeyi Huang and Xiaoting Zhang for their insightful discussions and suggestions. This work was supported in part by  the National Program of China (2020YFC1523201, 62172365, 19ZDA197), Zhejiang Elite Program (2022C01222), and Key Technologies and Product Research and Development Projects for Cultural Relics Protection and Trading Circulation.}

{\small
\bibliographystyle{ieee_fullname}
\bibliography{egbib}
}

\clearpage

\appendix

\section*{Supplementary Material}

\begin{itemize}
	\setlength{\itemsep}{3pt}
	\setlength{\parsep}{0pt}
	\setlength{\partopsep}{0pt}
	\setlength{\parskip}{0pt}
	
	\item Section~\ref{sec:impact}: Discussion of societal impact.
	\item Section~\ref{sec:asset}: List of used assets.
	\item Section~\ref{sec:remove}: Details of style removal process.
	\item Section~\ref{sec:finetune}: Details of StyleDiffusion fine-tuning.
	\item Section~\ref{sec:time}: More timing and resource information.
	\item Section~\ref{sec:moreablation}: More ablation study and analyses.
	\item Section~\ref{sec:extension}: Extensions to photo-realistic style transfer, multi-modal style manipulation, and diversified style transfer.
	\item Section~\ref{sec:cmp}: More comparison results with state-of-the-art style transfer methods.
	\item Section~\ref{sec:res}: Additional results synthesized by the proposed method.
	\item Section~\ref{sec:limit}: Limitation and discussion.
\end{itemize}

\section{Societal Impact}
\label{sec:impact}
{\bf Positive Impact.} There may be three positive impacts of the proposed method. (1) The proposed method may help the workers engaged in artistic creation or creative design improve the efficiency and quality of their work. (2) The proposed method may inspire researchers in similar fields to design more effective and superior approaches in the future. (3) The proposed method may help the common users obtain more satisfactory creative results.

{\bf Negative Impact.} The proposed method may be used for generating counterfeit artworks. To mitigate this, further research on identification of generated content is needed.

\section{Used Assets}
\label{sec:asset}

We used the following assets to (1) conduct the comparison experiments [1.-10.] and (2) train the proposed style transfer networks [11.-12.]. To the best of our knowledge, these assets have no ethical concerns.

\begin{enumerate}
	\setlength{\itemsep}{3pt}
	\setlength{\parsep}{0pt}
	\setlength{\partopsep}{0pt}
	\setlength{\parskip}{0pt}
	\item Gatys~\cite{gatys2016image}: \url{https://github.com/leongatys/PytorchNeuralStyle Transfer}, MIT License.
	\item EFDM~\cite{zhang2022exact}: \url{https://github.com/YBZh/EFDM}, MIT License.
	\item StyTr$^2$~\cite{deng2022stytr2}: \url{https://github.com/diyiiyiii/StyTR-2}, No License.
	\item ArtFlow~\cite{an2021artflow}: \url{https://github.com/pkuanjie/ArtFlow}, No License.
	\item AdaAttN~\cite{liu2021adaattn}: \url{https://github.com/Huage001/AdaAttN}, No License.
	\item IECAST~\cite{chen2021artistic}: \url{https://github.com/HalbertCH/IEContraAST}, MIT License.
	\item MAST~\cite{deng2020arbitrary}: \url{https://github.com/diyiiyiii/Arbitrary-Style-Transfer-via-Multi-Adaptation-Network}, No License.
	\item TPFR~\cite{svoboda2020two}: \url{https://github.com/nnaisense/conditional-style-transfer}, View License in repository.
	\item Johnson~\cite{johnson2016perceptual}: \url{https://github.com/abhiskk/fast-neural-style}, MIT License.
	\item LapStyle~\cite{lin2021drafting}: \url{https://github.com/PaddlePaddle/PaddleGAN/blob/develop/docs/en\_US/tutorials/lap\_style.md}, Apache-2.0 License.
	\item ADM~\cite{dhariwal2021diffusion}: \url{https://github.com/openai/guided-diffusion}, MIT License.
	\item ImageNet~\cite{russakovsky2015imagenet}: \url{https://image-net.org/}, Unknown License.
\end{enumerate}

\section{Details of Style Removal Process}
\label{sec:remove}
As detailed in Algorithm~\ref{alg:style_remove}, the style removal process consists of two steps. In the first step, we remove the color of the input image $I$ using a color removal operation $\mathcal{R}_{color}$ (\eg, the commonly used ITU-R 601-2 luma transform~\cite{gonzalez2009digital}), obtaining grayscale image $I'$. In the second step, we use the pre-trained diffusion model $\epsilon_{\theta}$ and adopt the deterministic DDIM forward and reverse processes to gradually remove the style information. To accelerate the process without sacrificing much performance, we use fewer discretization steps $\{t_s\}_{s=1}^{S_{for}}$ such that $t_1=0$ and $t_{S_{for}}=T_{remov}$. We set $S_{for} = 40$ for forward process and $S_{rev} = 40$ for reverse process in all experiments. While using larger $S_{for}$ or $S_{rev}$ could reconstruct the high-frequency details better, we found the current setting is enough for our task. {\em For more details about their effects, we suggest the readers refer to~\cite{kim2022diffusionclip}.} After $K_r$ iterations (we set $K_r=5$ for all experiments) of forward and reverse processes, the style characteristics of $I'$ will be dispelled, and thus we obtain the content $I^c$ of the input image.

\begin{algorithm}[t]
	\caption{Style Removal Process.}
	\label{alg:style_remove}
	\KwIn{ pre-trained model $\epsilon_\theta$, input image $I$, return step $T_{remov}$, forward step $S_{for}$, reverse step $S_{rev}$, iteration $K_r$
	}
	\KwOut{ input image's content $I^c$}  
	\BlankLine
	\tcp{\small \color{red} {Remove color}}
	$I' = \mathcal{R}_{color}(I)$

	\tcp{\small \color{red} {Diffusion-based style removal}}
	Compute $\{t_s\}^{S_{for}}_{s=1}$ s.t. $t_1 = 0$, $t_{S_{for}}=T_{remov}$
	
	$x_0 \leftarrow I$
	
	\For{$k=1:K_r$}{
		
		\For{$s=1:S_{for}-1$}{
			\footnotesize
			$x_{t_{s+1}} \leftarrow \sqrt{\bar{\alpha}_{t_{s+1}}}f_\theta(x_{t_s},t_s) + \sqrt{1-\bar{\alpha}_{t_{s+1}}} \epsilon_\theta(x_{t_s},t_s)$
		}

		$x_{t_{S_{rev}}} \leftarrow x_{t_{S_{for}}}$
		
		\For{$s=S_{rev}:2$}{
			{\footnotesize $x_{t_{s-1}} \leftarrow \sqrt{\bar{\alpha}_{t_{s-1}}}f_{\hat{\theta}}(x_{t_s},t_s) + \sqrt{1-\bar{\alpha}_{t_{s-1}}} \epsilon_{\hat{\theta}}(x_{t_s},t_s)$}
		}
	}
	
	$I^c \leftarrow x_0$
	
\end{algorithm}

\begin{algorithm}[tbhp]
	\caption{StyleDiffusion Fine-tuning.}
	\label{alg:fine-tune}
	\KwIn{ pre-trained model $\epsilon_\theta$, content images' contents $\{I_{ci}^c\}^N_{i=1}$, style image's content $I_s^c$, style image $I_s$, return step $T_{trans}$, forward step $S_{for}$, reverse step $S_{rev}$, fine-tuning epoch $K$, style reconstruction iteration $K_s$
	}
	\KwOut{ fine-tuned model $\epsilon_{\hat{\theta}}$}  
	\BlankLine
	\tcp{\small \color{red} {Precompute content latents}}
	
	Compute $\{t_s\}^{S_{for}}_{s=1}$ s.t. $t_1 = 0$, $t_{S_{for}}=T_{trans}$
	
	\For{$i=1:N$}{
		$x_0 \leftarrow I_{ci}^c$
		
		\For{$s=1:S_{for}-1$}{
			\footnotesize
			$x_{t_{s+1}} \leftarrow \sqrt{\bar{\alpha}_{t_{s+1}}}f_\theta(x_{t_s},t_s) + \sqrt{1-\bar{\alpha}_{t_{s+1}}} \epsilon_\theta(x_{t_s},t_s)$
		}
		Save the latent $x^{ci} \leftarrow x_{t_{S_{for}}}$
	}
	
	\tcp{\small \color{red} {Precompute style latent}}
	Compute $\{t_s\}^{S_{for}}_{s=1}$ s.t. $t_1 = 0$, $t_{S_{for}}=T_{trans}$
	
	$x_0 \leftarrow I_s^c$
	
	\For{$s=1:S_{for}-1$}{
		\footnotesize
		$x_{t_{s+1}} \leftarrow \sqrt{\bar{\alpha}_{t_{s+1}}}f_\theta(x_{t_s},t_s) + \sqrt{1-\bar{\alpha}_{t_{s+1}}} \epsilon_\theta(x_{t_s},t_s)$
	}
	Save the latent $x^s \leftarrow x_{t_{S_{for}}}$

	\tcp{\small \color{red} {Fine-tune the diffusion model}}
	
	Initialize $\epsilon_{\hat{\theta}} \leftarrow \epsilon_\theta$
	
	Compute $\{t_s\}^{S_{rev}}_{s=1}$ s.t. $t_1 = 0$, $t_{S_{rev}}=T_{trans}$
	
	\For{$k=1:K$}{
		
		\tcp{\scriptsize  \color{blue} {Optimize the style reconstruction loss}}
		
		\For{$i=1:K_s$}{
			$x_{t_{S_{rev}}} \leftarrow x^s$
			
			\For{$s=S_{rev}:2$}{
				{\footnotesize $x_{t_{s-1}} \leftarrow \sqrt{\bar{\alpha}_{t_{s-1}}}f_{\hat{\theta}}(x_{t_s},t_s) + \sqrt{1-\bar{\alpha}_{t_{s-1}}} \epsilon_{\hat{\theta}}(x_{t_s},t_s)$}
				
				$I_{ss}\leftarrow f_{\hat{\theta}}(x_{t_s},t_s)$
				
				$\mathcal{L} \leftarrow \mathcal{L}_{SR}(I_{ss}, I_s)$
				
				Take a gradient step on $\nabla_{\hat{\theta}} \mathcal{L}$

			}
		}
		
		\tcp{\scriptsize  \color{blue} {Optimize the style disentanglement loss}}
		
		\For{$i=1:N$}{
			$x_{t_{S_{rev}}} \leftarrow x^{ci}$
			
			\For{$s=S_{rev}:2$}{
				{\footnotesize $x_{t_{s-1}} \leftarrow \sqrt{\bar{\alpha}_{t_{s-1}}}f_{\hat{\theta}}(x_{t_s},t_s) + \sqrt{1-\bar{\alpha}_{t_{s-1}}} \epsilon_{\hat{\theta}}(x_{t_s},t_s)$}
				
				$I_{cs} \leftarrow f_{\hat{\theta}}(x_{t_s},t_s)$

				$\mathcal{L} \leftarrow \mathcal{L}_{SD}(I_{ci}^c, I_{cs}, I_s^c, I_s)$
				
				Take a gradient step on $\nabla_{\hat{\theta}} \mathcal{L}$
				
			}
			
		}
	}
\end{algorithm}

\section{Details of StyleDiffusion Fine-tuning}
\label{sec:finetune}
Similar to~\cite{kim2022diffusionclip} and detailed in Algorithm~\ref{alg:fine-tune}, we first precompute the content latents $\{x^{ci}\}_{i=1}^N$ using the deterministic DDIM forward process of the pre-trained diffusion model $\epsilon_{\theta}$. {\em The precomputed content latents can be stored and reused for fine-tuning other styles.}  In our experiments, we fine-tune the diffusion models for all styles using the same precomputed latents of 50 content images sampled from ImageNet~\cite{russakovsky2015imagenet}. Fine-tuning with more content images may improve the results but also increases the time cost. Thus, we made a trade-off and found the current setting could work well for most cases. To accelerate the process, we use fewer discretization steps $\{t_s\}_{s=1}^{S_{for}}$ such that $t_1=0$ and $t_{S_{for}}=T_{trans}$. We set $S_{for} = 40$ for forward process and $S_{rev} = 6$ for reverse process in all experiments. We found $S_{rev} = 6$ is enough to reconstruct clear content structures during style transfer.

In the second step, we precompute the style latent $x^s$ with the same process as above. The style latent will be used to optimize the style reconstruction loss.

In the third step, we copy $\epsilon_{\theta}$ to $\epsilon_{\hat{\theta}}$ and start to update $\epsilon_{\hat{\theta}}$ in two substeps. In the first substep, we feed the style latent $x^s$ and generate the stylized image $I_{ss}$ through the deterministic DDIM reverse process. The model is updated under the guidance of the style reconstruction loss $\mathcal{L}_{SR}$. The first substep is repeated $K_s$ times (we set $K_s=50$ for all experiments) until converged. In the second substep, we feed each content latent in $\{x^{ci}\}_{i=1}^N$ and generate the stylized image through the deterministic DDIM reverse process. The model is updated under the guidance of the style disentanglement loss $\mathcal{L}_{SD}$. At last, we repeat the whole third step $K$ epochs (we set $K=5$ for all experiments) until converged.

\begin{figure}[b]
	\centering
	\includegraphics[width=0.8\linewidth]{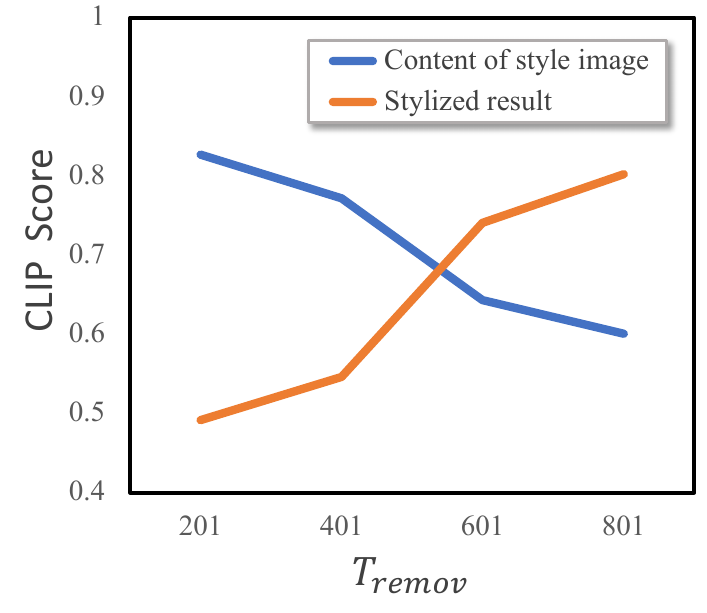}
	\caption{{\bf C-S disentanglement of style image} achieved by adjusting the return step $T_{remov}$ of the {\em style removal module}. CLIP score (averaged on 384 image pairs) measures the style similarity with the style image. When more style information is removed (blue line), it will be transferred to the stylized result (orange line).
	}
	\label{fig:disent_quantity}
\end{figure}

\section{Timing and Resource Information}
\label{sec:time}
Here, we provide more details on the timing and resource information of our StyleDiffusion using an Nvidia Tesla A100 GPU when stylizing $512 \times 512$ size images.

{\bf Style Removal.} When we use the default setting $(S_{for},S_{rev}) = (40, 40)$, the forward and reverse processes each takes around $4.921$ seconds. Therefore, the whole style removal process takes around $2\times 4.921 \times 5 = 49.21$ seconds. It requires about 11GB of GPU memory to run at resolution $512 \times 512$ pixels.

{\bf Fine-tuning.} As illustrated in Algorithm~\ref{alg:fine-tune}, the StyleDiffusion fine-tuning process consists of a latent precomputing stage and a model updating stage. The latent precomputing stage is carried out just once and can be reused for fine-tuning other styles. When we use $S_{for}=40$ as default, the forward process takes around $4.921$ seconds. 
Therefore, when we precompute the latents from 50 images, it takes around $50\times 4.921=246.05$ seconds and requires about 11GB GPU memory. For the model updating stage, when the batch size is 1 and $S_{rev}=6$, the first substep (optimizing the style reconstruction loss $\mathcal{L}_{SR}$) takes around $2.092$ seconds for each repeat, and the second substep (optimizing the style disentanglement loss $\mathcal{L}_{SD}$) takes around 3.351 seconds for each content latent. Therefore, one epoch with 50 repeated first substep and 50 precomputed content latents for the second substep takes around $50\times 2.092 + 50\times 3.351 = 272.15$ seconds. When we fine-tune the model with 5 epochs, it takes around 23 minutes in total. The fine-tuning process requires about 26GB of GPU memory.

\begin{figure}[b]
	\centering
	\includegraphics[width=0.75\linewidth]{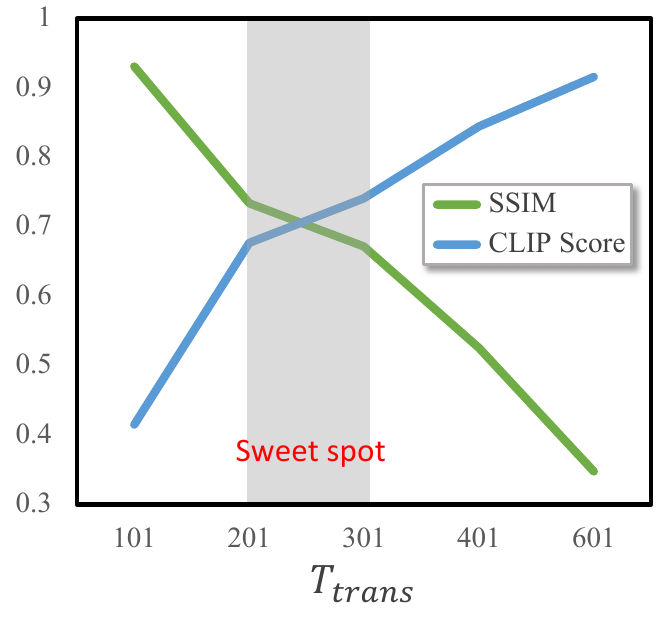}
	\caption{{\bf C-S trade-off} achieved by adjusting the return step $T_{trans}$ of the {\em style transfer module} at the {\bf training} stage while fixing $T_{trans}=301$ at the testing stage. SSIM and CLIP score (averaged on 384 image pairs) measure the content similarity and the style similarity, respectively.
	}
	\label{fig:tradeoff-train}
\end{figure}

\begin{figure}[b]
	\centering
	\includegraphics[width=0.75\linewidth]{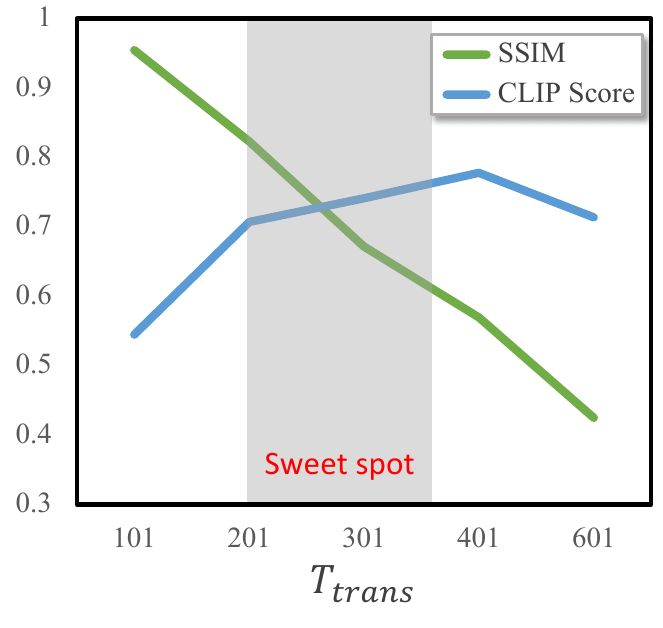}
	\caption{{\bf C-S trade-off} achieved by adjusting the return step $T_{trans}$ of the {\em style transfer module} at the {\bf testing} stage while fixing $T_{trans}=301$ at the training stage. SSIM and CLIP score (averaged on 384 image pairs) measure the content similarity and the style similarity, respectively.
	}
	\label{fig:tradeoff-test}
\end{figure}

\begin{figure*}[t]
	\centering
	\setlength{\tabcolsep}{0.02cm}
	\renewcommand\arraystretch{0.4}
	\begin{tabular}{ccp{0.15em}|p{0.15em}ccp{0.15em}|p{0.15em}ccc}
		
		\includegraphics[width=0.135\linewidth]{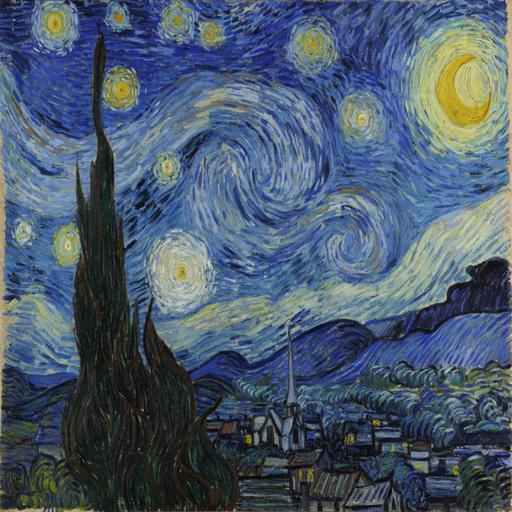}&
		\includegraphics[width=0.135\linewidth]{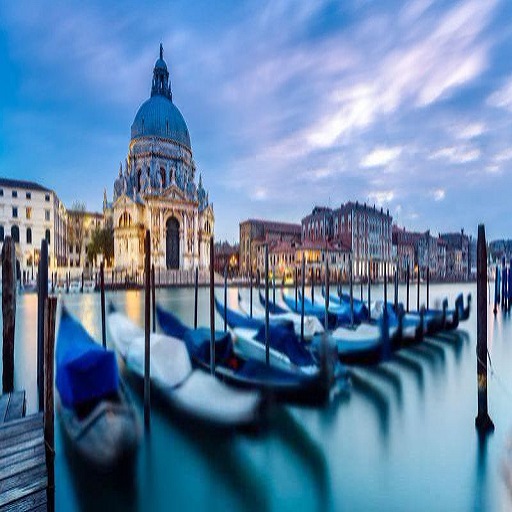}&&&
		\includegraphics[width=0.135\linewidth]{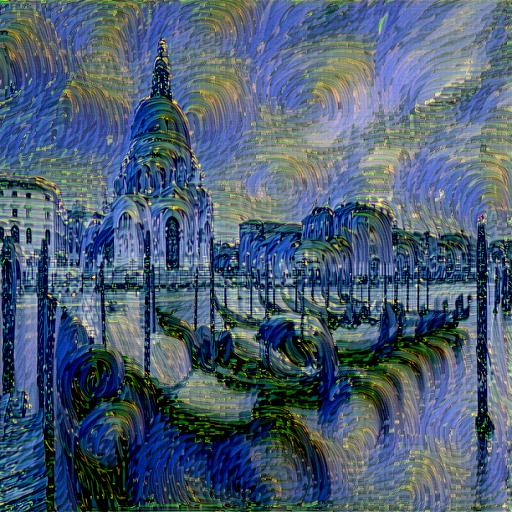}&
		\includegraphics[width=0.135\linewidth]{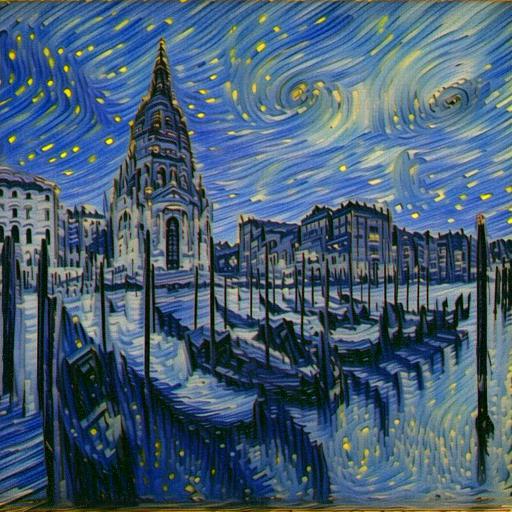}&&&
		\includegraphics[width=0.135\linewidth]{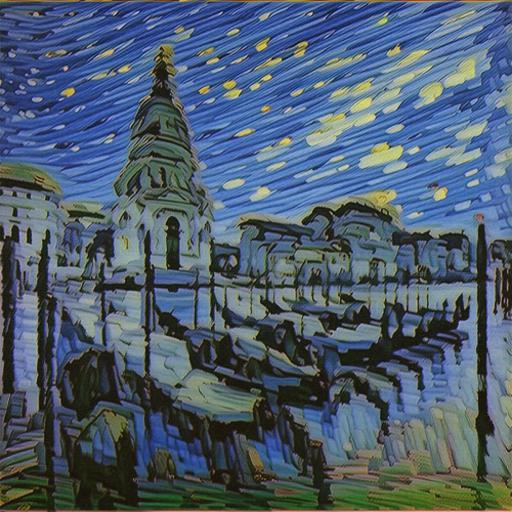}&
		\includegraphics[width=0.135\linewidth]{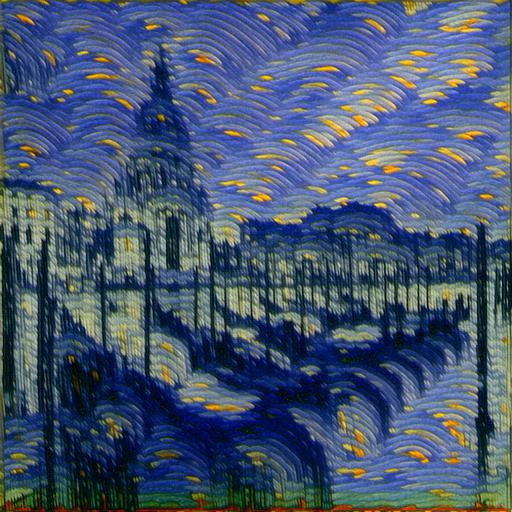}&
		\includegraphics[width=0.135\linewidth]{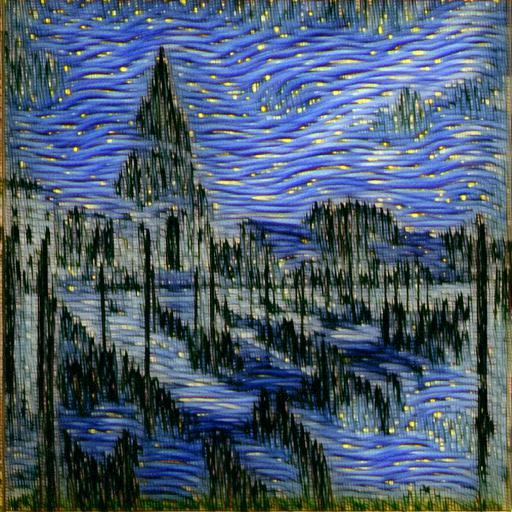}
		\\
		
		\includegraphics[width=0.135\linewidth]{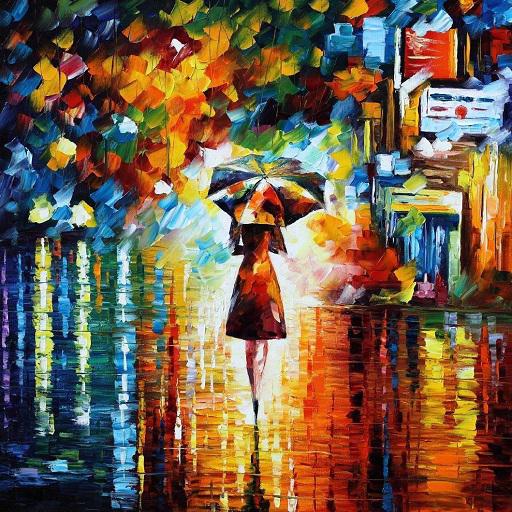}&
		\includegraphics[width=0.135\linewidth]{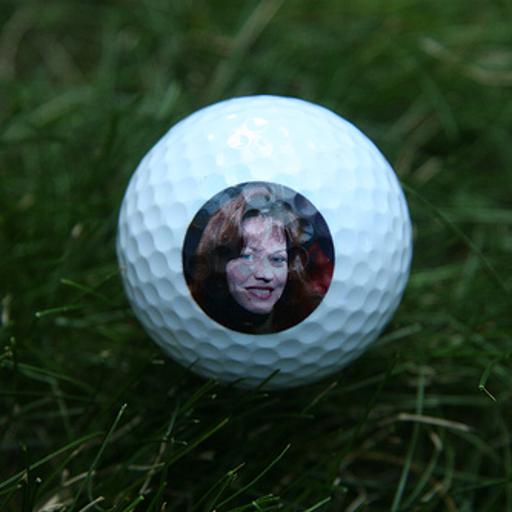}&&&
		\includegraphics[width=0.135\linewidth]{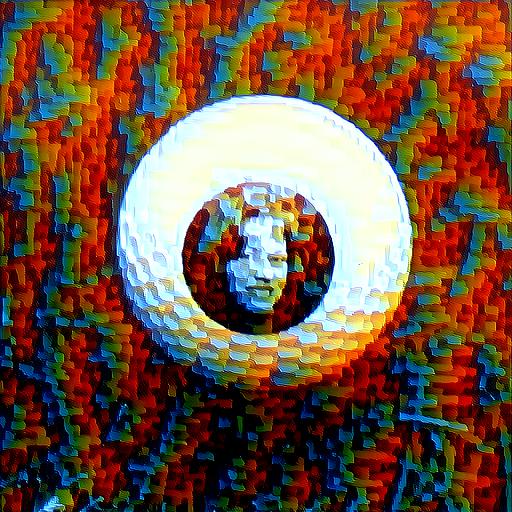}&
		\includegraphics[width=0.135\linewidth]{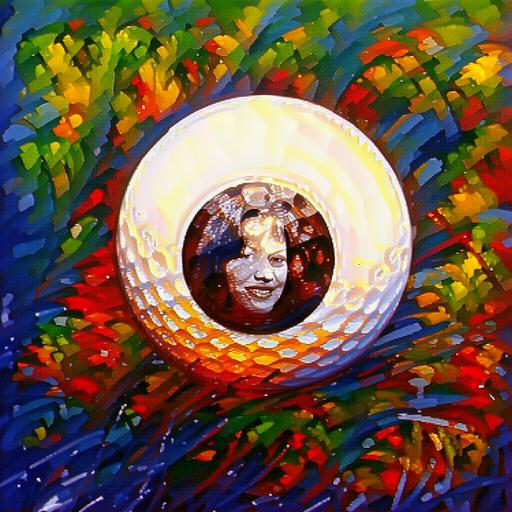}&&&
		\includegraphics[width=0.135\linewidth]{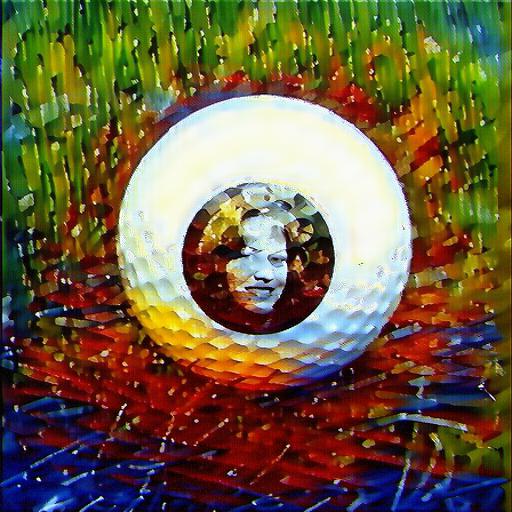}&
		\includegraphics[width=0.135\linewidth]{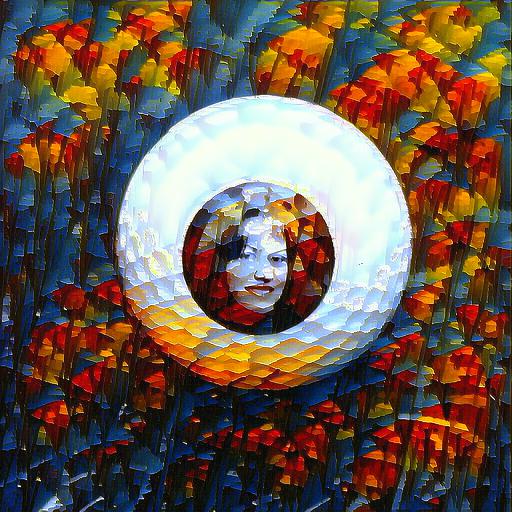}&
		\includegraphics[width=0.135\linewidth]{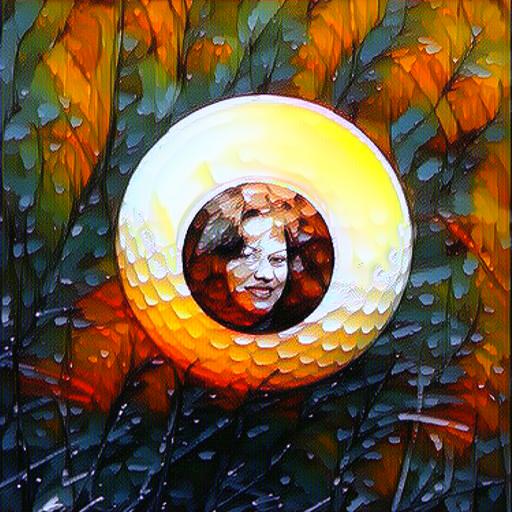}
		\\
		
		\footnotesize  Style & \footnotesize  Content &&& \footnotesize (a) ImageNet-VGG & \footnotesize (b) CLIP-ViT-B/16* &&& \footnotesize (c) CLIP-ViT-B/32 & \footnotesize (d) CLIP-RN50 & \footnotesize (e) CLIP-RN101
		
		\\
		\vspace{-0.3em}
		\\
		\hline
		\vspace{-0.3em}
		\\
		\multicolumn{2}{c}{ SSIM / CLIP Score:} &&&   0.541 / 0.602   & 0.672 / 0.741  &&&  0.591 / 0.705  &  0.529 / 0.677  &  0.582 / 0.719
		
	\end{tabular}
	\vspace{0.5em}
	\caption{ {\bf Ablation study} on different {\bf disentanglement space} (VGG vs. CLIP, columns (a-b)) and {\bf CLIP image encoders} (columns (b-e)). * denotes our default setting. Zoom-in for better comparison.}
	\label{fig:arch}
\end{figure*} 

{\bf Inference.} When we use the default setting $(S_{for}, S_{rev}) = (40, 6)$, the forward process takes around $4.921$ seconds and the reverse process takes around 0.691 seconds. Therefore, the total inference time is $4.921+0.691=5.612$ seconds. The inference process requires about 13GB of GPU memory.

Currently, {\em we have not optimized the model size and GPU memory consumption here}. We believe there is substantial room for improvement, and we would like to elaborate on that in future work.

\section{More Ablation Study and Analyses}
\label{sec:moreablation}

{\bf Quantitative Analyses of C-S Disentanglement.} Here, we provide more quantitative results to analyze the C-S Disentanglement achieved by our StyleDiffusion. As shown in Fig.~\ref{fig:disent_quantity}, we observe that the style is well disentangled from the content in the style image by adjusting the return step $T_{remov}$ of the style removal module. As such, when more style information is removed (blue line), it will be transferred to the corresponding stylized result (orange line). The quantitative analyses are consistent with the qualitative results displayed in Fig.~\ref{fig:disent} of our main paper.

{\bf Quantitative Analyses of C-S Trade-off.} We also provide more quantitative results to analyze the C-S trade-off achieved by our StyleDiffusion. As shown in Fig.~\ref{fig:tradeoff-train} and Fig.~\ref{fig:tradeoff-test}, we can flexibly control the C-S trade-off at both the training stage (Fig.~\ref{fig:tradeoff-train}) and the testing stage (Fig.~\ref{fig:tradeoff-test}) by adjusting the return step $T_{trans}$ of diffusion models. The sweet spot areas are highlighted in the figures, which are the most probable for obtaining satisfactory results. Overall, the quantitative analyses are consistent with the qualitative results displayed in Fig.~\ref{fig:tradeoff} of our main paper.

{\bf CLIP Space vs. VGG Space.} As discussed in our main paper, we leverage the open-domain CLIP~\cite{radford2021learning} space to formulate the style disentanglement. The pre-trained CLIP space integrates rich cross-domain image (and supplementarily, text) knowledge and thus can measure the ``style distance'' more accurately. As shown in Fig.~\ref{fig:arch}~(a-b), we compare it with the ImageNet~\cite{russakovsky2015imagenet} pre-trained VGG-19~\cite{simonyan2014very}, which has been widely adopted in prior arts~\cite{gatys2016image,huang2017arbitrary} to extract the style information. As is evident, using the CLIP space recovers the style information more sufficiently and realistically, significantly outperforming the VGG space (which is also validated by the bottom quantitative scores). It may be attributed to the fact that the VGG is pre-trained on ImageNet and therefore lacks a sufficient understanding of artistic styles. In contrast, the CLIP space encapsulates a myriad of knowledge of not only the photograph domain but also the artistic domain, which is more powerful in depicting the style of an image. Besides, it is worth noting that the CLIP space naturally provides multi-modal compatibility, which can facilitate users to control the style transfer with multi-modal signals, \eg, image and text (see later Sec.~\ref{sec:extension}).

{\bf Different CLIP Image Encoders.} We also investigate the effects of different CLIP~\cite{radford2021learning} image encoders to conduct the style disentanglement. As shown in Fig.~\ref{fig:arch}~(b-e), in general, ViTs~\cite{dosovitskiy2020image} achieve better visual results than ResNets (RN)~\cite{he2016deep}, \eg, the brushstrokes are more natural in the top row, and the colors are more vivid in the bottom row. And ViT-B/16 performs better than ViT-B/32 in capturing more fine-grained styles. Interestingly, our findings coincide with the reported performance of these image encoders on high-level vision tasks (\eg, classification) in the original CLIP paper~\cite{radford2021learning}. It indicates that our stylization performance is closely related to the high-level semantic representations learned by the image encoder, which also gives evidence to the correlations between high-level vision tasks and low-level vision tasks. 

\begin{figure*}
	\centering
	\setlength{\tabcolsep}{0.1cm}
	\renewcommand\arraystretch{0.8}
	
	\begin{tabular}{ccccp{0.1em}|p{0.1em}c}
		
		\includegraphics[width=0.14\linewidth]{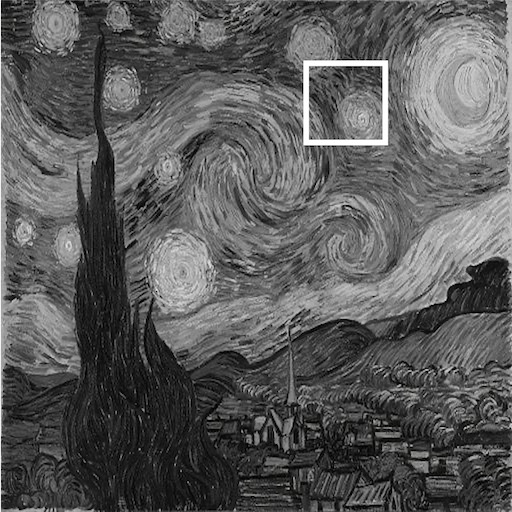}  &&& &&&
		\includegraphics[width=0.14\linewidth]{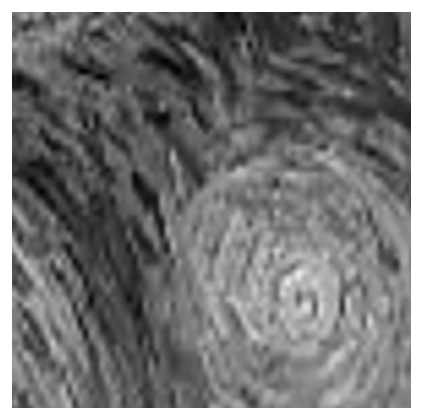}
		
		\\
		Input   &&&& && Input
		\\
		
		\includegraphics[width=0.14\linewidth]{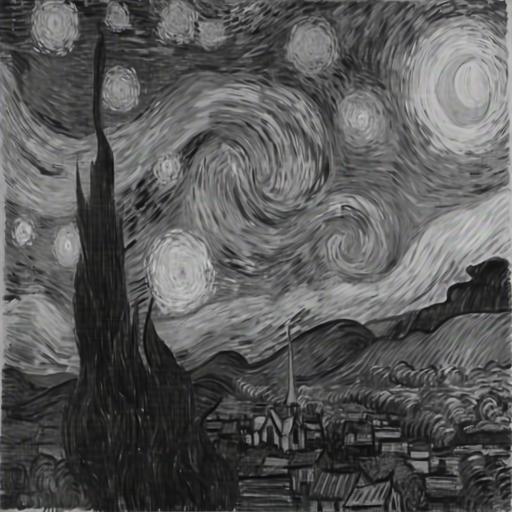}&
		\includegraphics[width=0.14\linewidth]{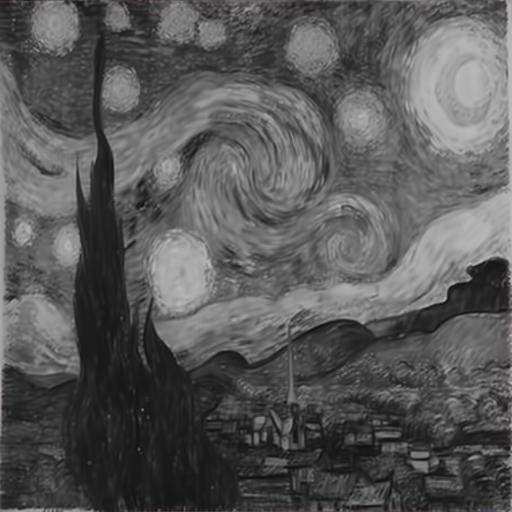}&
		\includegraphics[width=0.14\linewidth]{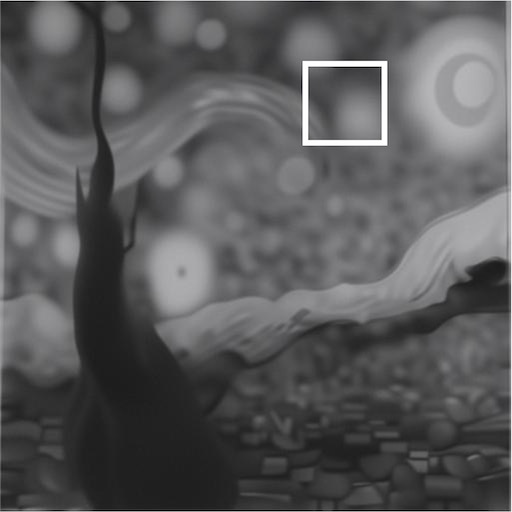}&
		\includegraphics[width=0.14\linewidth]{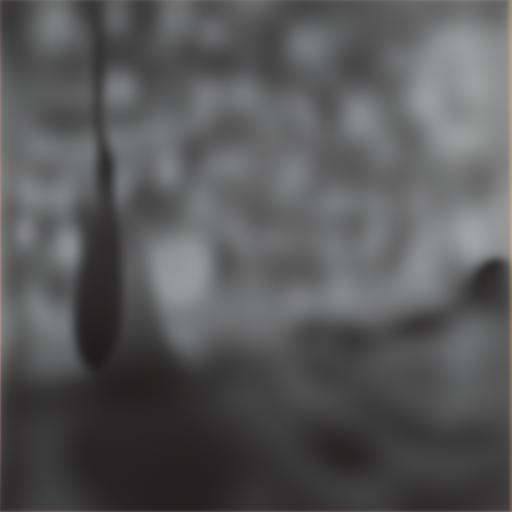} &&&
		\includegraphics[width=0.14\linewidth]{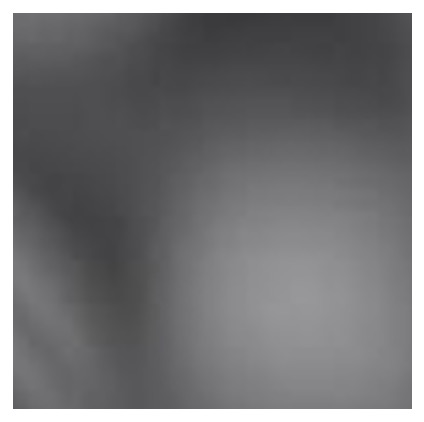}
		\\
		\multicolumn{4}{c}{Diffusion-based: $T_{remov}$ = {201, 401, 601, 801} } &&& $T_{remov}$ = 601
		\\
		
		\includegraphics[width=0.14\linewidth]{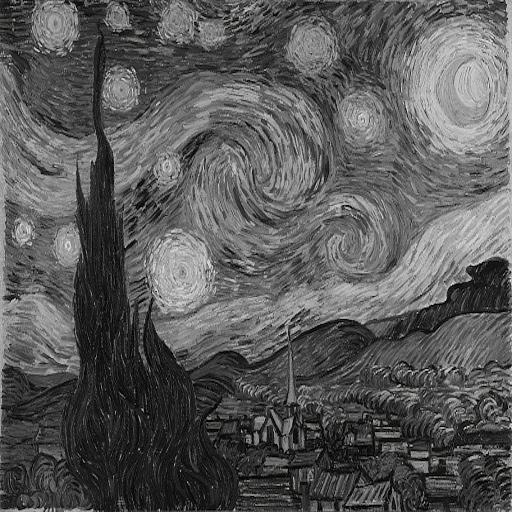}&
		\includegraphics[width=0.14\linewidth]{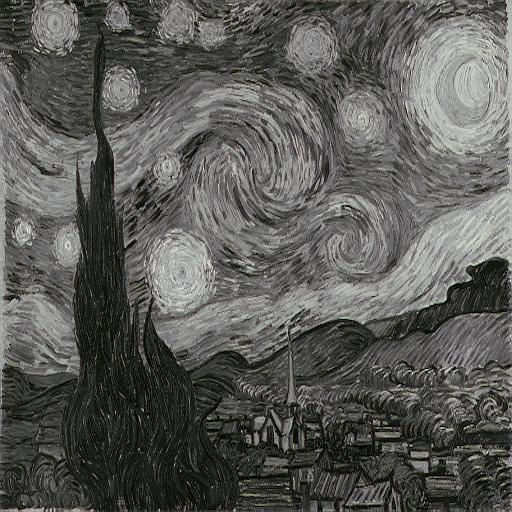}&
		\includegraphics[width=0.14\linewidth]{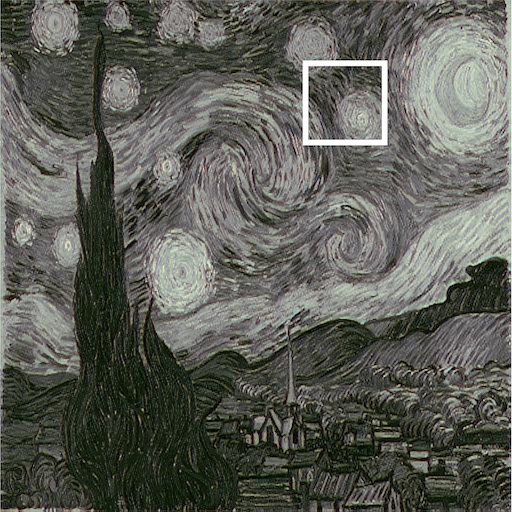}&
		\includegraphics[width=0.14\linewidth]{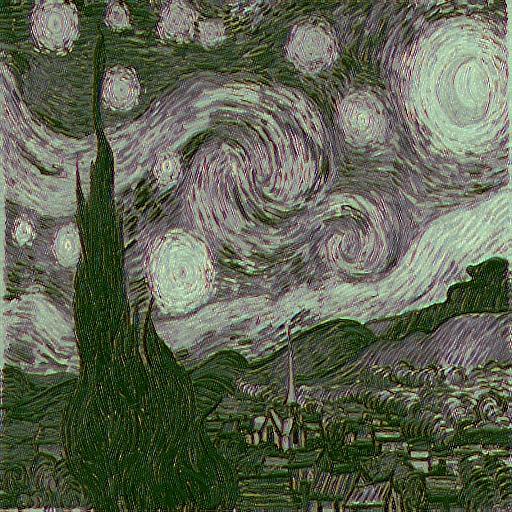} &&&
		\includegraphics[width=0.14\linewidth]{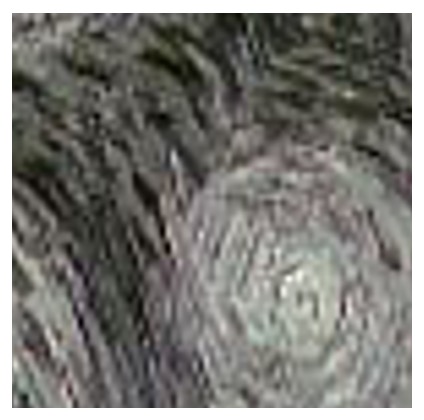}
		\\
		\multicolumn{4}{c}{AE1-based: $Iter$ = {1, 5, 10, 20} } &&& $Iter$ = 10
		\\
		
		\includegraphics[width=0.14\linewidth]{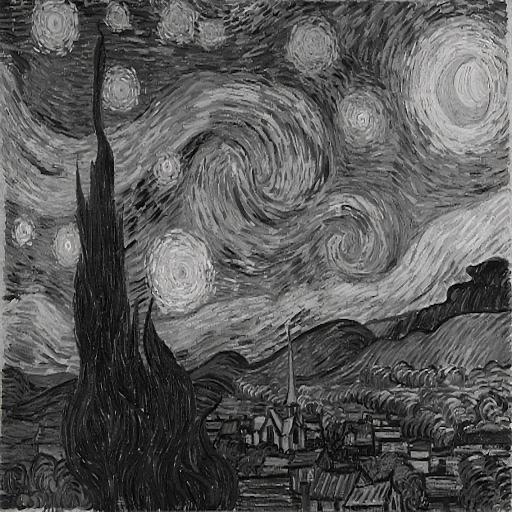}&
		\includegraphics[width=0.14\linewidth]{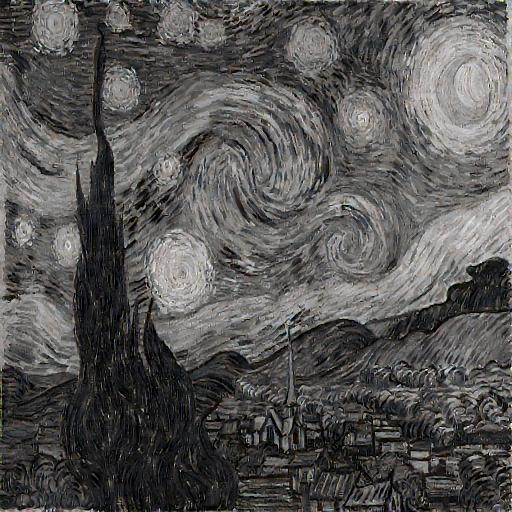}&
		\includegraphics[width=0.14\linewidth]{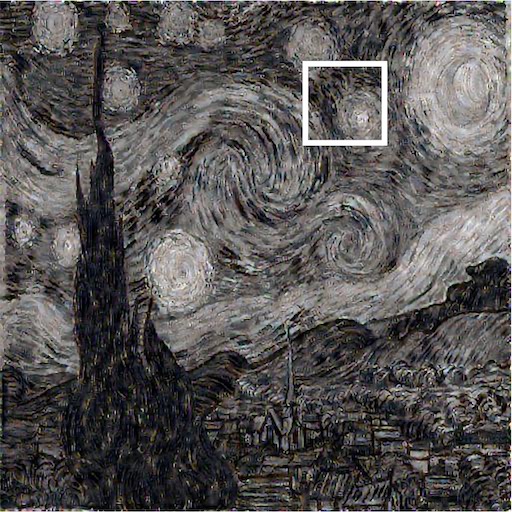}&
		\includegraphics[width=0.14\linewidth]{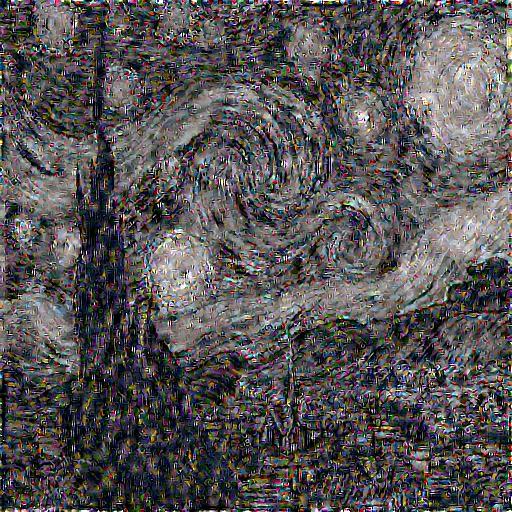} &&&
		\includegraphics[width=0.14\linewidth]{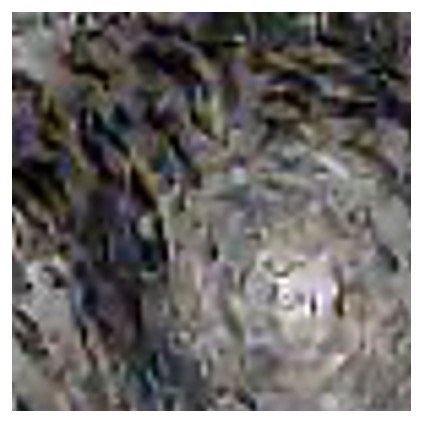}
		\\
		\multicolumn{4}{c}{AE2-based: $Iter$ = {1, 5, 10, 20} } &&& $Iter$ = 10
		\\
		
		\includegraphics[width=0.14\linewidth]{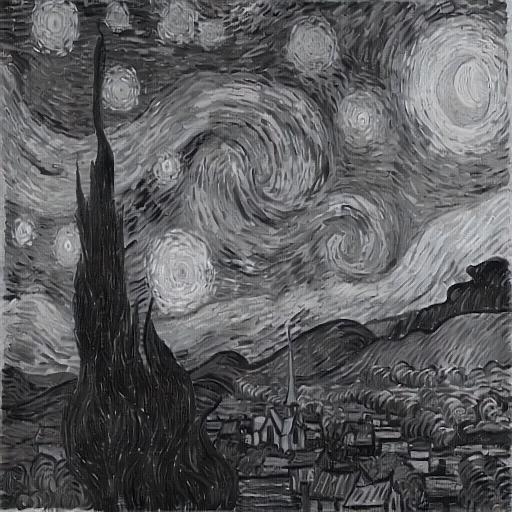}&
		\includegraphics[width=0.14\linewidth]{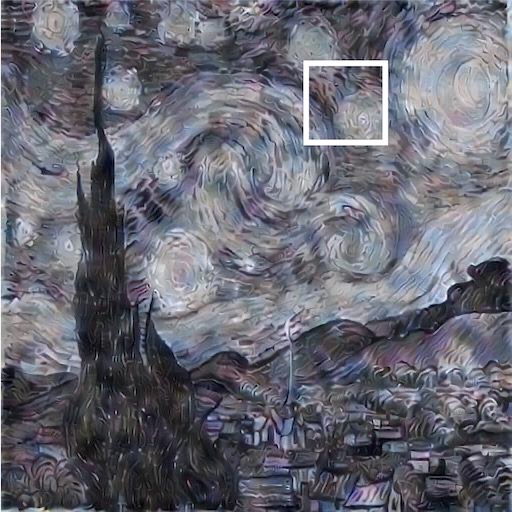}&
		\includegraphics[width=0.14\linewidth]{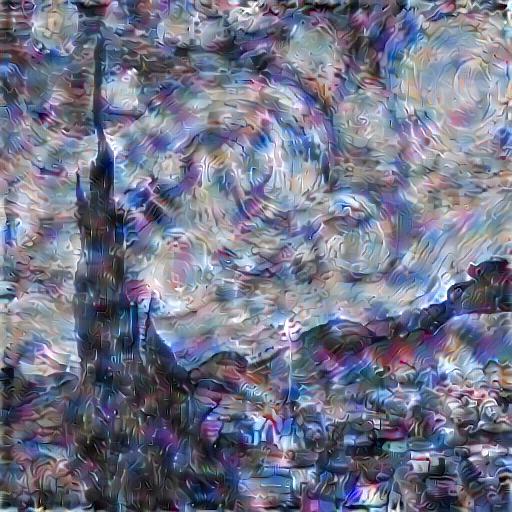}&
		\includegraphics[width=0.14\linewidth]{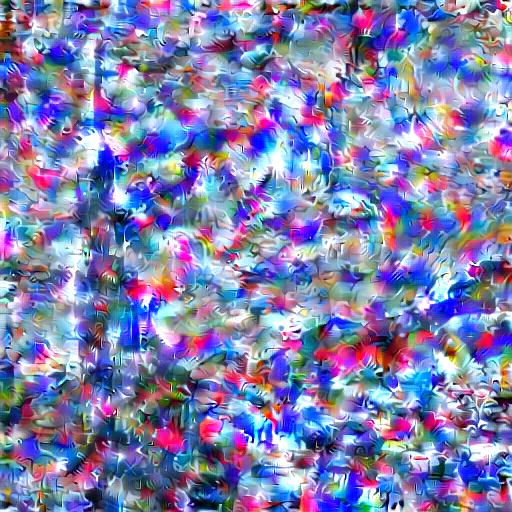} &&&
		\includegraphics[width=0.14\linewidth]{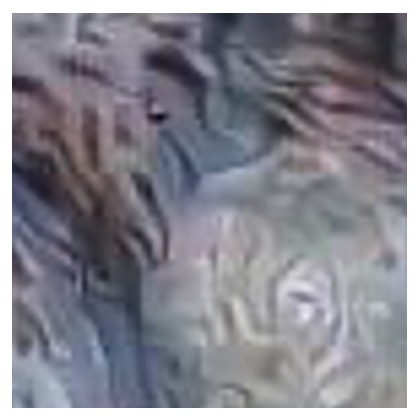}
		\\
		\multicolumn{4}{c}{AE3-based: $Iter$ = {1, 5, 10, 20} }  &&& $Iter$ = 5
		\\
		
		\includegraphics[width=0.14\linewidth]{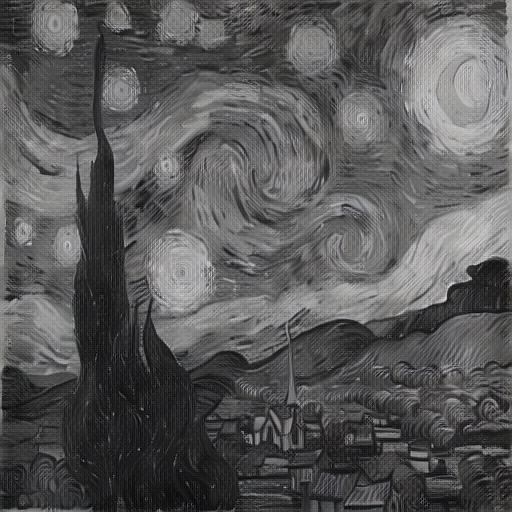}&
		\includegraphics[width=0.14\linewidth]{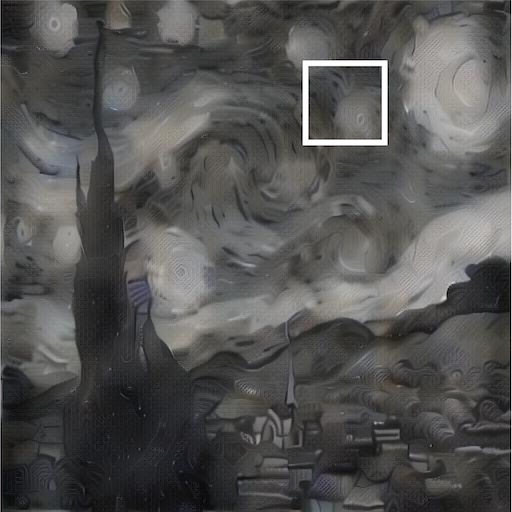}&
		\includegraphics[width=0.14\linewidth]{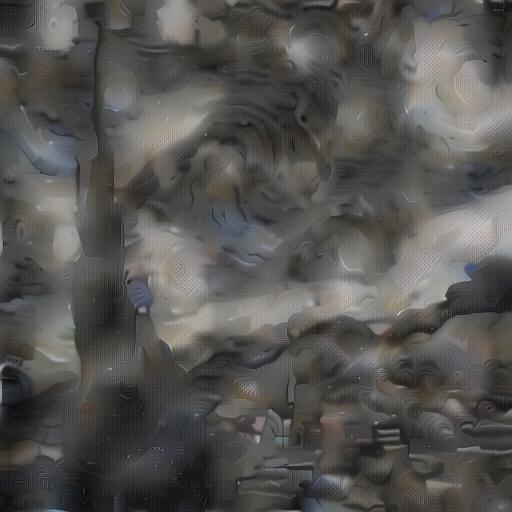}&
		\includegraphics[width=0.14\linewidth]{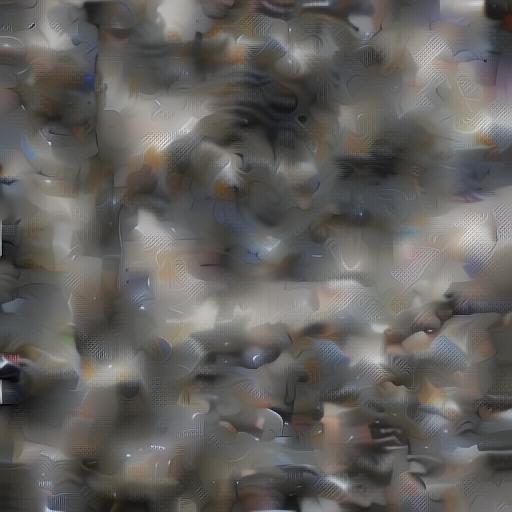} &&&
		\includegraphics[width=0.14\linewidth]{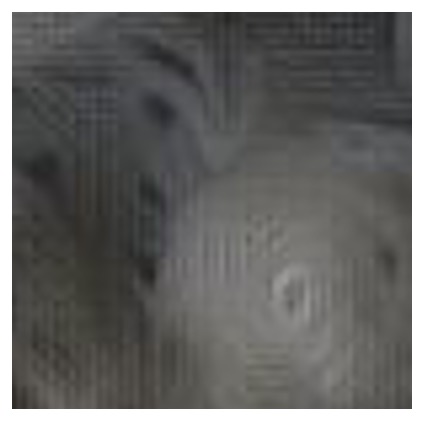}
		\\
		\multicolumn{4}{c}{AE4-based: $Iter$ = {1, 5, 10, 20} } &&& $Iter$ = 5
		
		\\
		
		\includegraphics[width=0.14\linewidth]{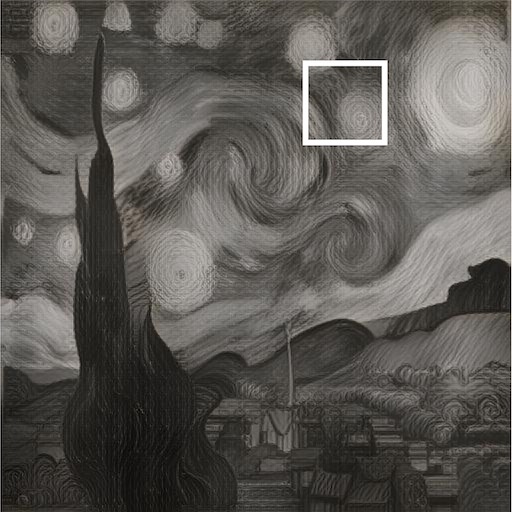}&
		\includegraphics[width=0.14\linewidth]{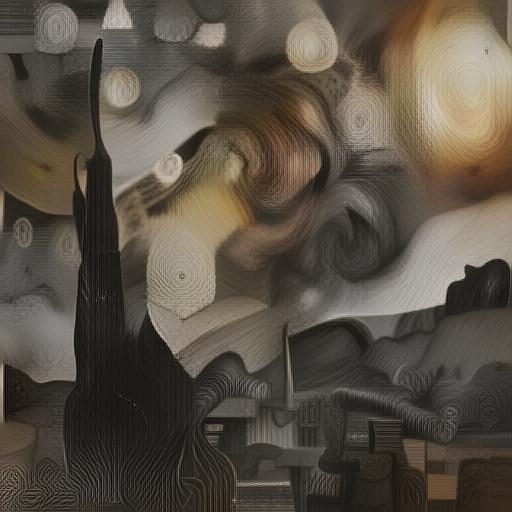}&
		\includegraphics[width=0.14\linewidth]{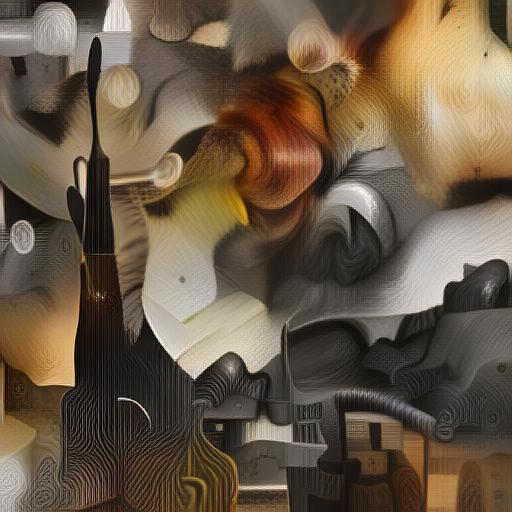}&
		\includegraphics[width=0.14\linewidth]{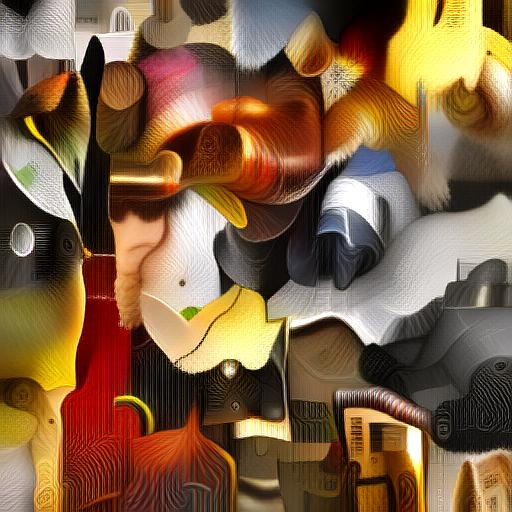} &&&
		\includegraphics[width=0.14\linewidth]{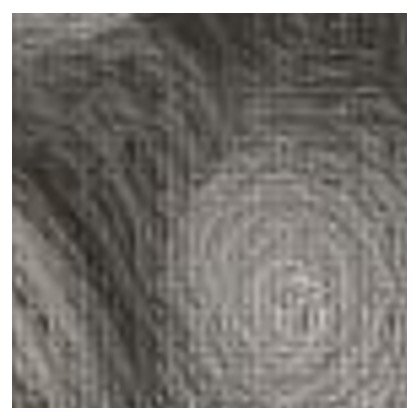}
		\\
		\multicolumn{4}{c}{AE5-based: $Iter$ = {1, 5, 10, 20} } &&& $Iter$ = 1
		\\

	\end{tabular}
	\vspace{0.5em}
	\caption{ {\bf Diffusion-based style removal vs. AE-based style removal.} The last column shows the enlarged areas of the corresponding best style removed results manually selected in each row. As can be observed, diffusion-based style removal can better remove the detailed style of the style image while preserving the main content structures. In contrast, AE-based style removal cannot plausibly remove the detailed style and often introduces color noises/artifacts and destroys the content structures.}
	\label{fig:style_removal}
\end{figure*}

\begin{figure*}[tbhp]
	\centering
	\setlength{\tabcolsep}{0.02cm}
	\renewcommand\arraystretch{0.4}
	
	\begin{tabular}{ccccc}
		
		\includegraphics[width=0.195\linewidth]{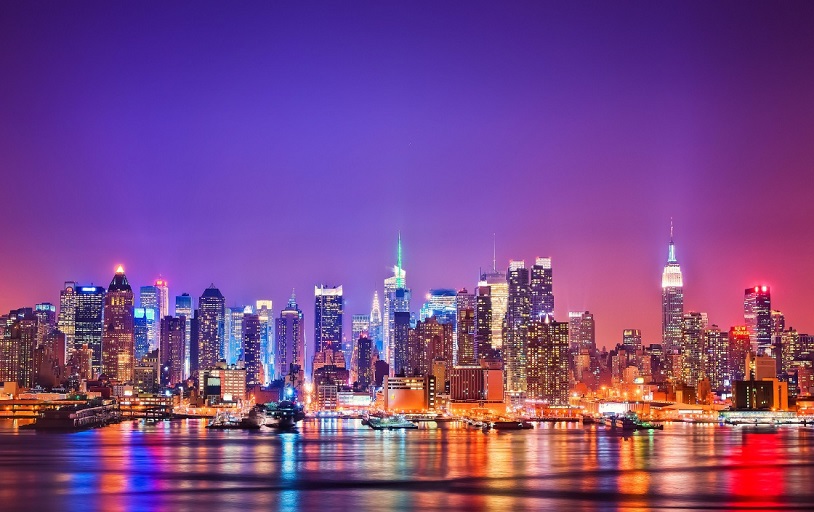}&
		\includegraphics[width=0.195\linewidth]{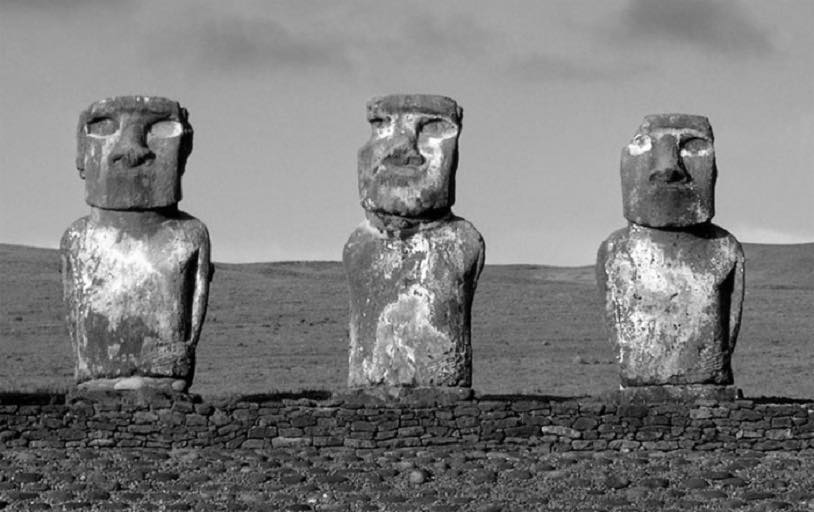}&
		\includegraphics[width=0.195\linewidth]{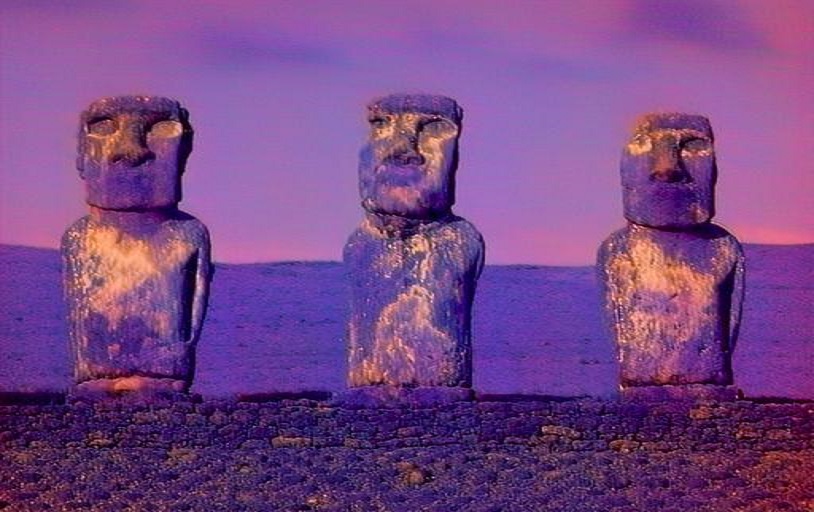}&
		\includegraphics[width=0.195\linewidth]{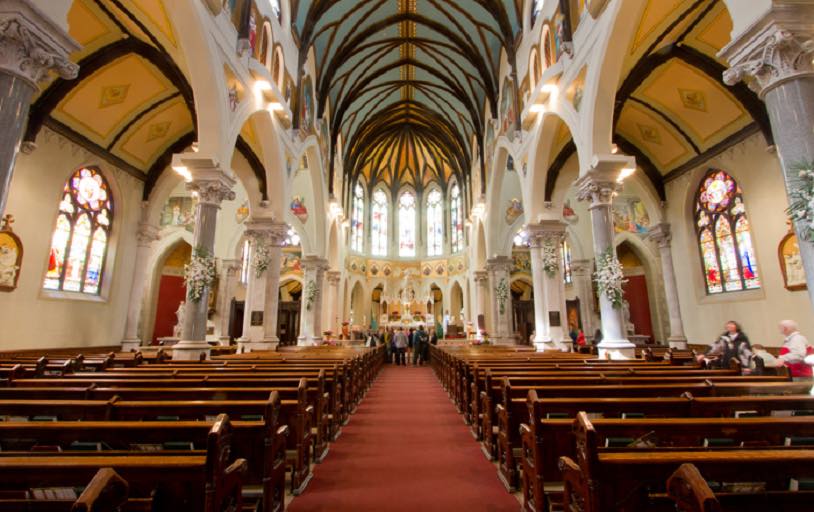}&
		\includegraphics[width=0.195\linewidth]{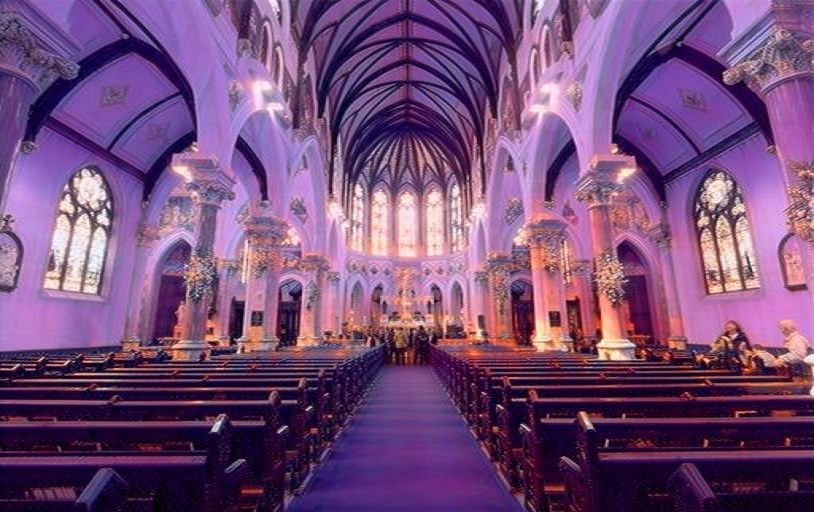}
		
		\\
		\includegraphics[width=0.195\linewidth]{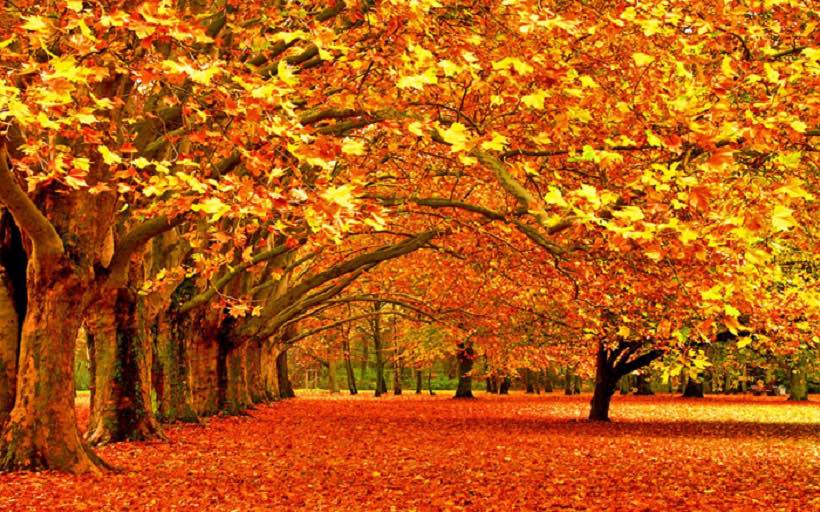}&
		\includegraphics[width=0.195\linewidth]{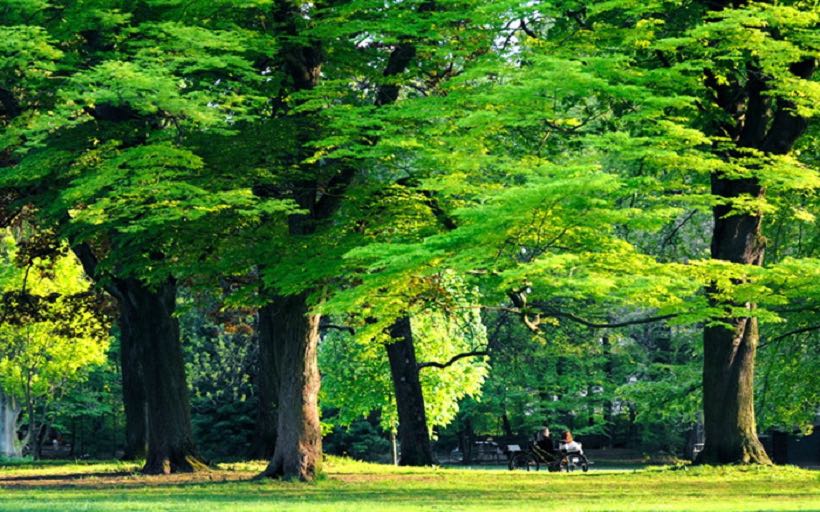}&
		\includegraphics[width=0.195\linewidth]{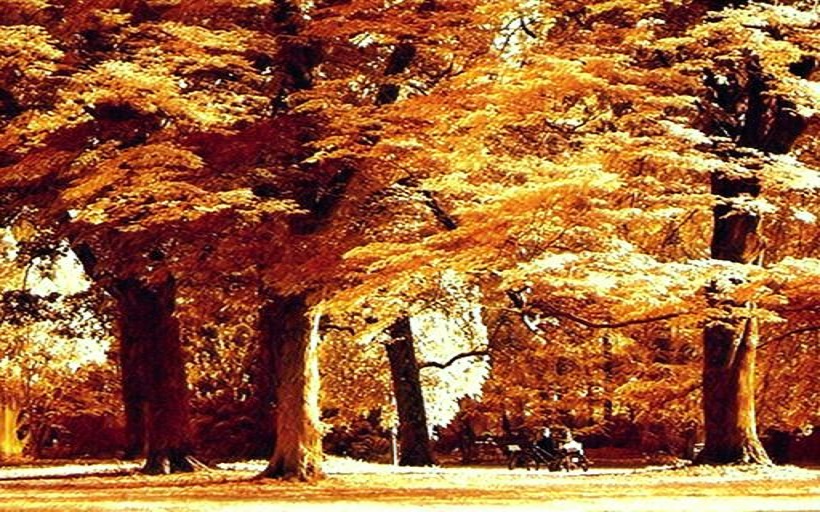}&
		\includegraphics[width=0.195\linewidth]{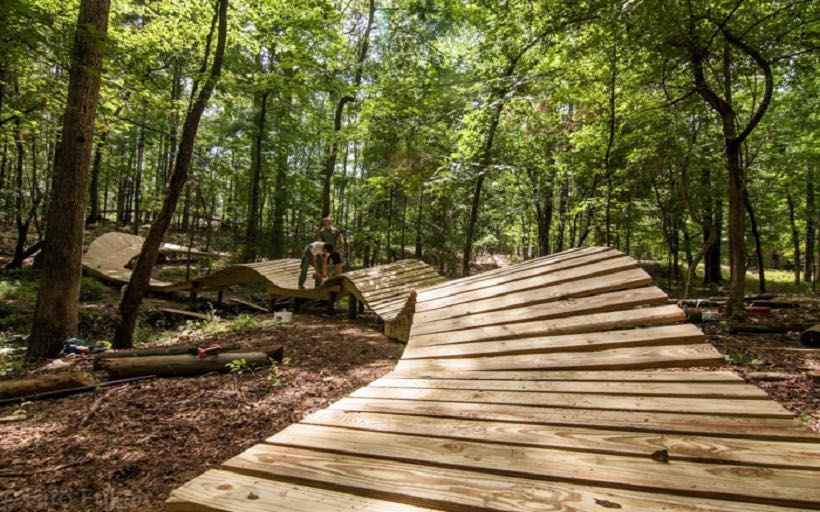}&
		\includegraphics[width=0.195\linewidth]{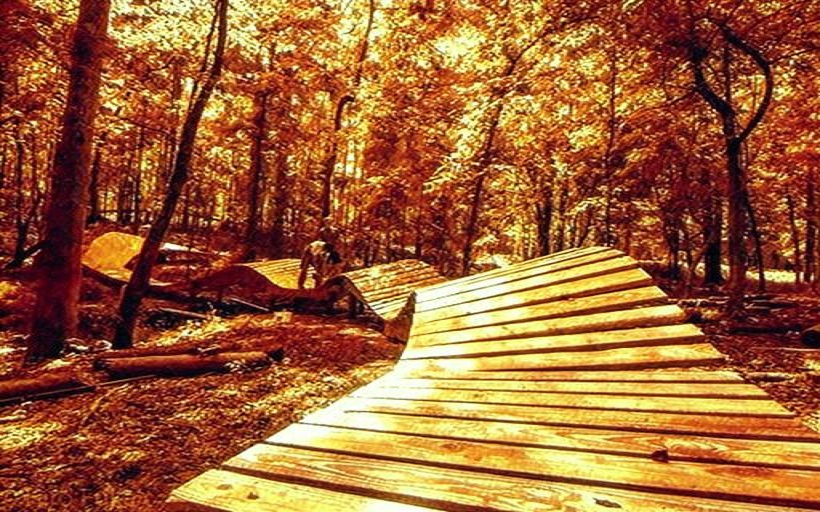}
		
		\\
		Style & Content 1 & Result 1 & Content 2 & Result 2

	\end{tabular}
	\vspace{0.5em}
	\caption{ {\bf Photo-realistic style transfer} achieved by our StyleDiffusion. We set $T_{remov}=401$ and $T_{trans}=101$ for this task.}
	\label{fig:photoreal}
\end{figure*}

\begin{figure*}[tbhp]
	\centering
	\setlength{\tabcolsep}{0.02cm}
	\renewcommand\arraystretch{0.4}
	
	\begin{tabular}{cccp{0.05em}|p{0.05em}cccc}
		\includegraphics[width=0.14\linewidth]{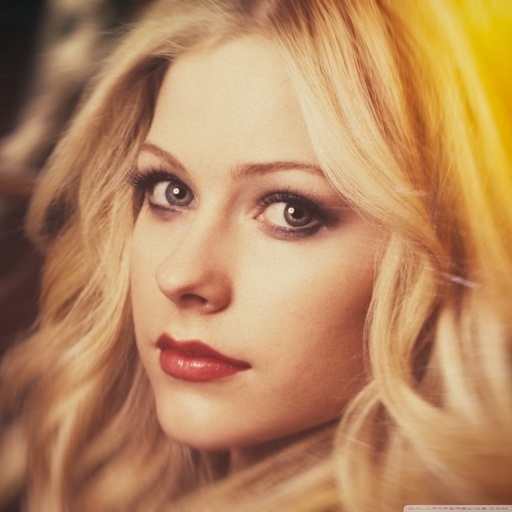}&
		\includegraphics[width=0.14\linewidth]{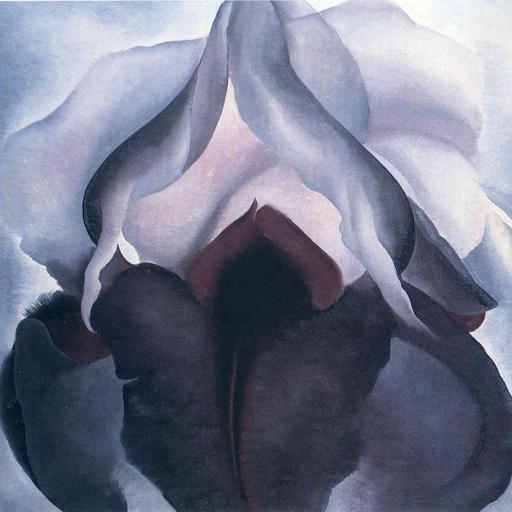}&
		\includegraphics[width=0.14\linewidth]{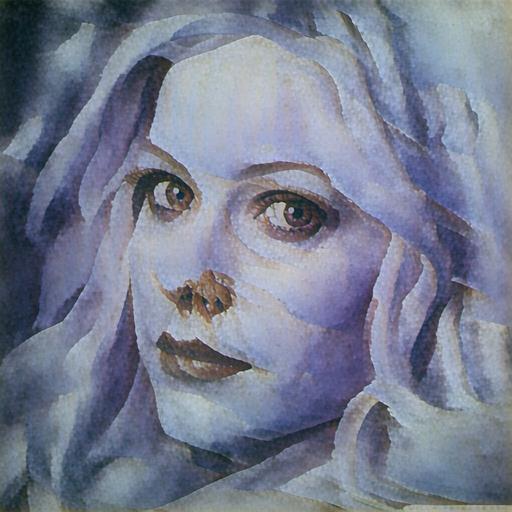}&&&
		\includegraphics[width=0.14\linewidth]{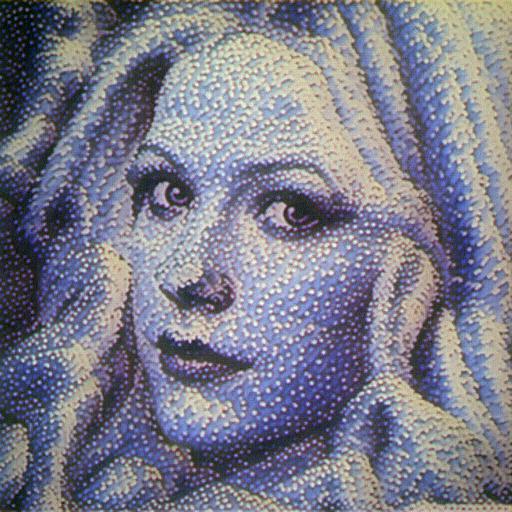}&
		\includegraphics[width=0.14\linewidth]{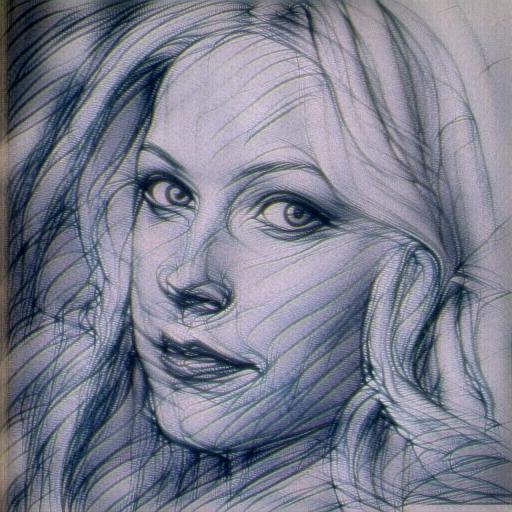}&
		\includegraphics[width=0.14\linewidth]{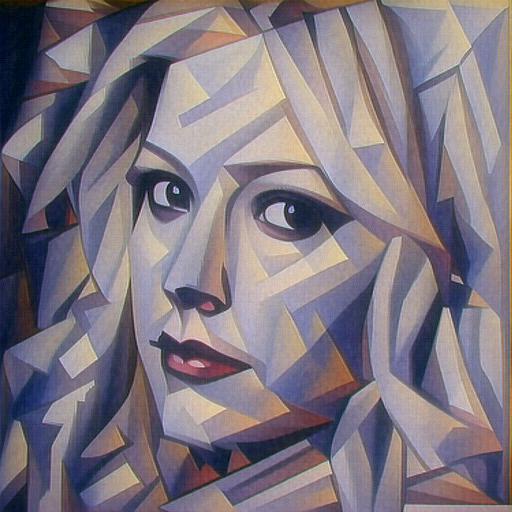}&
		\includegraphics[width=0.14\linewidth]{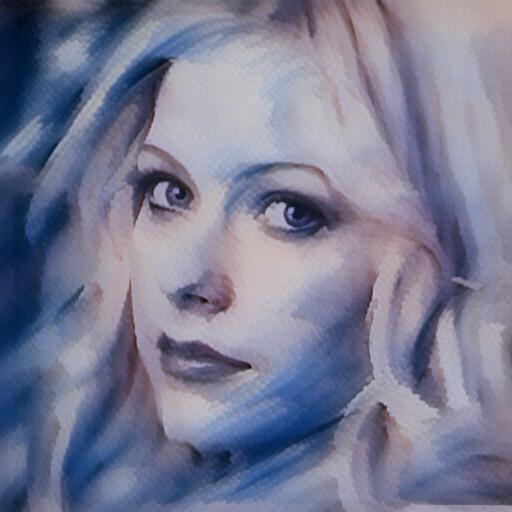}
		
		\\
		
		Content & Style & Result &&& + ``Pointillism'' & + ``Sketch'' & + ``Cubism'' & + ``Watercolor''

	\end{tabular}
	\vspace{0.5em}
	\caption{ {\bf Multi-modal style manipulation.} Our framework is compatible with image and text modulation signals, which provides users with a more flexible way to manipulate the style of images.}
	\label{fig:multimodal}
\end{figure*}

{\bf Diffusion-based Style Removal vs. AE-based Style Removal.} To demonstrate the superiority of diffusion-based style removal, we compare it with a possible alternative, \ie, Auto-Encoders (AEs), since one may argue that the diffusion model is a special kind of (Variational) Auto-Encoder network~\cite{luo2022understanding}. We directly use the AEs released by Li~\etal~\cite{li2017universal}, which employ the VGG-19 network~\cite{simonyan2014very} as the encoders, fix them and train decoder networks for inverting VGG features to the original images. They select feature maps at five layers of the VGG-19, i.e., Relu\_X\_1 (X=1,2,3,4,5), and train five decoders accordingly, which we denote as AEX (X=1,2,3,4,5) in the following. When used for style removal, we iteratively perform the encoding and decoding processes of AEs for the input images. The comparison results are shown in Fig.~\ref{fig:style_removal}. As can be observed in the bottom five rows, AE-based style removal cannot plausibly remove the detailed style and often introduces color noises/artifacts and destroys the content structures, which is undesirable for style removal. By contrast, diffusion-based style removal can smoothly remove the style details while preserving the main content structures, significantly outperforming AE-based style removal.

\begin{figure*}[t!]
	\centering
	\setlength{\tabcolsep}{0.02cm}
	\renewcommand\arraystretch{0.4}
	
	\begin{tabular}{ccp{0.05em}|p{0.05em}ccccc}
		\multirow{1}{*}[1.2in]{\includegraphics[width=0.14\linewidth]{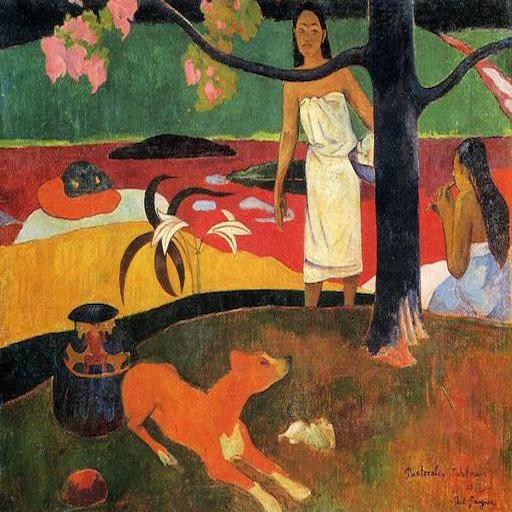}}&
		\includegraphics[width=0.14\linewidth]{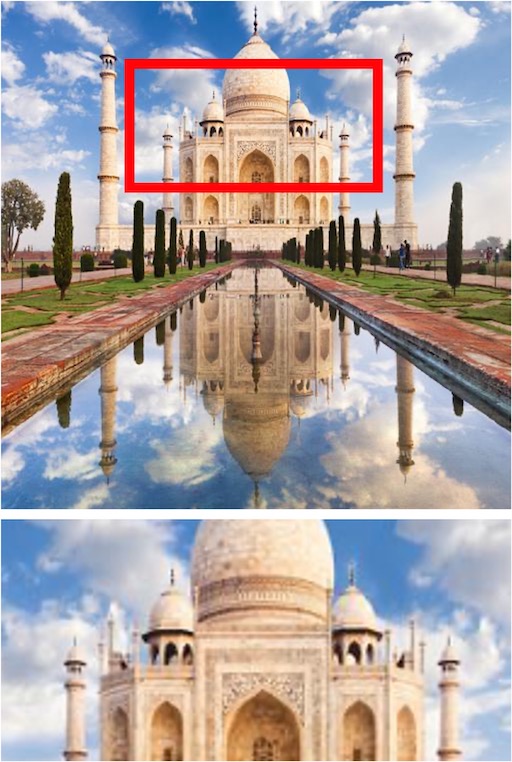}&&&
		\includegraphics[width=0.14\linewidth]{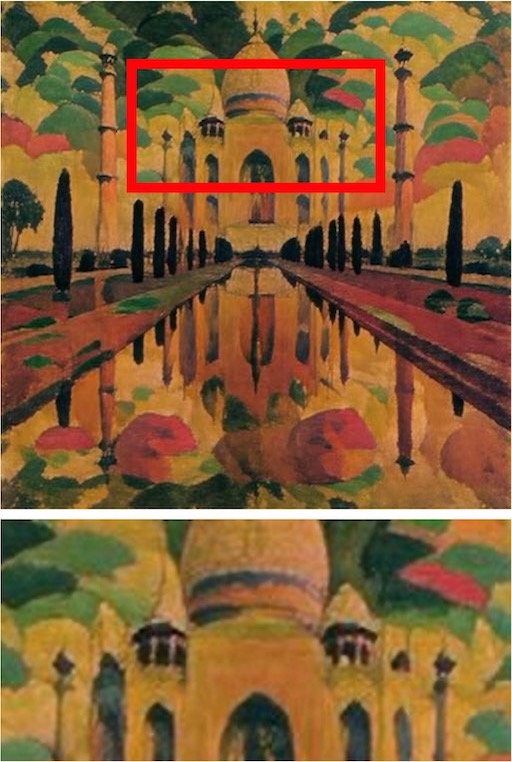}&
		\includegraphics[width=0.14\linewidth]{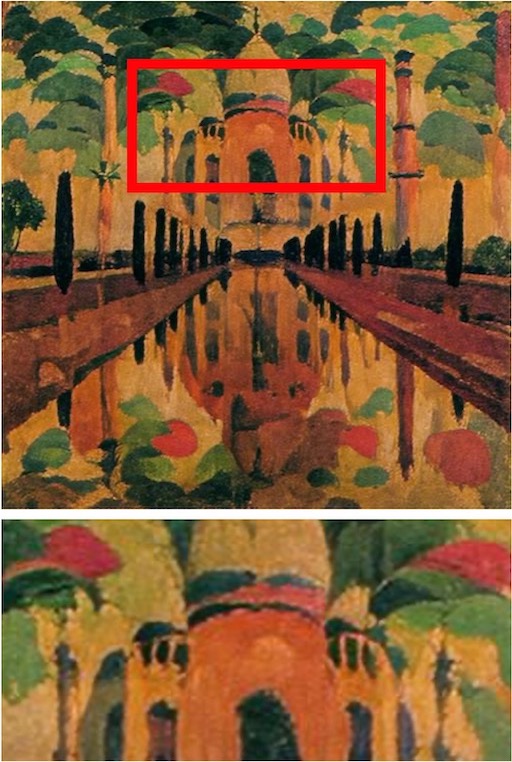}&
		\includegraphics[width=0.14\linewidth]{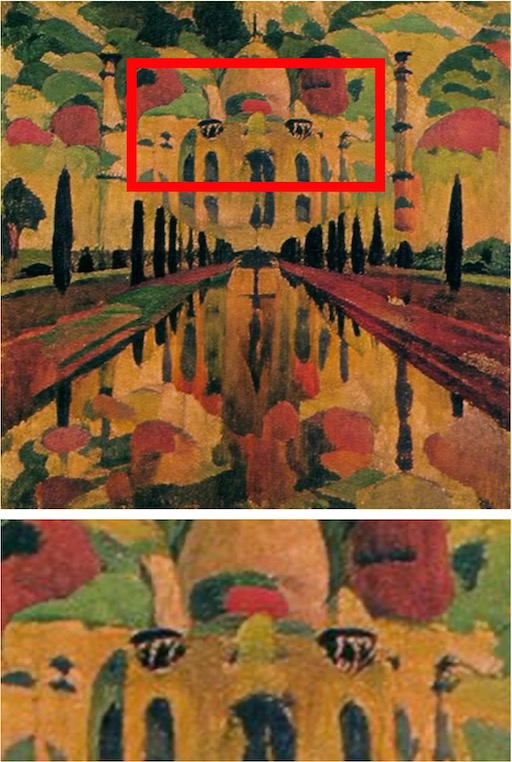}&
		\includegraphics[width=0.14\linewidth]{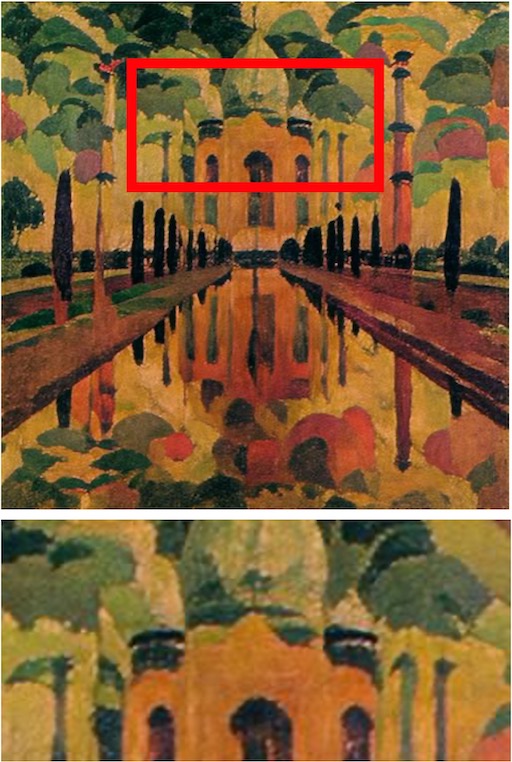}&
		\multirow{1}{*}[1.2in]{\animategraphics[width=0.14\linewidth, autoplay=False]{1.1}{figs_SM/div/1/div/}{1}{10}}
		
		\\
		
		Style & Content &&& Sample 1 & Sample 2 & Sample 3 & Sample 4 & Animation

	\end{tabular}
	\vspace{0.5em}
	\caption{ {\bf Diversified style transfer.} Our framework can easily achieve diversified style transfer during inference by directly adopting the stochastic DDPM~\cite{ho2020denoising} forward process. Click on the last image to see animation using Adobe Reader.}
	\label{fig:div}
\end{figure*}

\section{Extensions}
\label{sec:extension}

{\bf Photo-realistic Style Transfer.}
Our StyleDiffusion successfully separates style from content in a controllable manner. Thus, it can easily achieve photo-realistic style transfer~\cite{luan2017deep} by adjusting the content extraction of the style removal module. Specifically, since the style of a photo is mainly reflected by the low-level and high-frequency features such as colors and brightness, we reduce $T_{remov}$ to a relatively smaller value, \eg, 401. Moreover, to better preserve the content structures, we adjust the style transfer process and reduce $T_{trans}$ to 101. We show some photo-realistic style transfer results synthesized by our StyleDiffusion in Fig.~\ref{fig:photoreal}.

{\bf Multi-modal Style Manipulation.} As our framework leverages the open-domain CLIP~\cite{radford2021learning} space to measure the ``style distance'', it is naturally compatible with image and text modulation signals. By adding a directional CLIP loss term~\cite{gal2022stylegan,kim2022diffusionclip} to our total loss, our framework can easily achieve multi-modal style manipulation, as shown in Fig.~\ref{fig:multimodal}. {\em As far as we know, our framework is the first unified framework to achieve both image and text guided style transfer.}

{\bf Diversified Style Transfer.} In the fine-tuning, our style transfer module adopts the deterministic DDIM~\cite{song2020denoising} forward and reverse processes (Eq.~(\ref{eq:ddim}) and Eq.~(\ref{eq:ddim_ode}) in the main paper). However, during inference, we can directly replace the deterministic DDIM forward process with the stochastic DDPM~\cite{ho2020denoising} forward process (Eq.~(\ref{eq:ddpmf}) in the main paper) to achieve diversified style transfer~\cite{wang2020diversified}, as shown in Fig.~\ref{fig:div}. The users can easily trade off the diversity and quality by adjusting the return step or iteration of the DDPM forward process. The diverse results can give users endless choices to obtain more satisfactory results~\cite{wang2020diversified,wang2022divswapper}.

\section{More Comparison Results}
\label{sec:cmp}
\begin{figure*}
	\centering
	\setlength{\tabcolsep}{0.02cm}
	\renewcommand\arraystretch{0.6}
	\begin{tabular}{cp{0.1em}|p{0.1em}cccccc}
		
		\includegraphics[width=0.14\linewidth]{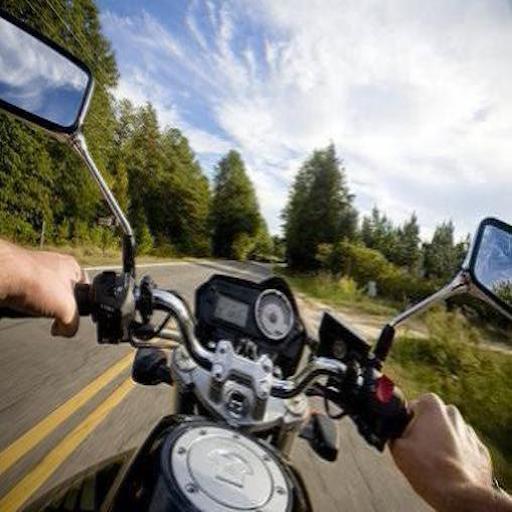}& &&
		\includegraphics[width=0.14\linewidth]{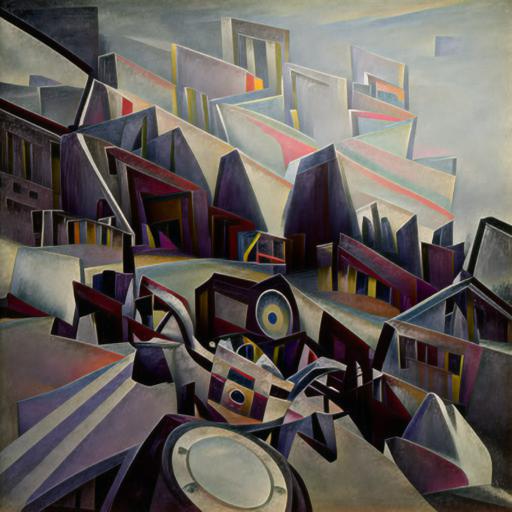}&
		\includegraphics[width=0.14\linewidth]{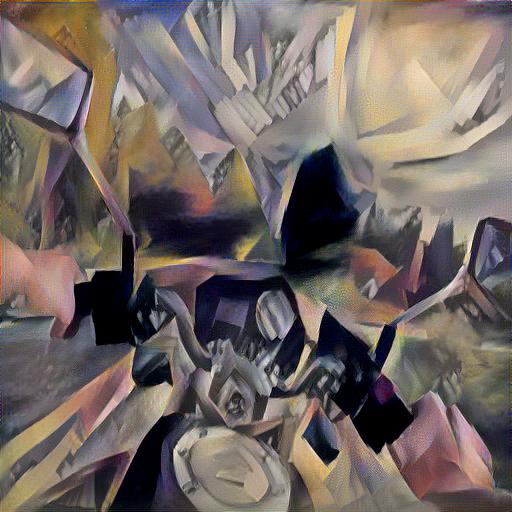}&
		\includegraphics[width=0.14\linewidth]{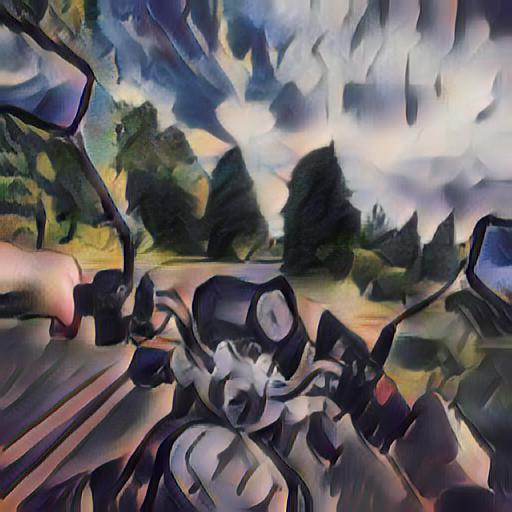}&
		\includegraphics[width=0.14\linewidth]{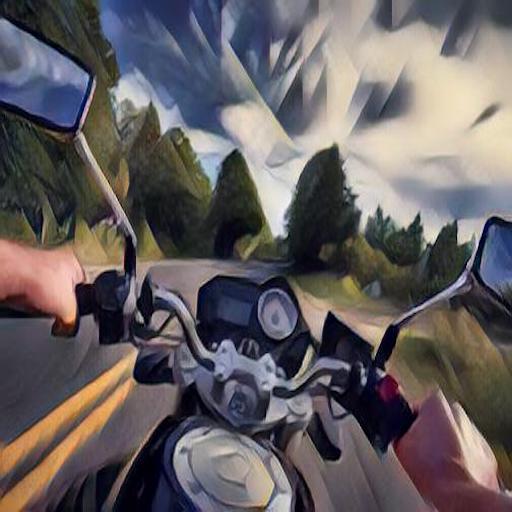}&
		\includegraphics[width=0.14\linewidth]{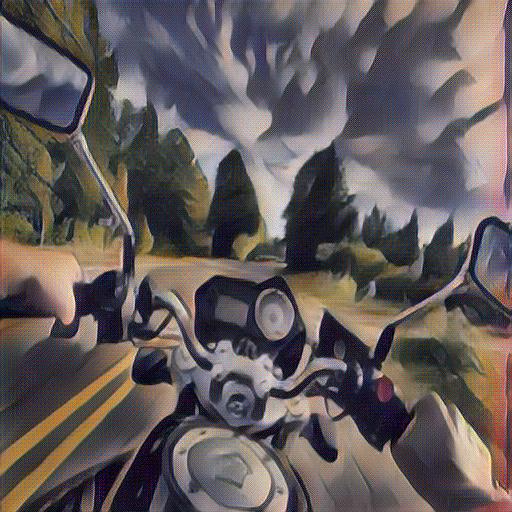}&
		\includegraphics[width=0.14\linewidth]{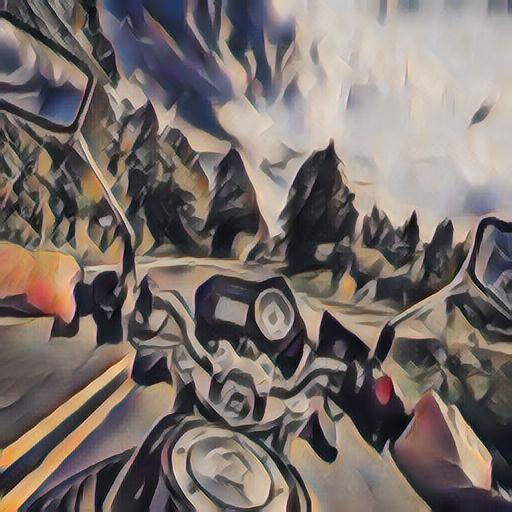}
		\\
		Content &&& \bf Ours & Gatys~\cite{gatys2016image} & EFDM~\cite{zhang2022exact} & StyTR$^2$~\cite{deng2022stytr2} & ArtFlow~\cite{an2021artflow} & AdaAttN~\cite{liu2021adaattn}
		\\
		\includegraphics[width=0.14\linewidth]{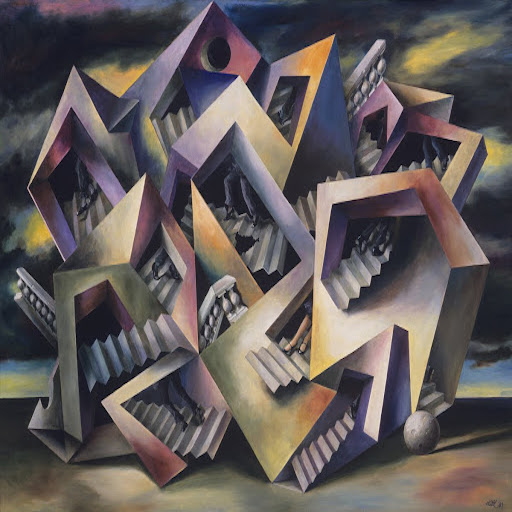}& &&
		
		\includegraphics[width=0.14\linewidth]{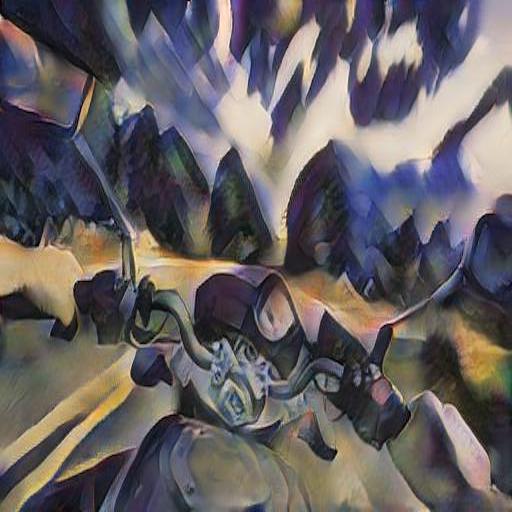}&
		\includegraphics[width=0.14\linewidth]{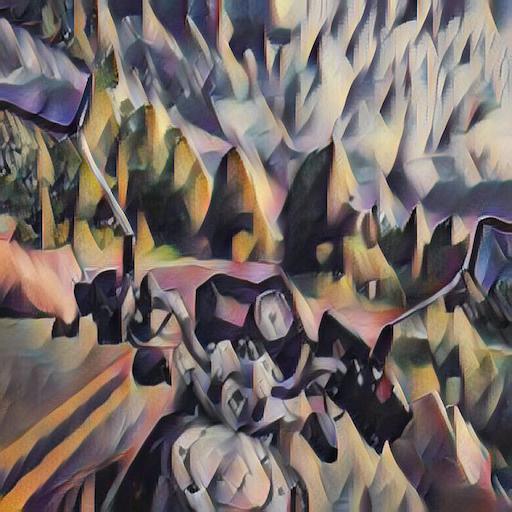}&
		\includegraphics[width=0.14\linewidth]{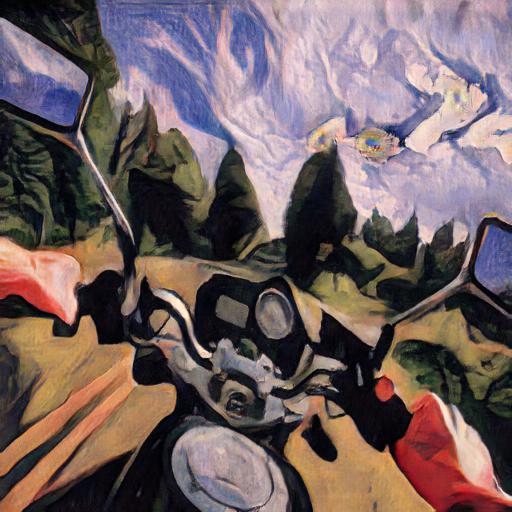}&
		\includegraphics[width=0.14\linewidth]{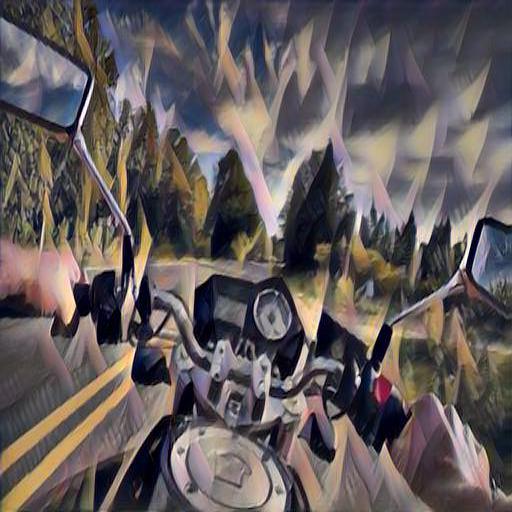}&
		\includegraphics[width=0.14\linewidth]{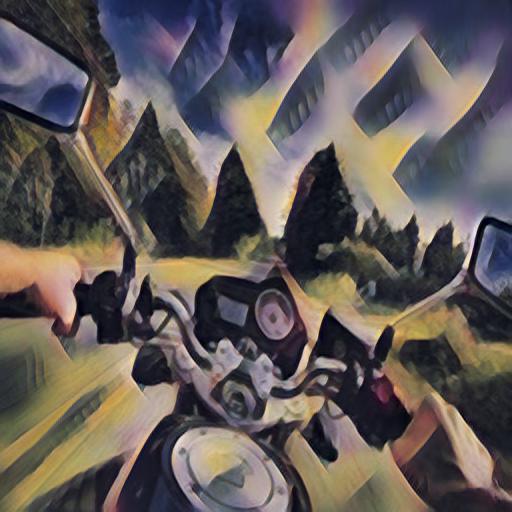}
		\\
		Style&&&  IECAST~\cite{chen2021artistic} & MAST~\cite{deng2020arbitrary} & TPFR~\cite{svoboda2020two} & Johnson~\cite{johnson2016perceptual} & LapStyle~\cite{lin2021drafting} 
		
		\\
		\vspace{0.1cm}
		\\
		\\
		\vspace{0.1cm}
		\\
		
		\includegraphics[width=0.14\linewidth]{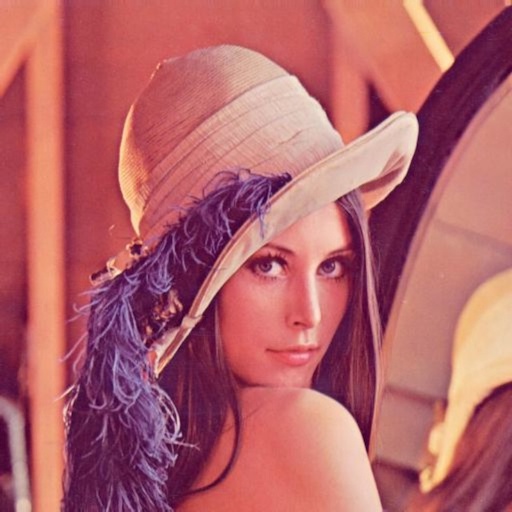}& &&
		\includegraphics[width=0.14\linewidth]{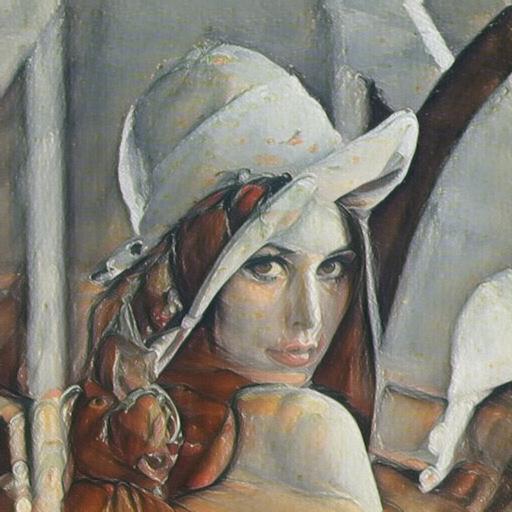}&
		\includegraphics[width=0.14\linewidth]{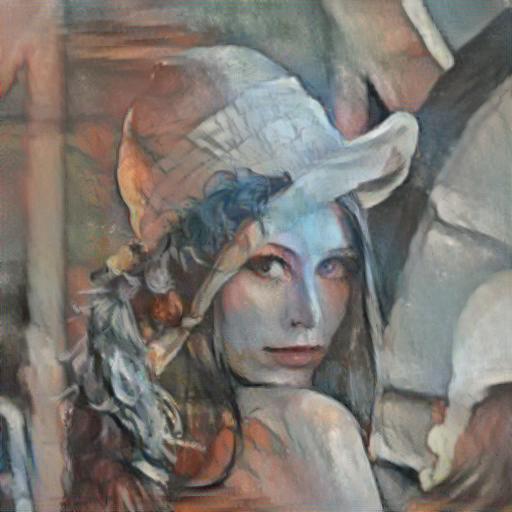}&
		\includegraphics[width=0.14\linewidth]{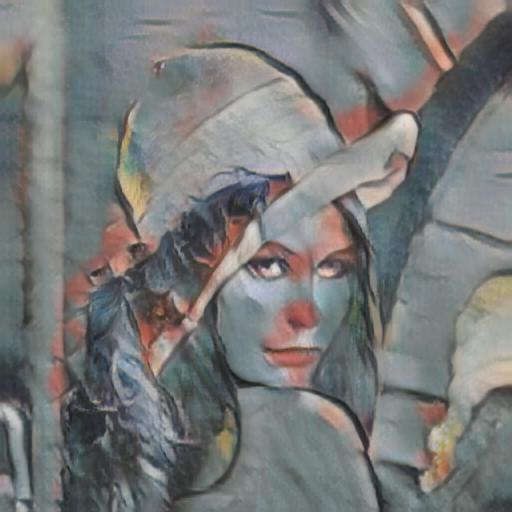}&
		\includegraphics[width=0.14\linewidth]{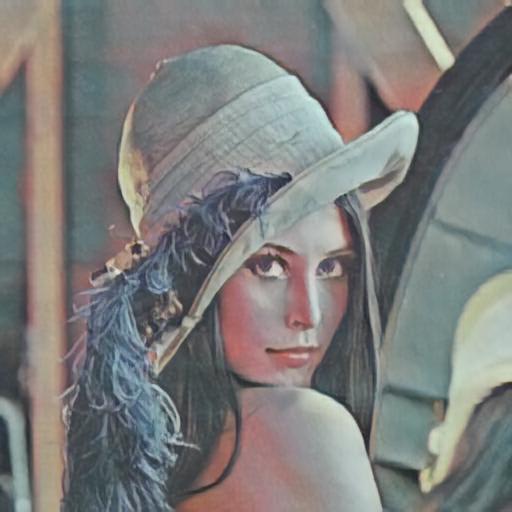}&
		\includegraphics[width=0.14\linewidth]{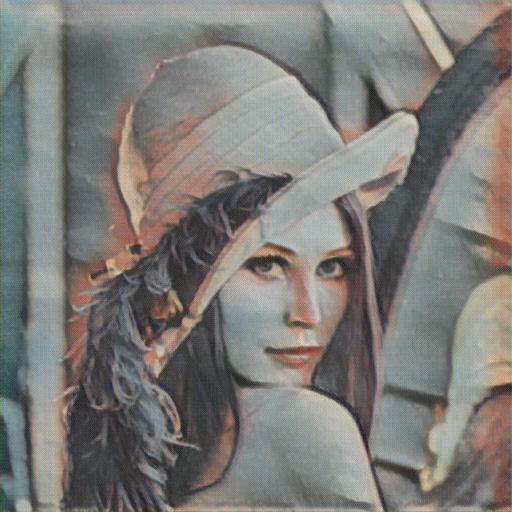}&
		\includegraphics[width=0.14\linewidth]{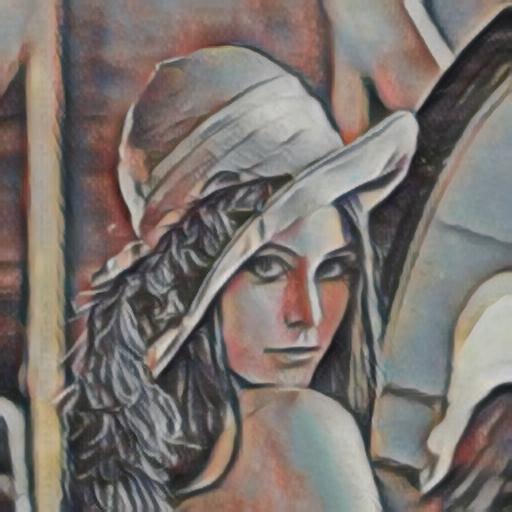}
		\\
		Content &&& \bf Ours & Gatys~\cite{gatys2016image} & EFDM~\cite{zhang2022exact} & StyTR$^2$~\cite{deng2022stytr2} & ArtFlow~\cite{an2021artflow} & AdaAttN~\cite{liu2021adaattn}
		\\
		\includegraphics[width=0.14\linewidth]{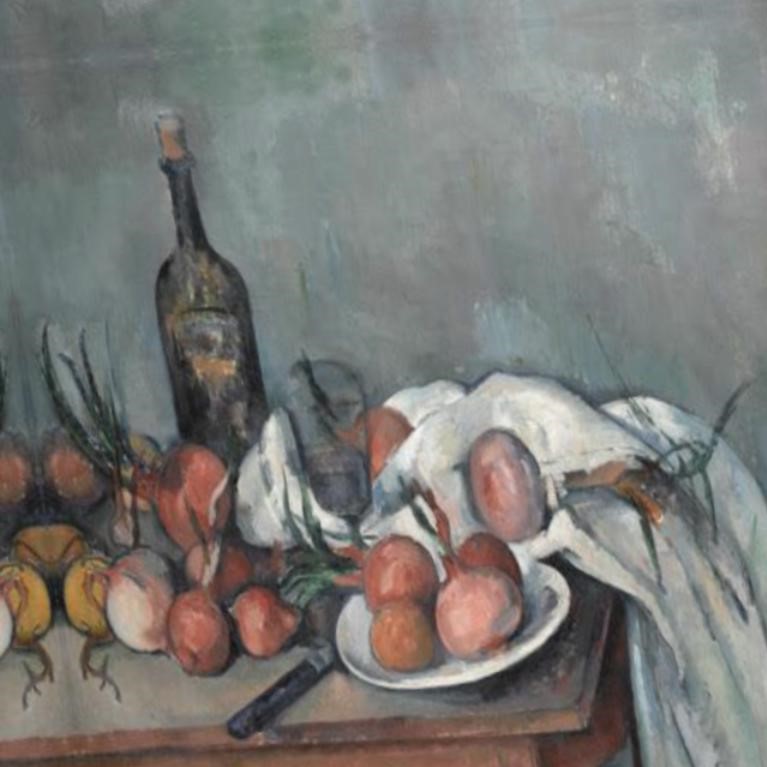}& &&
		
		\includegraphics[width=0.14\linewidth]{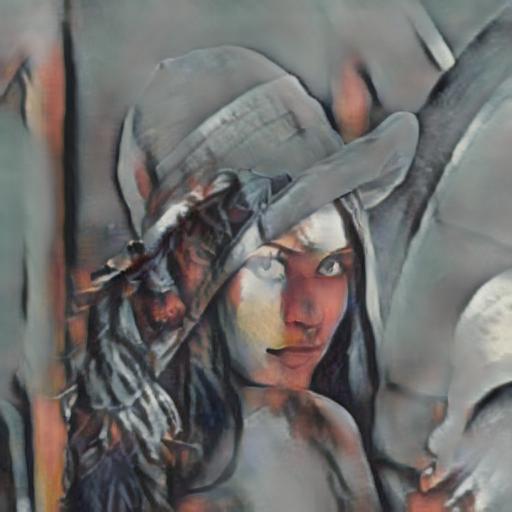}&
		\includegraphics[width=0.14\linewidth]{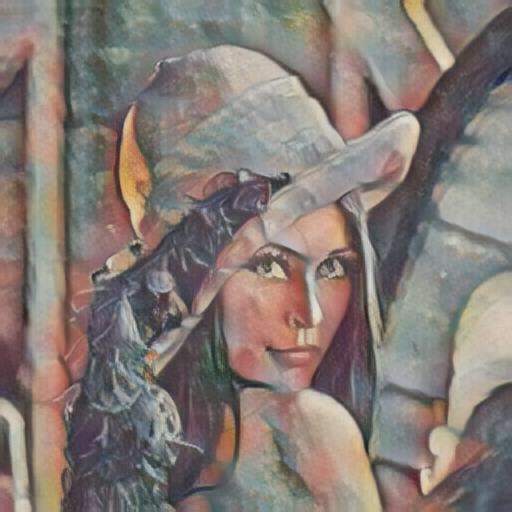}&
		\includegraphics[width=0.14\linewidth]{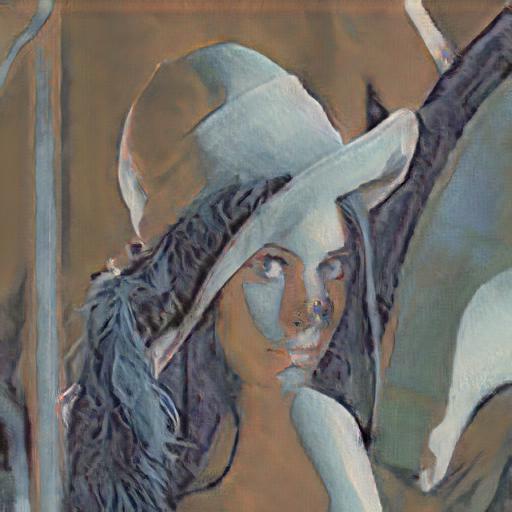}&
		\includegraphics[width=0.14\linewidth]{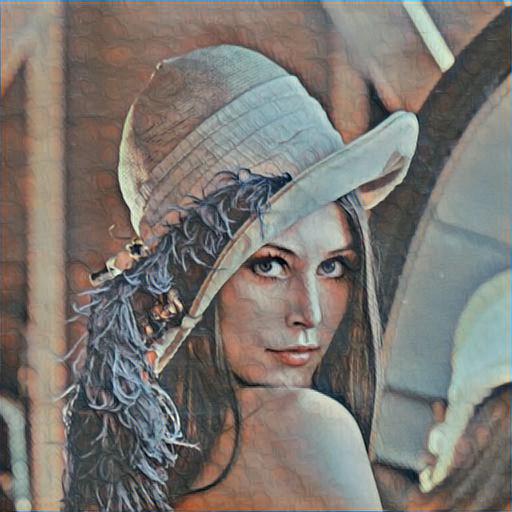}&
		\includegraphics[width=0.14\linewidth]{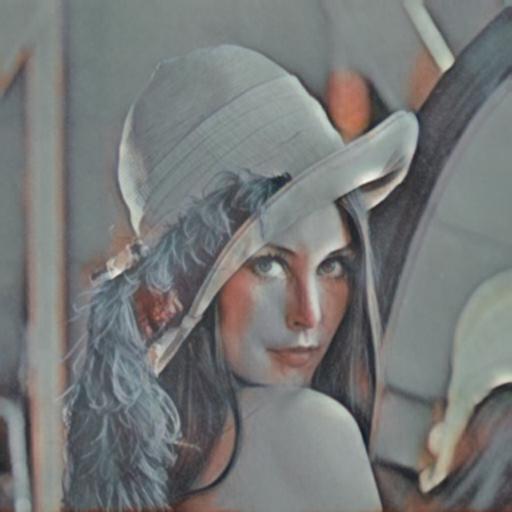}
		\\
		Style&&&  IECAST~\cite{chen2021artistic} & MAST~\cite{deng2020arbitrary} & TPFR~\cite{svoboda2020two} & Johnson~\cite{johnson2016perceptual} & LapStyle~\cite{lin2021drafting} 
		
		\\
		\vspace{0.1cm}
		\\
		\\
		\vspace{0.1cm}
		\\
		
		\includegraphics[width=0.14\linewidth]{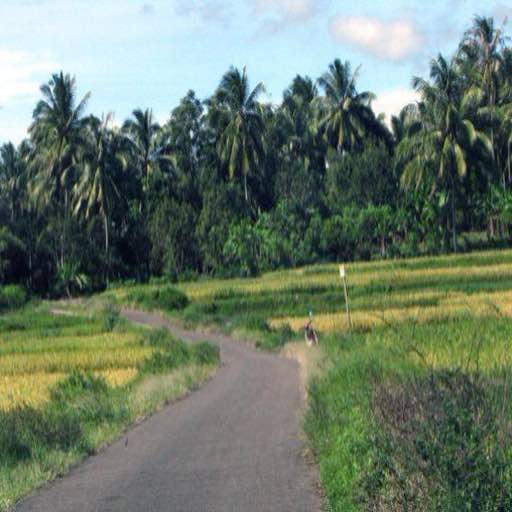}& &&
		\includegraphics[width=0.14\linewidth]{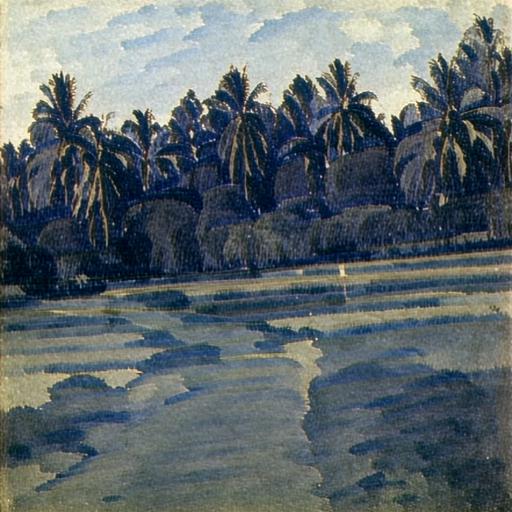}&
		\includegraphics[width=0.14\linewidth]{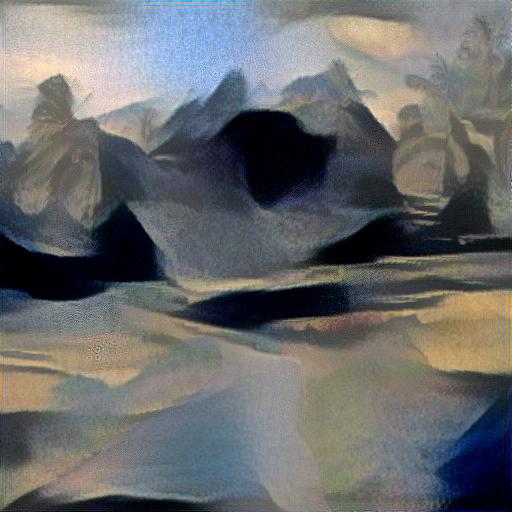}&
		\includegraphics[width=0.14\linewidth]{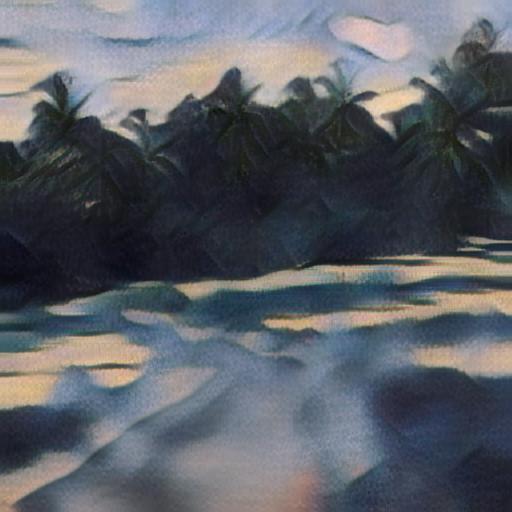}&
		\includegraphics[width=0.14\linewidth]{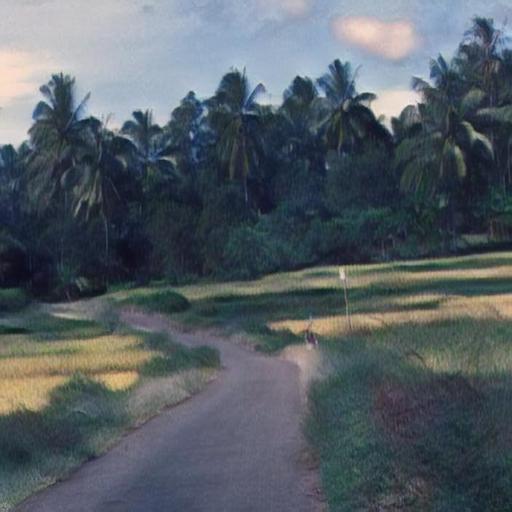}&
		\includegraphics[width=0.14\linewidth]{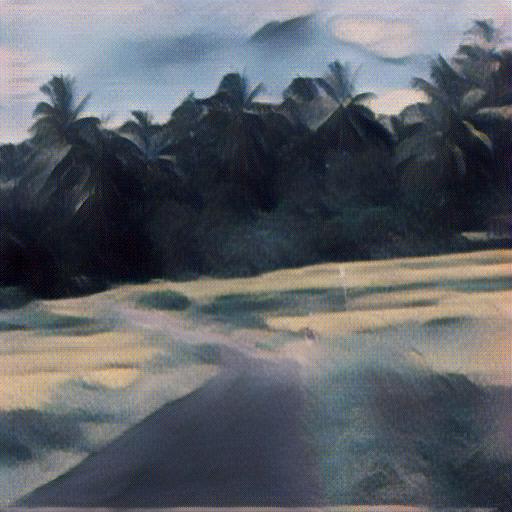}&
		\includegraphics[width=0.14\linewidth]{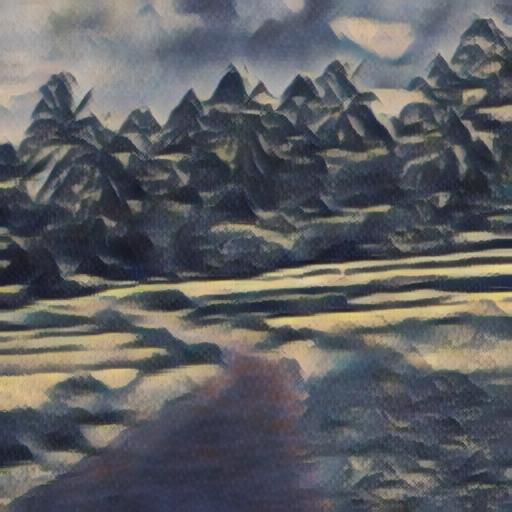}
		\\
		Content &&& \bf Ours & Gatys~\cite{gatys2016image} & EFDM~\cite{zhang2022exact} & StyTR$^2$~\cite{deng2022stytr2} & ArtFlow~\cite{an2021artflow} & AdaAttN~\cite{liu2021adaattn}
		\\
		\includegraphics[width=0.14\linewidth]{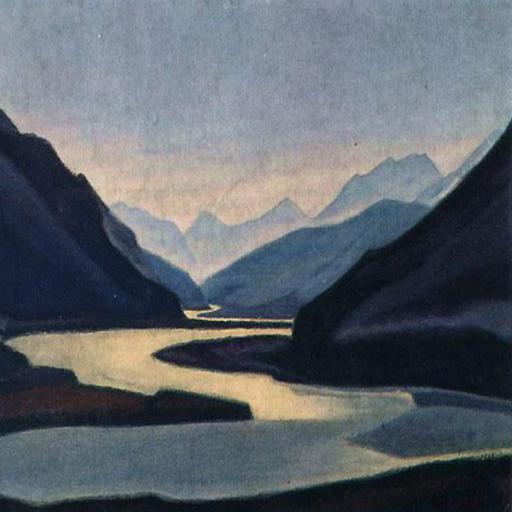}& &&
		
		\includegraphics[width=0.14\linewidth]{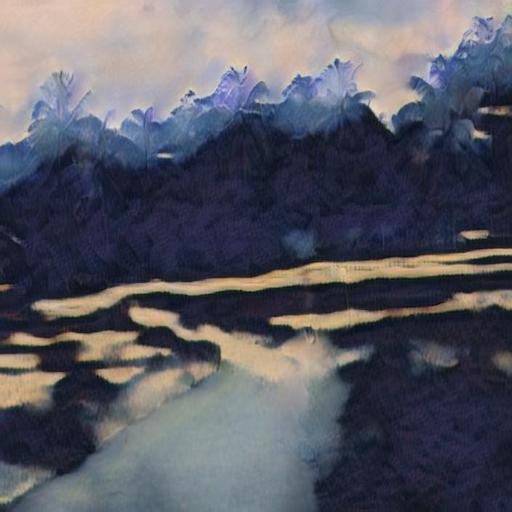}&
		\includegraphics[width=0.14\linewidth]{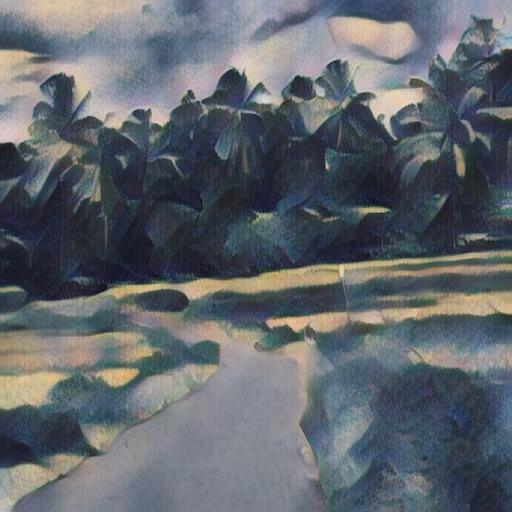}&
		\includegraphics[width=0.14\linewidth]{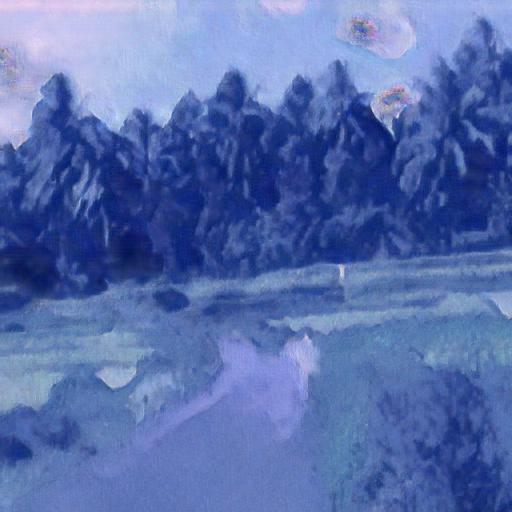}&
		\includegraphics[width=0.14\linewidth]{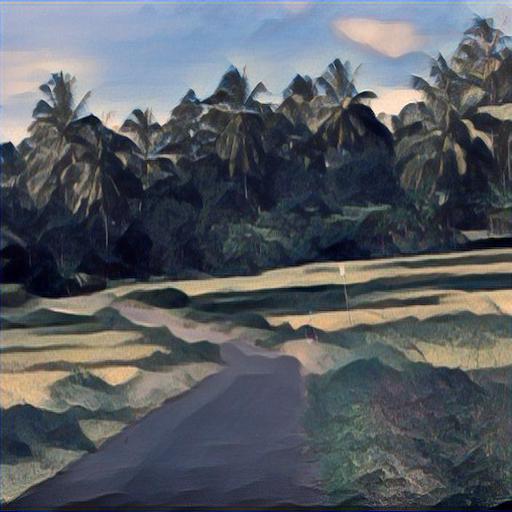}&
		\includegraphics[width=0.14\linewidth]{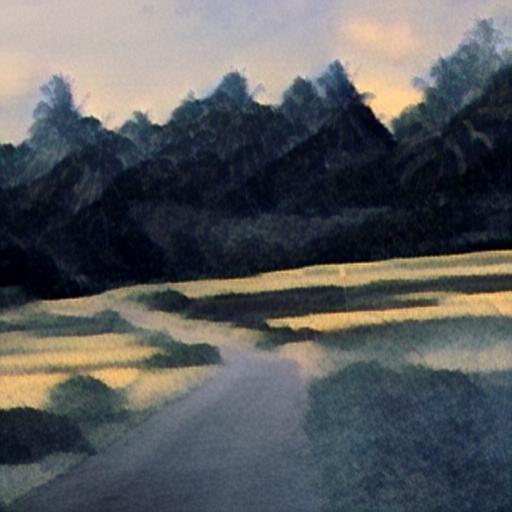}
		\\
		Style&&&  IECAST~\cite{chen2021artistic} & MAST~\cite{deng2020arbitrary} & TPFR~\cite{svoboda2020two} & Johnson~\cite{johnson2016perceptual} & LapStyle~\cite{lin2021drafting}

	\end{tabular}
	\vspace{0.5em}
	\caption{ {\bf More qualitative comparison results (set 1)} with state of the art. Zoom-in for better comparison.}
	\label{fig:cmp1}
\end{figure*}

\begin{figure*}
	\centering
	\setlength{\tabcolsep}{0.02cm}
	\renewcommand\arraystretch{0.6}
	\begin{tabular}{cp{0.1em}|p{0.1em}cccccc}
		
		\includegraphics[width=0.14\linewidth]{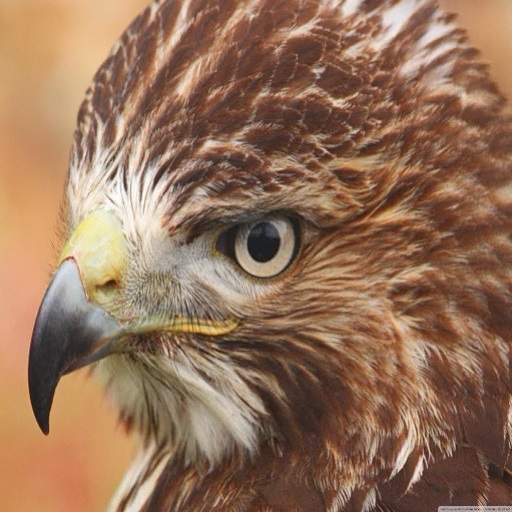}& &&
		\includegraphics[width=0.14\linewidth]{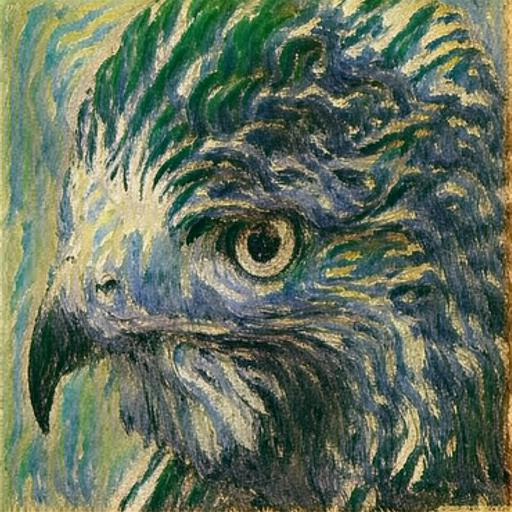}&
		\includegraphics[width=0.14\linewidth]{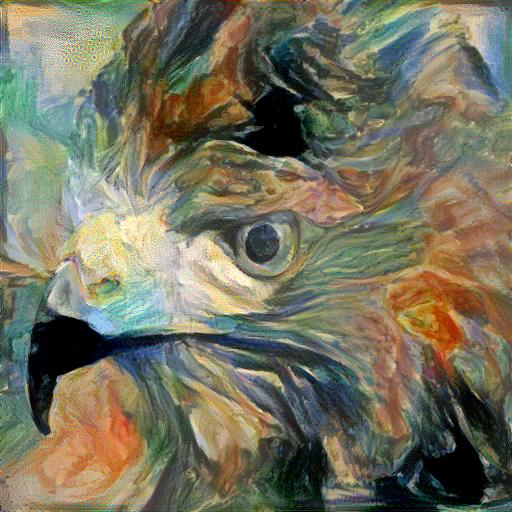}&
		\includegraphics[width=0.14\linewidth]{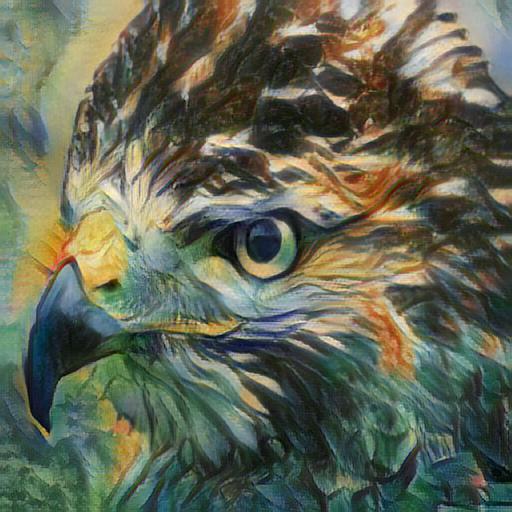}&
		\includegraphics[width=0.14\linewidth]{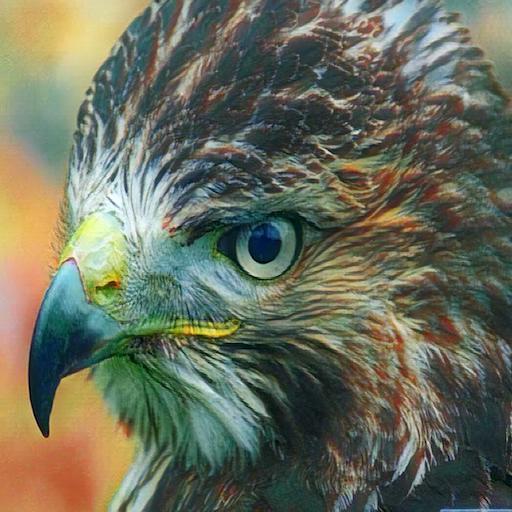}&
		\includegraphics[width=0.14\linewidth]{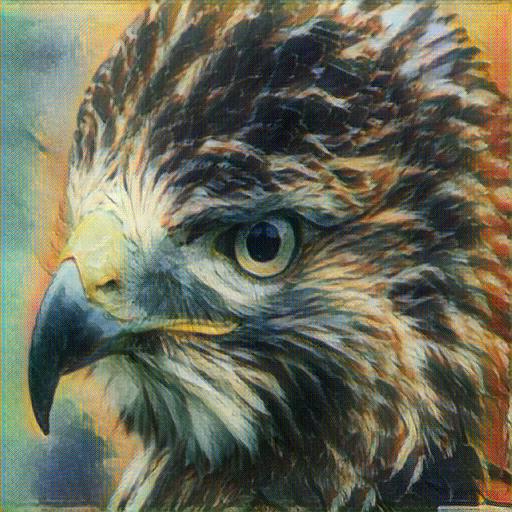}&
		\includegraphics[width=0.14\linewidth]{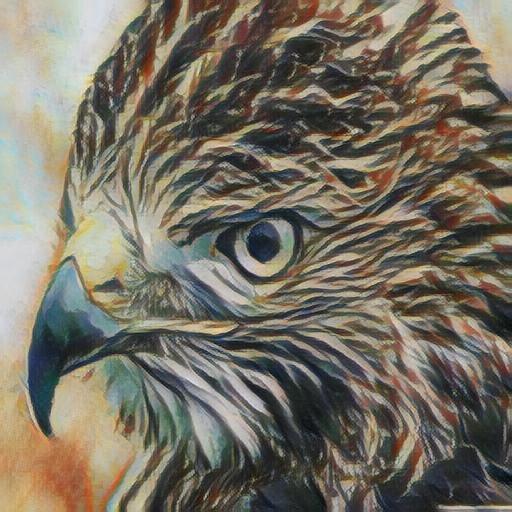}
		\\
		Content &&& \bf Ours & Gatys~\cite{gatys2016image} & EFDM~\cite{zhang2022exact} & StyTR$^2$~\cite{deng2022stytr2} & ArtFlow~\cite{an2021artflow} & AdaAttN~\cite{liu2021adaattn}
		\\
		\includegraphics[width=0.14\linewidth]{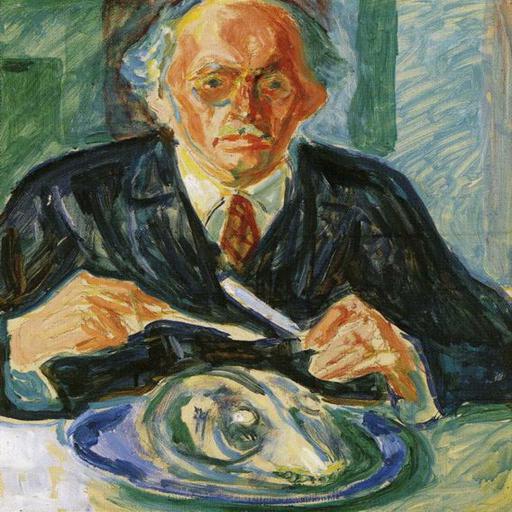}& &&
		
		\includegraphics[width=0.14\linewidth]{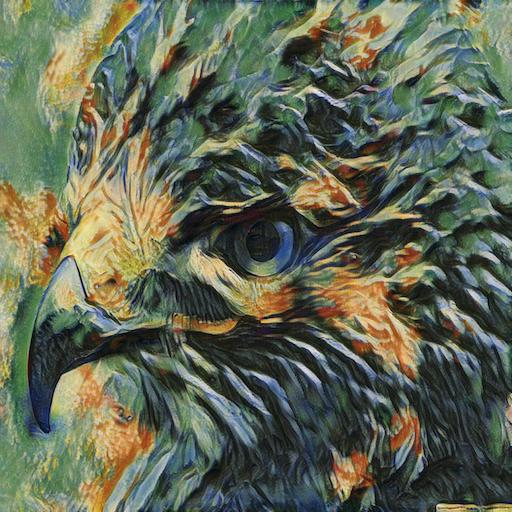}&
		\includegraphics[width=0.14\linewidth]{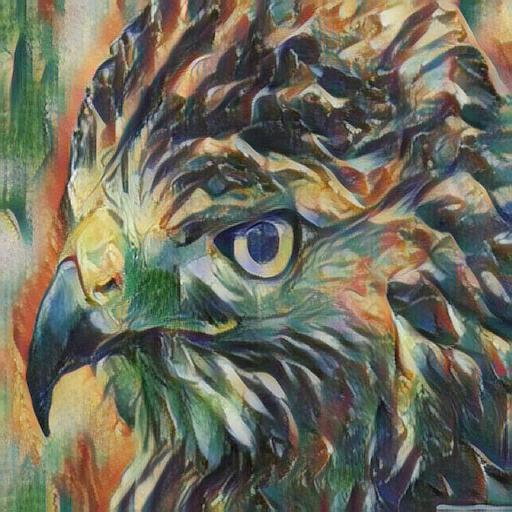}&
		\includegraphics[width=0.14\linewidth]{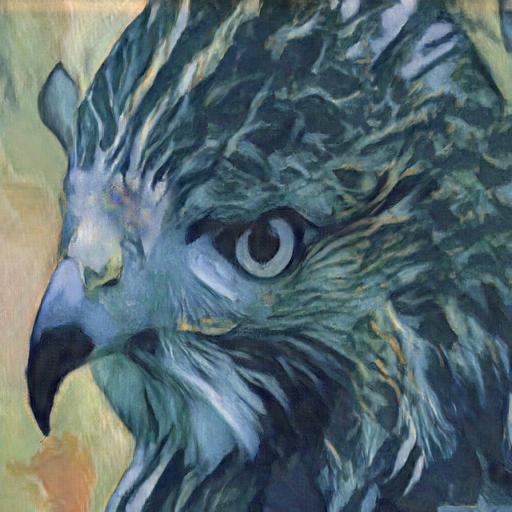}&
		\includegraphics[width=0.14\linewidth]{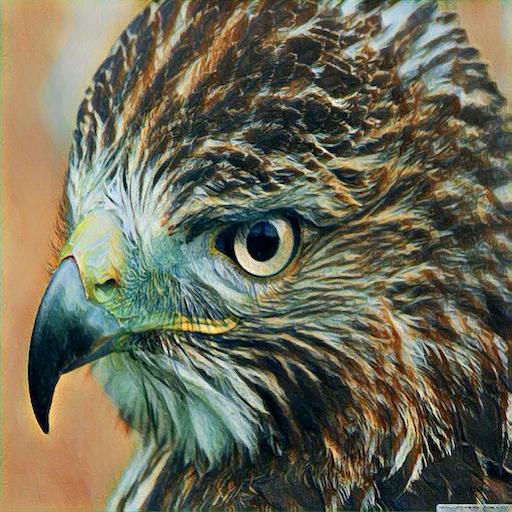}&
		\includegraphics[width=0.14\linewidth]{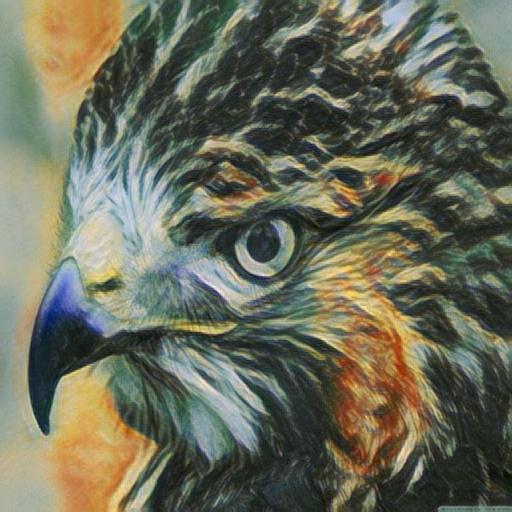}
		\\
		Style&&&  IECAST~\cite{chen2021artistic} & MAST~\cite{deng2020arbitrary} & TPFR~\cite{svoboda2020two} & Johnson~\cite{johnson2016perceptual} & LapStyle~\cite{lin2021drafting} 
		
		\\
		\vspace{0.1cm}
		\\
		\\
		\vspace{0.1cm}
		\\
		
		\includegraphics[width=0.14\linewidth]{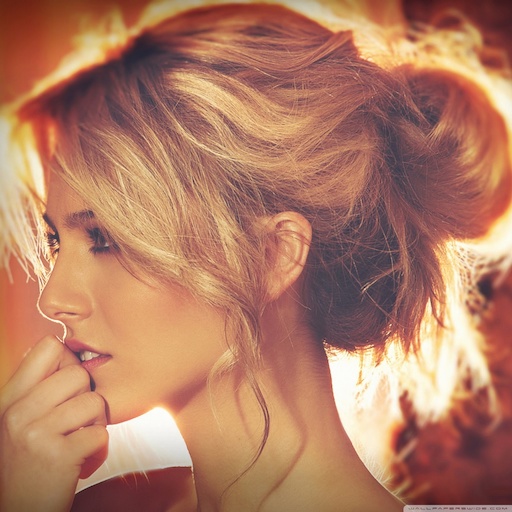}& &&
		\includegraphics[width=0.14\linewidth]{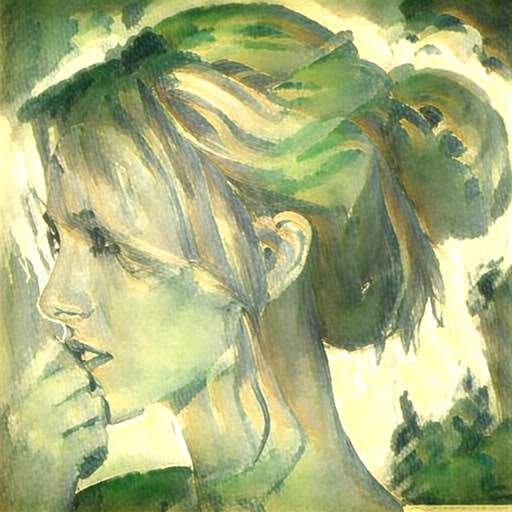}&
		\includegraphics[width=0.14\linewidth]{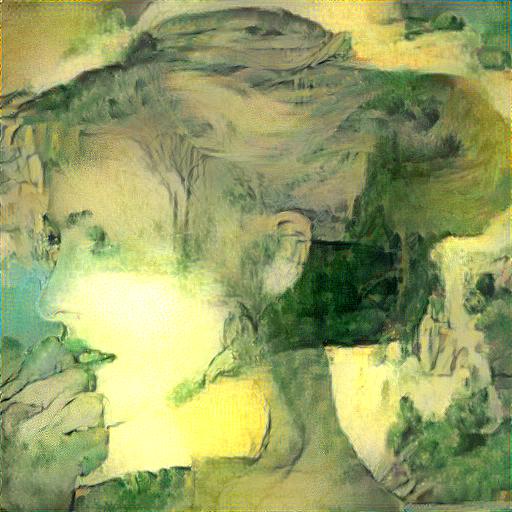}&
		\includegraphics[width=0.14\linewidth]{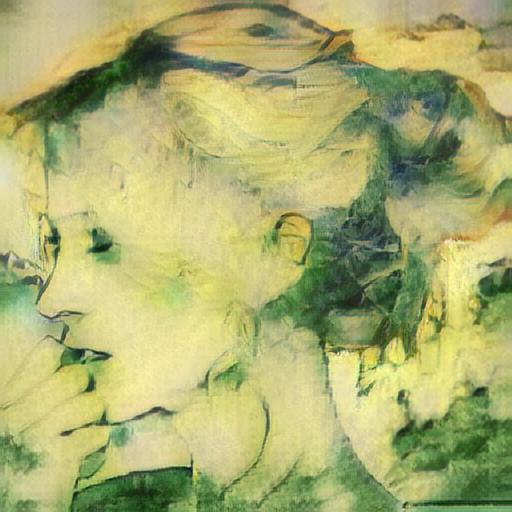}&
		\includegraphics[width=0.14\linewidth]{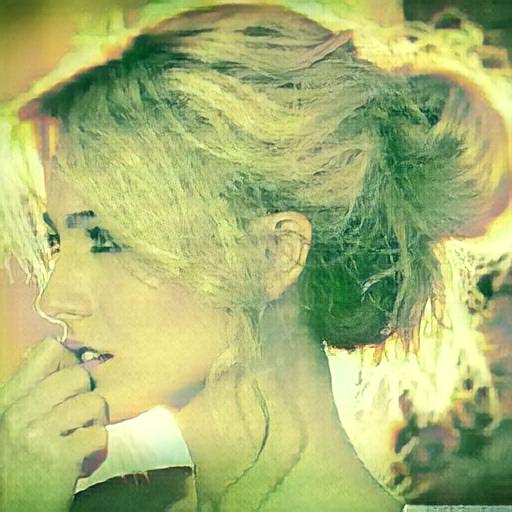}&
		\includegraphics[width=0.14\linewidth]{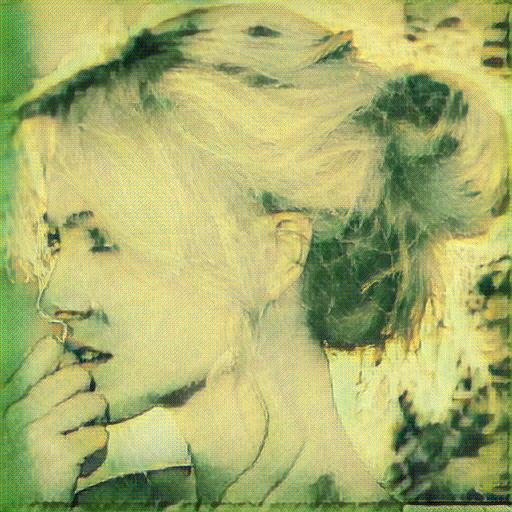}&
		\includegraphics[width=0.14\linewidth]{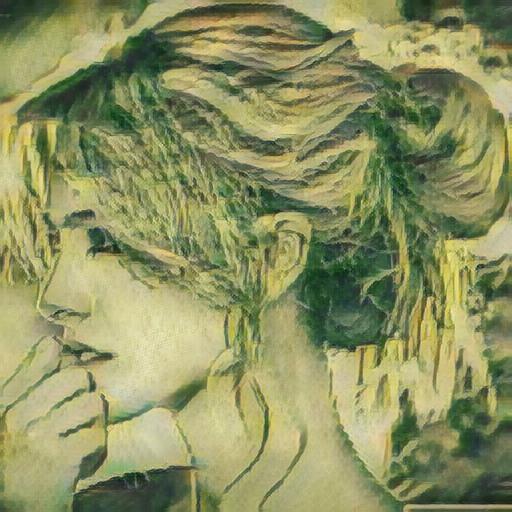}
		\\
		Content &&& \bf Ours & Gatys~\cite{gatys2016image} & EFDM~\cite{zhang2022exact} & StyTR$^2$~\cite{deng2022stytr2} & ArtFlow~\cite{an2021artflow} & AdaAttN~\cite{liu2021adaattn}
		\\
		\includegraphics[width=0.14\linewidth]{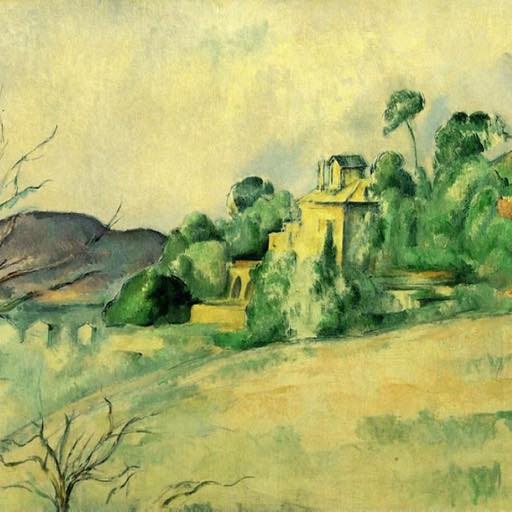}& &&
		
		\includegraphics[width=0.14\linewidth]{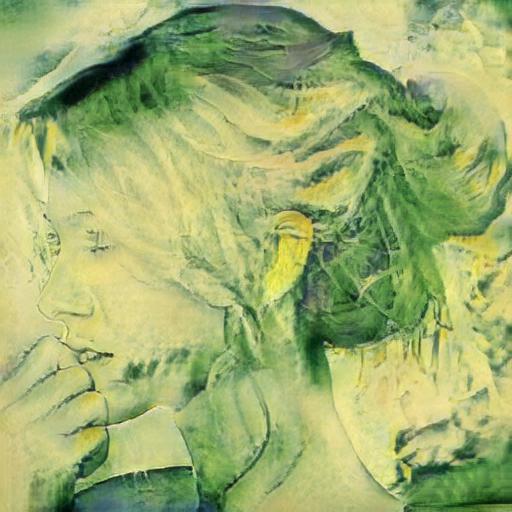}&
		\includegraphics[width=0.14\linewidth]{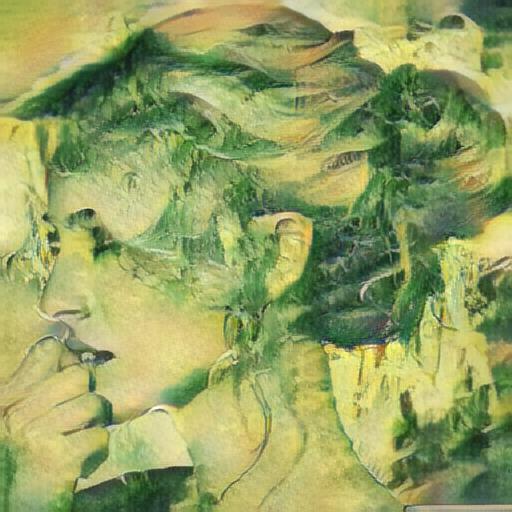}&
		\includegraphics[width=0.14\linewidth]{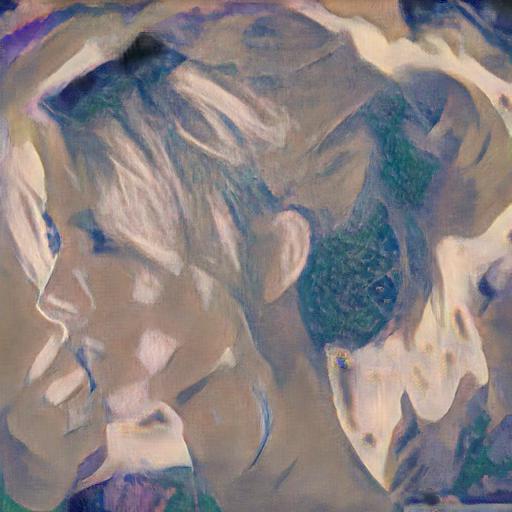}&
		\includegraphics[width=0.14\linewidth]{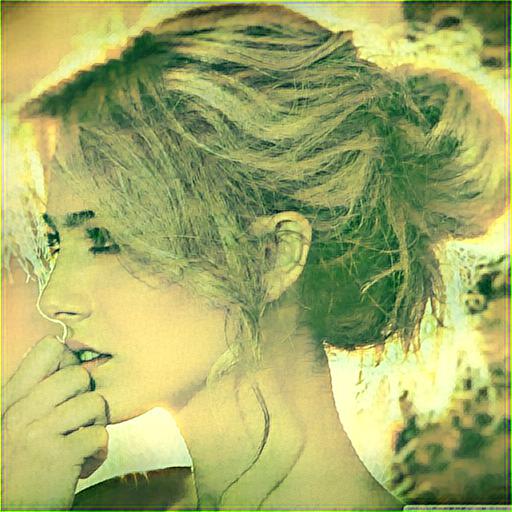}&
		\includegraphics[width=0.14\linewidth]{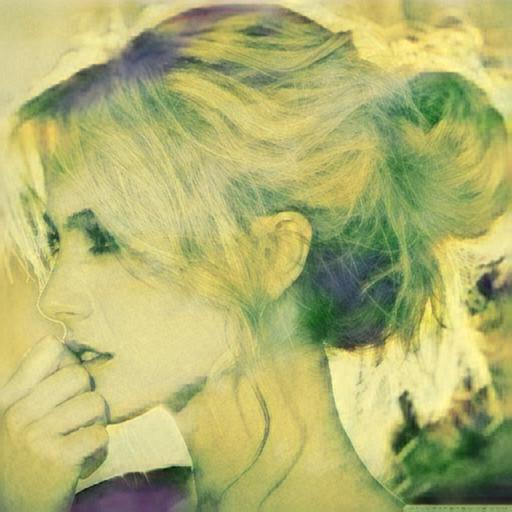}
		\\
		Style&&&  IECAST~\cite{chen2021artistic} & MAST~\cite{deng2020arbitrary} & TPFR~\cite{svoboda2020two} & Johnson~\cite{johnson2016perceptual} & LapStyle~\cite{lin2021drafting} 
		
		\\
		\vspace{0.1cm}
		\\
		\\
		\vspace{0.1cm}
		\\
		
		\includegraphics[width=0.14\linewidth]{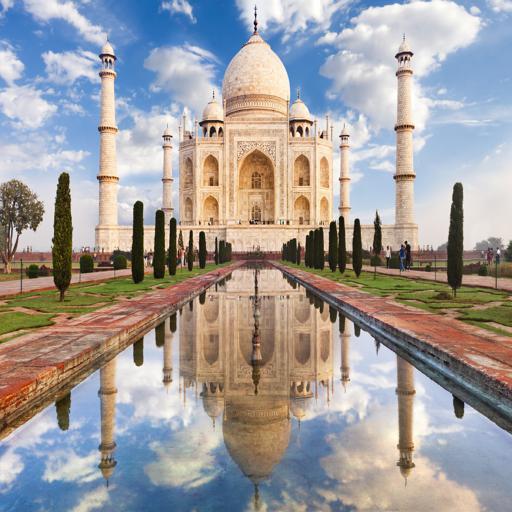}& &&
		\includegraphics[width=0.14\linewidth]{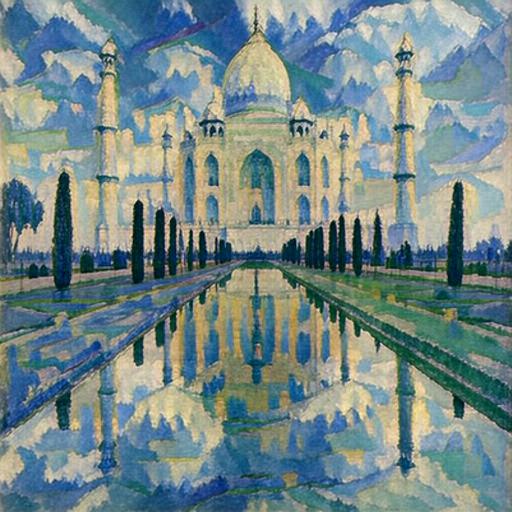}&
		\includegraphics[width=0.14\linewidth]{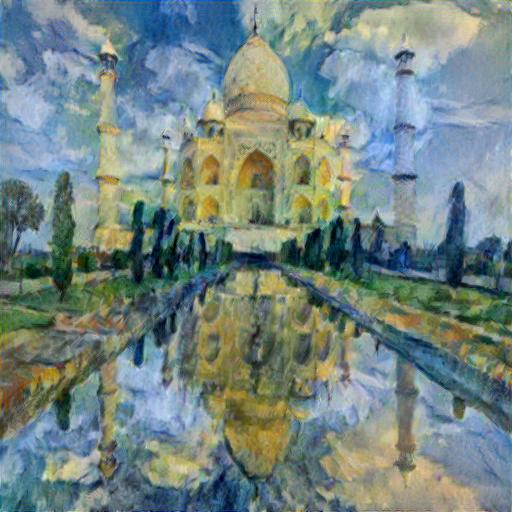}&
		\includegraphics[width=0.14\linewidth]{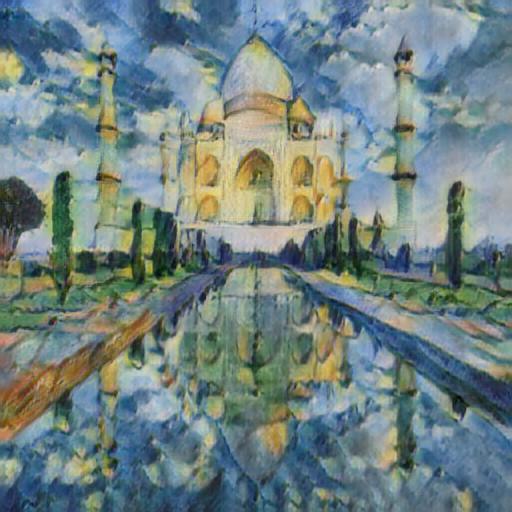}&
		\includegraphics[width=0.14\linewidth]{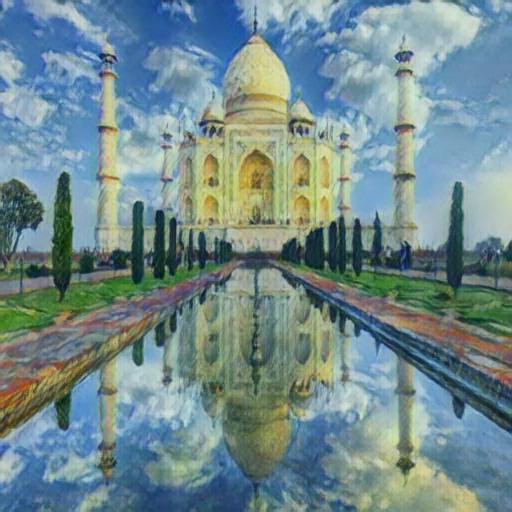}&
		\includegraphics[width=0.14\linewidth]{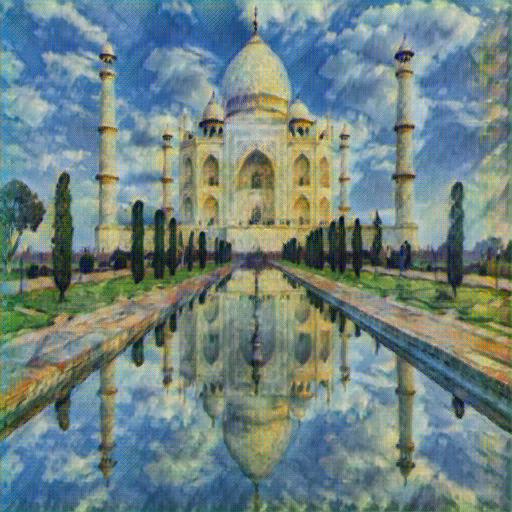}&
		\includegraphics[width=0.14\linewidth]{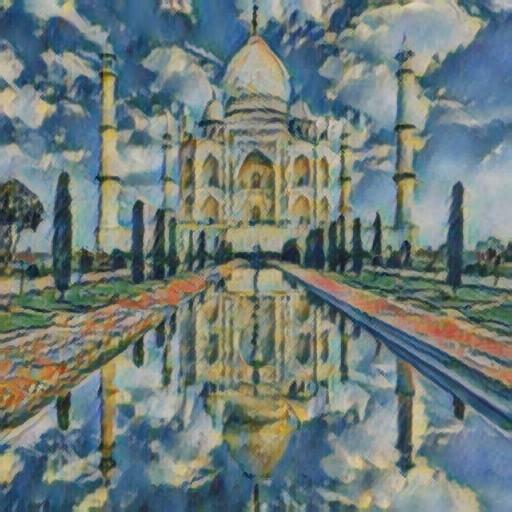}
		\\
		Content &&& \bf Ours & Gatys~\cite{gatys2016image} & EFDM~\cite{zhang2022exact} & StyTR$^2$~\cite{deng2022stytr2} & ArtFlow~\cite{an2021artflow} & AdaAttN~\cite{liu2021adaattn}
		\\
		\includegraphics[width=0.14\linewidth]{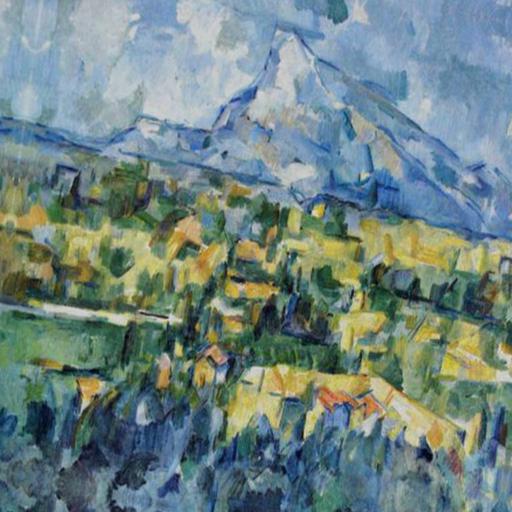}& &&
		
		\includegraphics[width=0.14\linewidth]{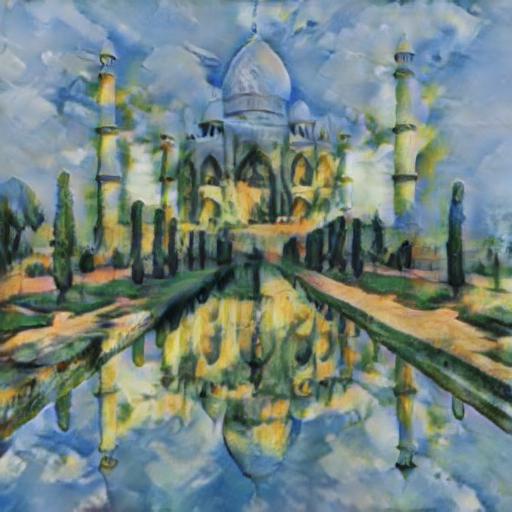}&
		\includegraphics[width=0.14\linewidth]{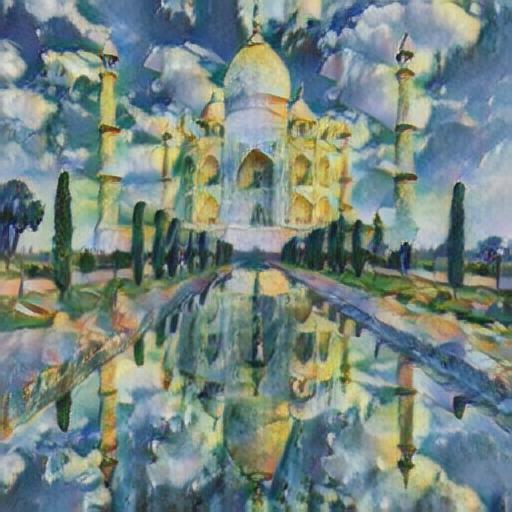}&
		\includegraphics[width=0.14\linewidth]{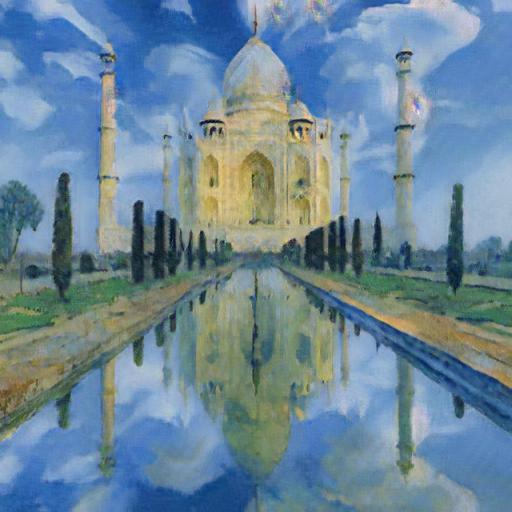}&
		\includegraphics[width=0.14\linewidth]{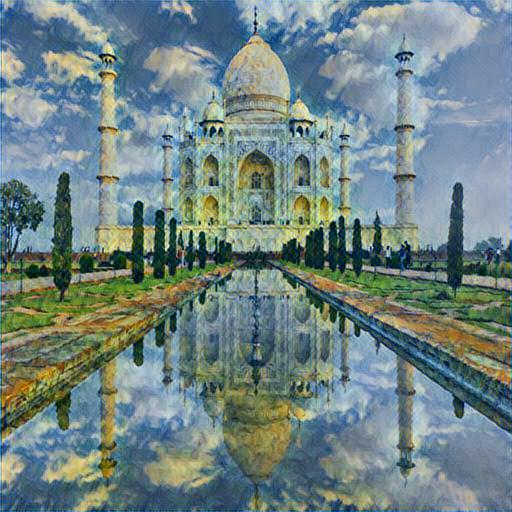}&
		\includegraphics[width=0.14\linewidth]{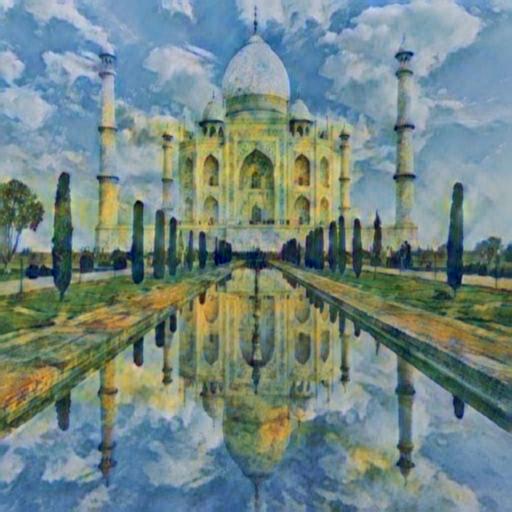}
		\\
		Style&&&  IECAST~\cite{chen2021artistic} & MAST~\cite{deng2020arbitrary} & TPFR~\cite{svoboda2020two} & Johnson~\cite{johnson2016perceptual} & LapStyle~\cite{lin2021drafting}

	\end{tabular}
	\vspace{0.5em}
	\caption{ {\bf More qualitative comparison results (set 2)} with state of the art. Zoom-in for better comparison.}
	\label{fig:cmp2}
\end{figure*}

In Fig.~\ref{fig:cmp1} and~\ref{fig:cmp2}, we provide more qualitative comparison results with state-of-the-art style transfer methods.

\section{Additional Stylized Results}
\label{sec:res}
\begin{figure*}
	\centering
	\setlength{\tabcolsep}{0.1cm}
	\renewcommand\arraystretch{5}
	\begin{tabular}{ccccccc}
		
		&
		\includegraphics[width=0.13\linewidth]{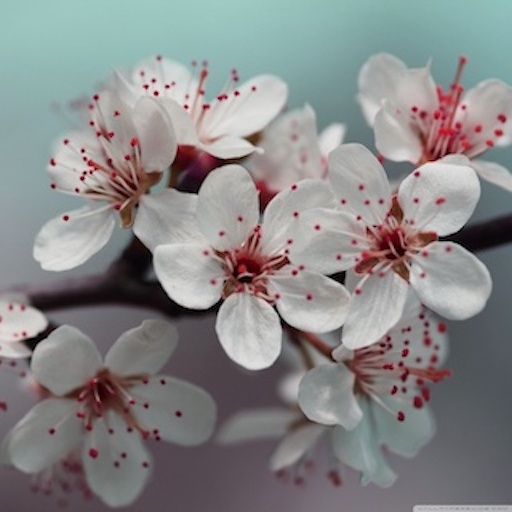}&
		\includegraphics[width=0.13\linewidth]{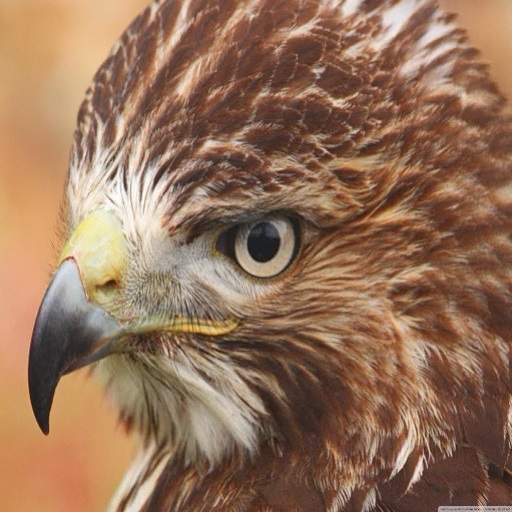}&
		\includegraphics[width=0.13\linewidth]{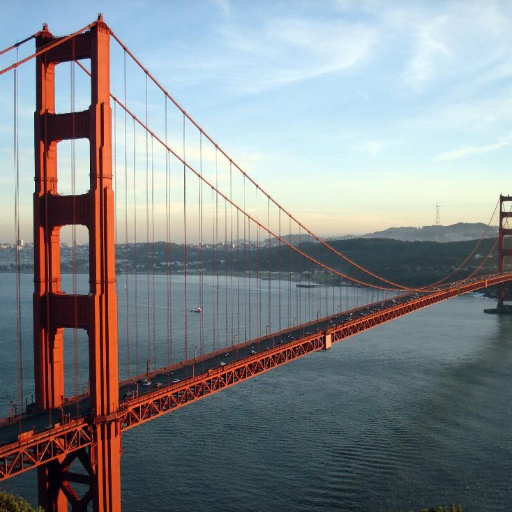}&
		\includegraphics[width=0.13\linewidth]{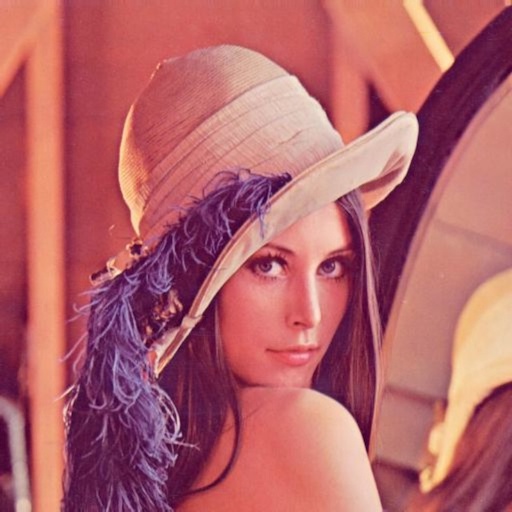}&
		\includegraphics[width=0.13\linewidth]{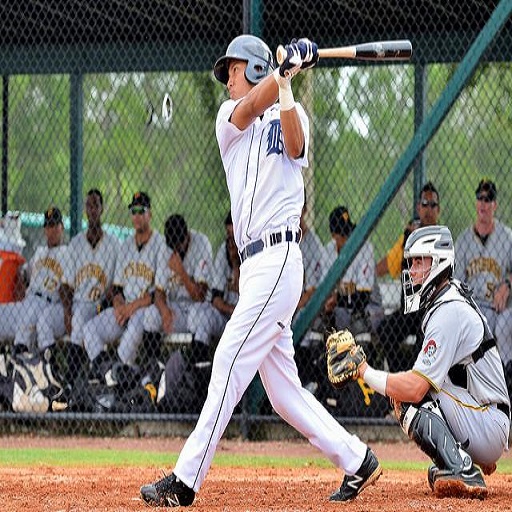}&
		\includegraphics[width=0.13\linewidth]{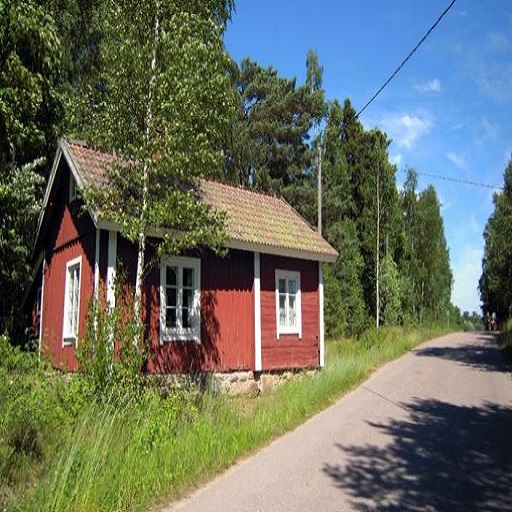}
		
		\\
		
		\includegraphics[width=0.13\linewidth]{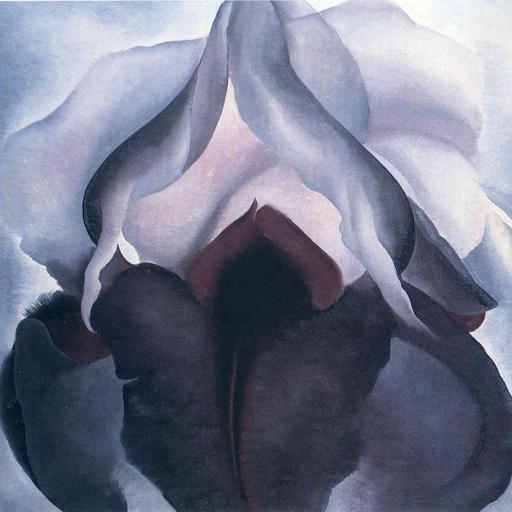}&
		\includegraphics[width=0.13\linewidth]{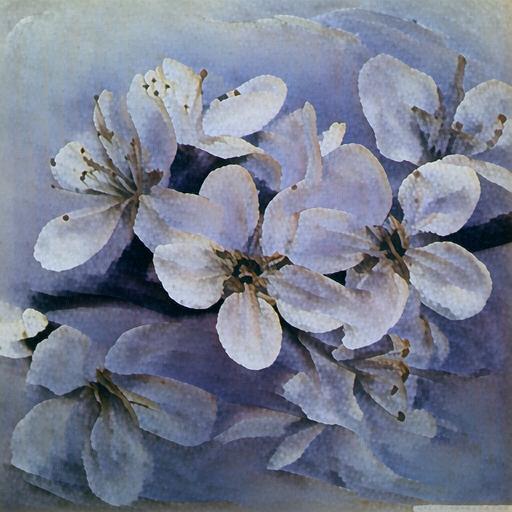}&
		\includegraphics[width=0.13\linewidth]{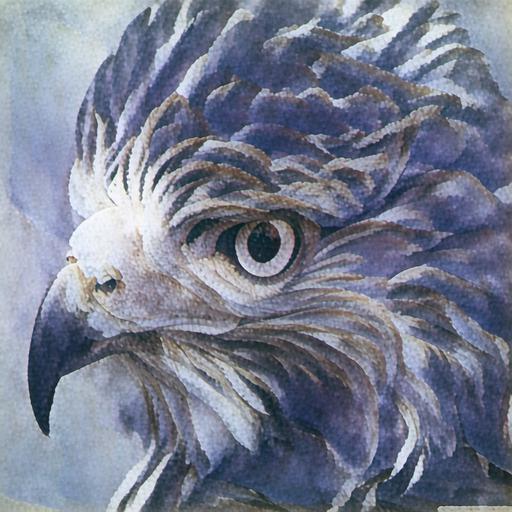}&
		\includegraphics[width=0.13\linewidth]{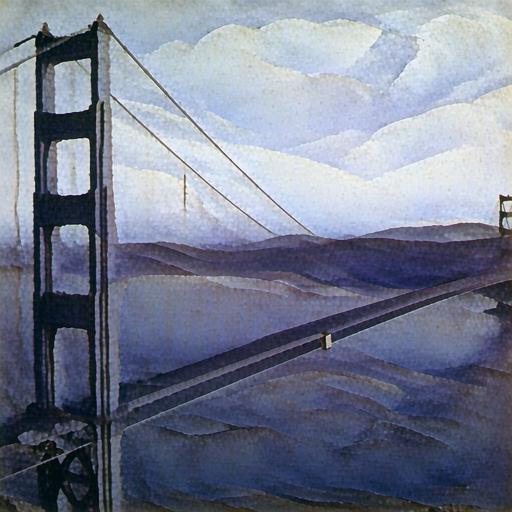}&
		\includegraphics[width=0.13\linewidth]{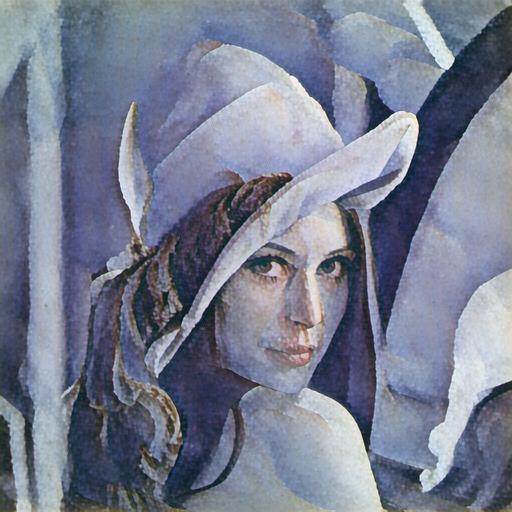}&
		\includegraphics[width=0.13\linewidth]{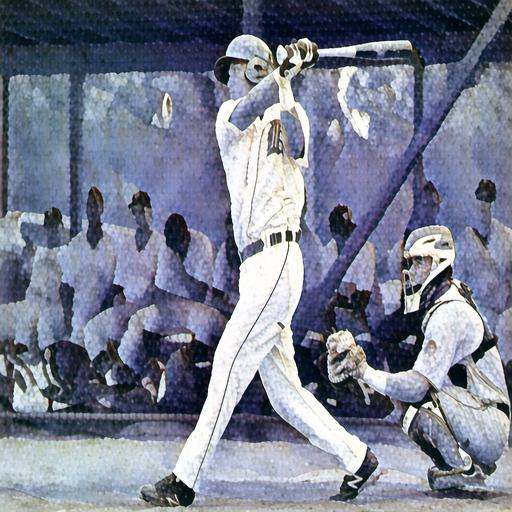}&
		\includegraphics[width=0.13\linewidth]{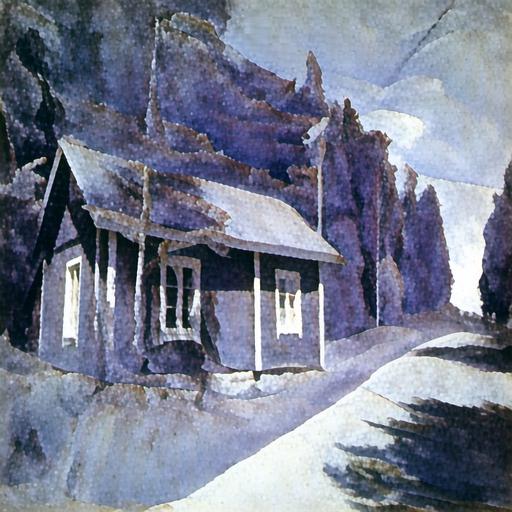}
		
		\\
		\includegraphics[width=0.13\linewidth]{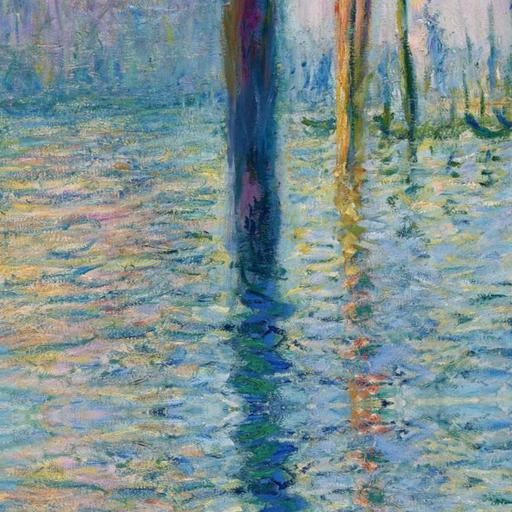}&
		\includegraphics[width=0.13\linewidth]{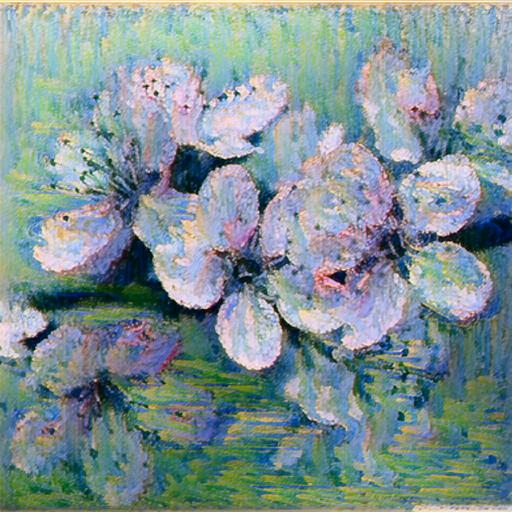}&
		\includegraphics[width=0.13\linewidth]{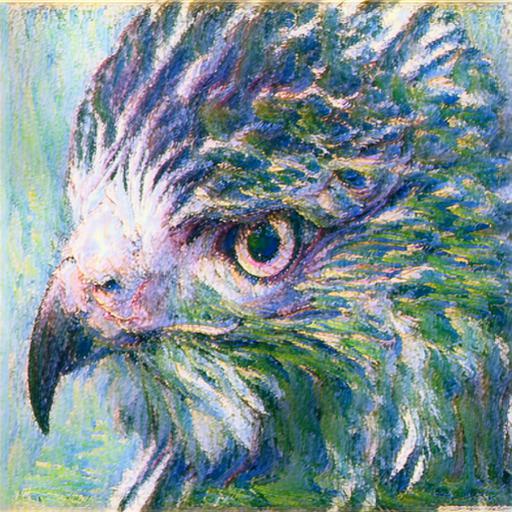}&
		\includegraphics[width=0.13\linewidth]{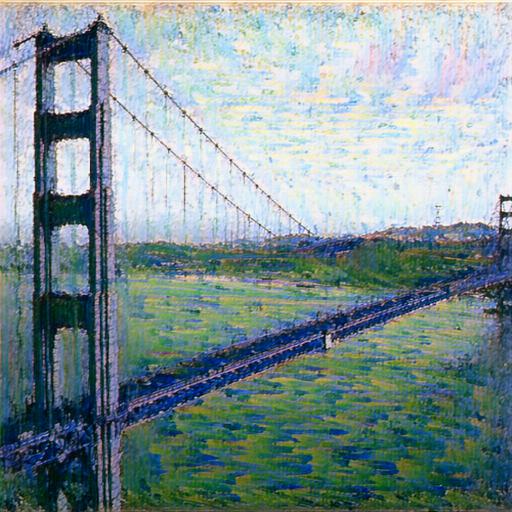}&
		\includegraphics[width=0.13\linewidth]{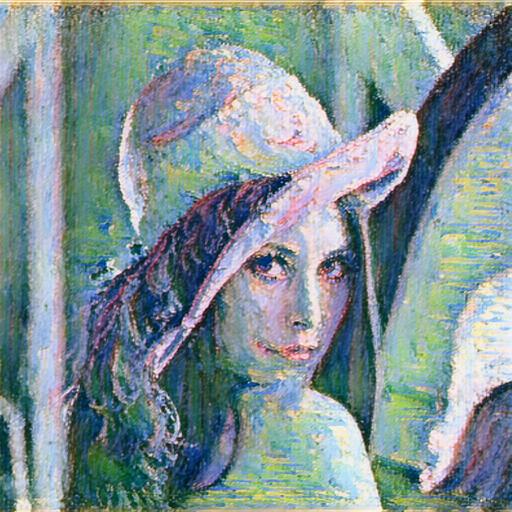}&
		\includegraphics[width=0.13\linewidth]{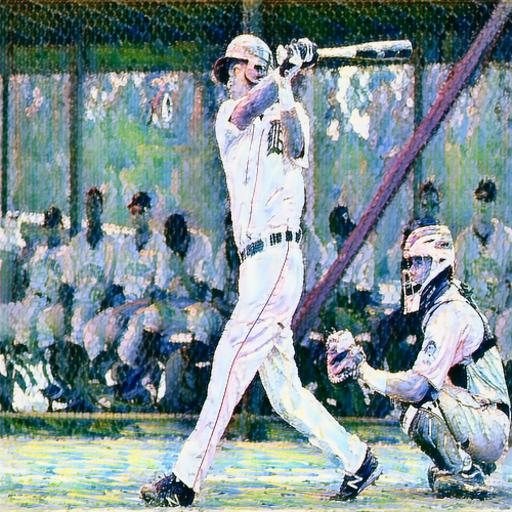}&
		\includegraphics[width=0.13\linewidth]{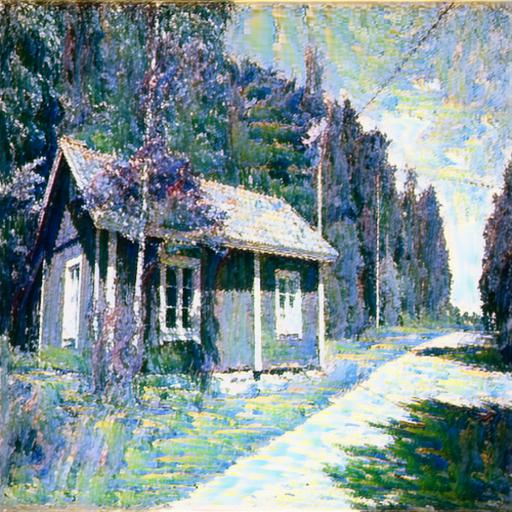}
		
		\\
		\includegraphics[width=0.13\linewidth]{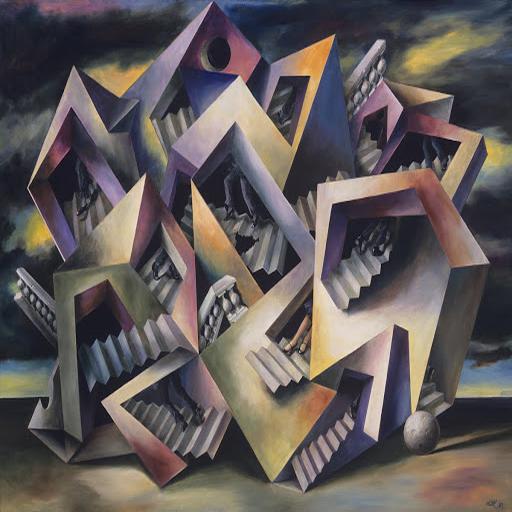}&
		\includegraphics[width=0.13\linewidth]{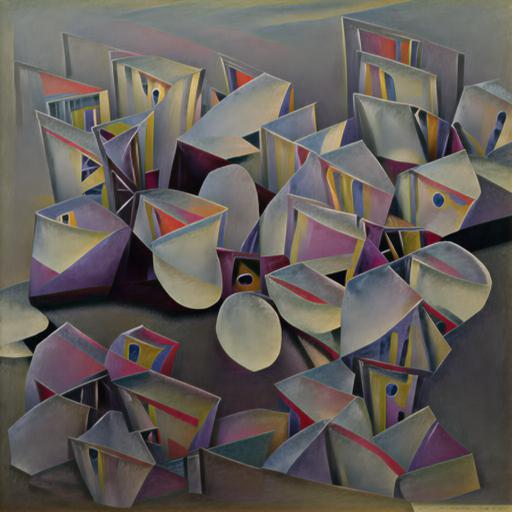}&
		\includegraphics[width=0.13\linewidth]{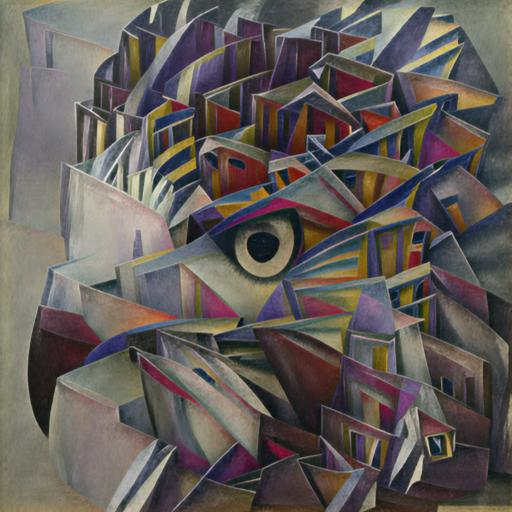}&
		\includegraphics[width=0.13\linewidth]{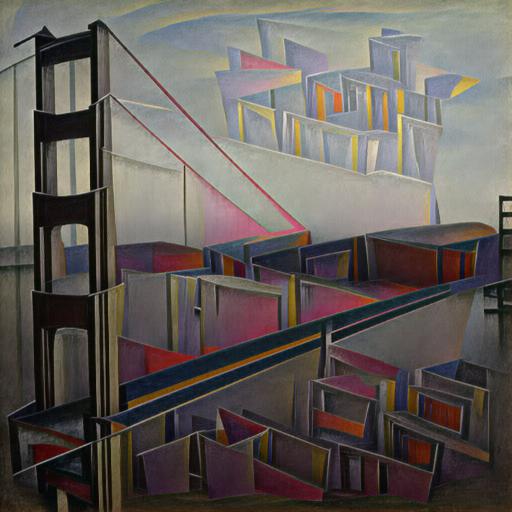}&
		\includegraphics[width=0.13\linewidth]{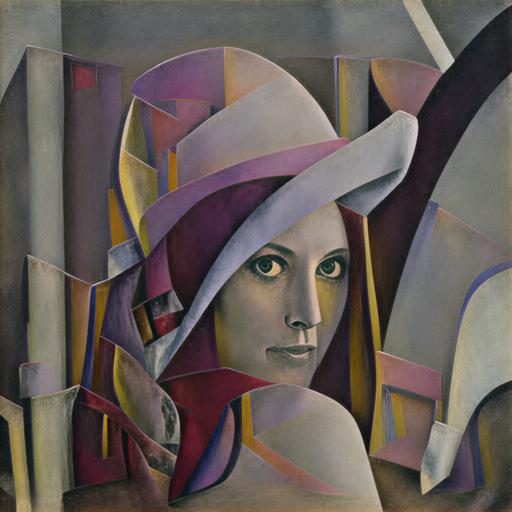}&
		\includegraphics[width=0.13\linewidth]{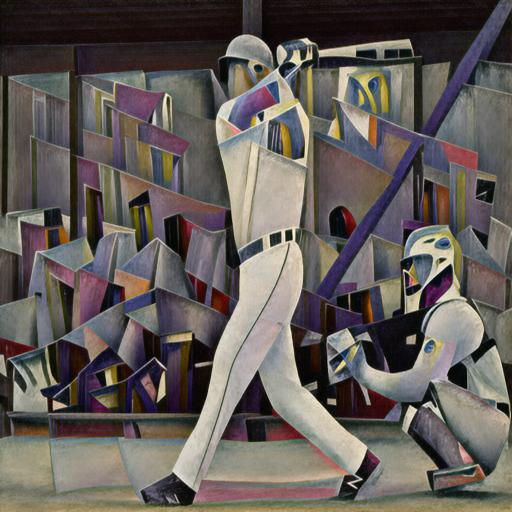}&
		\includegraphics[width=0.13\linewidth]{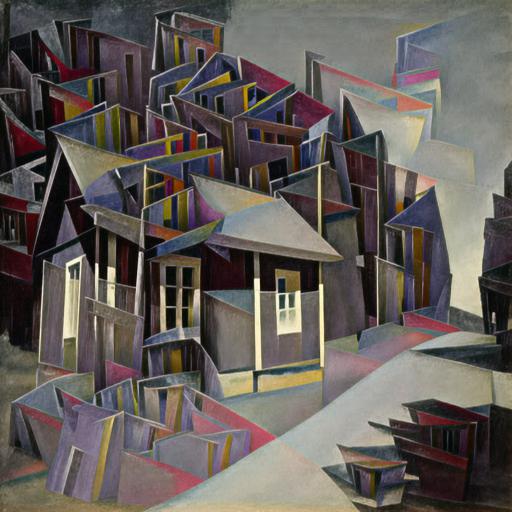}
		
		\\
		\includegraphics[width=0.13\linewidth]{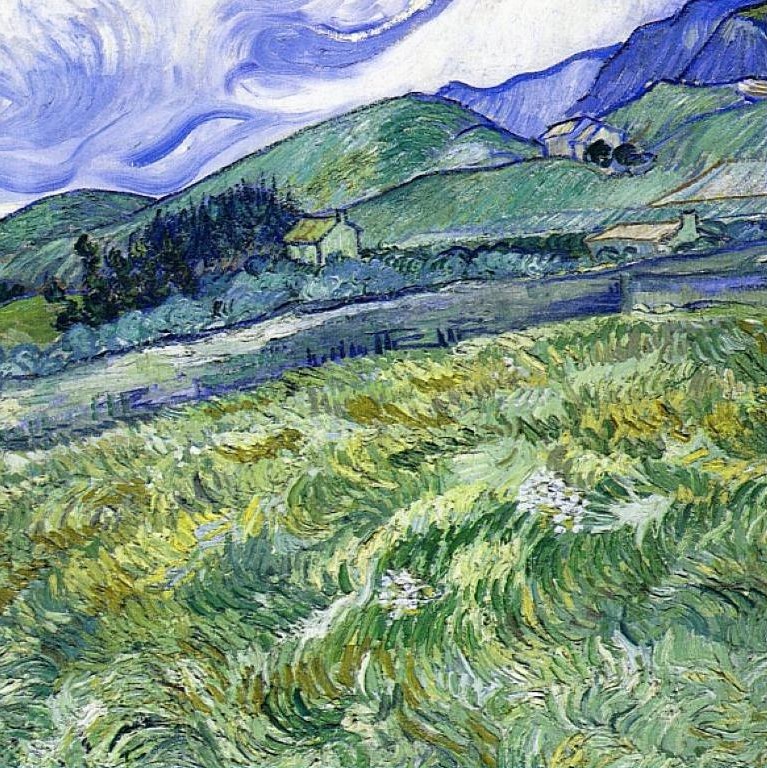}&
		\includegraphics[width=0.13\linewidth]{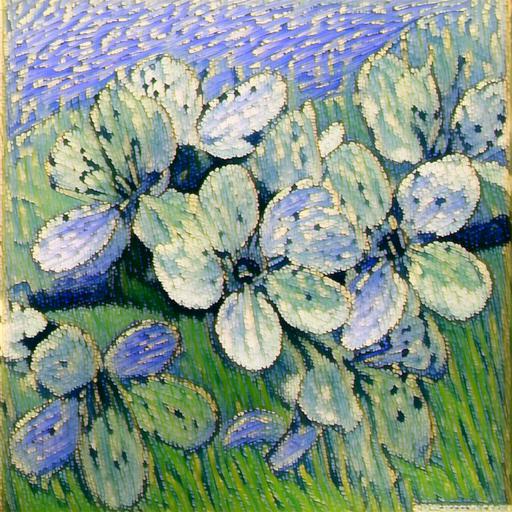}&
		\includegraphics[width=0.13\linewidth]{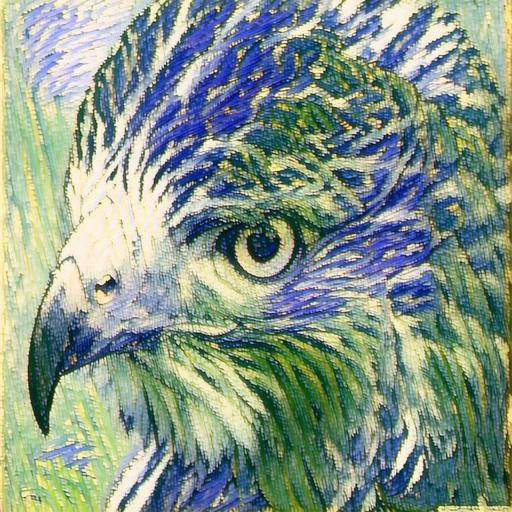}&
		\includegraphics[width=0.13\linewidth]{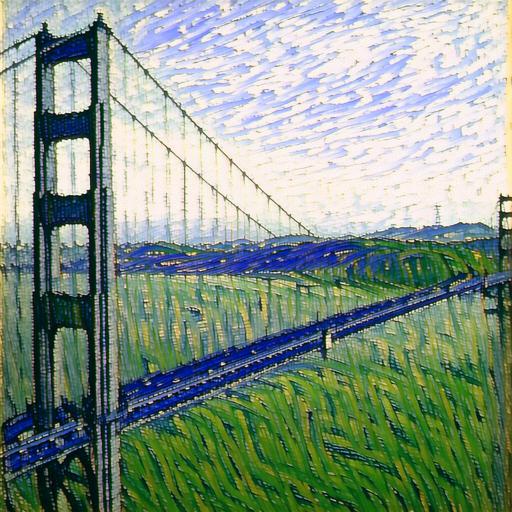}&
		\includegraphics[width=0.13\linewidth]{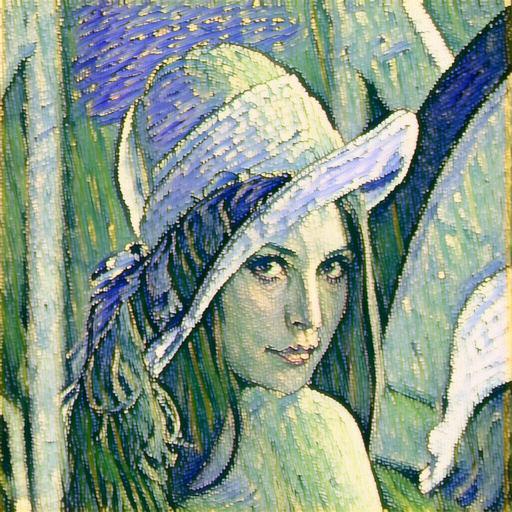}&
		\includegraphics[width=0.13\linewidth]{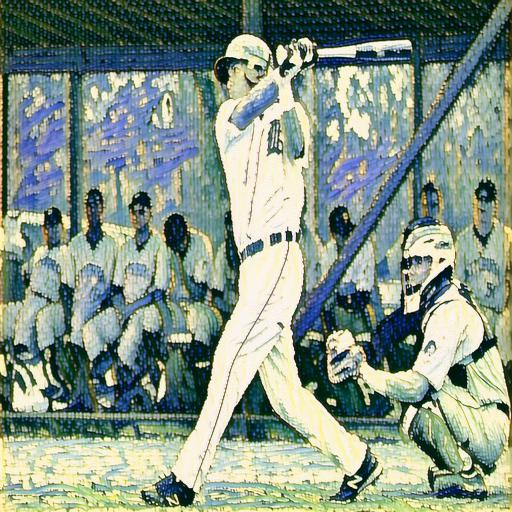}&
		\includegraphics[width=0.13\linewidth]{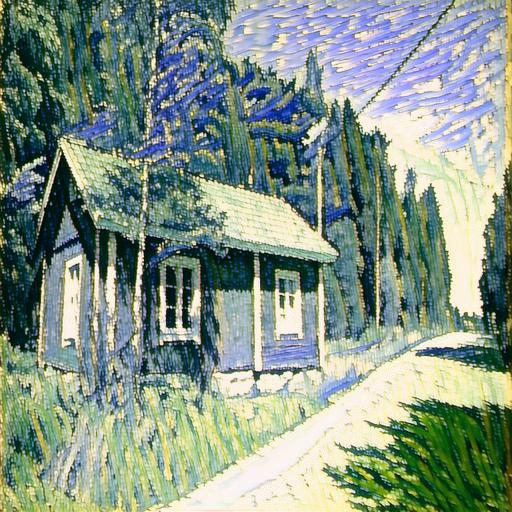}
		
		\\
		\includegraphics[width=0.13\linewidth]{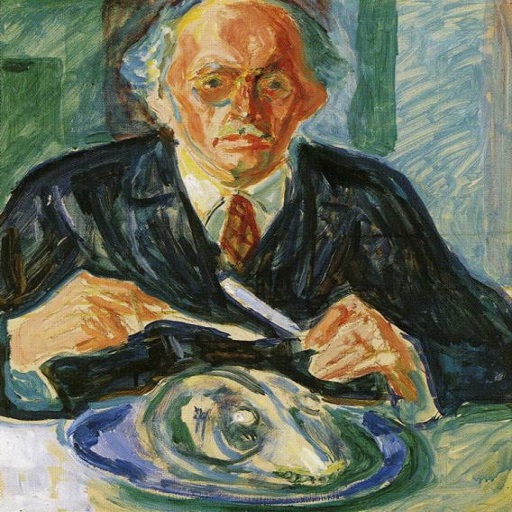}&
		\includegraphics[width=0.13\linewidth]{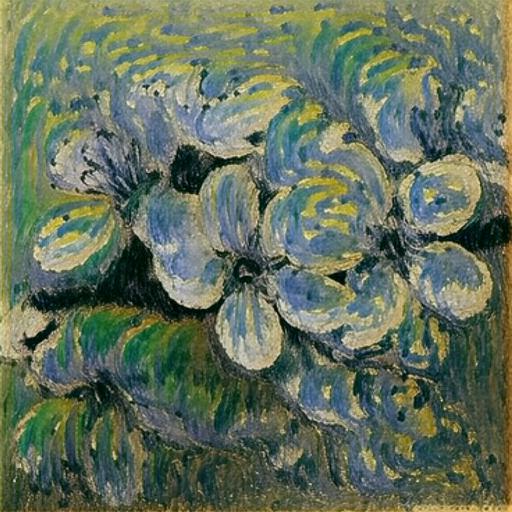}&
		\includegraphics[width=0.13\linewidth]{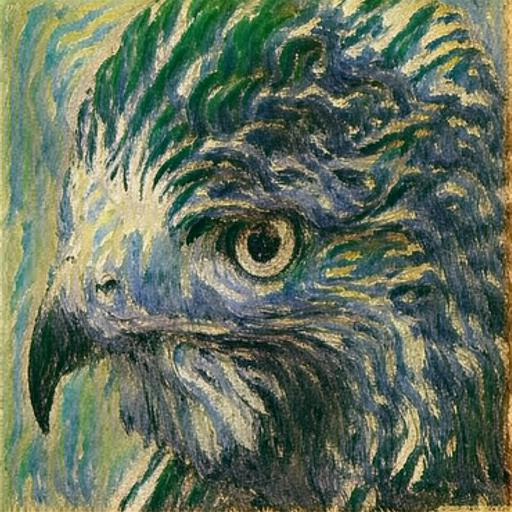}&
		\includegraphics[width=0.13\linewidth]{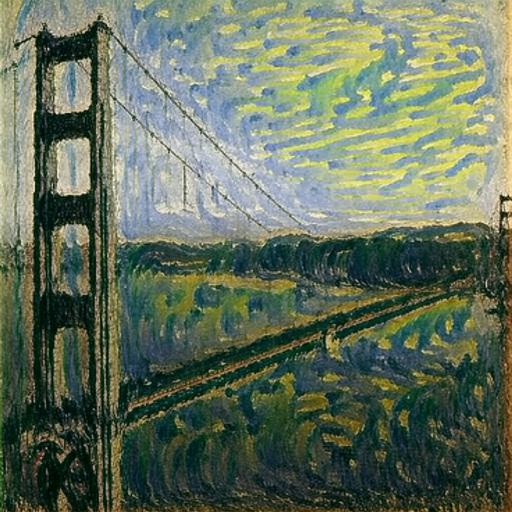}&
		\includegraphics[width=0.13\linewidth]{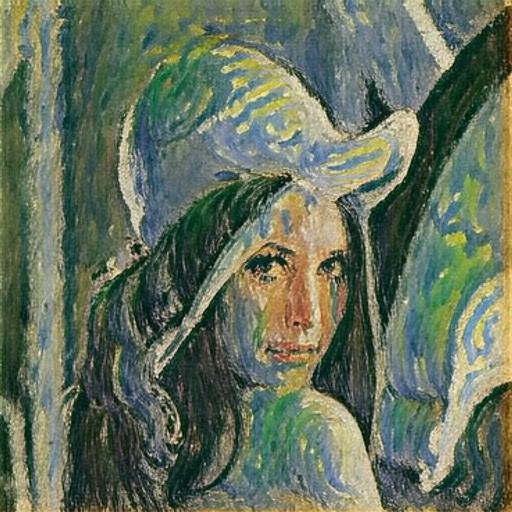}&
		\includegraphics[width=0.13\linewidth]{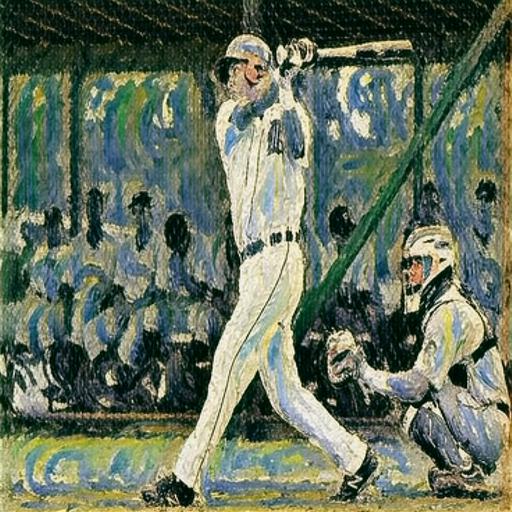}&
		\includegraphics[width=0.13\linewidth]{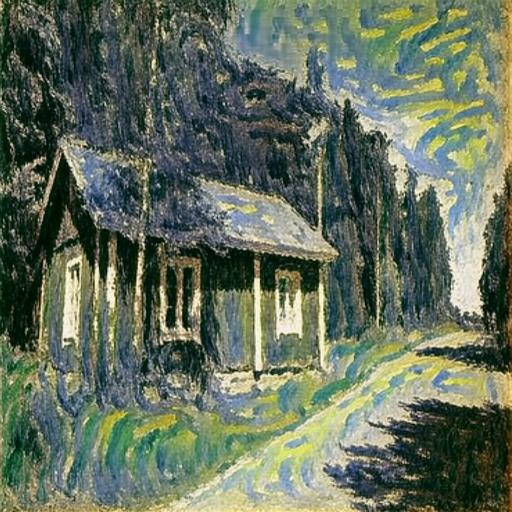}

	\end{tabular}
	\caption{ {\bf Additional stylized results (set 1)} synthesized by our proposed StyleDiffusion. The first row shows content images and the first column shows style images.}
	\label{fig:res1}
\end{figure*}

\begin{figure*}
	\centering
	\setlength{\tabcolsep}{0.1cm}
	\renewcommand\arraystretch{5}
	\begin{tabular}{ccccccc}
		
		&
		\includegraphics[width=0.13\linewidth]{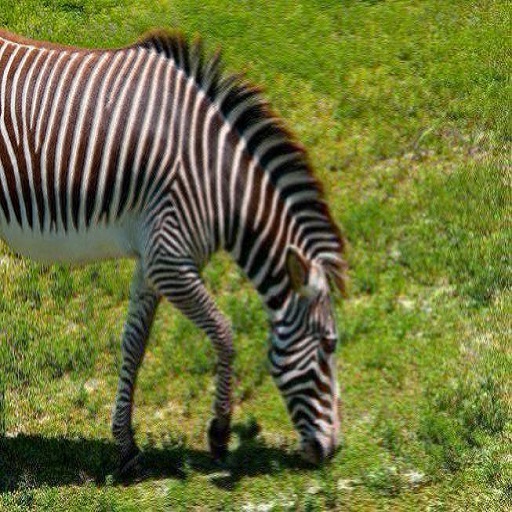}&
		\includegraphics[width=0.13\linewidth]{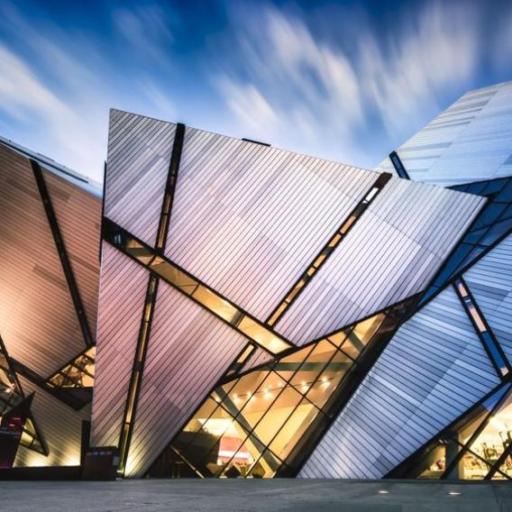}&
		\includegraphics[width=0.13\linewidth]{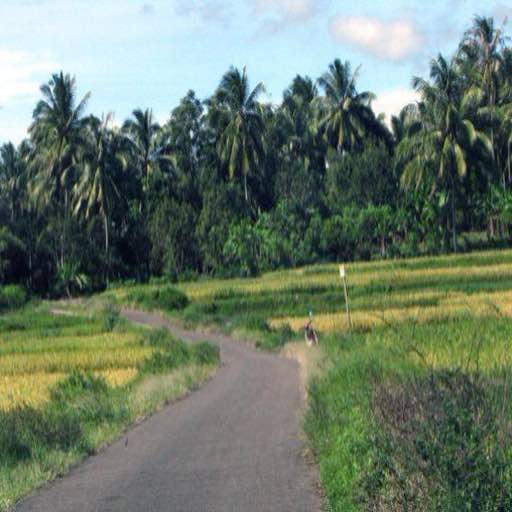}&
		\includegraphics[width=0.13\linewidth]{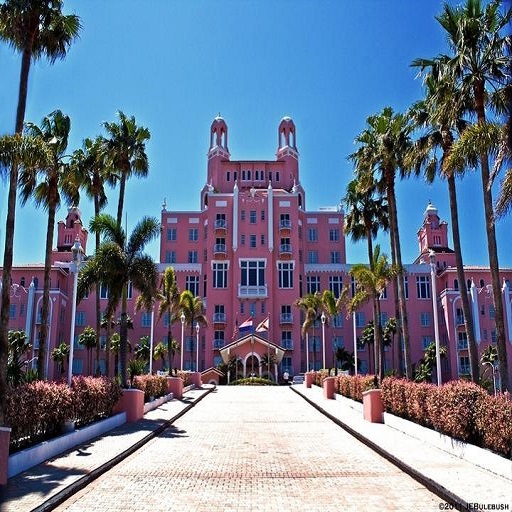}&
		\includegraphics[width=0.13\linewidth]{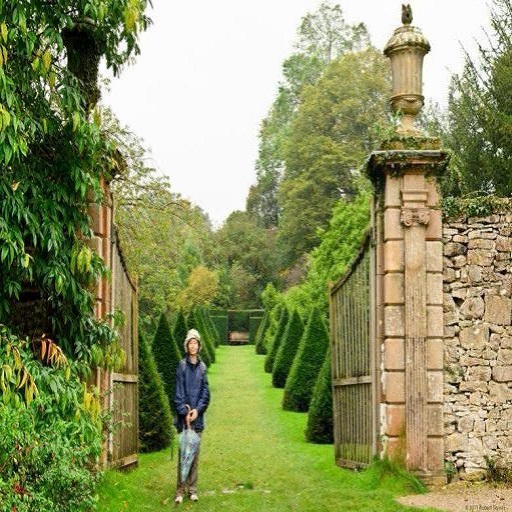}&
		\includegraphics[width=0.13\linewidth]{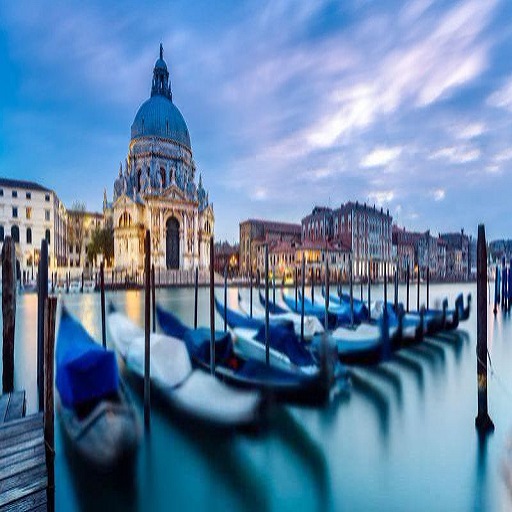}
		
		\\
		
		\includegraphics[width=0.13\linewidth]{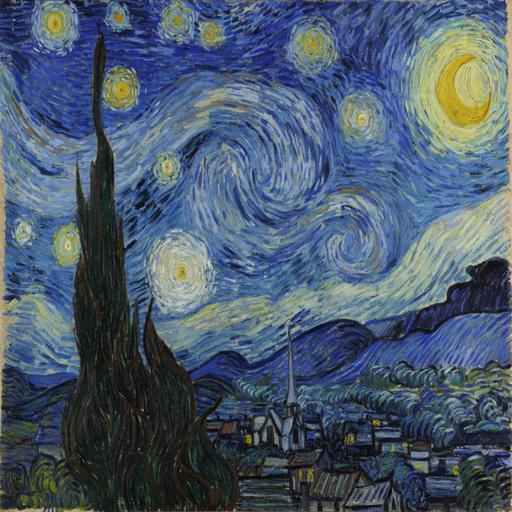}&
		\includegraphics[width=0.13\linewidth]{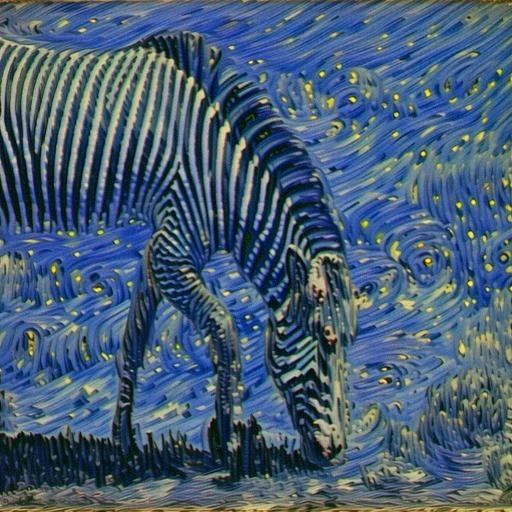}&
		\includegraphics[width=0.13\linewidth]{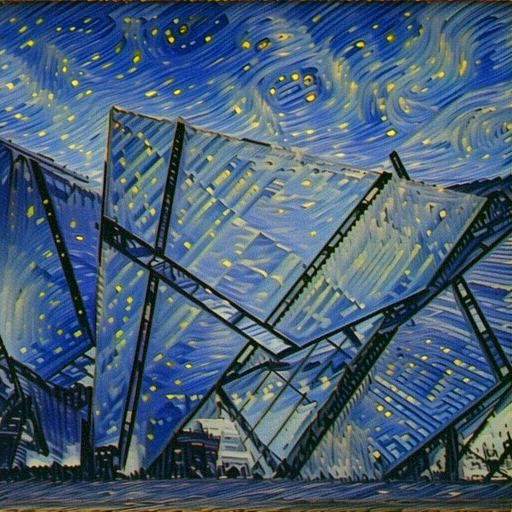}&
		\includegraphics[width=0.13\linewidth]{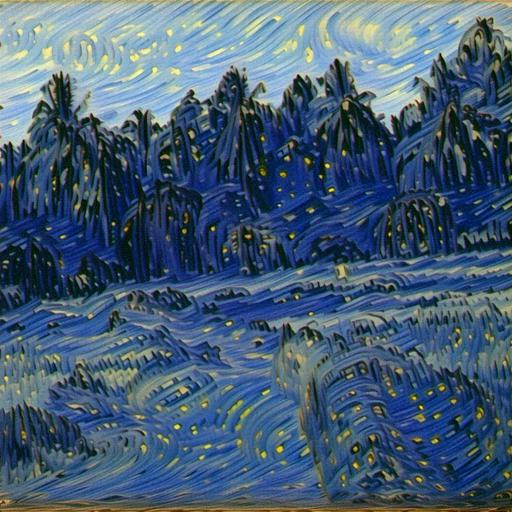}&
		\includegraphics[width=0.13\linewidth]{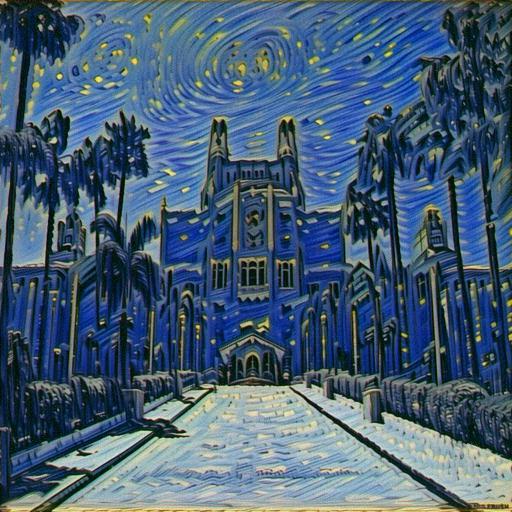}&
		\includegraphics[width=0.13\linewidth]{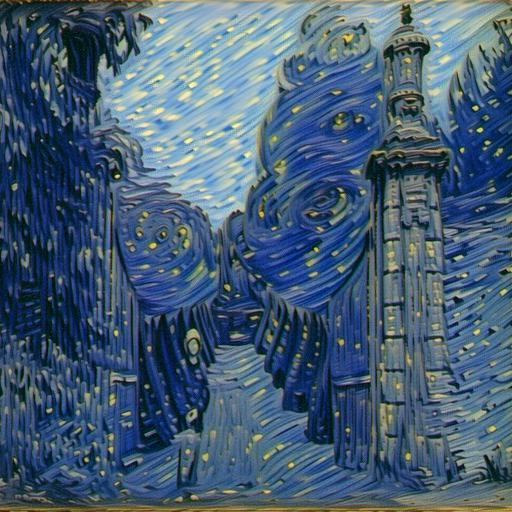}&
		\includegraphics[width=0.13\linewidth]{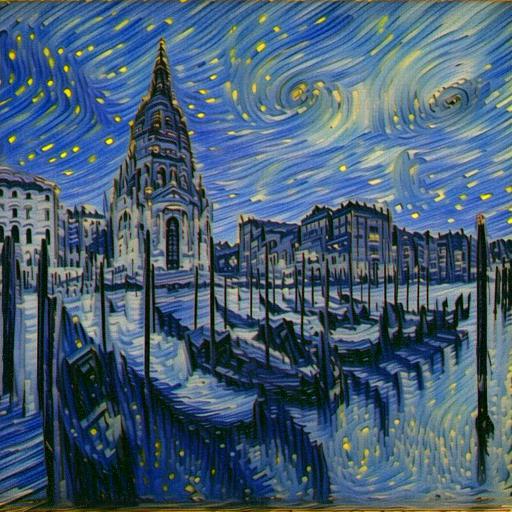}
		
		\\
		
		\includegraphics[width=0.13\linewidth]{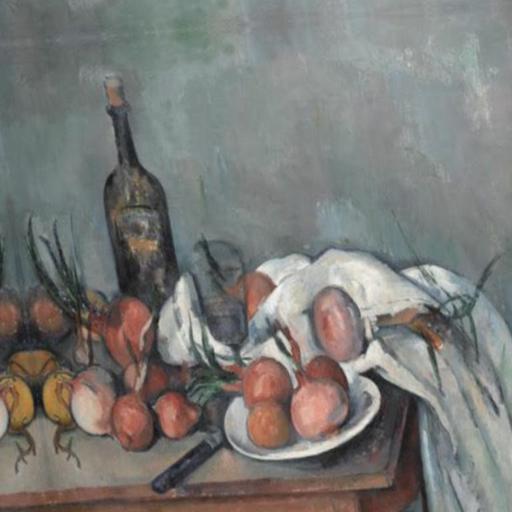}&
		\includegraphics[width=0.13\linewidth]{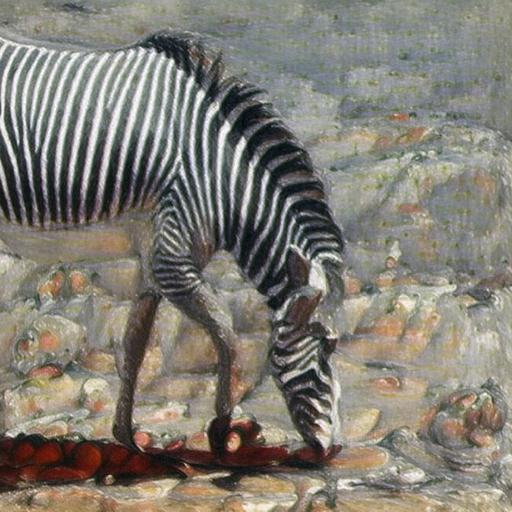}&
		\includegraphics[width=0.13\linewidth]{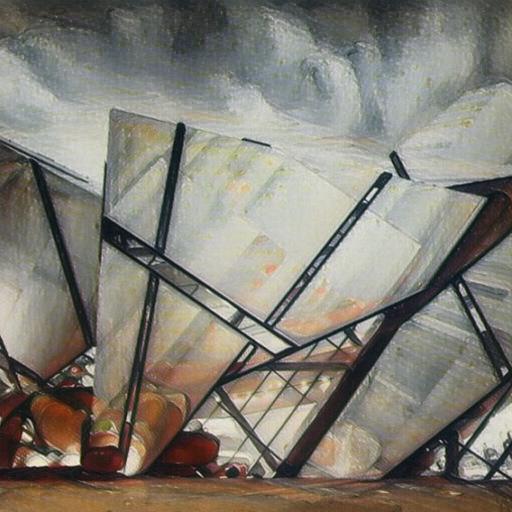}&
		\includegraphics[width=0.13\linewidth]{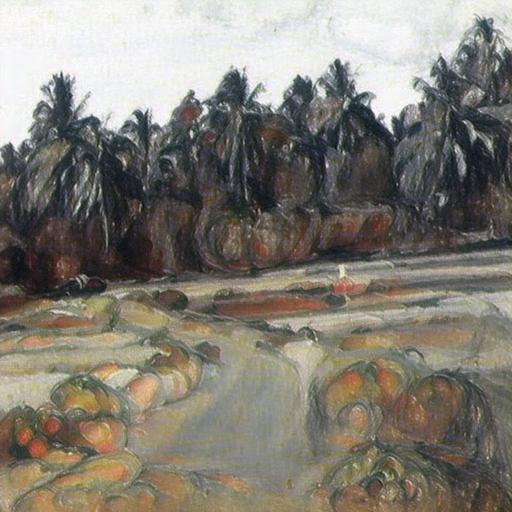}&
		\includegraphics[width=0.13\linewidth]{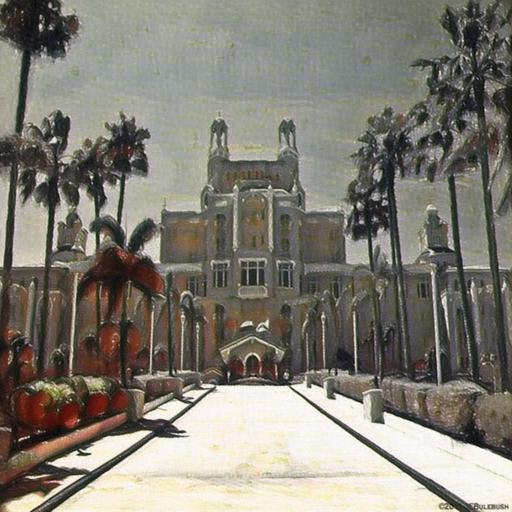}&
		\includegraphics[width=0.13\linewidth]{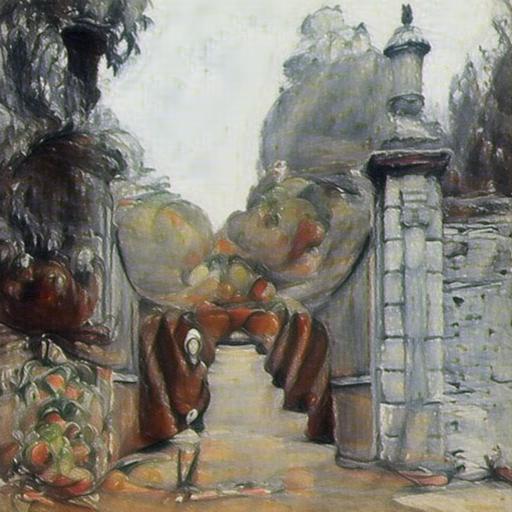}&
		\includegraphics[width=0.13\linewidth]{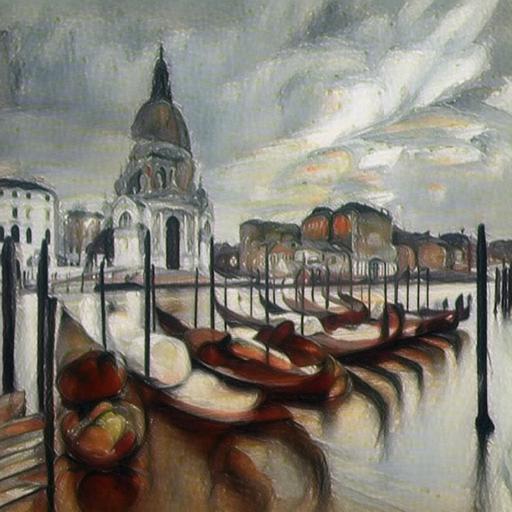}
		
		\\
		
		\includegraphics[width=0.13\linewidth]{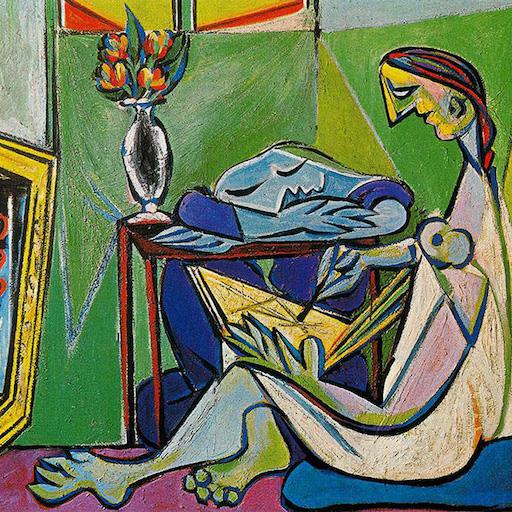}&
		\includegraphics[width=0.13\linewidth]{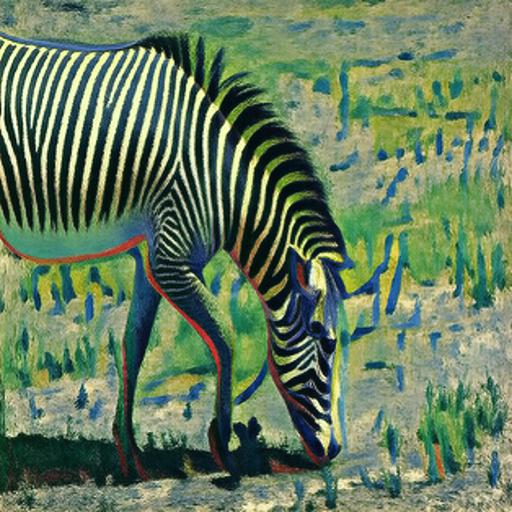}&
		\includegraphics[width=0.13\linewidth]{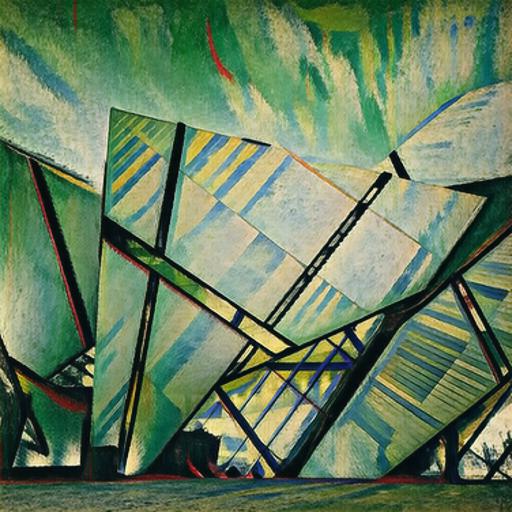}&
		\includegraphics[width=0.13\linewidth]{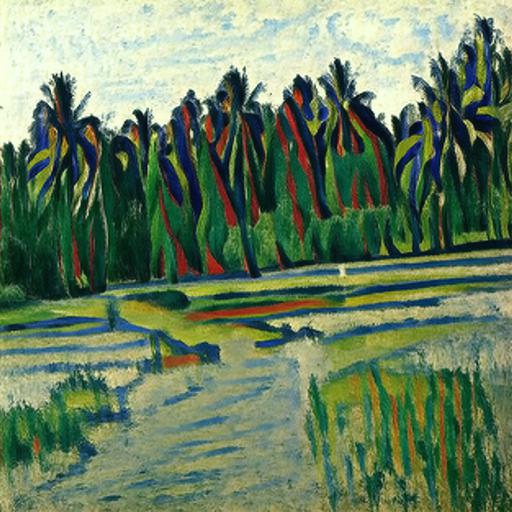}&
		\includegraphics[width=0.13\linewidth]{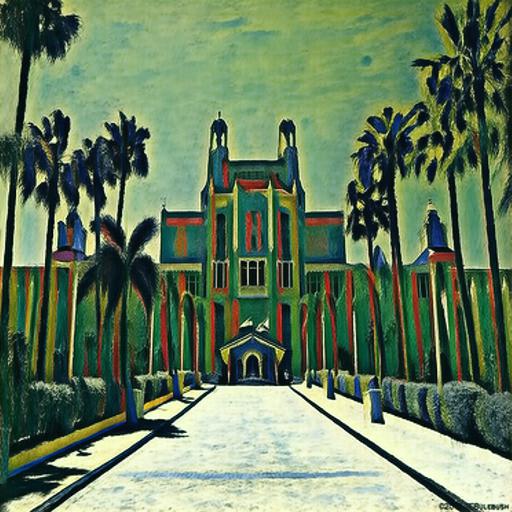}&
		\includegraphics[width=0.13\linewidth]{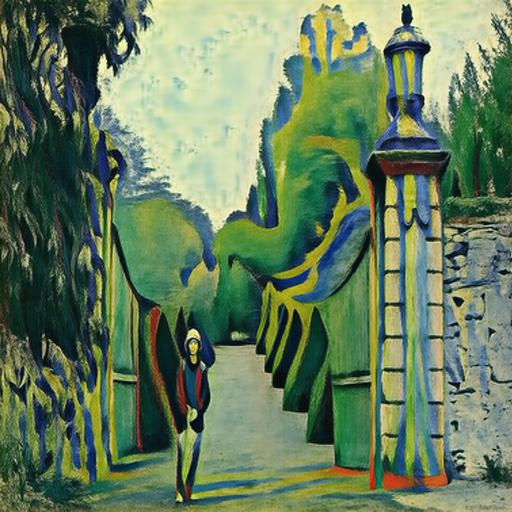}&
		\includegraphics[width=0.13\linewidth]{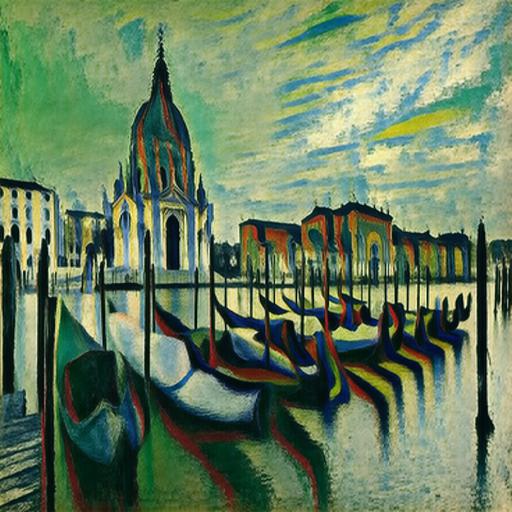}
		
		\\
		
		\includegraphics[width=0.13\linewidth]{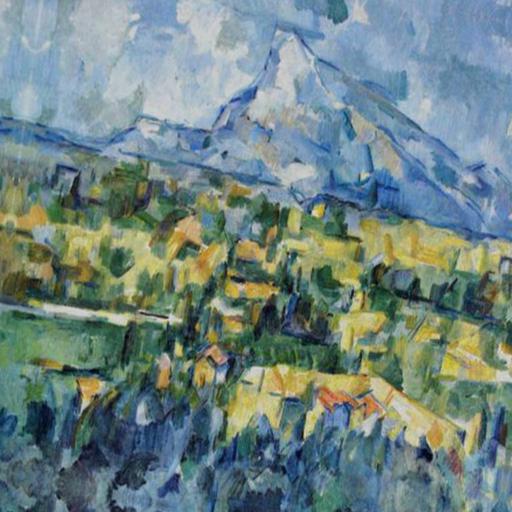}&
		\includegraphics[width=0.13\linewidth]{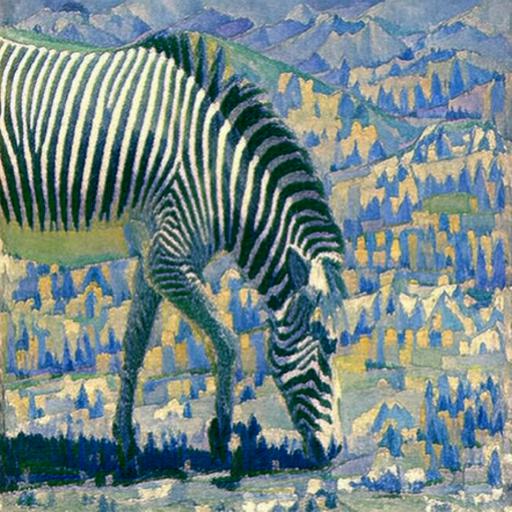}&
		\includegraphics[width=0.13\linewidth]{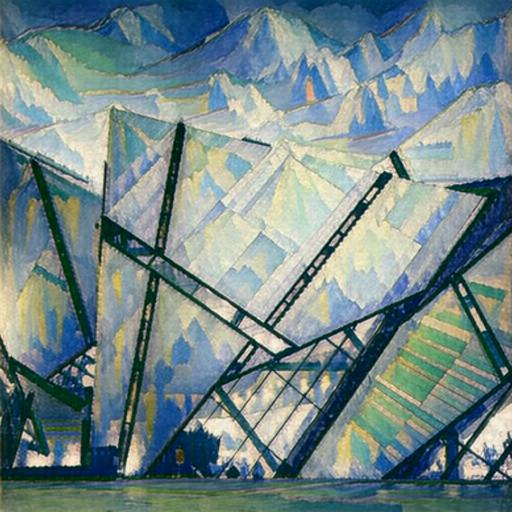}&
		\includegraphics[width=0.13\linewidth]{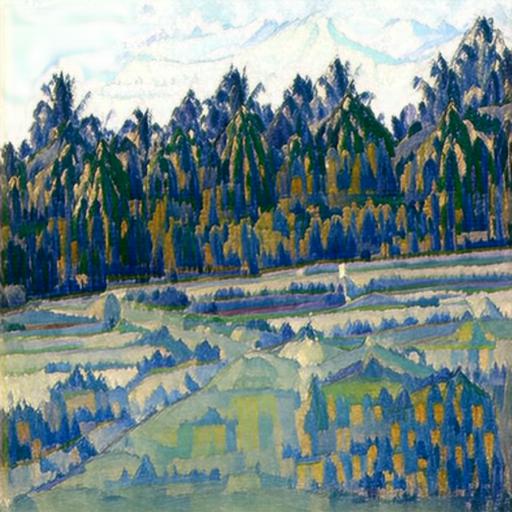}&
		\includegraphics[width=0.13\linewidth]{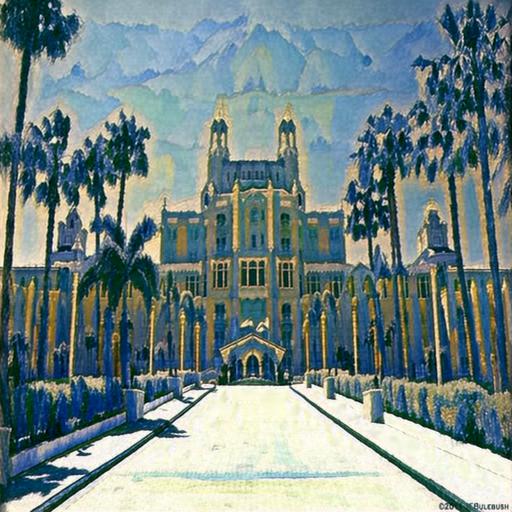}&
		\includegraphics[width=0.13\linewidth]{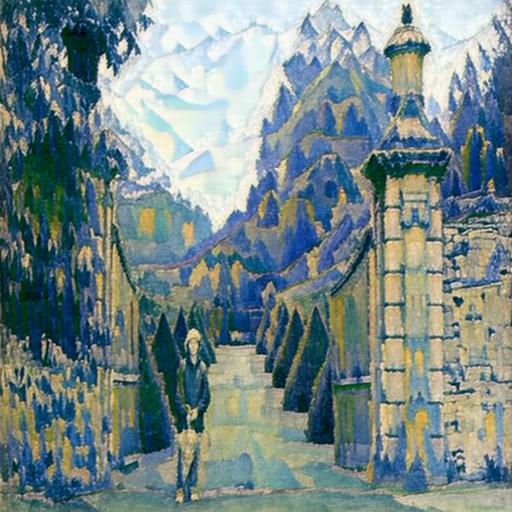}&
		\includegraphics[width=0.13\linewidth]{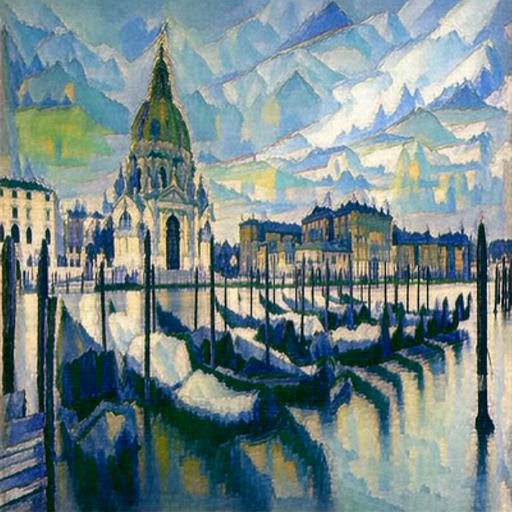}
		
		\\
		
		\includegraphics[width=0.13\linewidth]{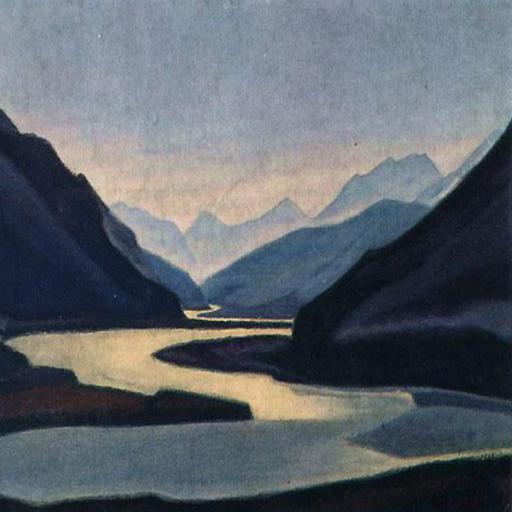}&
		\includegraphics[width=0.13\linewidth]{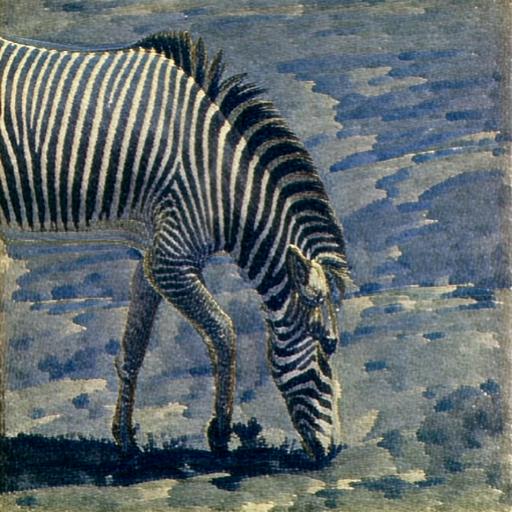}&
		\includegraphics[width=0.13\linewidth]{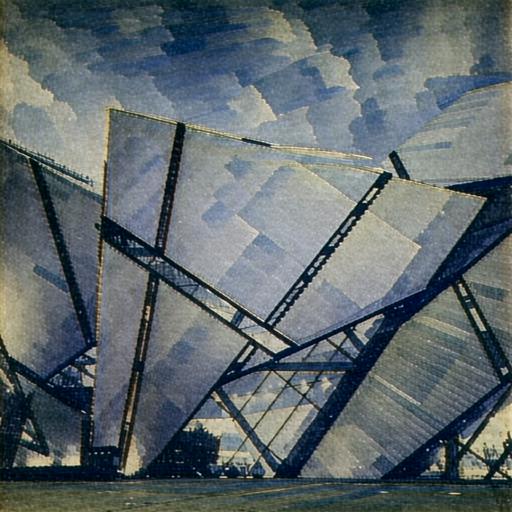}&
		\includegraphics[width=0.13\linewidth]{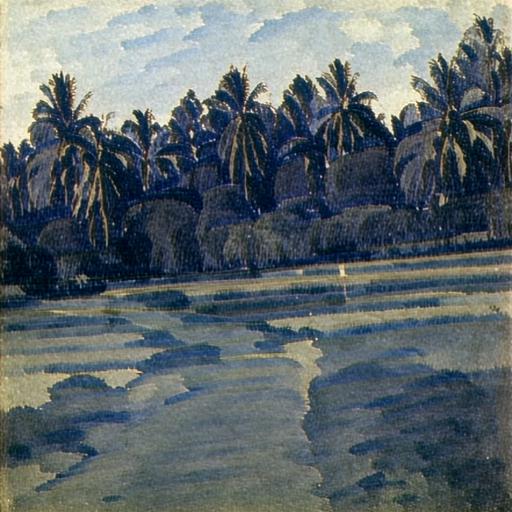}&
		\includegraphics[width=0.13\linewidth]{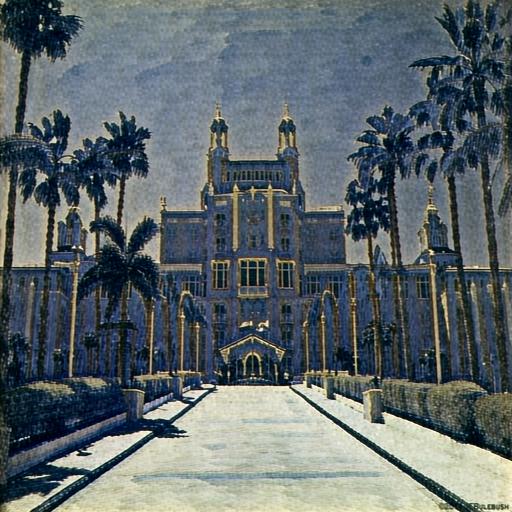}&
		\includegraphics[width=0.13\linewidth]{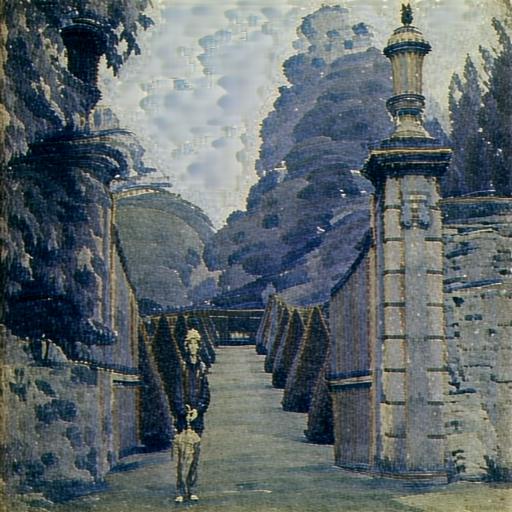}&
		\includegraphics[width=0.13\linewidth]{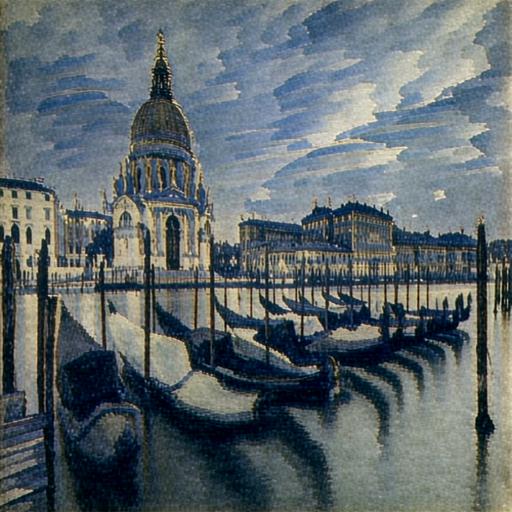}

	\end{tabular}
	\caption{ {\bf Additional stylized results (set 2)} synthesized by our proposed StyleDiffusion. The first row shows content images and the first column shows style images.}
	\label{fig:res2}
\end{figure*}

In Fig.~\ref{fig:res1} and~\ref{fig:res2}, we provide additional stylized results synthesized by our proposed StyleDiffusion.

\section{Limitation and Discussion}
\label{sec:limit}
Except for the limitations we have discussed in the main paper, here we provide some failure cases and analyze the reasons behind them. Further, we also discuss the possible solutions to address them, which may help inspire future improvements to our framework.

\begin{figure}[t]
	\centering
	\setlength{\tabcolsep}{0.02cm}
	\renewcommand\arraystretch{0.4}
	
	\begin{tabular}{cccc}
		
		\includegraphics[width=0.242\linewidth]{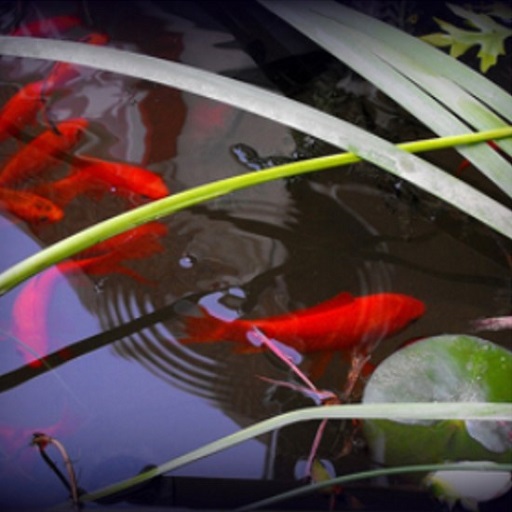}&
		\includegraphics[width=0.242\linewidth]{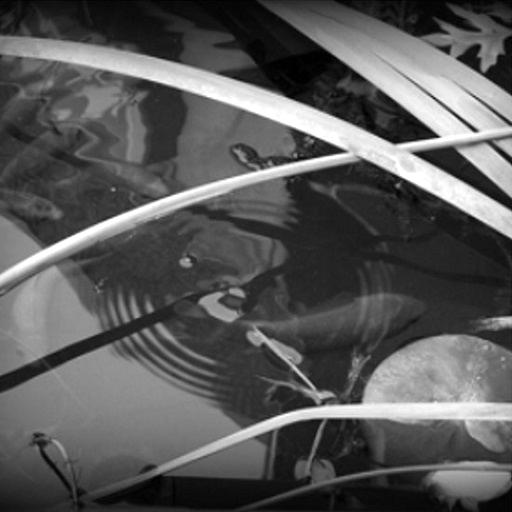}&
		\includegraphics[width=0.242\linewidth]{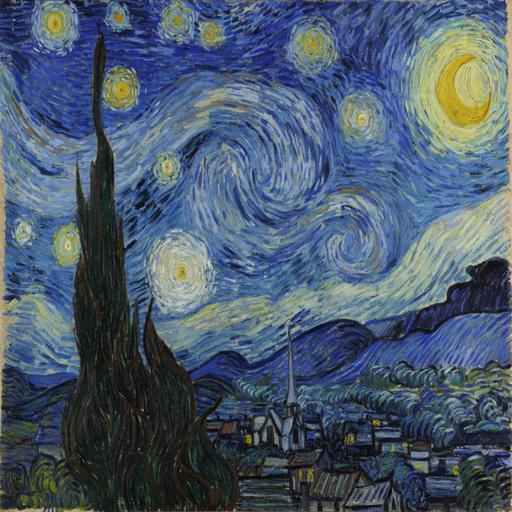}&
		\includegraphics[width=0.242\linewidth]{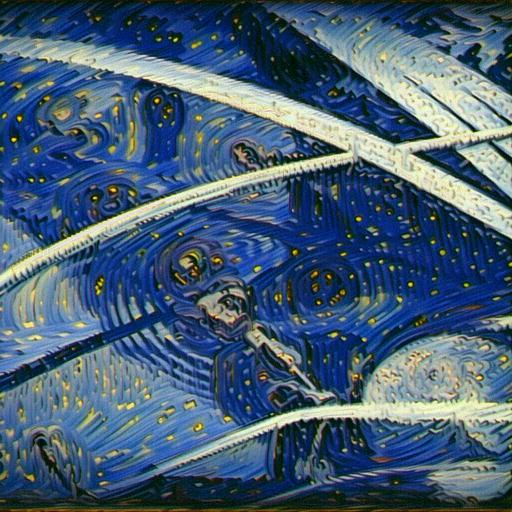}
		\\
		Content & Content gray & Style & Result

	\end{tabular}
	\vspace{0.5em}
	\caption{ {\bf Failure case of type 1: vanishing of salient content.} Some results generated by our method may vanish the salient content of the content image, \eg, the red carps.}
	\label{fig:fail1}
\end{figure}

{\bf Vanishing of Salient Content.} Some of our generated results may vanish the salient content of the content image, \eg, the red carps in Fig.~\ref{fig:fail1}. It can be attributed to the color removal operation used in our style removal module. The commonly used ITU-R 601-2 luma transform~\cite{gonzalez2009digital} may not well preserve the original RGB image's color contrast and color importance, as shown in column 2 of Fig.~\ref{fig:fail1}. We adopt it here mainly for its simplicity and fast speed. This problem may be addressed by using more advanced contrast-preserving decolorization techniques, like~\cite{lu2014contrast}.

\begin{figure}[t]
	\centering
	\setlength{\tabcolsep}{0.02cm}
	\renewcommand\arraystretch{0.4}
	
	\begin{tabular}{cccc}
		
		\includegraphics[width=0.242\linewidth]{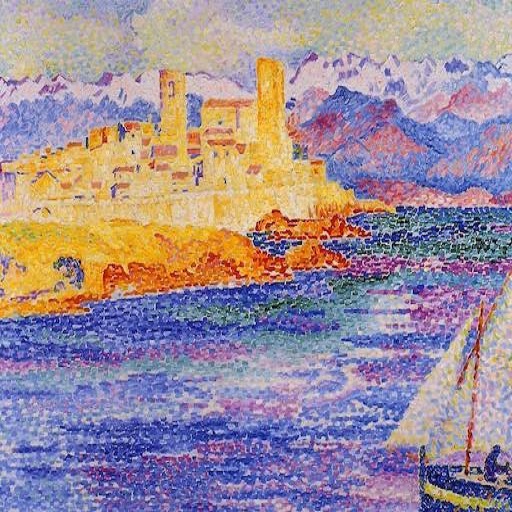}&
		\includegraphics[width=0.242\linewidth]{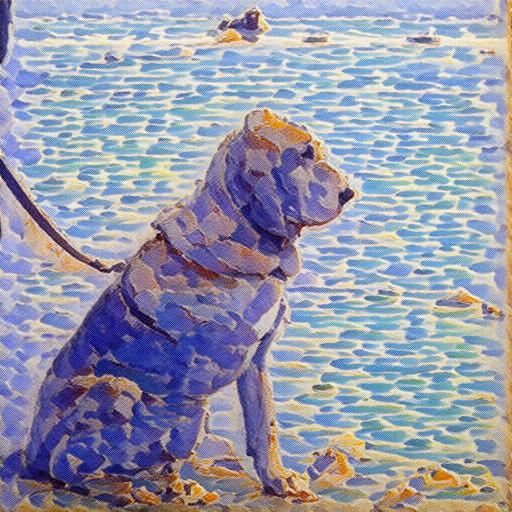}&
		\includegraphics[width=0.242\linewidth]{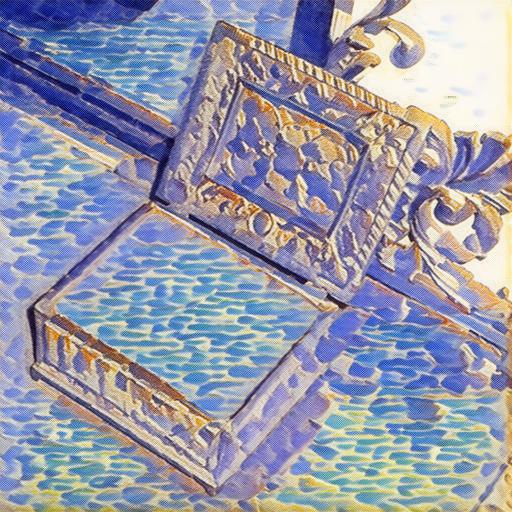}&
		\includegraphics[width=0.242\linewidth]{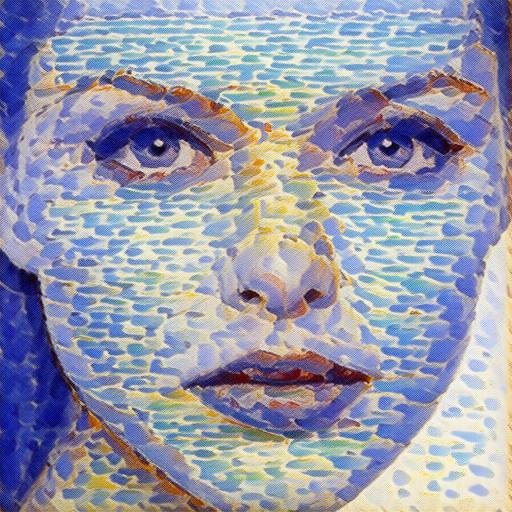}
		\\
		Style & Result 1 & Result 2 & Result 3
		
	\end{tabular}
	\vspace{0.5em}
	\caption{ {\bf Failure case of type 2: biased color distribution.} Our method may generate results that deviate from the color distribution of the style image.}
	\label{fig:fail2}
\end{figure}

{\bf Biased Color Distribution.} As shown in Fig.~\ref{fig:fail2}, though our method learns the challenging pointillism style well, the color distribution seems to stray from that of the style image. This problem can be alleviated by increasing the style reconstruction iteration $K_s$ (see Algorithm~\ref{alg:fine-tune}) to inject more style prior, but the training time also increases significantly. One may consider borrowing some ideas from existing color transfer approaches~\cite{he2018deep} to address this problem.

\begin{figure}[t]
	\centering
	\setlength{\tabcolsep}{0.02cm}
	\renewcommand\arraystretch{0.4}
	
	\begin{tabular}{cccc}
		
		\includegraphics[width=0.242\linewidth]{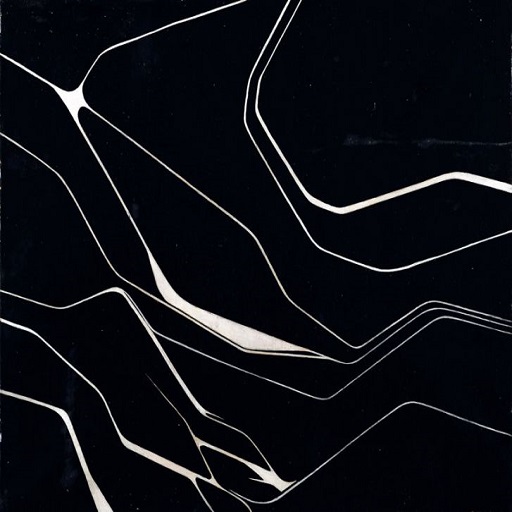}&
		\includegraphics[width=0.242\linewidth]{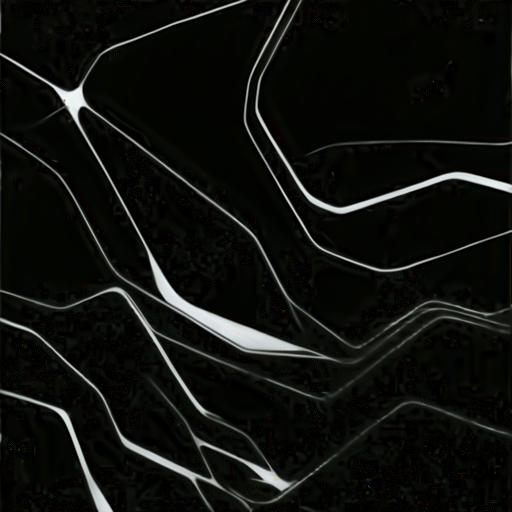}&
		\includegraphics[width=0.242\linewidth]{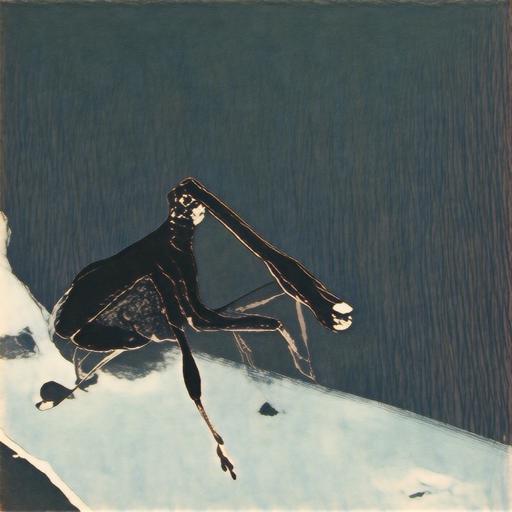}&
		\includegraphics[width=0.242\linewidth]{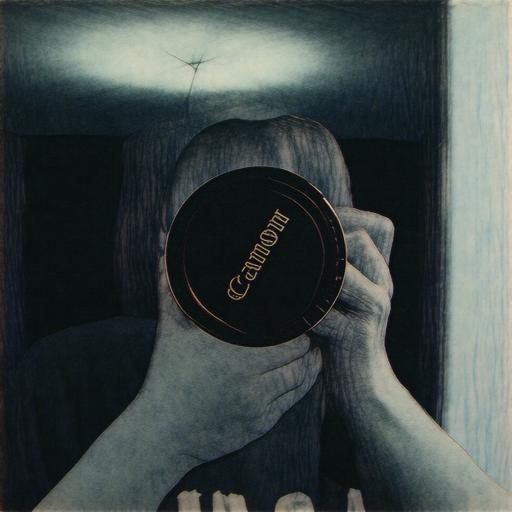}
		\\
		Style & Style removed & Result 1 & Result 2
		
	\end{tabular}
	\vspace{0.5em}
	\caption{ {\bf Failure case of type 3: inseparable content and style.} Our method is hard to transfer plausible style for style images with inseparable content and style. The second column shows the style removed result of the style image.}
	\label{fig:fail3}
\end{figure}

{\bf Inseparable Content and Style.} Our method is hard to achieve plausible style transfer for style images with inseparable content and style, \eg, the simple line art shown in Fig.~\ref{fig:fail3}. Since the content of line art is also its style, our framework is hard to separate them properly, as shown in column 2 of Fig.~\ref{fig:fail3}. One possible solution is to treat line art as the style only and increase the return step $T_{remov}$ of the style removal module to dispel as much style information as possible, or increase the return step $T_{trans}$ of the style transfer module to learn as sufficient line art style as possible.

\clearpage 

\clearpage

\clearpage

\clearpage

\clearpage

\end{document}